%% file: neurips_2022.tex
\documentclass{article}

\usepackage[final]{neurips_2022}

\usepackage[utf8]{inputenc} 
\usepackage[T1]{fontenc}    
\usepackage{hyperref}       
\usepackage{url}            
\usepackage{booktabs}       
\usepackage{amsfonts}       
\usepackage{nicefrac}       
\usepackage{microtype}      
\usepackage[dvipsnames]{xcolor}         

\usepackage{xargs}
\usepackage{amsmath, amsthm, amssymb, amsfonts}
\usepackage{bm}
\usepackage{mathtools}
\usepackage{algorithm, algorithmicx}
\usepackage{subfig}
\usepackage{cleveref}

\input{preamble}

\title{Consistency of Constrained Spectral Clustering under Graph Induced Fair Planted Partitions}

\author{%
  Shubham Gupta \thanks{Work done while the author was at the Indian Institute of Science, Bangalore.} \\
  IBM Research Paris-Saclay \\
  Orsay 91400, France. \\
  \texttt{shubhamg@iisc.ac.in} \\
  \And
  Ambedkar Dukkipati \\
  Computer Science and Automation\\ Indian Institute of Science, 
  Bangalore, India. \\
  \texttt{ambedkar@iisc.ac.in}
}

\begin{document}

\maketitle


\input{abstract}


\input{introduction}


\input{notation_preliminaries}


\input{rasc}


\input{analysis}


\input{numerical_results}


\input{conclusion}


\begin{ack}
    The authors would like to thank the Science and Engineering Research Board (SERB), Department of Science and Technology, Government of India, for their generous funding towards this work through the IMPRINT project: IMP/2019/000383. The authors also thank Muni Sreenivas Pydi for reviewing the manuscript and providing valuable suggestions.
\end{ack}


\bibliographystyle{plainnat}
\bibliography{bibliography}


\newpage
\appendix
\input{appendix_repr_constraint}


\input{appendix_normalized_sc}


\input{appendix_approximate_repsc}


\input{proofs}


\input{appendix_additional_experiments}


\input{appendix_ratio-cut_balance_separated}

\end{document}

%% file: preamble.tex
\newcommand{\bfe}{\mathbf{e}}

\newcommand{\bfh}{\mathbf{h}}

\newcommand{\bfu}{\mathbf{u}}

\newcommand{\bfx}{\mathbf{x}}
\newcommand{\bfy}{\mathbf{y}}

\newcommand{\bfA}{\mathbf{A}}

\newcommand{\bfD}{\mathbf{D}}

\newcommand{\bfG}{\mathbf{G}}
\newcommand{\bfH}{\mathbf{H}}
\newcommand{\bfI}{\mathbf{I}}
\newcommand{\bfJ}{\mathbf{J}}

\newcommand{\bfL}{\mathbf{L}}

\newcommand{\bfP}{\mathbf{P}}
\newcommand{\bfQ}{\mathbf{Q}}
\newcommand{\bfR}{\mathbf{R}}

\newcommand{\bfT}{\mathbf{T}}
\newcommand{\bfU}{\mathbf{U}}
\newcommand{\bfV}{\mathbf{V}}

\newcommand{\bfX}{\mathbf{X}}
\newcommand{\bfY}{\mathbf{Y}}
\newcommand{\bfZ}{\mathbf{Z}}

\newcommand{\rmE}{\mathrm{E}}

\newcommand{\rmP}{\mathrm{P}}

\newcommand{\calA}{\mathcal{A}}

\newcommand{\calC}{\mathcal{C}}
\newcommand{\calD}{\mathcal{D}}
\newcommand{\calE}{\mathcal{E}}

\newcommand{\calG}{\mathcal{G}}

\newcommand{\calJ}{\mathcal{J}}

\newcommand{\calL}{\mathcal{L}}

\newcommand{\calN}{\mathcal{N}}

\newcommand{\calP}{\mathcal{P}}
\newcommand{\calQ}{\mathcal{Q}}
\newcommand{\calR}{\mathcal{R}}

\newcommand{\calV}{\mathcal{V}}

\newcommand{\calX}{\mathcal{X}}
\newcommand{\calY}{\mathcal{Y}}
\newcommand{\calZ}{\mathcal{Z}}

\newcommand{\bbR}{\mathbb{R}}

\newcommand{\bmalpha}{\bm{\alpha}}

\newcommand{\bmmu}{\bm{\mu}}

\newcommand{\bmtheta}{\bm{\theta}}
\newcommand{\bmTheta}{\bm{\Theta}}
\newcommand{\bmone}{\bm{1}}

\newcommandx{\norm}[3][2={}, 3={}]{\vert\vert #1 \vert\vert_{#2}^{#3}}

\newcommand{\trace}[1]{\mathrm{trace}\{#1\}}
\newcommand{\nullspace}[1]{\mathrm{null}\{#1\}}
\newcommand{\rank}[1]{\mathrm{rank}\{#1\}}
\newcommand{\spanof}[1]{\mathrm{span}\{#1\}}
\newcommand{\lambdamax}[1]{\lambda_{\max}(#1)}
\newcommand{\lambdamin}[1]{\lambda_{\min}(#1)}

\newcommandx{\sumlim}[3][1={i}, 2={1}]{\sum_{#1 = #2}^{#3}}
\newcommandx{\prodlim}[3][1={i}, 2={1}]{\prod_{#1 = #2}^{#3}}

\newcommand{\abs}[1]{\vert #1 \vert}

\newcommand{\const}{\mathrm{const}}

\newtheorem{theorem}{Theorem}[section]
\newtheorem{lemma}{Lemma}[section]
\newtheorem{assumption}{Assumption}[section]
\newtheorem{corollary}{Corollary}[section]

\newtheorem{definition}{Definition}[section]

\theoremstyle{remark}

\newtheorem{remark}{Remark}

\newcommand{\rebuttal}[1]{\textcolor{black}{#1}}

%% file: abstract.tex
\begin{abstract}
    Spectral clustering is popular among practitioners and theoreticians alike. 
    While performance guarantees for spectral clustering are well understood, recent studies have focused on enforcing ``fairness'' in clusters, requiring them to be ``balanced'' with respect to a categorical sensitive node attribute (e.g. the race distribution in clusters must match the race distribution in the population).
    In this paper, we consider a setting where sensitive attributes indirectly manifest in an auxiliary \textit{representation graph} rather than being directly observed. This graph specifies node pairs that can represent each other with respect to sensitive attributes and is observed in addition to the usual \textit{similarity graph}. Our goal is to find clusters in the similarity graph while respecting a new individual-level fairness constraint encoded by the representation graph.
    We develop variants of unnormalized and normalized spectral clustering for this task and analyze their performance under a \emph{fair} planted partition model induced by the representation graph. This model uses both the cluster membership of the nodes and the structure of the representation graph to generate random similarity graphs.
    To the best of our knowledge, these are the first consistency results for constrained spectral clustering under an individual-level fairness constraint. Numerical results corroborate our theoretical findings.
\end{abstract}

%% file: introduction.tex
\section{Introduction}
\label{section:introduction}

Consider a recommendation service that groups news articles by finding clusters in a graph which connects these articles via cross-references. Unfortunately, as cross-references between articles with different political viewpoints are uncommon, this service risks forming ideological filter bubbles. To counter polarization, it must ensure that clusters on topics like finance and healthcare include a diverse range of opinions. This is an example of a constrained clustering problem. The popular spectral clustering algorithm \citep{NgEtAl:2001:OnSpectralClustering,Luxburg:2007:ATutorialOnSpectralClustering} has been adapted over the years to include constraints such as \textit{must-link} and \textit{cannot-link} constraints \citep{KamvarEtAl:2003:SpectralLearning, WangDavidson:2010:FlexibleConstrainedSpectralClustering}, size-balanced clusters \citep{BanerjeeGhosh:2006:ScalableClusteringAlgorithmsWithBalancingConstraints}, and statistical fairness \citep{KleindessnerEtAl:2019:GuaranteesForSpectralClusteringWithFairnessConstraints}. These constraints can be broadly divided into two categories: \textbf{(i)} \textit{Population level} constraints that must be satisfied by the clusters as a whole (e.g. size-balanced clusters and statistical fairness); and \textbf{(ii)} \textit{Individual level} constraints that must be satisfied at the level of individual nodes (e.g. must/cannot link constraints). To the best of our knowledge, the only known statistical consistency guarantees for constrained spectral clustering were studied in \citet{KleindessnerEtAl:2019:GuaranteesForSpectralClusteringWithFairnessConstraints} in the context of a \emph{population level} fairness constraint where the goal is to find clusters that are balanced with respect to a categorical sensitive node attribute. In this paper, we establish consistency guarantees for constrained spectral clustering under a new and more general \emph{individual level} fairness constraint.

\paragraph{Informal problem description:} We assume the availability of two graphs: a \textit{similarity graph} $\calG$ in which the clusters are to be found and a \textit{representation graph} $\calR$, defined on the same set of nodes as $\calG$, which encodes the ``is representative of'' relationship. Our goal is to find clusters in $\calG$ such that every node has a sufficient number of its representatives from $\calR$ in all clusters. For example, $\calG$ may be a graph of consumers based on the similarity of their purchasing habits and $\calR$ may be a graph based on the similarity of their sensitive attributes such as gender, race, and sexual orientation. This, for instance, would then be a step towards reducing discrimination in online marketplaces \citep{FismanLuca:2016:FixingDiscriminationInOnlineMarketplaces}.

\paragraph{Contributions and results:} \emph{First}, in \Cref{section:constraint}, we formalize our new individual level fairness constraint for clustering, called the \textit{representation constraint}. It is different from most existing fairness notions which either apply at the population level \citep{ChierichettiEtAl:2017:FairClusteringThroughFairlets,RosnerSchmidt:2018:PrivacyPreservingClusteringWithConstraints,BerceaEtAl:2019:OnTheCostOfEssentiallyFairClusterings,BeraEtAl:2019:FairAlgorithmsForClustering} or are hard to integrate with spectral clustering \citep{ChenEtAl:2019:ProportionallyFairClustering,MahabadiEtAl:2020:IndividualFairnessForKClustering,AndersonEtAl:2020:DistributionalIndividualFairnessInClustering}. Unlike these notions, our constraint can be used with multiple sensitive attributes of different types (categorical, numerical etc.) and only requires observing an abstract representation graph based on these attributes rather than requiring their actual values, thereby discouraging individual profiling. \Cref{appendix:constraint} discusses the utility of individual fairness notions.

\emph{Second}, in \Cref{section:unnormalized_repsc}, we develop the \textit{representation-aware} variant of unnormalized spectral clustering to find clusters that approximately satisfy the proposed constraint. An analogous variant for normalized spectral clustering is presented in \Cref{section:normalized_repsc}.

\emph{Third}, in \Cref{section:rsbm}, we introduce $\calR$-PP, a new representation-aware (or fair) planted partition model. This model generates random similarity graphs $\calG$ conditioned on both the cluster membership of nodes and a given representation graph $\calR$. Intuitively, $\calR$-PP plants the properties of $\calR$ in $\calG$. We show that this model generates ``hard'' problem instances and establish the weak consistency\footnote{An algorithm is called weakly consistent if it makes $o(N)$ mistakes with probability $1 - o(1)$, where $N$ is the number of nodes in the similarity graph $\calG$ \citep{Abbe:2018:CommunityDetectionAndStochasticBlockModels}} of our algorithms under this model for a class of $d$-regular representation graphs (\Cref{theorem:consistency_result_unnormalized,theorem:consistency_result_normalized}). To the best of our knowledge, these are the first consistency results for constrained spectral clustering under an individual-level constraint. In fact, we show that our results imply the only other similar consistency result (but for a population-level constraint) in \citet{KleindessnerEtAl:2019:GuaranteesForSpectralClusteringWithFairnessConstraints} as a special case (\Cref{appendix:constraint}).

Finally, \emph{fourth}, we present empirical results on both real and simulated data to corroborate our theoretical findings (\Cref{section:numerical_results}). In particular, our experiments show that our algorithms perform well in practice, even when the $d$-regularity assumption on $\calR$ is violated.

\paragraph{Related work:} Spectral clustering has been modified to satisfy individual level \emph{must-link} and \emph{cannot-link} constraints by pre-processing the similarity graph \citep{KamvarEtAl:2003:SpectralLearning}, post-processing the eigenvectors of the graph Laplacian \citep{LiEtAl:2009:ConstrainedClusteringViaSpectralRegularization}, and modifying its optimization problem \citep{YuShi:2001:GroupingWithBias,YuShi:2004:SegmentationGivenPartialGroupingConstraints, WangDavidson:2010:FlexibleConstrainedSpectralClustering,WangEtAl:2014:OnConstrainedSpectralClusteringAndItsApplications,CucuringuEtAl:2016:SimpleAndScalableConstrainedClustering}. It has also been extended to accommodate various population level constraints \citep{BanerjeeGhosh:2006:ScalableClusteringAlgorithmsWithBalancingConstraints,XuEtAl:2009:FastNormalizedCutWithLinearConstraints}. We are unaware of theoretical performance guarantees for any of these algorithms.

Of particular interest to us are the fairness constraints for clustering. One popular population level constraint requires sensitive attributes to be proportionally represented in clusters \citep{ChierichettiEtAl:2017:FairClusteringThroughFairlets,RosnerSchmidt:2018:PrivacyPreservingClusteringWithConstraints,BerceaEtAl:2019:OnTheCostOfEssentiallyFairClusterings,BeraEtAl:2019:FairAlgorithmsForClustering,EsmaeiliEtAl:2020:ProbabilisticFairClustering,EsmaeiliEtAl:2021:FairClusteringUnderABoundedCost}. For example, if $50\%$ of the population is female then the same proportion should be respected in all clusters. Several efficient algorithms for discovering such clusters have been proposed \citep{SchmidtEtAl:2018:FairCoresetsAndStreamingAlgorithmsForFairKMeansClustering,AhmadianEtAl:2019:ClusteringWithoutOverRepresentation,HarbShan:2020:KFC}, though they almost exclusively focus on variants of $k$-means while we are interested in spectral clustering. \citet{KleindessnerEtAl:2019:GuaranteesForSpectralClusteringWithFairnessConstraints} deserve a special mention as they develop a spectral clustering algorithm for this fairness notion. However, we recover all the results presented in \citet{KleindessnerEtAl:2019:GuaranteesForSpectralClusteringWithFairnessConstraints} as a special case of our analysis as our proposed constraint interpolates between population level and individual level fairness based on the structure of $\calR$. While individual fairness notions for clustering have also been explored \citep{ChenEtAl:2019:ProportionallyFairClustering,MahabadiEtAl:2020:IndividualFairnessForKClustering,AndersonEtAl:2020:DistributionalIndividualFairnessInClustering,ChakrabartyNagahbani:2021:BetterAlgorithmsForIndividuallyFairKClustering}, none of them have previously been used with spectral clustering. See \citet{CatonHaas:2020:FairnessInMachineLearning} for a broader discussion on fairness.

A final line of relevant work concerns consistency results for variants of unconstrained spectral clustering. \citet{LuxburgEtAl:2008:ConsistencyOfSpectralClustering} established the weak consistency of spectral clustering assuming that the similarity graph $\calG$ encodes cosine similarity between examples using feature vectors drawn from a particular probability distribution. \citet{RoheEtAl:2011:SpectralClusteringAndTheHighDimensionalSBM} and \citet{LeiEtAl:2015:ConsistencyOfSpectralClusteringInSBM} assume that $\calG$ is sampled from variants of the Stochastic Block Model (SBM) \citep{HollandEtAl:1983:StochasticBlockmodelsFirstSteps}. \citet{ZhangEtAl:2014:DetectingOverlappingCommunitiesInNetworksUsingSpectralMethods} allow clusters to overlap. \citet{BinkiewiczEtAl:2017:CovariateAssistedSpectralClustering} consider auxiliary node attributes, though, unlike us, their aim is to find clusters that are well \emph{aligned} with these attributes. A faster variant of spectral clustering was analyzed by \citet{TremblayEtAl:2016:CompressiveSpectralClustering}. Spectral clustering has also been studied on other types of graphs such as hypergraphs \citep{GhoshdastidarDukkipati:2017:ConsistencyOfSpectralHypergraphPartitioningUnderPlantedPartitionModel,GhoshdastidarDukkipati:2017:UniformHypergraphPartitioning} and strong consistency guarantees are also known \citep{GaoEtAl:2017:AchievingOptimalMisclassificationProportionInStochasticBlockModels,LeiZhu:2017:AGenericSampleSplittingApproachForRefinedCommunityRecoveryInSBMs,VuEtAl:2018:ASimpleSVDAlgorithmForFindingHiddenPartitions}, albeit under stronger assumptions.

\paragraph{Notation:} Define $[n] \coloneqq \{1, 2, \dots, n\}$ for any integer $n$. Let $\calG = (\calV, \calE)$ denote a similarity graph, where $\calV = \{v_1, v_2, \dots, v_N\}$ is the set of $N$ nodes and $\calE \subseteq \calV \times \calV$ is the set of edges. Clustering aims to partition the nodes in $\calG$ into $K \geq 2$ non-overlapping clusters $\calC_1, \dots, \calC_K \subseteq \calV$. We assume the availability of another graph, called a \textit{representation graph} $\calR = (\calV, \hat{\calE})$, which is defined on the same set of vertices as $\calG$ but with different edges $\hat{\calE}$. The discovered clusters $\calC_1, \dots, \calC_K$ are required to satisfy a fairness constraint encoded by $\calR$, as described in Section \ref{section:constraint}. $\bfA, \bfR \in \{0, 1\}^{N \times N}$ denote the adjacency matrices of graphs $\calG$ and $\calR$, respectively. We assume that $\calG$ and $\calR$ are undirected and that $\calG$ has no self-loops.

%% file: notation_preliminaries.tex
\section{Unnormalized spectral clustering}
\label{section:unnormalized_spectral_clustering}

We begin with a brief review of unnormalized spectral clustering which will be useful in describing our algorithm in \Cref{section:unnormalized_repsc}. The normalized variants of traditional spectral clustering and our algorithm have been deferred to \Cref{appendix:normalized_variant_of_algorithm}. Given a similarity graph $\calG$, unnormalized spectral clustering finds clusters by approximately minimizing the following  metric known as ratio-cut \citep{Luxburg:2007:ATutorialOnSpectralClustering}
\begin{equation*}
    \mathrm{RCut}(\calC_1, \dots, \calC_K) = \sum_{i = 1}^K \frac{\mathrm{Cut}(\calC_i, \calV \backslash \calC_i)}{\abs{\calC_i}}.
\end{equation*}
Here, $\calV \backslash \calC_i$ is the set difference between $\calV$ and $\calC_i$. For any two subsets $\calX, \calY \subseteq \calV$, $\mathrm{Cut}(\calX, \calY) = \frac{1}{2} \sum_{v_i \in \calX, v_j \in \calY} A_{ij}$ counts the number of edges that have one endpoint in $\calX$ and another in $\calY$. Let $\bfD \in \bbR^{N \times N}$ be a diagonal degree matrix where $D_{ii} = \sum_{j = 1}^N A_{ij}$ for all $i \in [N]$. It is easy to verify that ratio-cut can be expressed in terms of the graph Laplacian $\bfL \coloneqq \bfD - \bfA$ and a cluster membership matrix $\bfH \in \bbR^{N \times K}$ as $\mathrm{RCut}(\calC_1, \dots, \calC_K) = \trace{\bfH^\intercal \bfL \bfH}$, where
\begin{equation}
    \label{eq:H_def}
    H_{ij} = \begin{cases}
        \frac{1}{\sqrt{\abs{\calC_j}}} & \text{ if }v_i \in \calC_j \\
        0 & \text{ otherwise.}
    \end{cases}
\end{equation}
Thus, to find good clusters, one can minimize $\trace{\bfH^\intercal \bfL \bfH}$ over all $\bfH$ that have the form given in \eqref{eq:H_def}. However, the combinatorial nature of this constraint makes this problem NP-hard \citep{WagnerWagner:1993:BetweenMinCutAndGraphBisection}. Unnormalized spectral clustering instead solves the following relaxed problem:
\begin{equation}
    \label{eq:opt_problem_normal}
    \min_{\bfH \in \bbR^{N \times K}}  \,\,\,\, \trace{\bfH^\intercal \bfL \bfH} \,\,\,\, \text{s.t.} \,\,\,\, \bfH^\intercal \bfH = \bfI.
\end{equation}
Note that $\bfH$ in \eqref{eq:H_def} satisfies $\bfH^\intercal \bfH = \bfI$. The above relaxation is often referred to as the spectral relaxation. By Rayleigh-Ritz theorem \citep[Section 5.2.2]{Lutkepohl:1996:HandbookOfMatrices}, the optimal matrix $\bfH^*$ is such that it has $\bfu_1, \bfu_2, \dots, \bfu_K \in \bbR^N$ as its columns, where $\bfu_i$ is the eigenvector corresponding to the $i^{th}$ smallest eigenvalue of $\bfL$ for all $i \in [K]$. The algorithm clusters the rows of $\bfH^*$ into $K$ clusters using $k$-means clustering \citep{Lloyd:1982:LeastSquaresQuantisationInPCM} to return $\hat{\calC}_1, \dots, \hat{\calC}_K$. Algorithm \ref{alg:unnormalized_spectral_clustering} summarizes this procedure. Unless stated otherwise, we will use spectral clustering (without any qualification) to refer to unnormalized spectral clustering.

\begin{algorithm}[t]
    \begin{algorithmic}[1]
        \State \textbf{Input:} Adjacency matrix $\bfA$, number of clusters $K \geq 2$
        \State Compute the Laplacian matrix $\bfL = \bfD - \bfA$.
        \State Compute the first $K$ eigenvectors $\bfu_1, \dots, \bfu_K$ of $\bfL$. Let $\bfH^* \in \bbR^{N \times K}$ be a matrix that has $\bfu_1, \dots, \bfu_K$ as its columns.
        \State Let $\bfh^*_i$ denote the $i^{th}$ row of $\bfH^*$. Cluster $\bfh^*_1, \dots, \bfh^*_N$ into $K$ clusters using $k$-means clustering.
        \State \textbf{Output:} Clusters $\hat{\calC}_1, \dots, \hat{\calC}_K$, \textrm{s.t.} $\hat{\calC}_i = \{v_j \in \calV : \bfh^*_j \text{ was assigned to the }i^{th} \text{ cluster}\}$.
    \end{algorithmic}
    \caption{Unnormalized spectral clustering}
    \label{alg:unnormalized_spectral_clustering}
\end{algorithm}

%% file: rasc.tex

\section{Representation constraint and representation-aware spectral clustering}
\label{section:algorithms}

In this section, we first describe our individual level fairness constraint in \Cref{section:constraint} and then develop \textit{Unnormalized Representation-Aware Spectral Clustering} in \Cref{section:unnormalized_repsc} to find clusters that approximately satisfy this constraint. See \Cref{appendix:normalized_variant_of_algorithm} for the normalized variant of the algorithm.


\subsection{Representation constraint}
\label{section:constraint}

A representation graph $\calR$ connects nodes that represent each other based on sensitive attributes (e.g. political opinions). Let $\calN_{\calR}(i) = \{v_j \; : \; R_{ij} = 1\}$ be the set of neighbors of node $v_i$ in $\calR$. The size of $\calN_{\calR}(i) \cap \calC_k$ specifies node $v_i$'s representation in cluster $\calC_k$. To motivate our constraint, consider the following notion of \textit{balance} $\rho_i$ of clusters defined from the perspective of a particular node $v_i$:
\begin{equation}
    \label{eq:balance}
    \rho_i = \min_{k, \ell \in [K]} \;\; \frac{\abs{\calC_k \cap \calN_{\calR}(i)}}{\abs{\calC_\ell \cap \calN_{\calR}(i)}}
\end{equation}

It is easy to see that $0 \leq \rho_i \leq 1$ and higher values of $\rho_i$ indicate that node $v_i$ has an adequate representation in all clusters. Thus, one objective could be to find clusters $\calC_1, \dots, \calC_K$ that solve the following optimization problem.
\begin{equation}
    \label{eq:general_optimization_problem}
    \min_{\calC_1, \dots, \calC_K} \;\; f(\calC_1, \dots, \calC_K) \;\;\;\; \text{s.t.} \;\;\;\; \rho_i \geq \alpha, \; \forall \; i \in [N],
\end{equation}
where $f(\cdot)$ is inversely proportional to the quality of clusters (such as $\mathrm{RCut}$) and $\alpha \in [0, 1]$ is a user specified threshold. However, it is not clear how this approach can be combined with spectral clustering to develop a consistent algorithm. We take a different approach described below.

First, note that $\min_{i \in [N]} \rho_i \leq \min_{k, \ell \in [K]} \; \frac{\abs{\calC_k}}{\abs{\calC_\ell}}$. Therefore, the balance $\rho_i$ of the least balanced node $v_i$ is maximized when its representatives $\calN_{\calR}(i)$ are split across clusters $\calC_1, \dots, \calC_K$ in proportion to their sizes. Representation constraint requires this condition to be satisfied for each node in the graph.

\begin{definition}[Representation constraint]
    \label{def:representation_constraint}
    Given a representation graph $\calR$, clusters $\calC_1, \dots, \calC_K$ in $\calG$ satisfy the representation constraint if $\abs{\calC_k \cap \calN_{\calR}(i)} \propto \abs{\calC_k}$ for all $i \in [N]$ and $k \in [K]$, i.e.,
    \begin{equation}
        \label{eq:representation_constraint}
        \frac{\abs{\calC_k \cap \calN_{\calR}(i)}}{\abs{\calC_k}} = \frac{\abs{\calN_{\calR}(i)}}{N}, \;\; \forall k \in [K], \; \forall i \in [N].
    \end{equation}
\end{definition}

In other words, the representation constraint requires the representatives of any given node to have a proportional membership in all clusters. For example, if $v_i$ is connected to $30\%$ of all nodes in $\calR$, then it must have $30\%$ representation in all clusters discovered in $\calG$. It is important to note that this constraint applies at the level of individual nodes unlike population level constraints \citep{ChierichettiEtAl:2017:FairClusteringThroughFairlets}. 

While \eqref{eq:general_optimization_problem} can always be solved for a small enough value of $\alpha$ (with the convention that $0/0 = 1$), the constraint in \Cref{def:representation_constraint} may not always be feasible. For example, \eqref{eq:representation_constraint} can never be satisfied if a node has only two representatives (i.e., $\abs{\calN_{\calR}(i)} = 2$) and there are $K > 2$ clusters. However, as exactly satisfying constraints in clustering problems is often NP-hard \citep{DavidsonRavi:2005:ClusteringWithConstraints}, most approaches look for approximate solutions. In the same spirit, our algorithms use spectral relaxation to approximately satisfy \eqref{eq:representation_constraint}, ensuring their wide applicability even when exact satisfaction is impossible.

In practice, $\calR$ can be obtained by computing similarity between nodes based on one or more sensitive attributes (say by taking $k$-nearest neighbors). These attributes can have different types as opposed to existing notions that expect categorical attributes (\Cref{appendix:constraint}). Moreover, once $\calR$ has been calculated, the values of sensitive attributes need not be exposed to the algorithm, thus adding privacy. \Cref{appendix:constraint} presents a toy example to demonstrate the utility of individual level fairness and shows that \eqref{eq:representation_constraint} recovers the population level constraint from \citet{ChierichettiEtAl:2017:FairClusteringThroughFairlets} and \citet{KleindessnerEtAl:2019:GuaranteesForSpectralClusteringWithFairnessConstraints} for particular configurations of $\calR$, thus recovering all results from \citet{KleindessnerEtAl:2019:GuaranteesForSpectralClusteringWithFairnessConstraints} as a special case of our analysis.

Finally, while individual fairness notions have conventionally required similar individuals to be treated similarly \citep{DworkEtAl:2012:FairnessThroughAwareness}, our constraint requires similar individuals (neighbors in $\calR$) to be spread across different clusters (\Cref{def:representation_constraint}). This new type of individual fairness constraint may be of independent interest to the community. Next, we describe one of the proposed algorithms.


\subsection{Unnormalized representation-aware spectral clustering (\textsc{URepSC})}
\label{section:unnormalized_repsc}

The lemma below identifies a sufficient condition that implies the representation constraint and can be added to the optimization problem \eqref{eq:opt_problem_normal} solved by spectral clustering. See \Cref{appendix:proof_of_technical_lemmas_from_algorithms} for the proof.

\begin{lemma}
    \label{lemma:constraint_matrix_unnorm}
    Let $\bfH \in \bbR^{N \times K}$ have the form specified in \eqref{eq:H_def}. The condition 
    \begin{equation}
      \label{eq:matrix_fairness_criteria}
      \bfR \left( \bfI - \frac{1}{N}\bmone\bmone^\intercal \right) \bfH = \mathbf{0}
    \end{equation}
    implies that the corresponding clusters $\calC_1, \dots, \calC_K$ satisfy the constraint in \eqref{eq:representation_constraint}. Here, $\bfI$ is the $N \times N$ identity matrix and $\bmone$ is a $N$-dimensional all-ones vector.
\end{lemma}

With the unnormalized graph Laplacian $\bfL$ defined in \Cref{section:unnormalized_spectral_clustering}, we add the condition from \Cref{lemma:constraint_matrix_unnorm} to the optimization problem after spectral relaxation in \eqref{eq:opt_problem_normal} and solve
\begin{equation}
    \label{eq:opt_problem_with_eq_constraint}
    \min_{\bfH} \;\;\;\; \trace{\bfH^\intercal \bfL \bfH} \;\;\;\; \text{s.t.} \;\;\;\; \bfH^\intercal \bfH = \bfI; \;\;\;\; \bfR \left( \bfI - \frac{1}{N}\bmone\bmone^\intercal \right)\bfH = \mathbf{0}.
\end{equation}
Clearly, the columns of any feasible $\bfH$ must belong to the null space of $\bfR(\bfI - \bmone\bmone^\intercal / N)$. Thus, any feasible $\bfH$ can be expressed as $\bfH = \bfY \bfZ$ for some matrix $\bfZ \in \bbR^{N - r \times K}$, where $\bfY \in \bbR^{N \times N - r}$ is an orthonormal matrix containing the basis vectors for the null space of $\bfR(\bfI - \bmone\bmone^\intercal / N)$ as its columns. Here, $r$ is the rank of $\bfR(\bfI - \bmone\bmone^\intercal / N)$. Because $\bfY^\intercal \bfY = \bfI$, $\bfH^\intercal \bfH = \bfZ^\intercal \bfY^\intercal \bfY \bfZ = \bfZ^\intercal \bfZ$. Thus, $\bfH^\intercal \bfH = \bfI \Leftrightarrow \bfZ^\intercal \bfZ = \bfI$. The following problem is equivalent to \eqref{eq:opt_problem_with_eq_constraint} by setting $\bfH = \bfY \bfZ$.
\begin{equation}
    \label{eq:optimization_problem}
    \min_{\bfZ} \;\;\;\; \trace{\bfZ^\intercal \bfY^\intercal \bfL \bfY \bfZ} \;\;\;\; \text{s.t.} \;\;\;\; \bfZ^\intercal \bfZ = \bfI.
\end{equation}
As in standard spectral clustering, the solution to \eqref{eq:optimization_problem} is given by the $K$ leading eigenvectors of $\bfY^\intercal \bfL \bfY$. Of course, for $K$ eigenvectors to exist, $N - r$ must be at least $K$ as $\bfY^\intercal \bfL \bfY$ has dimensions $N - r \times N - r$. The clusters can then be recovered by using $k$-means clustering to cluster the rows of $\bfH = \bfY \bfZ$, as in \Cref{alg:unnormalized_spectral_clustering}. \Cref{alg:urepsc} summarizes this procedure. We refer to this algorithm as unnormalized representation-aware spectral clustering (\textsc{URepSC}). We make three important remarks before proceeding with the theoretical analysis.

\begin{remark}[Spectral relaxation]
    \label{remark:spectral_relaxation}
    As $\bfR(\bfI - \bmone\bmone^\intercal / N) \bfH = \mathbf{0}$ implies the satisfaction of the representation constraint only when $\bfH$ has the form given in \eqref{eq:H_def}, a feasible solution to \eqref{eq:opt_problem_with_eq_constraint} may not necessarily result in \textit{representation-aware} clusters. In fact, even in the unconstrained case, there are no general guarantees that bound the difference between the optimal solution of \eqref{eq:opt_problem_normal} and the original NP-hard ratio-cut problem \citep{KleindessnerEtAl:2019:GuaranteesForSpectralClusteringWithFairnessConstraints}. Thus, the representation-aware nature of the clusters discovered by solving \eqref{eq:optimization_problem} cannot be guaranteed in general (as is the case with \citep{KleindessnerEtAl:2019:GuaranteesForSpectralClusteringWithFairnessConstraints}). Nonetheless, we show in \Cref{section:analysis} that the discovered clusters indeed satisfy the constraint under certain additional assumptions.
\end{remark}

\begin{remark}[Computational complexity]
    \label{remark:computational_complexity}
    \Cref{alg:urepsc} has a time complexity of $O(N^3)$ and space complexity of $O(N^2)$. Finding the null space of $\bfR(\bfI - \bmone \bmone^\intercal / N)$ to calculate $\bfY$ and computing the eigenvectors of appropriate matrices are the computationally dominant steps. This matches the worst-case complexity of \Cref{alg:unnormalized_spectral_clustering}. For small $K$, several approximations can reduce this complexity, but most such techniques require $K = 2$ \citep{YuShi:2004:SegmentationGivenPartialGroupingConstraints,XuEtAl:2009:FastNormalizedCutWithLinearConstraints}.
\end{remark}

\begin{remark}[Approximate \textsc{URepSC}]
    \label{remark:approximate_urepsc}
    \Cref{alg:urepsc} requires $\rank{\bfR} \leq N - K$ to ensure the existence of $K$ orthonormal eigenvectors of $\bfY^\intercal \bfL \bfY$. When a graph $\calR$ violates this assumption, we instead use the best rank $R$ approximation of its adjacency matrix $\bfR$ ($R \leq N - K$) and refer to this algorithm as \textsc{URepSC (approx.)}. This approximation of $\bfR$ need not have binary elements, but it works well in practice (\Cref{section:numerical_results}). \Cref{appendix:a_note_on_approximate_repsc} provides more intuition behind this low rank approximation, contrasts this strategy with clustering $\calR$ to recover latent sensitive groups that can be reused with existing population level notions, and highlights the challenges associated with finding theoretical guarantees for \textsc{URepSC (approx.)}, which is an interesting direction for future work.
\end{remark}

\begin{algorithm}[t]
    \begin{algorithmic}[1]
        \State \textbf{Input: }Adjacency matrix $\bfA$, representation graph $\bfR$, number of clusters $K \geq 2$
        \State Compute $\bfY$ containing orthonormal basis vectors of $\nullspace{\bfR(\bfI - \frac{1}{N}\bmone\bmone^\intercal)}$
        \State Compute Laplacian $\bfL = \bfD - \bfA$
        \State Compute leading $K$ eigenvectors of $\bfY^\intercal \bfL \bfY$. Let $\bfZ$ contain these vectors as its columns.
        \State Apply $k$-means clustering to rows of $\bfH = \bfY \bfZ$ to get clusters $\hat{\calC}_1, \hat{\calC}_2, \dots, \hat{\calC}_K$
        \State \textbf{Return:} Clusters $\hat{\calC}_1, \hat{\calC}_2, \dots, \hat{\calC}_K$
    \end{algorithmic}
    \caption{\textsc{URepSC}}
    \label{alg:urepsc}
\end{algorithm}

%% file: analysis.tex
\section{Analysis}
\label{section:analysis}

This section shows that \Cref{alg:urepsc,alg:nrepsc} (see \Cref{appendix:normalized_variant_of_algorithm}) recover ground truth clusters with high probability under certain assumptions on the representation graph. We begin by introducing the representation-aware planted partition model in \Cref{section:rsbm}.


\subsection{$\calR$-PP model}
\label{section:rsbm}

The well known Planted Partition random graph model independently connects two nodes in $\calV$ with probability $p$ if they belong to the same cluster and $q$ otherwise, where the ground truth cluster memberships are specified by a function $\pi: \calV \rightarrow [K]$. Below, we define a variant of this model \emph{with respect to} a representation graph $\calR$ and refer to it as the Representation-Aware (or Fair) Planted Partition model or $\calR$-PP.

\begin{definition}[$\calR$-PP]
    \label{def:conditioned_sbm}
    A $\calR$-PP is defined by the tuple $(\pi, \calR, p, q, r, s)$, where $\pi: \calV \rightarrow [K]$ maps nodes in $\calV$ to clusters, $\calR$ is a representation graph, and $1 \geq p \geq q \geq r \geq s \geq 0$ are probabilities used for sampling edges. Under this model, for all $i > j$,
    \begin{equation}
        \label{eq:sbm_specification}
        \rmP(A_{ij} = 1) = \begin{cases}
          p & \text{if } \pi(v_i) = \pi(v_j) \text{ and } R_{ij} = 1, \\
          q & \text{if } \pi(v_i) \neq \pi(v_j) \text{ and } R_{ij} = 1, \\
          r & \text{if } \pi(v_i) = \pi(v_j) \text{ and } R_{ij} = 0,  \\
          s & \text{if } \pi(v_i) \neq \pi(v_j) \text{ and } R_{ij} = 0.
        \end{cases}
      \end{equation}
\end{definition}

Similarity graphs $\calG$ sampled from $\calR$-PP have two interesting properties: \textbf{(i)} Everything else being equal, nodes have a higher tendency of connecting with other nodes in the same cluster ($p \geq q$ and $r \geq s$); and \textbf{(ii)} Nodes connected in $\calR$ have a higher probability of connecting in $\calG$ ($p \geq r$ and $q \geq s$). Thus, $\calR$-PP plants both the clusters in $\pi$ and the properties of $\calR$ into the sampled graph $\calG$.

\begin{remark}[$\calR$-PP and ``hard'' problem instances] 
    Clusters satisfying \eqref{eq:representation_constraint} must proportionally distribute the nodes connected in $\calR$ amongst themselves. However, $\calR$-PP makes nodes connected in $\calR$ more likely to connect in $\calG$, even if they belong to different clusters ($q \geq r$). In this sense, graphs sampled from $\calR$-PP are ``hard'' instances for our algorithms. 
\end{remark}

When $\calR$ itself has latent groups, there are two natural ways to cluster the nodes: \textbf{(i)} Based on the clusters specified by $\pi$; and \textbf{(ii)} Based on the clusters in $\calR$. The clusters based on option \textbf{(ii)} are likely to not satisfy \eqref{eq:representation_constraint} as tightly connected nodes in $\calR$ will be assigned to the same cluster. We show in the next section that, under certain assumptions, $\pi$ can be defined so that the clusters encoded by it satisfy \eqref{eq:representation_constraint} by construction. Recovering these ground truth clusters (instead of other natural choices like option \textbf{(ii)}) then amounts to recovering \textit{representation-aware} clusters.


\subsection{Consistency results}
\label{section:consistency_results}

As noted in \Cref{section:constraint}, some representation graphs lead to constraints that cannot be satisfied. For our theoretical analysis, we restrict our focus to a case where the constraint in \eqref{eq:representation_constraint} is feasible. Towards this end, an additional assumption on $\calR$ is required.

\begin{assumption}
    \label{assumption:R_is_d_regular}
    $\calR$ is a $d$-regular graph for $K \leq d \leq N$. Moreover, $R_{ii} = 1$ for all $i \in [N]$ and each node in $\calR$ is connected to $d / K$ nodes from cluster $\calC_j$ for all $j \in [K]$ (including the self-loop).
\end{assumption}

\Cref{assumption:R_is_d_regular} ensures the existence of a $\pi$ for which the ground-truth clusters satisfy \eqref{eq:representation_constraint}. Namely, assuming equal-sized clusters, set $\pi(v_i) = k$ if $(k - 1) \frac{N}{K} \leq i \leq k \frac{N}{K}$ for all $i \in [N]$ and $k \in [K]$. Before presenting our main results, we need additional notation. Let $\bmTheta \in \{0, 1\}^{N \times K}$ indicate the ground-truth cluster memberships encoded by $\pi$ (i.e., $\Theta_{ij} = 1 \Leftrightarrow v_i \in \calC_j$) and $\hat{\bmTheta} \in \{0, 1\}^{N \times K}$ indicate the clusters returned by the algorithm ($\hat{\Theta}_{ij} = 1 \Leftrightarrow v_i \in \hat{\calC}_j$). With $\calJ$ as the set of all $K \times K$ permutation matrices, the fraction of misclustered nodes is defined as $M(\bmTheta, \hat{\bmTheta}) = \min_{\bfJ \in \calJ} \frac{1}{N} \norm{\bmTheta - \hat{\bmTheta} \bfJ}[0]$ \citep{LeiEtAl:2015:ConsistencyOfSpectralClusteringInSBM}. \Cref{theorem:consistency_result_unnormalized,theorem:consistency_result_normalized} use the eigenvalues of the Laplacian matrix in the expected case, defined as $\calL = \calD - \calA$, where $\calA = \rmE[\bfA]$ is the expected adjacency matrix of a graph sampled from $\calR$-PP and $\calD \in \bbR^{N \times N}$ is its corresponding degree matrix. The next two results establish high-probability upper bounds on the fraction of misclustered nodes for \textsc{URepSC} and \textsc{NRepSC} (see \Cref{appendix:normalized_variant_of_algorithm}) for similarity graphs $\calG$ sampled from $\calR$-PP.

\begin{theorem}[Error bound for \textsc{URepSC}]
    \label{theorem:consistency_result_unnormalized}
    Let $\rank{\bfR} \leq N - K$ and assume that all clusters have equal sizes. Let $\mu_1 \leq \mu_2 \leq \dots \leq \mu_{N - r}$ denote the eigenvalues of $\bfY^\intercal \calL \bfY$, where $\bfY$ was defined in \Cref{section:unnormalized_repsc}. Define $\gamma = \mu_{K + 1} - \mu_{K}$. Under \Cref{assumption:R_is_d_regular}, there exists a universal constant $\const(C, \alpha)$, such that if $\gamma$ satisfies $\gamma^2 \geq \const(C, \alpha) (2 + \epsilon) p N K \ln N$ and $p \geq C \ln N / N$ for some $C > 0$, then $$M(\bmTheta, \hat{\bmTheta}) \leq \const(C, \alpha) \frac{(2 + \epsilon)}{\gamma^2} p N \ln N$$ for every $\epsilon > 0$ with probability at least $1 - 2 N^{-\alpha}$ when a $(1 + \epsilon)$-approximate algorithm for $k$-means clustering is used in Step 5 of \Cref{alg:urepsc}.
\end{theorem}

\begin{theorem}[Error bound for \textsc{NRepSC}]
    \label{theorem:consistency_result_normalized}
    Let $\rank{\bfR} \leq N - K$ and assume that all clusters have equal sizes. Let $\mu_1 \leq \mu_2 \leq \dots \leq \mu_{N - r}$ denote the eigenvalues of $\calQ^{-1} \bfY^\intercal \calL \bfY \calQ^{-1}$, where $\calQ = \sqrt{\bfY^\intercal \calD \bfY}$ and $\bfY$ was defined in \Cref{section:unnormalized_repsc}. Define $\gamma = \mu_{K + 1} - \mu_{K}$ and $\lambda_1 = qd + s(N - d) + (p - q) \frac{d}{K} + (r - s) \frac{N - d}{K}$. Under \Cref{assumption:R_is_d_regular}, there are universal constants $\const_1(C, \alpha)$, $\const_2(C, \alpha)$, and $\const_3(C, \alpha)$ such that if:
    \begin{enumerate}
        \item $\left(\frac{\sqrt{p N \ln N}}{\lambda_1 - p}\right) \left(\frac{\sqrt{p N \ln N}}{\lambda_1 - p} + \frac{1}{6\sqrt{C}}\right) \leq \frac{1}{16(\alpha + 1)}$,
        \item $\frac{\sqrt{p N \ln N}}{\lambda_1 - p} \leq \const_2(C, \alpha)$, and
        \item $16(2 + \epsilon)\left[ \frac{8 \const_3(C, \alpha) \sqrt{K}}{\gamma} + \const_1(C, \alpha)\right]^2 \frac{p N^2 \ln N}{(\lambda_1 - p)^2} < \frac{N}{K}$,
    \end{enumerate}
    and $p \geq C \ln N / N$ for some $C > 0$, then,
    $$M(\bmTheta, \hat{\bmTheta}) \leq 32(2 + \epsilon)\left[ \frac{8 \const_3(C, \alpha) \sqrt{K}}{\gamma} + \const_1(C, \alpha)\right]^2 \frac{p N \ln N}{(\lambda_1 - p)^2},$$
    for every $\epsilon > 0$ with probability at least $1 - 2 N^{-\alpha}$ when a $(1 + \epsilon)$-approximate algorithm for $k$-means clustering is used in Step 6 of \Cref{alg:nrepsc}.
\end{theorem}

All proofs have been deferred to \Cref{section:proof_of_theorems}. Briefly, we show that the top $K$ eigenvectors of $\calL$ \textbf{(i)} recover ground-truth clusters in the expected case (\Crefrange{lemma:introducing_uks}{lemma:orthonormal_eigenvectors_y2_yK}) and \textbf{(ii)} lie in the null space of $\bfR(\bfI - \bmone \bmone^\intercal / N)$ and hence are also the top $K$ eigenvectors of $\bfY^\intercal \calL \bfY$ (\Cref{lemma:first_K_eigenvectors_of_L}). Matrix perturbation arguments then establish a high probability mistake bound in the general case when the graph $\calG$ is sampled from a $\calR$-PP (Lemmas \ref{lemma:bound_on_D-calD}--\ref{lemma:k_means_error}). Next, we discuss our assumptions and use the error bounds above to establish the weak consistency of our algorithms. 

\subsection{Discussion}
\label{section:discussion}

Note that $\bfI - \bmone \bmone^\intercal / N$ is a projection matrix and $\bmone$ is its eigenvector with eigenvalue $0$. Any vector orthogonal to $\bmone$ is an eigenvector with eigenvalue $1$. Thus, $\rank{\bfI - \bmone \bmone^\intercal / N} = N - 1$. Because $\rank{\bfR (\bfI - \bmone \bmone^\intercal / N)} \leq \min(\rank{\bfR}, \rank{\bfI - \bmone \bmone^\intercal / N})$, requiring $\rank{\bfR} \leq N - K$ ensures that $\rank{\bfR(\bfI - \bmone \bmone^\intercal / N)} \leq N - K$, which is necessary for \eqref{eq:optimization_problem} to have a solution. The assumption on the size of the clusters and the $d$-regularity of $\calR$ allows us to compute the smallest $K$ eigenvalues of the Laplacian matrix in the expected case. This is a crucial step in our proof. 

\rebuttal{In \Cref{theorem:consistency_result_normalized}, $\lambda_1$ is defined such that the largest eigenvalue of the expected adjacency matrix under the $\calR$-PP model is given by $\lambda_1 - p$ (see \eqref{eq:eigenvector_A_tildeA} and \Cref{lemma:uk_eigenvector_of_tildeA}). The three assumptions in Theorem 4.2 essentially control the minimum rate at which $\lambda_1$ must grow with $N$. For instance, the first two assumptions can be replaced by a simpler but slightly stronger (yet practical) condition: $\lambda_1 - p = \omega(\sqrt{p N \ln N})$. This, for example, is satisfied in a realistic setting where $K = O(\ln N)$ and $d = O(\ln N)$. The third assumption also controls the rate of growth of $\lambda_1$, but in the context of the community structure contained in $\mathcal{Q}^{-1} \mathbf{Y}^T \mathcal{L} \mathbf{Y} \mathcal{Q}^{-1}$, as encoded by the eigengap $\gamma$. So, $\lambda_1$  must not increase at the expense of the community structure (e.g., by setting $p=q=r=s=1$).}

\begin{remark}
    In practice, \Cref{alg:urepsc,alg:nrepsc} only require the rank assumption on $\bfR$ to ensure the feasibility of the corresponding optimization problems. The assumptions on the size of clusters and $d$-regularity of $\calR$ are only needed for our theoretical analysis.
\end{remark}

The next two corollaries establish the weak consistency of our algorithms as a direct consequence of \Cref{theorem:consistency_result_unnormalized,theorem:consistency_result_normalized}.

\begin{corollary}[Weak consistency of \textsc{URepSC}]
    \label{corollary:weak_consistency_urepsc}
    Under the same setup as \Cref{theorem:consistency_result_unnormalized}, for \textsc{URepSC}, $M(\bmTheta, \hat{\bmTheta}) = o(1)$ with probability $1 - o(1)$ if $\gamma = \omega(\sqrt{pN K \ln N})$.
\end{corollary}

\begin{corollary}[Weak consistency of \textsc{NRepSC}]
    \label{corollary:weak_consistency_nrepsc}
    Under the same setup as \Cref{theorem:consistency_result_normalized}, for \textsc{NRepSC}, $M(\bmTheta, \hat{\bmTheta}) = o(1)$ with probability $1 - o(1)$ if $\gamma = \omega(\sqrt{pN K \ln N} / (\lambda_1 - p))$.
\end{corollary}

The conditions on $\gamma$ are satisfied in many interesting cases. For example, when there are $P$ \textit{protected groups} as in \citet{ChierichettiEtAl:2017:FairClusteringThroughFairlets}, the equivalent representation graph has $P$ cliques that are not connected to each other (see Appendix \ref{appendix:constraint}). \citet{KleindessnerEtAl:2019:GuaranteesForSpectralClusteringWithFairnessConstraints} show that $\gamma = \theta(N/K)$ in this case (for the unnormalized variant), which satisfies the criterion given above if $K$ is not too large.

Finally, Theorems \ref{theorem:consistency_result_unnormalized} and \ref{theorem:consistency_result_normalized} require a $(1 + \epsilon)$-approximate solution to $k$-means clustering. Several efficient algorithms have been proposed in the literature for this task \citep{KumarEtAl:2004:ASimpleLinearTimeApproximateAlgorithmForKMeansClusteringInAnyDimension,ArthurVassilvitskii:2007:KMeansTheAdvantagesOfCarefulSeeding,AhmadianEtAl:2017:BetterGuaranteesForKMeansAndEuclideanKMedianByPrimalDualAlgorithms}. Such algorithms are also available in commonly used software packages like MATLAB and scikit-learn\footnote{\rebuttal{The algorithm in \citet{KumarEtAl:2004:ASimpleLinearTimeApproximateAlgorithmForKMeansClusteringInAnyDimension} runs in linear time in $N$ only when $K$ is a constant. When $K$ grows with $N$, one can instead use other practical variants whose average time complexity is linear in both $N$ and $K$ (e.g., in scikit-learn). These variants are often run with multiple seeds in practice to avoid local minima.}}. The assumption that $p \geq C \ln N / N$ controls the sparsity of the graph and is required in the consistency proofs for standard spectral clustering as well \citep{LeiEtAl:2015:ConsistencyOfSpectralClusteringInSBM}.


%% file: numerical_results.tex
\section{Numerical results}
\label{section:numerical_results}

We experiment with three types of graphs: synthetically generated $d$-regular and non-$d$-regular representation graphs and a real-world dataset. See \Cref{appendix:additional_experiments} for analogous results for \textsc{NRepSC}, the normalized variant of our algorithm. Before proceeding further, we make an important remark.

\begin{figure}[t]
    \centering
    \subfloat[][Accuracy vs no. of nodes]{\includegraphics[width=0.33\textwidth]{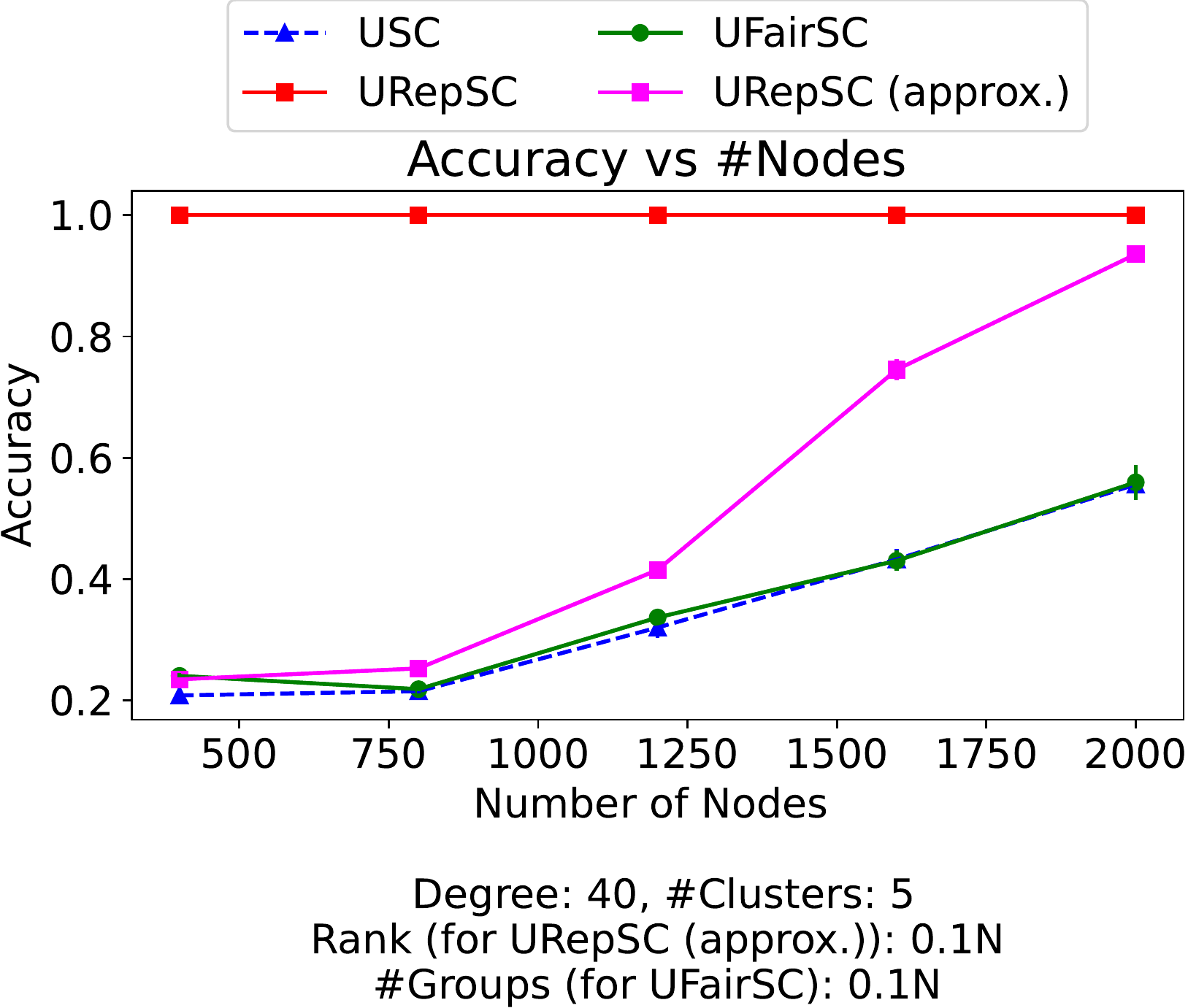}\label{fig:d_reg_unnorm:vs_N}}%
    \subfloat[][Accuracy vs no. of clusters]{\includegraphics[width=0.33\textwidth]{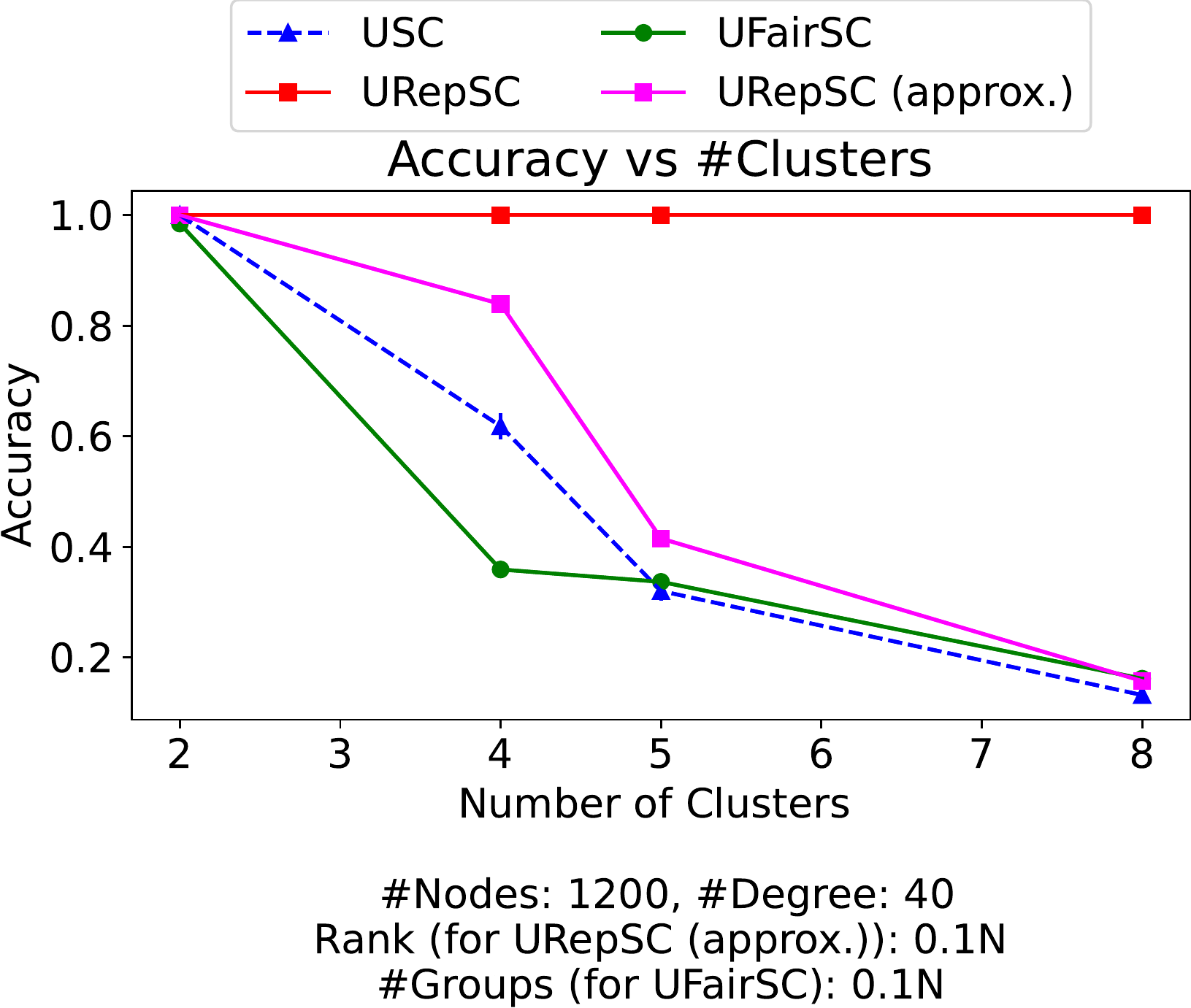}\label{fig:d_reg_unnorm:vs_K}}%
    \subfloat[][Accuracy vs degree of $\calR$]{\includegraphics[width=0.33\textwidth]{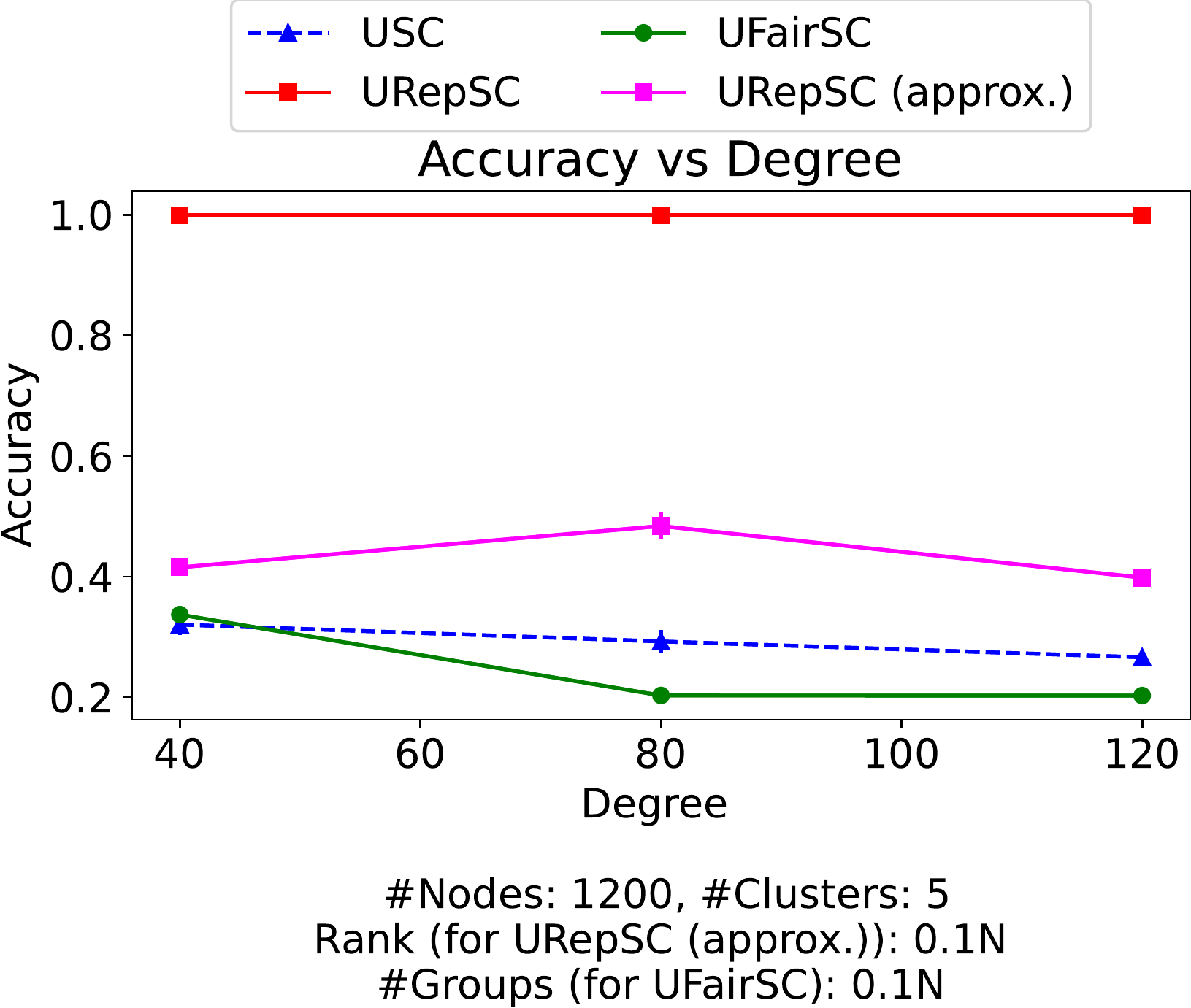}\label{fig:d_reg_unnorm:vs_d}}
    \caption{Comparing \textsc{URepSC} with other ``unnormalized'' algorithms using synthetically generated $d$-regular representation graphs.}
    \label{fig:d_reg_unnorm}
\end{figure}

\remark[Comparison with \citet{KleindessnerEtAl:2019:GuaranteesForSpectralClusteringWithFairnessConstraints}] We refer to the unnormalized variant of the algorithm in \citet{KleindessnerEtAl:2019:GuaranteesForSpectralClusteringWithFairnessConstraints} as \textsc{UFairSC}. It assumes that each node belongs to one of the $P$ \textit{protected groups} $\calP_1, \dots, \calP_P \subseteq \calV$ that are observed by the learner. \textsc{URepSC} recovers \textsc{UFairSC} as a special case when $\calR$ is block diagonal (\Cref{appendix:constraint}). To demonstrate the generality of \textsc{URepSC}, we only experiment with $\calR$'s that are not block diagonal. As \textsc{UFairSC} is not directly applicable in this setting, to compare with it, we approximate the protected groups by clustering the nodes in $\calR$ using standard spectral clustering. Each discovered cluster is then treated as a protected group (also see \Cref{appendix:a_note_on_approximate_repsc}).


\paragraph{$d$-regular representation graphs:} We sampled similarity graphs from $\calR$-PP using $p=0.4$, $q=0.3$, $r=0.2$, and $s=0.1$ for various values of $d$, $N$, and $K$, while ensuring that the underlying $\calR$ satisfies Assumption \ref{assumption:R_is_d_regular} and $\rank{\bfR} \leq N - k$. \Cref{fig:d_reg_unnorm} compares \textsc{URepSC} with unnormalized spectral clustering (\textsc{USC}) (\Cref{alg:unnormalized_spectral_clustering}) and \textsc{UFairSC}. \Cref{fig:d_reg_unnorm:vs_N} shows the effect of varying $N$ for a fixed $d = 40$ and $K=5$. \Cref{fig:d_reg_unnorm:vs_K} varies $K$ and keeps $N = 1200$ and $d = 40$ fixed. Similarly, \Cref{fig:d_reg_unnorm:vs_d} keeps $N = 1200$ and $K = 5$ fixed and varies $d$. In all cases, we use $R = P = N/10$. The figures plot the accuracy on $y$-axis and report the mean and standard deviation across $10$ independent executions. As the ground truth clusters satisfy \Cref{def:representation_constraint} by construction, a high accuracy implies that the algorithm returns representation-aware clusters. In \Cref{fig:d_reg_unnorm:vs_N}, it appears that even \textsc{USC} will return representation-aware clusters for a large enough graph. However, this is not true if the number of clusters increases with $N$ (\Cref{fig:d_reg_unnorm:vs_K}), as is common in practice.


\begin{figure}[t]
    \centering
    \subfloat[][$N = 1000$, $K = 4$]{\includegraphics[width=0.48\textwidth]{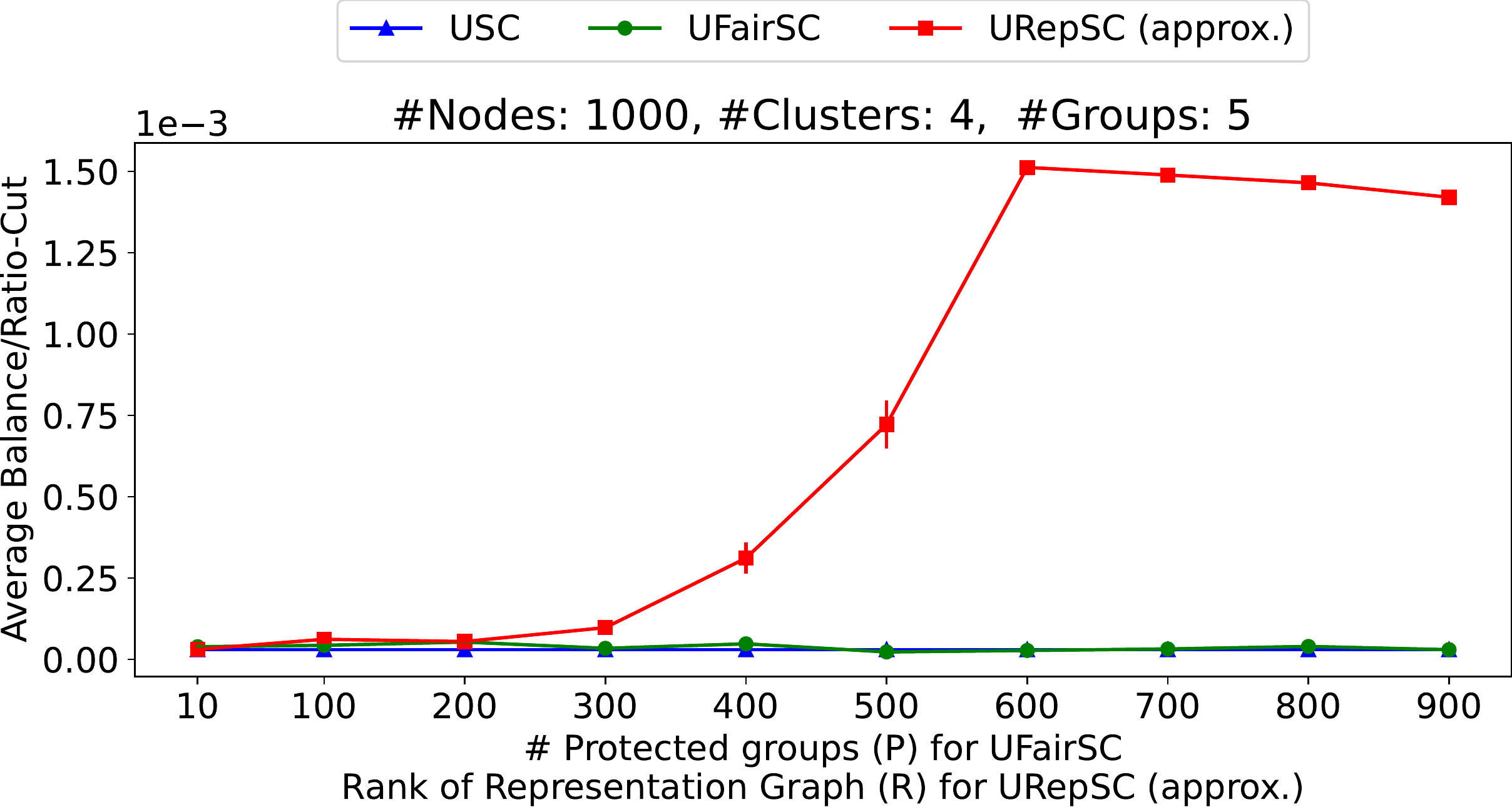}}%
    \hspace{0.5cm}\subfloat[][$N = 1000$, $K = 8$]{\includegraphics[width=0.48\textwidth]{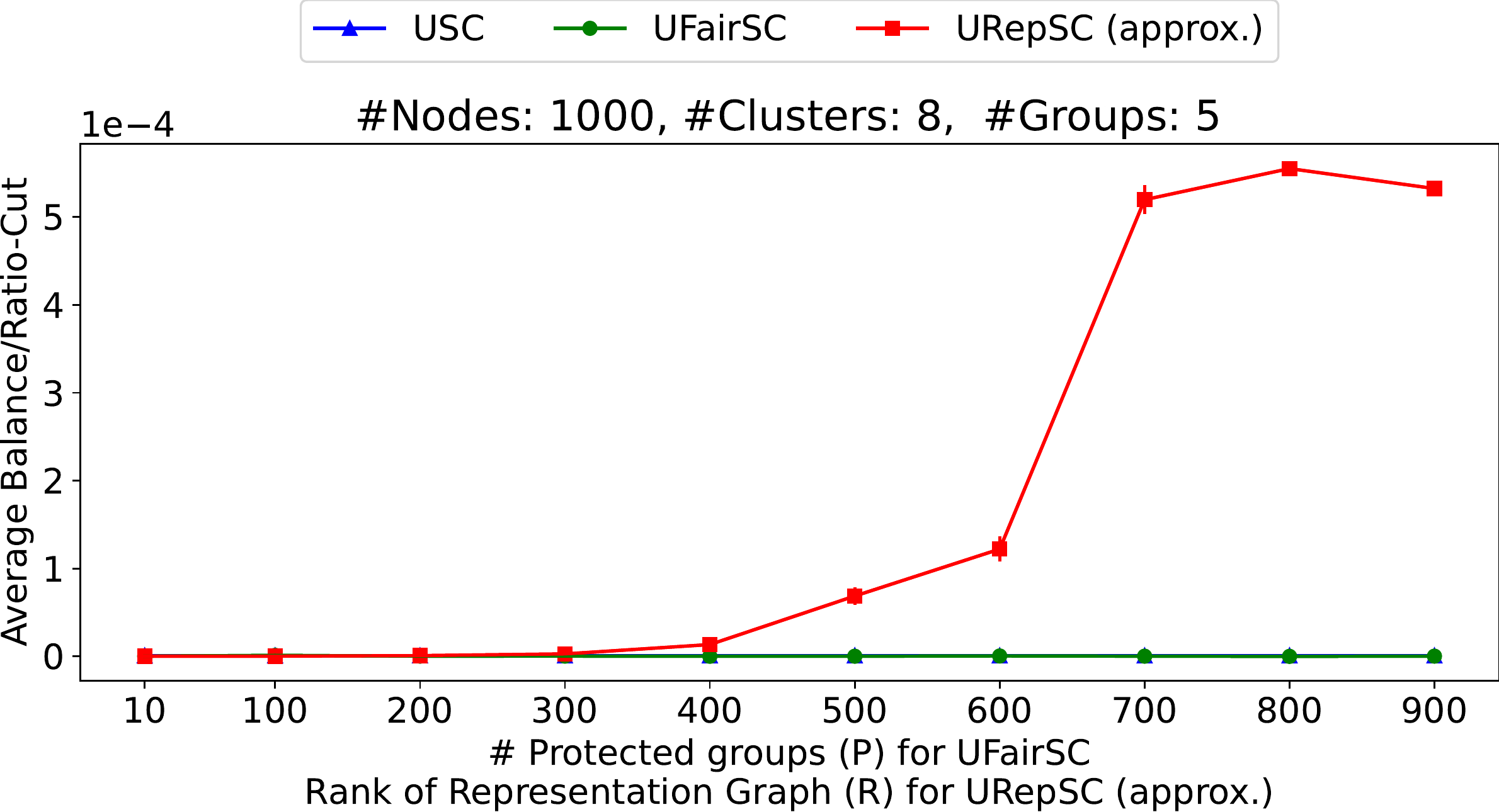}}
    \caption{Comparing \textsc{URepSC (approx.)} with \textsc{UFairSC} using synthetically generated representation graphs sampled from an SBM.}
    \label{fig:sbm_comparison_unnorm}
\end{figure}

\paragraph{Representation graphs sampled from planted partition model:} We divide the nodes into $P = 5$ protected groups and sample a representation graph $\calR$ using a (traditional) planted partition model with within (resp. across) protected group(s) connection probability given by $p_{\mathrm{in}} = 0.8$ (resp. $p_{\mathrm{out}} = 0.2$). Conditioned on $\calR$, we then sample similarity graphs from $\calR$-PP using the same parameters as before. We only experiment with \textsc{URepSC (approx.)} as an $\calR$ generated this way may violate the rank assumption. Moreover, because such an $\calR$ may not be $d$-regular, high accuracy no longer implies representation awareness, and we instead report the ratio of average individual balance $\bar{\rho} = \frac{1}{N} \sum_{i = 1}^N \rho_i$ (see \eqref{eq:balance}) to the value of $\mathrm{RCut}$ in \Cref{fig:sbm_comparison_unnorm}. Higher values indicate high quality clusters that are also balanced from the perspective of individuals. \Cref{fig:sbm_comparison_unnorm} shows a trade-off between accuracy and representation awareness. One can choose an appropriate value of $R$ in \textsc{URepSC (approx.)} to get good quality clusters with a high balance.


\begin{figure}[t]
    \centering
    \subfloat[][$K = 2$]{\includegraphics[width=0.48\textwidth]{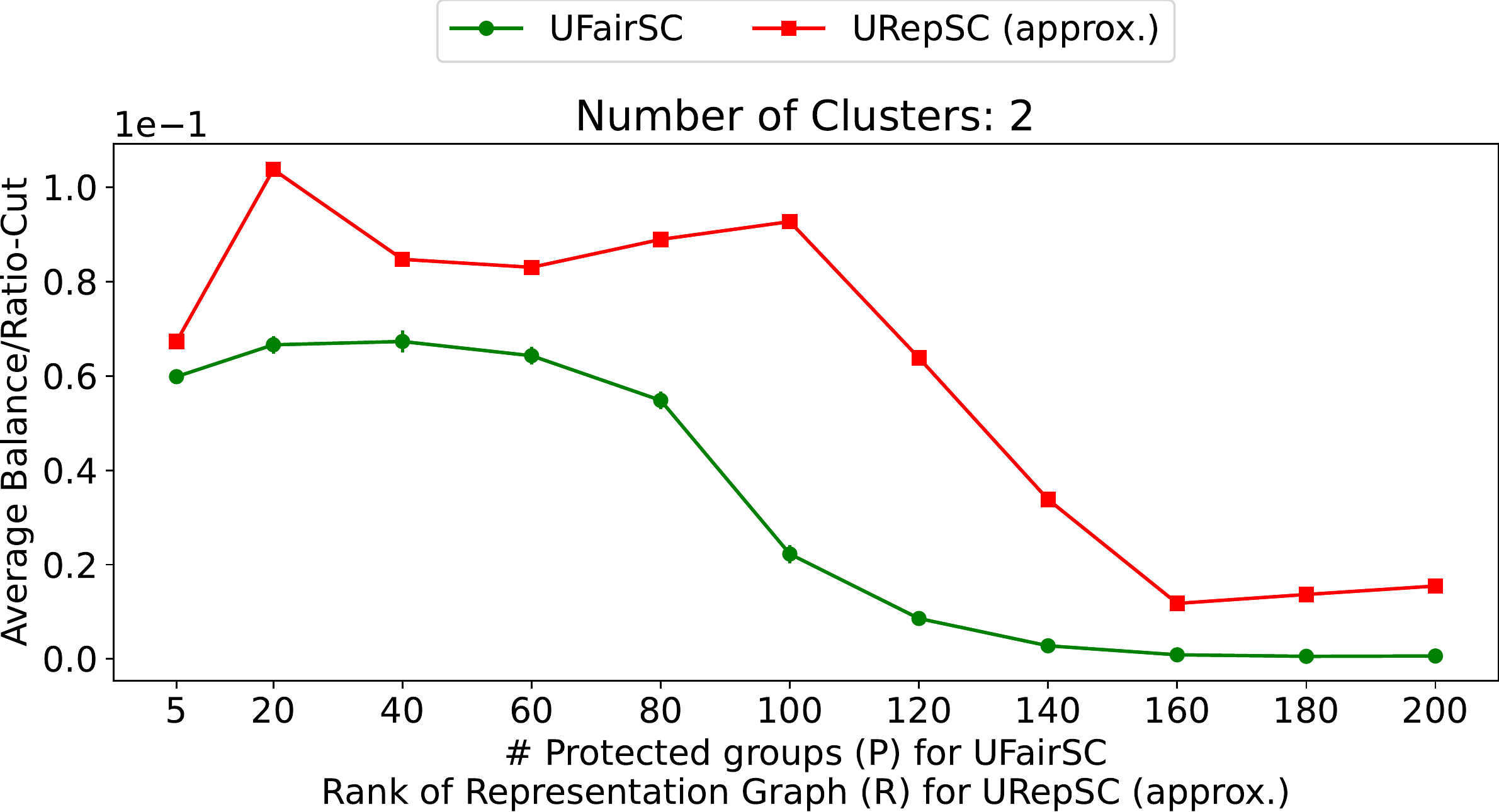}}%
    \hspace{0.5cm}\subfloat[][$K = 4$]{\includegraphics[width=0.48\textwidth]{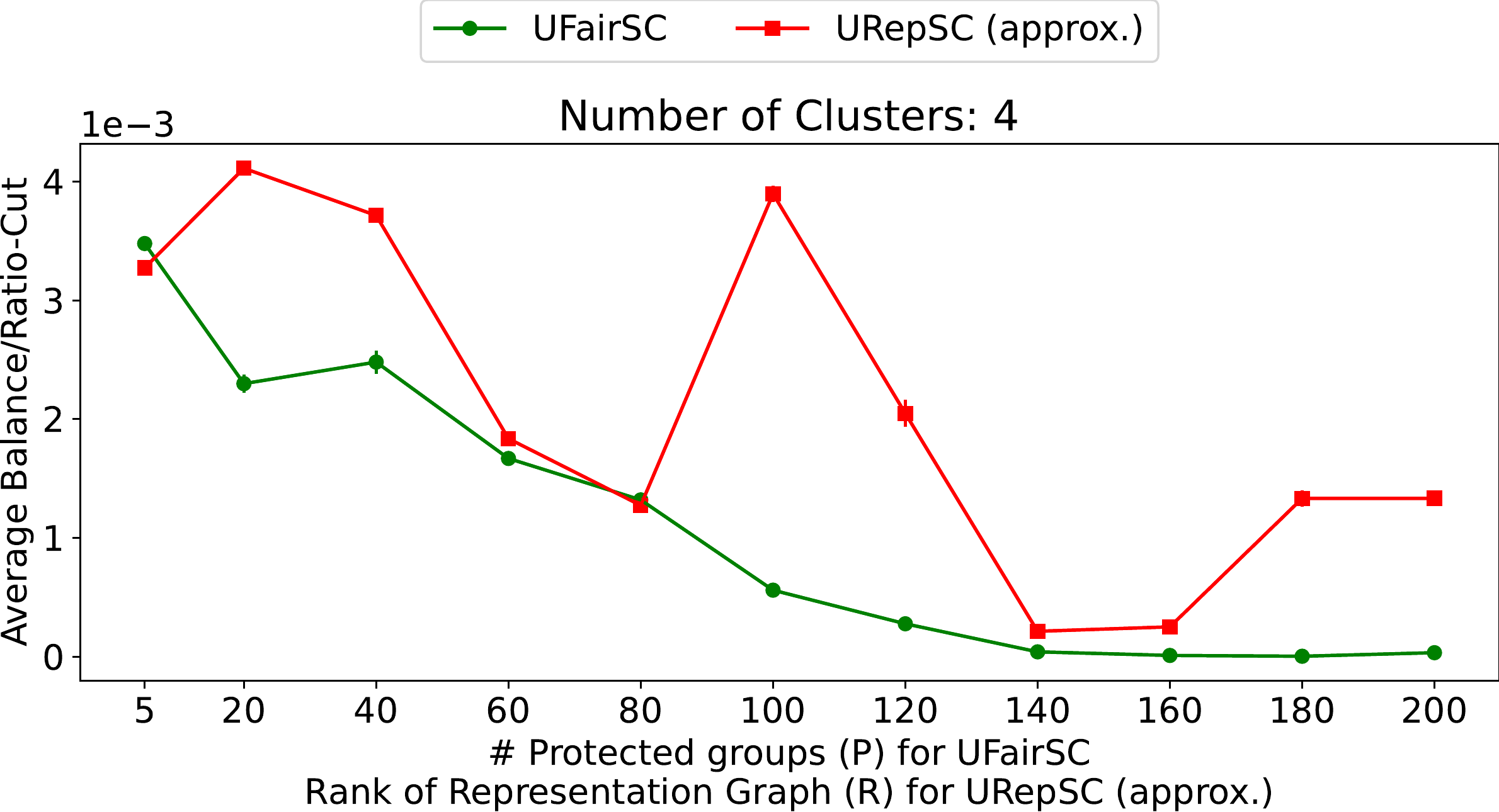}}
    \caption{Comparing \textsc{URepSC (approx.)} with \textsc{UFairSC} on FAO trade network.}
    \label{fig:real_data_comparison_unnorm}
\end{figure}

\paragraph{A real-world network:} The FAO trade network from United Nations is a multiplex network with $214$ nodes representing countries and $364$ layers representing commodities like coffee and barley \citep{DomenicoEtAl:2015:StructuralReducibilityOfMultilayerNetworks}. Edges corresponding to the volume of trade between countries. We connect each node to its five nearest neighbors in each layer and make the edges undirected. The first $182$ layers are aggregated to construct $\calR$ (connecting nodes that are connected in at least one layer) and the next $182$ layers are used for constructing $\calG$. Note that $\calR$ constructed this way is not $d$-regular. To motivate this experiment further, note that clusters based only on $\calG$ only consider the trade of commodities $183$--$364$. However, countries also have other trade relations in $\calR$, leading to shared economic interests. If the members of each cluster jointly formulate their economic policies, countries have an incentive to influence the economic policies of all clusters by having their representatives in them.

As before, we use the low-rank approximation for the representation graph in \textsc{URepSC (approx.)}. \Cref{fig:real_data_comparison_unnorm} compares \textsc{URepSC (approx.)} with \textsc{UFairSC} and has the same semantics as \Cref{fig:sbm_comparison_unnorm}. Different plots in \Cref{fig:real_data_comparison_unnorm} correspond to different choices of $K$. \textsc{URepSC (approx.)} achieves a higher ratio of average balance to ratio-cut. In practice, a user would choose $R$ by assessing the relative importance of a quality metric like ratio-cut and representation metric like average balance.

\rebuttal{\Cref{appendix:additional_experiments} contains results for \textsc{NRepSC}, experiments with another real-world network, a few additional experiments related to \textsc{URepSC}, and a numerical validation of our time-complexity analysis. \Cref{appendix:separated_plots} contains plots that show both average balance and ratio-cut instead of their ratio.}

%% file: conclusion.tex
\section{Conclusion}
\label{section:conclusion}

We studied the consistency of constrained spectral clustering under a new individual level fairness constraint, called the representation constraint, using a novel $\calR$-PP random graph model. Our work naturally generalizes existing population level constraints \citep{ChierichettiEtAl:2017:FairClusteringThroughFairlets} and associated spectral algorithms \citep{KleindessnerEtAl:2019:GuaranteesForSpectralClusteringWithFairnessConstraints}. Four important avenues for future work include \textbf{(i)} the relaxation of the $d$-regularity assumption in our analysis (needed to ensure representation awareness of ground-truth clusters), \textbf{(ii)} better theoretical understanding of \textsc{URepSC (approx.)}, \textbf{(iii)} improvement of the computational complexity of our algorithms, and \textbf{(iv)} exploring relaxed variants of our constraint and other (possibly non-spectral) algorithms for finding representation-aware clusters under such a relaxed constraint.

%% file: appendix_repr_constraint.tex
\begin{figure}[t]
    \centering
    \subfloat[][Protected groups]{\includegraphics[width=0.3\textwidth]{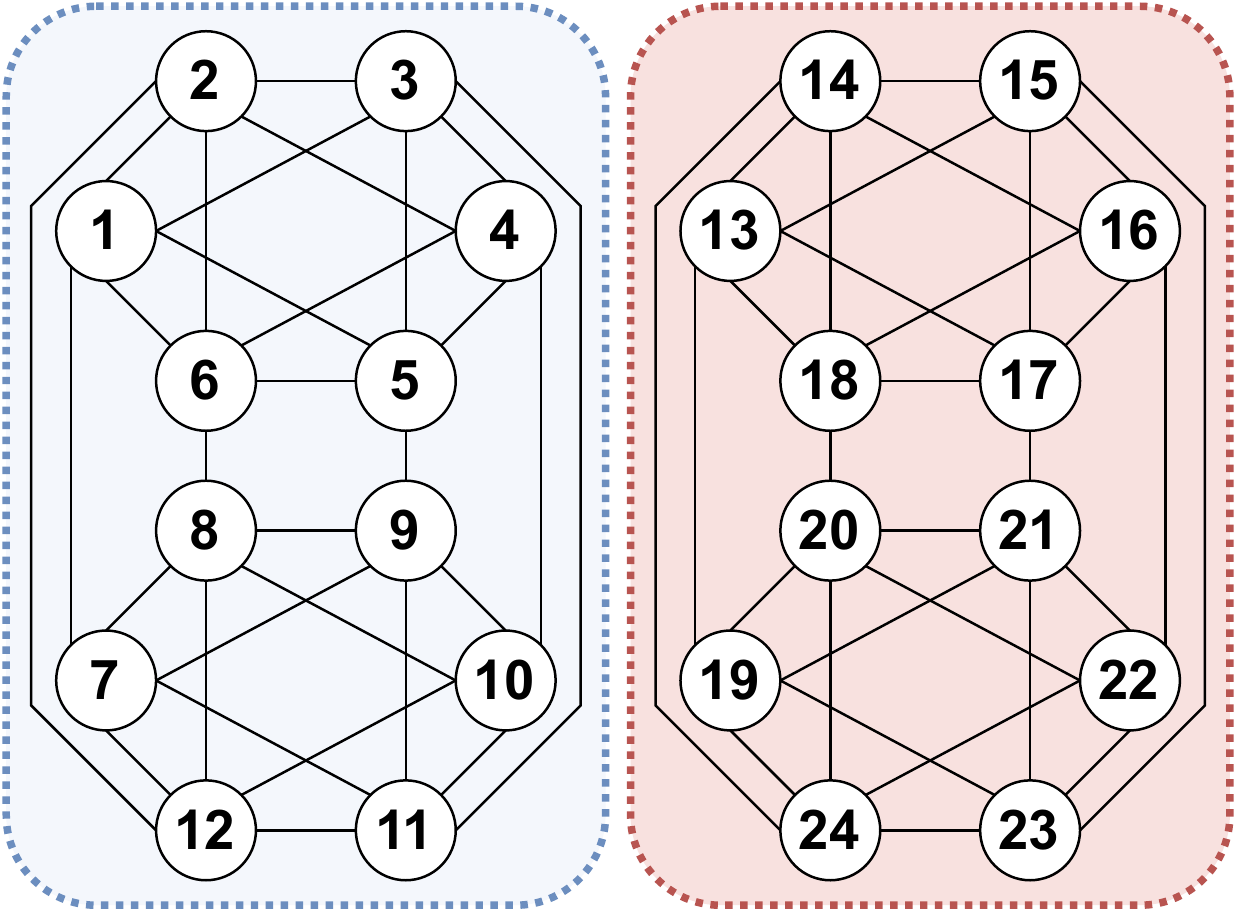}\label{fig:toy_example:protected_groups}}%
    \hspace{5mm}\subfloat[][Statistically fair clusters]{\includegraphics[width=0.3\textwidth]{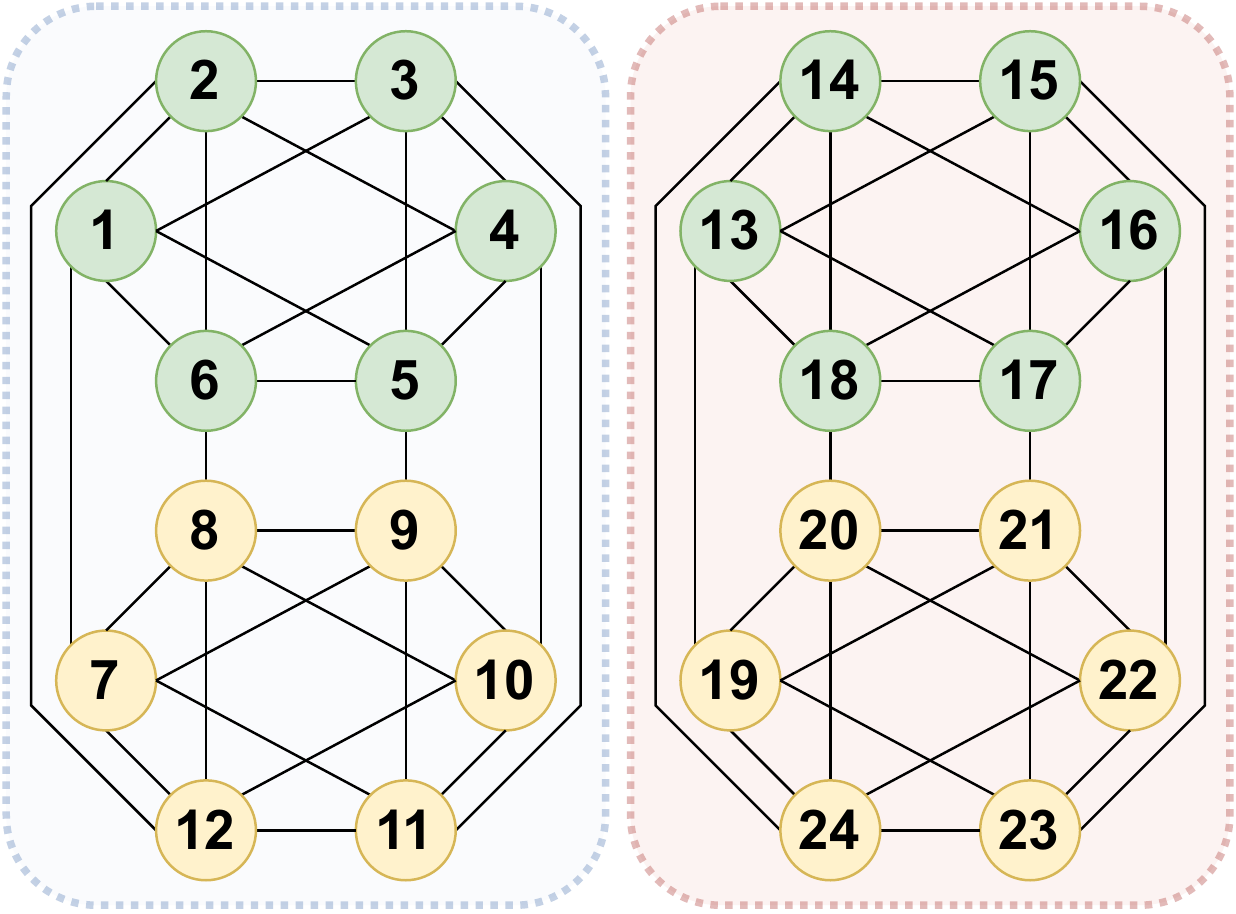}\label{fig:toy_example:fairsc}}%
    \hspace{5mm}\subfloat[][Individually fair clusters]{\includegraphics[width=0.3\textwidth]{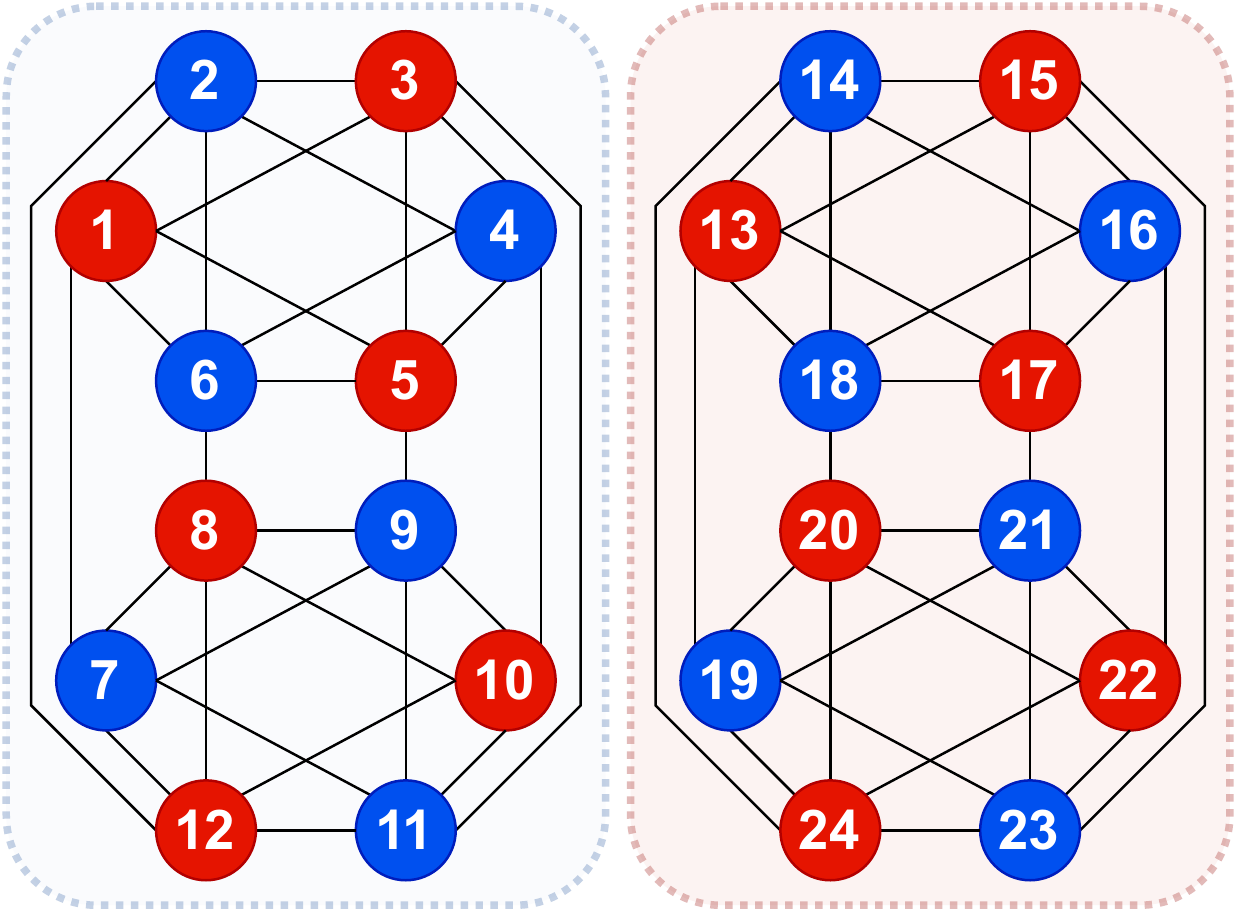}\label{fig:toy_example:repfairsc}}
    \caption{An example representation graph $\calR$. Panel (a) shows the protected groups recovered from $\calR$. Panel (b) shows the clusters recovered by a statistically fair clustering algorithm. Panel (c) shows the ideal individually fair clusters. (Best viewed in color)}
    \label{fig:fairness:toy_example}
\end{figure}

\section{Representation constraint: Additional details}
\label{appendix:constraint}

In this section, we make three additional remarks about the properties of the proposed constraint.

\paragraph{Need for individual fairness:} To understand the need for individual fairness notions, consider the representation graph $\calR$ specified in Figure \ref{fig:toy_example:protected_groups}. All the nodes have a self-loop associated with them that has not been shown for clarity. In this example, $N = 24$, $K = 2$, and every node is connected to $d=6$ nodes (including the self-loop). To use a statistical fairness notion \citep{ChierichettiEtAl:2017:FairClusteringThroughFairlets}, one would begin by clustering the nodes in $\calR$ to approximate the protected groups as the members of these protected groups will be each other's representatives to the first order of approximation. A natural choice is to have two protected groups, as shown in Figure~\ref{fig:toy_example:protected_groups} using different colors. However, clustering nodes based on these protected groups can produce the green and yellow clusters shown in Figure~\ref{fig:toy_example:fairsc}. It is easy to verify that these clusters satisfy the statistical fairness criterion as they have an equal number of members from both protected groups. However, these clusters are very "unfair" from the perspective of each individual. For example, node $v_1$ does not have enough representation in the yellow cluster as only one of its six representatives is in this cluster, despite the equal size of both the clusters. A similar argument can be made for every other node in this graph. This example highlights an extreme case where a statistically fair clustering is highly unfair from the perspective of each individual. Figure~\ref{fig:toy_example:repfairsc} shows another clustering assignment and it is easy to verify that each node in this assignment has the same representation in both red and blue clusters, making it individually fair with respect to $\calR$. Our goal is to develop algorithms that prefer the clusters in Figure~\ref{fig:toy_example:repfairsc} over the clusters in Figure~\ref{fig:toy_example:fairsc}.

\paragraph{Statistical fairness as a special case:} Recall that our constraint specifies an individual fairness notion. Contrast this with several existing approaches that assign each node to one of the $P$ \textit{protected groups} $\calP_1, \dots, \calP_P \subseteq \calV$ \citep{ChierichettiEtAl:2017:FairClusteringThroughFairlets} and require these protected groups to have a proportional representation in all clusters, i.e., 
\begin{equation*}
    \frac{\abs{\calP_i \cap \calC_j}}{\abs{\calC_j}} = \frac{\abs{\calP_i}}{N}, \,\, \forall i \in [P],\,\, j \in [K].
\end{equation*}
This is an example of \textit{statistical fairness}. In the previous paragraph, we argued that statistical fairness may not be enough in some cases. We now show that the constraint in \Cref{def:representation_constraint} is equivalent to a statistical fairness notion for an appropriately constructed representation graph $\calR$ from the given protected groups $\calP_1, \dots, \calP_P$. Namely, let $\calR$ be such that $R_{ij} = 1$ if and only if $v_i$ and $v_j$ belong to the same protected group. In this case, it is easy to verify that the constraint in \Cref{def:representation_constraint} reduces to the statistical fairness criterion given above. In general, for other configurations of the representation graph, we strictly generalize the statistical fairness notion. We also strictly generalize the approach presented in \citet{KleindessnerEtAl:2019:GuaranteesForSpectralClusteringWithFairnessConstraints}, where the authors use spectral clustering to produce statistically fair clusters. Also noteworthy is the assumption made by statistical fairness, namely that every pair of vertices in a protected group can represent each others' interests ($R_{ij} = 1 \Leftrightarrow v_i$ and $v_j$ are in the same protected group) or they are very similar with respect to some sensitive attributes. This assumption becomes unreasonable as protected groups grow in size.

\paragraph{Sensitive attributes and protected groups:} Viewed as a fairness notion, the proposed constraint only requires a representation graph $\calR$. It has two advantages over existing fairness criteria: \textbf{(i)} it does not require observable sensitive attributes (such as age, gender, and sexual orientation), and \textbf{(ii)} even if sensitive attributes are provided, one need not specify the number of protected groups or explicitly compute them. This ensures data privacy and helps against individual profiling. Our constraint only requires access to the representation graph $\calR$. This graph can either be directly elicited from the individuals or derived as a function of several sensitive attributes. In either case, once $\calR$ is available, we no longer need to expose any sensitive attributes to the clustering algorithm. For example, individuals in $\calR$ may be connected if their age difference is less than five years and if they went to the same school. Crucially, the sensitive attributes used to construct $\calR$ may be numerical, binary, categorical, etc.

%% file: appendix_normalized_sc.tex
\section{Normalized variant of the algorithm}
\label{appendix:normalized_variant_of_algorithm}

\Cref{section:normalized_spectral_clustering} presents the normalized variant of the traditional spectral clustering algorithm. \Cref{section:normalized_repsc} describes our algorithm.


\subsection{Normalized spectral clustering}
\label{section:normalized_spectral_clustering}

The ratio-cut objective divides $\mathrm{Cut}(\calC_i, \calV \backslash \calC_i)$ by the number of nodes in $\calC_i$ to balance the size of the clusters. The volume of a cluster $\calC \subseteq \calV$, defined as $\mathrm{Vol}(\calC) = \sum_{v_i \in \calC} D_{ii}$, is another popular notion of its size. The normalized cut or $\mathrm{NCut}$ objective divides $\mathrm{Cut}(\calC_i, \calV \backslash \calC_i)$ by $\mathrm{Vol}(\calC_i)$, and is defined as
\begin{equation*}
    \mathrm{NCut}(\calC_1, \dots, \calC_K) = \sum_{i = 1}^K \frac{\mathrm{Cut}(\calC_i, \calV \backslash \calC_i)}{\mathrm{Vol}(\calC_i)}.
\end{equation*}
As before, one can show that $\mathrm{NCut}(\calC_1, \dots, \calC_K) = \trace{\bfT^\intercal \bfL \bfT}$ \citep{Luxburg:2007:ATutorialOnSpectralClustering}, where $\bfT \in \bbR^{N \times K}$ is specified below.
\begin{equation}
    \label{eq:T_def}
    T_{ij} = \begin{cases}
        \frac{1}{\sqrt{\mathrm{Vol}(\calC_j)}} & \text{ if }v_i \in \calC_j \\
        0 & \text{ otherwise.}
    \end{cases}
\end{equation}
Note that $\bfT^\intercal \bfD \bfT = \bfI$. Thus, the optimization problem for minimizing the NCut objective is
\begin{equation}
    \label{eq:normalized_ideal_opt_problem}
    \min_{\bfT \in \bbR^{N \times K}}  \,\,\,\, \trace{\bfT^\intercal \bfL \bfT} \,\,\,\, \text{s.t.} \,\,\,\, \bfT^\intercal \bfD \bfT = \bfI \text{ and } \bfT \text{ is of the form \eqref{eq:T_def}.}
\end{equation}
As before, this optimization problem is hard to solve, and normalized spectral clustering solves a relaxed variant of this problem. Let $\bfH = \bfD^{1/2} \bfT$ and define the normalized graph Laplacian as $\bfL_{\mathrm{norm}} = \bfI - \bfD^{-1/2} \bfA \bfD^{-1/2}$. Normalized spectral clustering solves the following relaxed problem:
\begin{equation}
    \label{eq:opt_problem_normal_normalized}
    \min_{\bfH \in \bbR^{N \times K}}  \,\,\,\, \trace{\bfH^\intercal \bfL_{\mathrm{norm}} \bfH} \,\,\,\, \text{s.t.} \,\,\,\, \bfH^\intercal \bfH = \bfI.
\end{equation}
Note that $\bfH^\intercal \bfH = \bfI \Leftrightarrow \bfT^\intercal \bfD \bfT = \bfI$. This is again the standard form of the trace minimization problem that can be solved using the Rayleigh-Ritz theorem. Algorithm \ref{alg:normalized_spectral_clustering} summarizes the normalized spectral clustering algorithm.

\begin{algorithm}[t]
    \begin{algorithmic}[1]
        \State \textbf{Input:} Adjacency matrix $\bfA$, number of clusters $K \geq 2$
        \State Compute the normalized Laplacian matrix $\bfL_{\mathrm{norm}} = \bfI - \bfD^{-1/2}\bfA \bfD^{-1/2}$.
        \State Compute the first $K$ eigenvectors $\bfu_1, \dots, \bfu_K$ of $\bfL_{\mathrm{norm}}$. Let $\bfH^* \in \bbR^{N \times K}$ be a matrix that has $\bfu_1, \dots, \bfu_K$ as its columns.
        \State Let $\bfh^*_i$ denote the $i^{th}$ row of $\bfH^*$. Compute $\tilde{\bfh}^*_i = \frac{\bfh^*_i}{\norm{\bfh^*_i}[2]}$ for all $i = 1, 2, \dots, N$.
        \State Cluster $\tilde{\bfh}^*_1, \dots, \tilde{\bfh}^*_N$ into $K$ clusters using $k$-means clustering.
        \State \textbf{Output:} Clusters $\hat{\calC}_1, \dots, \hat{\calC}_K$, \textrm{s.t.} $\hat{\calC}_i = \{v_j \in \calV : \tilde{\bfh}^*_j \text{ was assigned to the }i^{th} \text{ cluster}\}$.
    \end{algorithmic}
    \caption{Normalized spectral clustering}
    \label{alg:normalized_spectral_clustering}
\end{algorithm}


\subsection{Normalized representation-aware spectral clustering (\textsc{NRepSC})}
\label{section:normalized_repsc}

We use a similar strategy as in \Cref{section:unnormalized_repsc} to develop the normalized variant of our algorithm. Recall from \Cref{section:normalized_spectral_clustering} that normalized spectral clustering approximately minimizes the $\mathrm{NCut}$ objective. The lemma below is a counterpart of \Cref{lemma:constraint_matrix_unnorm}. It formulates a sufficient condition that implies our constraint in \eqref{eq:representation_constraint}, but this time in terms of the matrix $\bfT$ defined in \eqref{eq:T_def}.

\begin{lemma}
    \label{lemma:constraint_matrix_norm}
    Let $\bfT \in \bbR^{N \times K}$ have the form specified in \eqref{eq:T_def}. The condition 
    \begin{equation}
      \label{eq:normalized_matrix_fairness_criteria}
      \bfR \left( \bfI - \frac{1}{N}\bmone\bmone^\intercal \right) \bfT = \mathbf{0}
    \end{equation}
    implies that the corresponding clusters $\calC_1, \dots, \calC_K$ satisfy \eqref{eq:representation_constraint}. Here, $\bfI$ is the $N \times N$ identity matrix and $\bmone$ is a $N$ dimensional all-ones vector.
\end{lemma}

For \textsc{NRepSC}, we assume that the similarity graph $\calG$ is connected so that the diagonal entries of $\bfD$ are strictly positive. We proceed as before to incorporate constraint \eqref{eq:normalized_matrix_fairness_criteria} in optimization problem \eqref{eq:normalized_ideal_opt_problem}. After applying the spectral relaxation, we get
\begin{equation}
    \label{eq:optimization_problem_normalized}
    \min_{\bfT} \;\;\;\; \trace{\bfT^\intercal  \bfL \bfT} \;\;\;\; \text{s.t.} \;\;\;\; \bfT^\intercal \bfD \bfT = \bfI; \;\;\;\; \bfR(\bfI - \bmone \bmone^\intercal / N) \bfT = \mathbf{0}.
\end{equation}
As before, $\bfT = \bfY \bfZ$ for some $\bfZ \in \bbR^{N - r \times K}$, where recall that columns of $\bfY$ contain orthonormal basis for $\nullspace{\bfR(\bfI - \bmone \bmone^\intercal / N)}$. This reparameterization yields
\begin{equation*}
    \min_{\bfZ} \;\;\;\; \trace{\bfZ^\intercal \bfY^\intercal  \bfL \bfY \bfZ} \;\;\;\; \text{s.t.} \;\;\;\; \bfZ^\intercal \bfY^\intercal \bfD \bfY \bfZ = \bfI.
\end{equation*}
Define $\bfQ \in \bbR^{N - r \times N - r}$ such that $\bfQ^2 = \bfY^\intercal \bfD \bfY$. Note that $\bfQ$ exists as the entries of $\bfD$ are non-negative. Let $\bfV = \bfQ \bfZ$. Then, $\bfZ = \bfQ^{-1} \bfV$ and $\bfZ^\intercal \bfQ^2 \bfZ = \bfV^\intercal \bfV$ as $\bfQ$ is symmetric. Reparameterizing again, we get
\begin{equation*}
    \min_{\bfV} \;\;\;\; \trace{\bfV^\intercal \bfQ^{-1} \bfY^\intercal  \bfL \bfY \bfQ^{-1} \bfV} \;\;\;\; \text{s.t.} \;\;\;\; \bfV^\intercal \bfV = \bfI.
\end{equation*}
This again is the standard form of the trace minimization problem and the optimal solution is given by the leading $K$ eigenvectors of $\bfQ^{-1} \bfY^\intercal  \bfL \bfY \bfQ^{-1}$. Algorithm \ref{alg:nrepsc} summarizes the normalized representation-aware spectral clustering algorithm, which we denote by \textsc{NRepSC}. Note that the algorithm assumes that $\bfQ$ is invertible, which requires the absence of isolated nodes in the similarity graph $\calG$. 

\begin{algorithm}[t]
    \begin{algorithmic}[1]
        \State \textbf{Input: }Adjacency matrix $\bfA$, representation graph $\bfR$, number of clusters $K \geq 2$
        \State Compute $\bfY$ containing orthonormal basis vectors of $\nullspace{\bfR(\bfI - \frac{1}{N}\bmone\bmone^\intercal)}$
        \State Compute Laplacian $\bfL = \bfD - \bfA$
        \State Compute $\bfQ = \sqrt{\bfY^\intercal \bfD \bfY}$ using the matrix square root
        \State Compute leading $K$ eigenvectors of $\bfQ^{-1} \bfY^\intercal  \bfL \bfY \bfQ^{-1}$. Set them as columns of $\bfV \in \bbR^{N-r \times K}$
        \State Apply $k$-means clustering to the rows of $\bfT = \bfY \bfQ^{-1} \bfV$ to get clusters $\hat{\calC}_1, \hat{\calC}_2, \dots, \hat{\calC}_K$
        \State \textbf{Return:} Clusters $\hat{\calC}_1, \hat{\calC}_2, \dots, \hat{\calC}_K$
    \end{algorithmic}
    \caption{\textsc{NRepSC}}
    \label{alg:nrepsc}
\end{algorithm}


%% file: appendix_approximate_repsc.tex
\section{A note on the approximate variants of our algorithms}
\label{appendix:a_note_on_approximate_repsc}

Recall the \textsc{URepSC (approx.)} algorithm from \Cref{section:unnormalized_repsc} that we use in our experiments when $\rank{\bfR} > N - K$. It first obtains a rank $R \leq N - K$ approximation of $\bfR$ and then uses the approximate $\bfR$ in \Cref{alg:unnormalized_spectral_clustering}. \textsc{NRepSC (approx.)} can be analogously defined for \textsc{NRepSC} in \Cref{alg:nrepsc}. In this section, we provide additional intuition behind this idea of using a low rank approximation of $\bfR$.

Existing (population-level) fairness notions for clustering assign a categorical value to each node and treat it as its sensitive attribute. Based on this sensitive attribute, the set of nodes $\calV$ can be partitioned into $P$ \textit{protected groups} $\calP_1, \dots, \calP_P \subseteq \calV$ such that all nodes in $\calP_i$ have the $i^{th}$ value for the sensitive attribute. Our guiding motivation behind representation graph was to connect nodes in $\calR$ based on their similarity with respect to \emph{multiple} sensitive attributes of different types (see \Cref{section:constraint}). Finding clusters in $\calR$ is then in the same spirit as clustering the original sensitive attributes and approximating their combined effect via a single categorical \textit{meta}-sensitive attribute (the cluster to which the node belongs in $\calR$). Indeed, as described in \Cref{section:numerical_results}, we do this while experimenting with \textsc{UFairSC} and \textsc{NFairSC} from \citet{KleindessnerEtAl:2019:GuaranteesForSpectralClusteringWithFairnessConstraints} in our experiments. 

\Cref{appendix:constraint} shows that a block diagonal $\bfR$ encodes protected groups $\calP_1, \dots, \calP_P$ defined above and reduces our representation constraint to the existing population level constraint \citep{ChierichettiEtAl:2017:FairClusteringThroughFairlets}, thereby recovering all existing results from \citet{KleindessnerEtAl:2019:GuaranteesForSpectralClusteringWithFairnessConstraints} as a special case of our analysis. A natural question to ask is how a low rank approximation of $\bfR$ is different from the block diagonal matrix described in \Cref{appendix:constraint} and why the approximate variants of \textsc{URepSC} and \textsc{NRepSC} are different from simply using \textsc{UFairSC} and \textsc{NFairSC} on clusters in $\calR$, as described in \Cref{section:numerical_results}.

To understand the differences, first note that a low rank approximation $\hat{\bfR}$ of $\bfR$ need not have a block diagonal structure with only ones and zeros. Entry $\hat{R}_{ij}$ approximates the strength of similarity between nodes $i$ and $j$. The constraint $\hat{\bfR}(\bfI - \bf1 \bf1^\intercal/N)$ translates to
\begin{align*}
    \sum_{j=1}^N \hat{R}_{ij} H_{jk} = \frac{1}{N} \left(\sum_{j=1}^N \hat{R}_{ij} \right) \left( \sum_{j=1}^N H_{jk} \right), \;\;\;\; \forall i \in [N], k \in [K].
\end{align*}
Let us look at a particular node $v_i$ and focus on a particular cluster $\calC_k$. If $\bfH$ has the form specified in \eqref{eq:H_def}, we get
\begin{align*}
    \frac{1}{\abs{\calC_k}} \sum_{j: v_j \in \calC_k} \hat{R}_{ij} = \frac{1}{N} \sum_{j=1}^N \hat{R}_{ij}.
\end{align*}
From the perspective of node $v_i$, this constraint requires the average similarity of $v_i$ to other nodes in $\calC_k$ to be same as the average similarity of node $v_i$ to all other nodes in the population. This must simultaneously hold for all clusters $\calC_k$ so that nodes similar to $v_i$ are present on an average in all clusters. One can see this as a continuous variant of our constraint in \Cref{def:representation_constraint}.

An important point to note is that this is still an individual level constraint and reduces to the existing population level constraint only when $\hat{\bfR}$ is binary and block diagonal (\Cref{appendix:constraint}). Thus, in general, using a low rank approximation of $\bfR$ is different from first clustering $\bfR$ and then using the resulting protected groups in \textsc{UFairSC} and \textsc{NFairSC}. Therefore, \textsc{URepSC (approx.)} and \textsc{NRepSC (approx.)} do not trivially reduce to \textsc{UFairSC} and \textsc{NFairSC} from \citet{KleindessnerEtAl:2019:GuaranteesForSpectralClusteringWithFairnessConstraints}. This is clearly visible in our experiments where the approximate variants perform better in terms of individual balance as compared to \textsc{UFairSC} and \textsc{NFairSC}.

Unfortunately, defining $\calR$-PP for an $\calR$ with continuous valued entries is not straightforward. This makes the analysis of the approximate variants more challenging. However, we believe that their practical utility outweighs the lack of theoretical guarantees and leave such an analysis for future work.

%% file: proofs.tex
\section{Proof of \Cref{theorem:consistency_result_unnormalized,theorem:consistency_result_normalized}}
\label{section:proof_of_theorems}

The proof of \Cref{theorem:consistency_result_unnormalized,theorem:consistency_result_normalized} follow the commonly used template for such results \citep{RoheEtAl:2011:SpectralClusteringAndTheHighDimensionalSBM, LeiEtAl:2015:ConsistencyOfSpectralClusteringInSBM}. In the context of \textsc{URepSC} (similar arguments work for \textsc{NRepSC} as well), we
\begin{enumerate}
    \item Compute the expected Laplacian matrix $\calL$ under $\calR$-PP and show that its top $K$ eigenvectors can be used to recover the ground-truth clusters (Lemmas \ref{lemma:introducing_uks}--\ref{lemma:orthonormal_eigenvectors_y2_yK}).
    \item Show that these top $K$ eigenvectors lie in the null space of $\bfR(\bfI - \bmone \bmone^\intercal / N)$ and hence are also the top $K$ eigenvectors of $\bfY^\intercal \calL \bfY$ (Lemma \ref{lemma:first_K_eigenvectors_of_L}). This implies that Algorithm \ref{alg:urepsc} returns the ground truth clusters in the expected case.
    \item Use matrix perturbation arguments to establish a high probability mistake bound in the general case when the graph $\calG$ is sampled from a $\calR$-PP (Lemmas \ref{lemma:bound_on_D-calD}--\ref{lemma:k_means_error}).
\end{enumerate}

We begin with a series of lemmas that highlight certain useful properties of eigenvalues and eigenvectors of the expected Laplacian $\calL$. These lemmas will be used in \Cref{section:proof_consistency_unnormalized,section:proof_consistency_normalized} to prove \Cref{theorem:consistency_result_unnormalized,theorem:consistency_result_normalized}, respectively. See \Cref{appendix:proof_of_technical_lemmas_from_algorithms} for the proofs of all technical lemmas. For the remainder of this section, we assume that all appropriate assumptions made in \Cref{theorem:consistency_result_unnormalized,theorem:consistency_result_normalized} are satisfied.

The first lemma shows that certain vectors that can be used to recover the ground-truth clusters indeed satisfy the representation constraint in \eqref{eq:matrix_fairness_criteria} and \eqref{eq:normalized_matrix_fairness_criteria}.

\begin{lemma}
\label{lemma:introducing_uks}
The $N$-dimensional vector of all ones, denoted by $\bmone$, is an eigenvector of $\bfR$ with eigenvalue $d$. Define $\bfu_k \in \bbR^N$ for $k \in [K - 1]$ as,
\begin{equation*}
    u_{ki} = \begin{cases}
    1 & \text{ if }v_i \in \calC_k \\
    -\frac{1}{K - 1} & \text{ otherwise,}
    \end{cases}
\end{equation*}
where $u_{ki}$ is the $i^{th}$ element of $\bfu_k$. Then, $\bmone, \bfu_1, \dots, \bfu_{K - 1} \in \nullspace{\bfR(\bfI - \frac{1}{N}\bmone\bmone^\intercal)}$. Moreover, $\bmone, \bfu_1, \dots, \bfu_{K - 1}$ are linearly independent.
\end{lemma}

Recall that we use $\calA \in \bbR^{N \times N}$ to denote the expected adjacency matrix of the similarity graph $\calG$. Clearly, $\calA = \tilde{\calA} - p \bfI$, where $\tilde{\calA}$ is such that $\tilde{\calA}_{ij} = P(A_{ij} = 1)$ if $i \neq j$ (see \eqref{eq:sbm_specification}) and $\tilde{\calA}_{ii} = p$ otherwise. Note that 
\begin{equation}
    \label{eq:eigenvector_A_tildeA}
    \tilde{\calA} \bfx = \lambda \bfx \,\,\,\, \Leftrightarrow \,\,\,\, \calA \bfx = (\lambda - p) \bfx.
\end{equation}
Simple algebra shows that $\tilde{\calA}$ can be written as
\begin{equation}
    \label{eq:tilde_cal_A_def}
    \tilde{\calA} = q \bfR + s (\bmone \bmone^\intercal - \bfR) + (p - q)\sum_{k = 1}^K \bfG_k \bfR \bfG_k + (r - s) \sum_{k = 1}^K \bfG_k (\bmone \bmone^\intercal - \bfR) \bfG_k,
\end{equation}
where, for all $k \in [K]$, $\bfG_k \in \bbR^{N \times N}$ is a diagonal matrix such that $(\bfG_k)_{ii} = 1$ if $v_i \in \calC_k$ and $0$ otherwise. The next lemma shows that $\bmone, \bfu_1, \dots, \bfu_{K - 1}$ defined in \Cref{lemma:introducing_uks} are eigenvectors of $\tilde{\calA}$.

\begin{lemma}
    \label{lemma:uk_eigenvector_of_tildeA}
    Let $\bmone, \bfu_1, \dots, \bfu_{K - 1}$ be as defined in \Cref{lemma:introducing_uks}. Then,
    \begin{eqnarray*}
        \tilde{\calA} \bmone &=& \lambda_1 \bmone \text{ where } \lambda_1 = qd + s(N - d) + (p - q) \frac{d}{K} + (r - s) \frac{N - d}{K}, \text{ and } \\
        \tilde{\calA} \bfu_k &=& \lambda_{1 + k} \bfu_k \text{ where } \lambda_{1 + k} = (p - q) \frac{d}{K} + (r - s) \frac{N - d}{K}.
    \end{eqnarray*}
\end{lemma}

Let $\calL = \calD - \calA$ be the expected Laplacian matrix, where $\calD$ is a diagonal matrix with $\calD_{ii} = \sum_{j=1}^N \calA_{ij}$ for all $i \in [N]$. It is easy to see that $\calD_{ii} = \lambda_1 - p$ for all $i \in [N]$ as $\calA \bmone = (\lambda_1 - p) \bmone$ by \eqref{eq:eigenvector_A_tildeA} and \Cref{lemma:uk_eigenvector_of_tildeA}. Thus, $\calD = (\lambda_1 - p) \bfI$ and hence any eigenvector of $\tilde{\calA}$ with eigenvalue $\lambda$ is also an eigenvector of $\calL$ with eigenvalue $\lambda_1 - \lambda$. That is, if $\tilde{\calA} \bfx = \lambda \bfx$,
\begin{equation}
    \label{eq:eigenvectors_of_L}
    \calL \bfx = (\calD - \calA)\bfx = ((\lambda_1 - p) \bfI - (\tilde{\calA} - p \bfI)) \bfx = (\lambda_1 - \lambda) \bfx.
\end{equation}
Hence, the eigenvectors of $\calL$ corresponding to the $K$ smallest eigenvalues are the same as the eigenvectors of $\tilde{\calA}$ corresponding to the $K$ largest eigenvalues.

Recall that the columns of the matrix $\bfY$ used in \Cref{alg:urepsc,alg:nrepsc} contain the orthonormal basis for the null space of $\bfR(\bfI - \bmone \bmone^\intercal/N)$. To solve \eqref{eq:optimization_problem} and \eqref{eq:optimization_problem_normalized}, we only need to optimize over vectors that belong to this null space. By \Cref{lemma:introducing_uks}, $\bmone, \bfu_1, \dots, \bfu_{K - 1} \in \nullspace{\bfR(\bfI - \bmone \bmone^\intercal/N)}$ and these vectors are linearly independent. However, we need an orthonormal basis to compute $\bfY$. Let $\bfy_1 = \bmone / \sqrt{N}$ and $\bfy_2, \dots, \bfy_K$ be orthonormal vectors that span the same space as $\bfu_1, \dots, \bfu_{K - 1}$. The next lemma computes such $\bfy_2, \dots, \bfy_K$. The matrix $\bfY \in \bbR^{N \times N - r}$ contains these vectors $\bfy_1, \dots, \bfy_K$ as its first $K$ columns.

\begin{lemma}
\label{lemma:orthonormal_eigenvectors_y2_yK}
Define $\bfy_{1 + k} \in \bbR^N$ for $k \in [K - 1]$ as
\begin{equation*}
    y_{1 + k, i} = \begin{cases}
    0 & \text{ if } v_i \in \calC_{k{'}} \text{ s.t. } k{'} < k \\
    \frac{K - k}{\sqrt{\frac{N}{K}(K - k)(K - k + 1)}} & \text{ if } v_i \in \calC_k \\
    -\frac{1}{\sqrt{\frac{N}{K}(K - k)(K - k + 1)}} & \text{ otherwise.}
    \end{cases}
\end{equation*}
Then, for all $k \in [K - 1]$, $\bfy_{1 + k}$ are orthonormal vectors that span the same space as $\bfu_1, \bfu_2, \dots, \bfu_{K - 1}$ and $\bfy_1^\intercal \bfy_{1 + k} = 0$. As before, $y_{1 + k, i}$ refers to the $i^{th}$ element of $\bfy_{1 + k}$.
\end{lemma}

Let $\bfX \in \bbR^{N \times K}$ be such that it has $\bfy_1, \dots, \bfy_{K}$ as its columns. If two nodes belong to the same cluster, the rows corresponding to these nodes in $\bfX \bfU$ will be identical for any $\bfU \in \bbR^{K \times K}$ such that $\bfU^\intercal \bfU = \bfU \bfU^\intercal = \bfI$. Thus, any $K$ orthonormal vectors belonging to the span of $\bfy_1, \dots, \bfy_K$ can be used to recover the ground truth clusters.  With the general properties of the eigenvectors and eigenvalues established in the lemmas above, we next move on to the proof of \Cref{theorem:consistency_result_unnormalized} in the next section and \Cref{theorem:consistency_result_normalized} in \Cref{section:proof_consistency_normalized}.


\subsection{Proof of \Cref{theorem:consistency_result_unnormalized}}
\label{section:proof_consistency_unnormalized}

Let $\calZ \in \bbR^{N - r \times K}$ be a solution to the optimization problem \eqref{eq:optimization_problem} in the expected case with $\calA$ as input. The next lemma shows that columns of $\bfY \calZ$ indeed lie in the span of $\bfy_1, \dots, \bfy_K$. Thus, the $k$-means clustering step in \Cref{alg:urepsc} will return the correct ground truth clusters when $\calA$ is passed as input.

\begin{lemma}
    \label{lemma:first_K_eigenvectors_of_L}
    Let $\bfy_1 = \bmone / \sqrt{N}$ and $\bfy_{1 + k}$ be as defined in \Cref{lemma:orthonormal_eigenvectors_y2_yK} for all $k \in [K - 1]$. Further, let $\calZ$ be the optimal solution of the optimization problem in \eqref{eq:optimization_problem} with $\bfL$ set to $\calL$. Then, the columns of $\bfY \calZ$ lie in the span of $\bfy_1, \bfy_2, \dots, \bfy_K$.
\end{lemma}

Next, we use arguments from matrix perturbation theory to show a high-probability bound on the number of mistakes made by the algorithm. In particular, we need an upper bound on $\norm{\bfY^\intercal \bfL \bfY - \bfY^\intercal \calL \bfY}$, where $\bfL$ is the Laplacian matrix for a graph randomly sampled from $\calR$-PP and $\norm{\bfP} = \sqrt{\lambdamax{\bfP^\intercal \bfP}}$ for any matrix $\bfP$. Note that  $\norm{\bfY} = \norm{\bfY^\intercal} = 1$ as $\bfY^\intercal \bfY = \bfI$. Thus, 
\begin{equation}
    \label{eq:reducing_YLY_to_L}
    \norm{\bfY^\intercal \bfL \bfY - \bfY^\intercal \calL \bfY} \leq \norm{\bfY^\intercal} \,\, \norm{\bfL - \calL} \,\, \norm{\bfY} = \norm{\bfL - \calL}.
\end{equation}
Moreover, $$\norm{\bfL - \calL} = \norm{\bfD - \bfA - (\calD - \calA)} \leq \norm{\bfD - \calD} + \norm{\bfA - \calA}.$$ The next two lemmas bound the two terms on the right hand side of the inequality above, thus providing an upper bound on $\norm{\bfL - \calL}$ and hence on $\norm{\bfY^\intercal \bfL \bfY - \bfY^\intercal \calL \bfY}$ by \eqref{eq:reducing_YLY_to_L}.

\begin{lemma}
    \label{lemma:bound_on_D-calD}
    Assume that $p \geq C \frac{\ln N}{N}$ for some constant $C > 0$. Then, for every $\alpha > 0$, there exists a constant $\const_1(C, \alpha)$ that only depends on $C$ and $\alpha$ such that $$\norm{\bfD - \calD} \leq \const_1(C, \alpha) \sqrt{p N \ln N}$$ with probability at-least $1 - N^{-\alpha}$.
\end{lemma}

\begin{lemma}
    \label{lemma:bound_on_A-calA}
    Assume that $p \geq C \frac{\ln N}{N}$ for some constant $C > 0$. Then, for every $\alpha > 0$, there exists a constant $\const_4(C, \alpha)$ that only depends on $C$ and $\alpha$ such that $$\norm{\bfA - \calA} \leq \const_4(C, \alpha) \sqrt{p N}$$ with probability at-least $1 - N^{-\alpha}$.
\end{lemma}

From \Cref{lemma:bound_on_D-calD,lemma:bound_on_A-calA}, we conclude that there is always a constant $\const_5(C, \alpha) = \max\{\const_1(C, \alpha), \const_4(C, \alpha)\}$ such that for any $\alpha > 0$, with probability at least $1 - 2N^{-\alpha}$, 
\begin{equation}
    \label{eq:L-calL_bound}
    \norm{\bfY^\intercal \bfL \bfY - \bfY^\intercal \calL \bfY} \leq \norm{\bfL - \calL} \leq \const_5(C, \alpha) \sqrt{p N \ln N}. 
\end{equation}

Let $\calZ$ and $\bfZ$ denote the optimal solution of \eqref{eq:optimization_problem} in the expected ($\bfL$ replaced with $\calL$) and observed case.
We use \eqref{eq:L-calL_bound} to show a bound on $\norm{\bfY \calZ - \bfY Z}[F]$ in \Cref{lemma:bound_on_eigenvector_diff} and then use this bound to argue that \Cref{alg:urepsc} makes a small number of mistakes when the graph is sampled from $\calR$-PP.

\begin{lemma}
\label{lemma:bound_on_eigenvector_diff}
    Let $\mu_1 \leq \mu_2 \leq \dots \leq \mu_{N - r}$ be eigenvalues of $\bfY^\intercal \calL \bfY$. Further, let the columns of $\calZ \in \bbR^{N - r \times K}$ and $\bfZ \in \bbR^{N - r \times K}$ correspond to the leading $K$ eigenvectors of $\bfY^\intercal \calL \bfY$ and $\bfY^\intercal \bfL \bfY$, respectively. Define $\gamma = \mu_{K + 1} - \mu_{K}$. Then, with probability at least $1 - 2N^{-\alpha}$, $$\inf_{\bfU \in \bbR^{K \times K} : \bfU\bfU^\intercal = \bfU^\intercal \bfU = \bfI} \norm{\bfY \calZ - \bfY \bfZ \bfU}[F] \leq \const_5(C, \alpha) \frac{4\sqrt{2K}}{\gamma} \sqrt{p N \ln N},$$ where $\const_5(C, \alpha)$ is from \eqref{eq:L-calL_bound}.
\end{lemma}

Recall that $\bfX \in \bbR^{N \times K}$ is a matrix that contains $\bfy_1, \dots, \bfy_K$ as its columns. Let $\bfx_i$ denote the $i^{th}$ row of $\bfX$. Simple calculation using \Cref{lemma:orthonormal_eigenvectors_y2_yK} shows that,
\begin{equation*}
    \norm{\bfx_i - \bfx_j}[2] = \begin{cases}
        0 & \text{ if }v_i \text{ and }v_j \text{ belong to the same cluster} \\
        \sqrt{\frac{2K}{N}} & \text{ otherwise.}
\end{cases}
\end{equation*}
By \Cref{lemma:first_K_eigenvectors_of_L}, $\calZ$ can be chosen such that $\bfY \calZ = \bfX$. Let $\bfU$ be the matrix that solves $\inf_{\bfU \in \bbR^{K \times K} : \bfU\bfU^\intercal = \bfU^\intercal \bfU = \bfI} \norm{\bfY \calZ - \bfY \bfZ \bfU}[F]$. As $\bfU$ is orthogonal, $\norm{\bfx_i^\intercal \bfU - \bfx_j^\intercal \bfU}[2] = \norm{\bfx_i - \bfx_j}[2]$. The following lemma is a direct consequence of Lemma 5.3 in \citet{LeiEtAl:2015:ConsistencyOfSpectralClusteringInSBM}.

\begin{lemma}
    \label{lemma:k_means_error}
    Let $\bfX$ and $\bfU$ be as defined above. For any $\epsilon > 0$, let $\hat{\bmTheta} \in \bbR^{N \times K}$ be the assignment matrix returned by a $(1 + \epsilon)$-approximate solution to the $k$-means clustering problem when rows of $\bfY \bfZ$ are provided as input features. Further, let $\hat{\bmmu}_1$, $\hat{\bmmu}_2$, \dots, $\hat{\bmmu}_K \in \bbR^{K}$ be the estimated cluster centroids. Define $\hat{\bfX} = \hat{\bmTheta} \hat{\bmmu}$ where $\hat{\bmmu} \in \bbR^{K \times K}$ contains $\hat{\bmmu}_1, \dots, \hat{\bmmu}_K$ as its rows. Further, define $\delta = \sqrt{\frac{2K}{N}}$, and $S_k = \{v_i \in \calC_k : \norm{\hat{\bfx}_i - \bfx_i} \geq \delta/2\}$. Then,
    \begin{equation}
        \label{eq:num_mistakes_bound}
        \delta^2 \sum_{k = 1}^K \abs{S_k} \leq 8(2 + \epsilon) \norm{\bfX \bfU^\intercal - \bfY \bfZ}[F][2].
    \end{equation}
    Moreover, if $\gamma$ from \Cref{lemma:bound_on_eigenvector_diff} satisfies $\gamma^2 > \const(C, \alpha) (2 + \epsilon) p NK \ln N$ for a universal constant $\const(C, \alpha)$, there exists a permutation matrix  $\bfJ \in \bbR^{K \times K}$ such that 
    \begin{equation}
        \label{eq:correct_solution_on_non-mistakes}
        \hat{\bmtheta}_i^\intercal \bfJ = \bmtheta_i^\intercal, \,\,\,\, \forall \,\, i \in [N] \backslash (\cup_{k=1}^K S_k).    
    \end{equation}
    Here, $\hat{\bmtheta}_i \bfJ$ and $\bmtheta_i$ represent the $i^{th}$ row of matrix $\hat{\bmTheta}\bfJ$ and $\bmTheta$ respectively.
\end{lemma}
    
By the definition of $M(\bmTheta, \hat{\bmTheta})$, for the matrix $\bfJ$ used in \Cref{lemma:k_means_error}, $M(\bmTheta, \hat{\bmTheta}) \leq \frac{1}{N} \norm{\bmTheta - \hat{\bmTheta} \bfJ}[0]$. But, according to \Cref{lemma:k_means_error},
$\norm{\bmTheta - \hat{\bmTheta} \bfJ}[0] \leq 2 \sum_{k = 1}^K \abs{S_k}$. Using \Cref{lemma:bound_on_eigenvector_diff,lemma:k_means_error}, we get
\begin{eqnarray*}
    M(\bmTheta, \hat{\bmTheta}) \leq \frac{1}{N} \norm{\bmTheta - \hat{\bmTheta} \bfJ}[0] \leq \frac{2}{N} \sum_{k = 1}^K \abs{S_k}
    &\leq& \frac{16(2 + \epsilon)}{N \delta^2} \norm{\bfX \bfU^\intercal - \bfY \bfZ}[F][2] \\
    &\leq& \const_5(C, \alpha)^2 \frac{512(2 + \epsilon)}{N \delta^2 \gamma^2} p N K \ln N.
\end{eqnarray*}
Noting that $\delta = \sqrt{\frac{2K}{N}}$ and setting $\const(C, \alpha) = 256 \times \const_5(C, \alpha)^2$ finishes the proof.


\subsection{Proof of \Cref{theorem:consistency_result_normalized}}
\label{section:proof_consistency_normalized}

Recall that $\bfQ = \sqrt{\bfY^\intercal \bfD \bfY}$ and analogously define $\calQ = \sqrt{\bfY^\intercal \calD \bfY}$, where $\calD$ is the expected degree matrix. It was shown after Lemma \ref{lemma:uk_eigenvector_of_tildeA} that $\calD = (\lambda_1 - p) \bfI$. Thus, $\calQ = \sqrt{\lambda_1 - p} \;\bfI$ as $\bfY^\intercal \bfY = \bfI$. Hence $\calQ^{-1} \bfY^\intercal \calL \bfY \calQ^{-1} = \frac{1}{\lambda_1 - p} \bfY^\intercal \calL \bfY$. Therefore, $\calQ^{-1} \bfY^\intercal \calL \bfY \calQ^{-1} \bfx = \frac{\lambda}{\lambda_1 - p} \bfx \; \Longleftrightarrow \;  \bfY^\intercal \calL \bfY \bfx = \lambda \bfx$. Let $\calZ \in \bbR^{N - r \times K}$ contain the leading $K$ eigenvectors of $\calQ^{-1} \bfY^\intercal \calL \bfY \calQ^{-1}$ as its columns. Algorithm \ref{alg:nrepsc} will cluster the rows of $\bfY \calQ^{-1} \calZ$ to recover the clusters in the expected case. As $\calQ^{-1} =\frac{1}{\sqrt{\lambda_1 - p}} \bfI$, we have $\bfY \calQ^{-1} \calZ = \frac{1}{\sqrt{\lambda_1 - p}} \bfY \calZ$. By Lemma \ref{lemma:first_K_eigenvectors_of_L}, $\calZ$ can always be chosen such that $\bfY \calZ = \bfX$, where recall that $\bfX \in \bbR^{N \times K}$ has $\bfy_1, \dots, \bfy_K$ as its columns. Because the rows of $\bfX$ are identical for nodes that belong to the same cluster, Algorithm \ref{alg:nrepsc} returns the correct ground truth clusters in the expected case. 

To bound the number of mistakes made by Algorithm \ref{alg:nrepsc}, we show that $\bfY \bfQ^{-1} \bfZ$ is close to $\bfY \calQ^{-1} \calZ$. Here, $\bfZ \in \bbR^{N - r \times K}$ contains the top $K$ eigenvectors of $\bfQ^{-1} \bfY^\intercal \bfL \bfY \bfQ^{-1}$. As in the proof of Lemma \ref{lemma:bound_on_eigenvector_diff}, we use Davis-Kahan theorem to bound this difference. This requires us to compute $\norm{\calQ^{-1} \bfY^\intercal \calL \bfY \calQ^{-1} - \bfQ^{-1} \bfY^\intercal \bfL \bfY \bfQ^{-1}}$. Note that:
\begin{align*}
    \norm{\calQ^{-1} \bfY^\intercal \calL \bfY \calQ^{-1} - \bfQ^{-1} \bfY^\intercal \bfL \bfY \bfQ^{-1}} = &\norm{\calQ^{-1} - \bfQ^{-1}} \cdot \norm{\bfY^\intercal \calL \bfY} \cdot \norm{\calQ^{-1}} + \\
    &\norm{\bfQ^{-1}} \cdot \norm{\bfY^\intercal \calL \bfY - \bfY^\intercal \bfL \bfY} \cdot \norm{\calQ^{-1}} + \\
    &\norm{\bfQ^{-1}} \cdot \norm{\bfY^\intercal \bfL \bfY} \cdot \norm{\calQ^{-1} - \bfQ^{-1}}.
\end{align*}
We already have a bound on $\norm{\bfY^\intercal \calL \bfY - \bfY^\intercal \bfL \bfY}$ in \eqref{eq:L-calL_bound}. Also, note that $\norm{\calQ^{-1}} = \frac{1}{\sqrt{\lambda_1 - p}}$ as $\calQ^{-1} = \frac{1}{\sqrt{\lambda_1 - p}} \bfI$. Similarly, as $\bfY^\intercal \bfY = \bfI$, $\norm{\bfY^\intercal \calL \bfY} \leq \norm{\calL} = \lambda_1 - \bar{\lambda}$, where $\bar{\lambda} = \lambdamin{\tilde{\calA}}$. Finally,
\begin{align*}
    \norm{\bfQ^{-1}} &\leq \norm{\calQ^{-1} - \bfQ^{-1}} + \norm{\calQ^{-1}} = \norm{\calQ^{-1} - \bfQ^{-1}} + \frac{1}{\sqrt{\lambda_1 - p}} \text{, and} \\
    \norm{\bfY^\intercal \bfL \bfY} &\leq \norm{\bfY^\intercal \calL \bfY - \bfY^\intercal \bfL \bfY} + \norm{\bfY^\intercal \calL \bfY} = \norm{\bfY^\intercal \calL \bfY - \bfY^\intercal \bfL \bfY} + \lambda_1 - \bar{\lambda}.
\end{align*}
Thus, to compute a bound on $\norm{\calQ^{-1} \bfY^\intercal \calL \bfY \calQ^{-1} - \bfQ^{-1} \bfY^\intercal \bfL \bfY \bfQ^{-1}}$, we only need a bound on $\norm{\calQ^{-1} - \bfQ^{-1}}$. The next lemma provides this bound.

\begin{lemma}
    \label{lemma:calQ-Q_bound}
    Let $\calQ = \sqrt{\bfY^\intercal \calD \bfY}$, $\bfQ = \sqrt{\bfY^\intercal \bfD \bfY}$, and assume that
    $$\left(\frac{\sqrt{pN \ln N}}{\lambda_1 - p}\right)\left(\frac{\sqrt{pN \ln N}}{\lambda_1 - p} + \frac{1}{6\sqrt{C}} \right) \leq \frac{1}{16(\alpha + 1)},$$
    where $C$ and $\alpha$ are used in $\const_1(C, \alpha)$ defined in Lemma \ref{lemma:bound_on_D-calD}. Then,
    $$\norm{\calQ^{-1} - \bfQ^{-1}} \leq \sqrt{\frac{2}{(\lambda_1 - p)^3}} \norm{\bfD - \calD}.$$
\end{lemma}

\noindent
Using the lemma above with \eqref{eq:L-calL_bound}, we get
{
\small
\begin{align}
    \label{eq:calQYcalLYcalQ-QYLYQ_bound}
    \begin{aligned}
    \norm{\calQ^{-1} \bfY^\intercal \calL \bfY \calQ^{-1} - \bfQ^{-1} \bfY^\intercal \bfL &\bfY \bfQ^{-1}} \leq \frac{2(\lambda_1 - \bar{\lambda})}{(\lambda_1 - p)^2} \left[ \sqrt{2} + \frac{\norm{\bfD - \calD}}{\lambda_1 - p}\right] \norm{\bfD - \calD} + \\
    & \frac{\const_5(C, \alpha)}{\lambda_1 - p} \left[\frac{2\sqrt{2} \norm{\bfD - \calD}}{\lambda_1 - p} + \frac{2 \norm{\bfD - \calD}[][2]}{(\lambda_1 - p)^2} + 1  \right] \sqrt{p N \ln N}.
    \end{aligned}
\end{align}
}

\noindent
The next lemma uses the bound above to show that $\bfY \bfQ^{-1} \bfZ$ is close to $\bfY \calQ^{-1} \calZ$.

\begin{lemma}
    \label{lemma:eigenvector_diff_bound_normalized}
    Let $\mu_1 \leq \mu_2 \leq \dots \leq \mu_{N - r}$ be eigenvalues of $\calQ^{-1} \bfY^\intercal \calL \bfY \calQ^{-1}$. Further, let the columns of $\calZ \in \bbR^{N - r \times K}$ and $\bfZ \in \bbR^{N - r \times K}$ correspond to the leading $K$ eigenvectors of $\calQ^{-1} \bfY^\intercal \calL \bfY \calQ^{-1}$ and $\bfQ^{-1} \bfY^\intercal \bfL \bfY \bfQ^{-1}$, respectively. Define $\gamma = \mu_{K + 1} - \mu_{K}$ and let there be a constant $\const_2(C, \alpha)$ such that 
    $\frac{\sqrt{p N \ln N}}{\lambda_1 - p} \leq \const_2(C, \alpha)$. Then, with probability at least $1 - 2N^{-\alpha}$, there exists a constant $\const_3(C, \alpha)$ such that
    \begin{align*}
        \inf_{\bfU : \bfU^\intercal \bfU = \bfU\bfU^\intercal = \bfI} \norm{\bfY \calQ^{-1} \calZ - &\bfY \bfQ^{-1} \bfZ \bfU}[F] \leq \\
        &\left[\frac{16 K \const_3(C, \alpha)}{\gamma (\lambda_1 - p)^{3/2}} + \frac{2 \const_1(C, \alpha) \sqrt{K}}{(\lambda_1 - p)^{3/2}} \right] \sqrt{p N \ln N},
    \end{align*}
    where $\const_1(C, \alpha)$ is defined in Lemma \ref{lemma:bound_on_D-calD}.
\end{lemma}

Recall that, by Lemma \ref{lemma:first_K_eigenvectors_of_L}, $\calZ$ can always be chosen such that $\bfY \calZ = \bfX$, where $\bfX$ contains $\bfy_1, \dots, \bfy_K$ as its columns. As $\calQ^{-1} = \frac{1}{\sqrt{\lambda_1 - p}} \bfI$, one can show that:
\begin{equation*}
    \norm{(\calQ^{-1} \bfX)_i - (\calQ^{-1} \bfX)_j}[2] = \begin{cases}
        0 & \text{ if } v_i \text{ and } v_j \text{ belong to the same cluster} \\
        \sqrt{\frac{2K}{N(\lambda_1 - p)}} & \text{ otherwise.}
    \end{cases}
\end{equation*}
Here, $(\calQ^{-1} \bfX)_i$ denotes the $i^{th}$ row of the matrix $\bfY \calQ^{-1} \calZ$. Let $\bfU$ be the matrix that solves $\inf_{\bfU \in \bbR^{K \times K} : \bfU\bfU^\intercal = \bfU^\intercal \bfU = \bfI} \norm{\bfY \calQ^{-1} \calZ - \bfY \bfQ^{-1} \bfZ \bfU}[F]$. As $\bfU$ is orthogonal, $\norm{(\calQ^{-1} \bfX)_i^\intercal \bfU - (\calQ^{-1} \bfX)_j^\intercal \bfU}[2] = \norm{(\calQ^{-1} \bfX)_i - (\calQ^{-1} \bfX)_j}[2]$. As in the previous case, the following lemma is a direct consequence of Lemma 5.3 in \citep{LeiEtAl:2015:ConsistencyOfSpectralClusteringInSBM}.

\begin{lemma}
    \label{lemma:k_means_error_normalized}
    Let $\bfX$ and $\bfU$ be as defined above. For any $\epsilon > 0$, let $\hat{\bmTheta} \in \bbR^{N \times K}$ be the assignment matrix returned by a $(1 + \epsilon)$-approximate solution to the $k$-means clustering problem when rows of $\bfY \bfQ^{-1} \bfZ$ are provided as input features. Further, let $\hat{\bmmu}_1$, $\hat{\bmmu}_2$, \dots, $\hat{\bmmu}_K \in \bbR^{K}$ be the estimated cluster centroids. Define $\hat{\bfX} = \hat{\bmTheta} \hat{\bmmu}$ where $\hat{\bmmu} \in \bbR^{K \times K}$ contains $\hat{\bmmu}_1, \dots, \hat{\bmmu}_K$ as its rows. Further, define $\delta = \sqrt{\frac{2K}{N(\lambda_1 - p)}}$, and $S_k = \{v_i \in \calC_k : \norm{\hat{\bfx}_i - \bfx_i} \geq \delta/2\}$. Then,
    \begin{equation*}
        \delta^2 \sum_{k = 1}^K \abs{S_k} \leq 8(2 + \epsilon) \norm{\bfX \bfU^\intercal - \bfY \bfQ^{-1} \bfZ}[F][2].
    \end{equation*}
    Moreover, if $\gamma$ from Lemma \ref{lemma:eigenvector_diff_bound_normalized} satisfies $$16(2 + \epsilon)\left[ \frac{8 \const_3(C, \alpha) \sqrt{K}}{\gamma} + \const_1(C, \alpha)\right]^2 \frac{p N^2 \ln N}{(\lambda_1 - p)^2} < \frac{N}{K},$$ then, there exists a permutation matrix  $\bfJ \in \bbR^{K \times K}$ such that 
    \begin{equation*}
        \hat{\bmtheta}_i^\intercal \bfJ = \bmtheta_i^\intercal, \,\,\,\, \forall \,\, i \in [N] \backslash (\cup_{k=1}^K S_k).    
    \end{equation*}
    Here, $\hat{\bmtheta}_i \bfJ$ and $\bmtheta_i$ represent the $i^{th}$ row of matrix $\hat{\bmTheta}\bfJ$ and $\bmTheta$ respectively.
\end{lemma}

The proof of Lemma \ref{lemma:k_means_error_normalized} is similar to that of Lemma \ref{lemma:k_means_error}, and has been omitted. The result follows by using a similar calculation as was done after Lemma \ref{lemma:k_means_error} in Section \ref{section:proof_consistency_unnormalized}.


\section{Proof of technical lemmas}
\label{appendix:proof_of_technical_lemmas_from_algorithms}


\subsection{Proof of \Cref{lemma:constraint_matrix_unnorm}}
Fix an arbitrary node $v_i \in \calV$ and $k \in [K]$. Because $\bfR(\bfI - \bmone \bmone^\intercal / N) \bfH = \mathbf{0}$,
\begin{align*}
    &\sum_{j=1}^N R_{ij} H_{jk} = \frac{1}{N} \left(\sum_{j=1}^N R_{ij} \right) \left(\sum_{j=1}^N H_{jk} \right) \\
    \Rightarrow& \,\, \frac{1}{\sqrt{\abs{\calC_k}}} \abs{\{v_j \in \calV : R_{ij} = 1 \land v_j \in \calC_k\}} = \frac{1}{N} \abs{\{v_j \in \calV : R_{ij} = 1\}} \frac{\abs{\calC_k}}{\sqrt{\abs{\calC_k}}} \\
    \Rightarrow&  \,\,\frac{\abs{\{v_j \in \calV : R_{ij} = 1 \land v_j \in \calC_k\}}}{\abs{\calC_k}} = \frac{\abs{\{v_j \in \calV : R_{ij} = 1\}}}{N}. 
\end{align*}
Because this holds for an arbitrary $v_i \in \calV$ and $k \in [K]$, $\bfR(\bfI - \bmone \bmone^\intercal / N) \bfH = \mathbf{0}$ implies the constraint in \Cref{def:representation_constraint}.


\subsection{Proof of \Cref{lemma:constraint_matrix_norm}}
Fix an arbitrary node $v_i \in \calV$ and $k \in [K]$. Because $\bfR(\bfI - \bmone \bmone^\intercal / N) \bfT = 0$,
\begin{align*}
  &\sum_{j=1}^N R_{ij} T_{jk} = \frac{1}{N}  \left(\sum_{j=1}^N R_{ij} \right) \left(\sum_{j=1}^N T_{jk} \right) \\
  \Rightarrow& \,\, \frac{1}{\sqrt{\mathrm{Vol}(\calC_k)}} \abs{\{v_j \in \calV : R_{ij} = 1 \land v_j \in \calC_k\}} = \frac{1}{N} \abs{\{v_j \in \calV : R_{ij} = 1\}} \frac{\abs{\calC_k}}{\sqrt{\mathrm{Vol}(\calC_k)}} \\
  \Rightarrow&  \,\,\frac{\abs{\{v_j \in \calV : R_{ij} = 1 \land v_j \in \calC_k\}}}{\abs{\calC_k}} = \frac{\abs{\{v_j \in \calV : R_{ij} = 1\}}}{N}. 
\end{align*}
Here, recall that $\mathrm{Vol}(\calC_k) = \sum_{v_i \in \calC_k} D_{ii}$ is the volume of the cluster $\calC_k$, which is used in \eqref{eq:T_def}. Because this holds for an arbitrary $v_i \in \calV$ and $k \in [K]$, $\bfR(\bfI - \bmone \bmone^\intercal / N) \bfT = \mathbf{0}$ implies the constraint in \Cref{def:representation_constraint}.


\subsection{Proof of Lemma \ref{lemma:introducing_uks}}

Because $\calR$ is a $d$-regular graph, it is easy to see that $\bfR \bmone = d \bmone$. Recall from \Cref{section:consistency_results} that $\bfI - \frac{1}{N}\bmone \bmone^\intercal$ is a projection matrix that removes the component of any vector $\bfx \in \bbR^N$ along the all ones vector $\bmone$. Thus, $(\bfI - \frac{1}{N}\bmone \bmone^\intercal) \bmone = 0$ and hence $\bmone \in \nullspace{\bfR(\bfI - \frac{1}{N}\bmone \bmone^\intercal)}$. Moreover, as all clusters have the same size, $$\bmone^\intercal \bfu_k = \sum_{i = 1}^N u_{ki} = \sum_{i \,:\, v_i \in \calC_k} u_{ki} + \sum_{i \,:\, v_i \notin \calC_k} u_{ki} = \frac{N}{K} - \frac{1}{K - 1} \left(N - \frac{N}{K}\right) = 0.$$

Thus, $\bfR(\bfI - \frac{1}{N} \bmone \bmone^\intercal) \bfu_k = \bfR \bfu_k$. Let us compute the $i^{th}$ element of the vector $\bfR\bfu_k$ for an arbitrary $i \in [N]$. $$(\bfR\bfu_k)_i = \sum_{j = 1}^N R_{ij} u_{kj} = \sum_{\substack{j \,:\, R_{ij} = 1 \\ \& v_j \in \calC_k}} 1 - \sum_{\substack{j \,:\, R_{ij} = 1 \\ \& v_j \notin \calC_k}} \frac{1}{K - 1} = \frac{d}{K} - \frac{1}{K - 1}\left(d - \frac{d}{K}\right) = 0.$$
Here, the second last equality follows from \Cref{assumption:R_is_d_regular} and the assumption that all clusters have the same size. Thus, $\bfR\bfu_k = 0$ and hence $\bfu_k \in \nullspace{\bfR(\bfI - \frac{1}{N} \bmone \bmone^\intercal)}$. 

Because $\bmone^\intercal \bfu_k = 0$ for all $k \in [K - 1]$, to show that $\bmone, \bfu_1, \dots, \bfu_{K - 1}$ are linearly independent, it is enough to show that $\bfu_1, \dots, \bfu_{K - 1}$ are linearly independent. Consider the $i^{th}$ component of $\sum_{k = 1}^{K - 1} \alpha_k \bfu_k$ for arbitrary $\alpha_1 \dots, \alpha_{K - 1} \in \bbR$ and $i \in [N]$. If $v_i \in \calC_K$, then
\begin{equation*}
  \left(\sum_{k = 1}^{K - 1} \alpha_k \bfu_k \right)_i = -\frac{1}{K - 1} \sum_{k = 1}^{K - 1} \alpha_k.
\end{equation*}
Similarly, when $v_i \in \calC_{k{'}}$ for some $k{'} \in [K - 1]$, we have,
\begin{equation*}
  \left(\sum_{k = 1}^{K - 1} \alpha_k \bfu_k \right)_i = \alpha_{k{'}} -\frac{1}{K - 1} \sum_{\substack{k = 1 \\ k \neq k{'}}}^{K - 1} \alpha_{k}.
\end{equation*}
Thus, $\sum_{k = 1}^{K - 1} \alpha_k \bfu_k = 0$ implies that $-\frac{1}{K - 1} \sum_{k = 1}^{K - 1} \alpha_k = 0$ and $\alpha_{k{'}} - \frac{1}{K - 1} \sum_{k = 1, k \neq k{'}}^{K - 1} \alpha_{k} = 0$ for all $k{'} \in [K - 1]$. Subtracting the first equation from the second gives $\alpha_{k{'}} + \frac{1}{K - 1} \alpha_{k{'}} = 0$, which in turn implies that $\alpha_{k{'}} = 0$ for all $k{'} \in [K - 1]$. Thus, $\bmone, \bfu_1, \dots, \bfu_{K - 1}$ are linearly independent.


\subsection{Proof of Lemma \ref{lemma:uk_eigenvector_of_tildeA}}

Using the representation of $\tilde{\calA}$ from \eqref{eq:tilde_cal_A_def}, Lemma \ref{lemma:introducing_uks}, and the assumption on equal size of the clusters, we get,
\begin{align*}
  \tilde{\calA} \bmone &= q \bfR \bmone + s (\bmone \bmone^\intercal - \bfR) \bmone + (p - q)\sum_{k = 1}^K \bfG_k \bfR \bfG_k \bmone + (r - s) \sum_{k = 1}^K \bfG_k (\bmone \bmone^\intercal - \bfR) \bfG_k \bmone \\
  &= qd \bmone + sN \bmone - sd \bmone + (r - s)\sum_{k = 1}^K \bfG_k \bmone \bmone^\intercal \bfG_k \bmone + [(p - q) - (r - s)] \sum_{k = 1}^K \bfG_k \bfR \bfG_k \bmone \\
  &= \left[qd + s(N - d) + (p - q) \frac{d}{K} + (r - s)\frac{N - d}{K} \right] \bmone.
\end{align*}
Similarly, for any $k{'} \in [K]$,
\small
{
\begin{align*}
  \allowdisplaybreaks
  \tilde{\calA} \bfu_{k{'}} &= q \bfR \bfu_{k{'}} + s (\bmone \bmone^\intercal - \bfR) \bfu_{k{'}} + (p - q)\sum_{k = 1}^K \bfG_k \bfR \bfG_k \bfu_{k{'}} + (r - s) \sum_{k = 1}^K \bfG_k (\bmone \bmone^\intercal - \bfR) \bfG_k \bfu_{k{'}} \\
  &= 0 + 0 + (r - s) \sum_{k = 1}^K \bfG_k \bmone \bmone^\intercal \bfG_k \bfu_{k{'}} + [(p - q) - (r - s)] \sum_{k = 1}^K \bfG_k \bfR \bfG_k \bfu_{k{'}} \\
  &= \left[ (p - q) \frac{d}{K} + (r - s) \frac{N - d}{K} \right] \bfu_{k{'}}.
\end{align*}
}


\subsection{Proof of Lemma \ref{lemma:orthonormal_eigenvectors_y2_yK}}

It is easy to verify that vectors $\bfy_2, \dots, \bfy_K$ are obtained by applying the Gram-Schmidt normalization process to the vectors $\bfu_1, \dots, \bfu_{K - 1}$. Thus, $\bfy_2, \dots, \bfy_K$ span the same space as $\bfu_1, \dots, \bfu_{K - 1}$. Recall that $\bfy_1 = \bmone / \sqrt{N}$. We start by showing that $\bfy_1^\intercal \bfy_{1 + k} = 0$.
\begin{align*}
  \bfy_1^\intercal \bfy_{1 + k} = \frac{1}{\sqrt{N}} \sum_{i = 1}^N y_{(1+k)i} &= \frac{1}{\sqrt{N}} \left[ \sum_{i : v_i \in \calC_k} (K - k) q_k - \sum_{i : v_i \in \calC_{k{'}}, k{'} > k} q_k \right] \\
  &= \frac{1}{\sqrt{N}} \left[ \frac{N}{K} (K - k) q_k - \left(N - k\frac{N}{K} \right)q_k \right] = 0.
\end{align*}
Here, $q_k = \frac{1}{\sqrt{\frac{N}{K} (K - k) (K - k + 1)}}$. Now consider $\bfy_{1 + k_1}^\intercal \bfy_{1 + k_2}$ for $k_1, k_2 \in [K - 1]$ such that $k_1 \neq k_2$. Assume without loss of generality that $k_1 < k_2$.
\begin{align*}
  \bfy_{1 + k_1}^\intercal \bfy_{1 + k_2} &= \sum_{i : v_i \in \calC_{k_2}} (-q_{k_1})(K - k_2)q_{k_2} + \sum_{i : v_i \in \calC_k, k > k_2} (-q_{k_1})(-q_{k_2}) \\
  &= -q_{k_1} q_{k_2} (K - k_2) \frac{N}{K} + q_{k_1} q_{k_2} \left(N - k_2 \frac{N}{K}\right) = 0.
\end{align*}
Finally, for any $k \in [K - 1]$,
\begin{align*}
  \bfy_{1 + k}^\intercal \bfy_{1 + k} = \sum_{i : v_i \in \calC_k} (K - k)^2 q_k^2 + \sum_{i : v_i \in \calC_{k{'}}, k{'} > k} q_k^2 = q_k^2 \left[ \frac{N}{K} (K - k)^2 + N - k\frac{N}{K} \right] = 1,
\end{align*}
where the last equality follows from the definition of $q_k$.


\subsection{Proof of Lemma \ref{lemma:first_K_eigenvectors_of_L}}

Note that the columns of $\calZ$ are also the dominant $K$ eigenvectors of $\bfY^\intercal \tilde{\calA} \bfY$, as $\calZ$ is the solution to \eqref{eq:optimization_problem} with $\bfL$ set to $\calL$. The calculations below show that for all $k \in [K]$, $\bfe_k \in \bbR^{N - r}$, the $k^{th}$ standard basis vector, is an eigenvector of $\bfY^\intercal \tilde{\calA} \bfY$ with eigenvalue $\lambda_k$, where $\lambda_1, \dots, \lambda_K$ are defined in Lemma \ref{lemma:uk_eigenvector_of_tildeA}.
\begin{equation*}
  \bfY^\intercal \tilde{\calA} \bfY \bfe_k = \bfY^\intercal \tilde{\calA} \bfy_k = \lambda_k \bfY^\intercal \bfy_k = \lambda_k \bfe_k.
\end{equation*}
The second equality follows from Lemma \ref{lemma:uk_eigenvector_of_tildeA} because $\bfy_2, \dots, \bfy_K \in \spanof{\bfu_1, \dots, \bfu_{K - 1}}$, and $\bfu_1, \dots, \bfu_{K - 1}$ are all eigenvectors of $\tilde{\calA}$ with the same eigenvalue. To show that the columns of $\bfY \calZ$ lie in the span of $\bfy_1, \dots, \bfy_K$, it is enough to show that $\bfe_1, \dots, \bfe_K$ are the dominant $K$ eigenvectors of $\bfY^\intercal \tilde{\calA} \bfY$.

Let $\bmalpha \in \bbR^{N - r}$ be an eigenvector of $\bfY^\intercal \tilde{\calA} \bfY$ such that $\bmalpha \notin \spanof{\bfe_1, \dots, \bfe_K}$ and $\norm{\bmalpha}[2][2] = 1$. Then, because $\bfY^\intercal \tilde{\calA} \bfY$ is symmetric, $\bmalpha^\intercal \bfy_1 = 0$, i.e. $\alpha_1 = 0$, where $\alpha_i$ denotes the $i^{th}$ element of $\bmalpha$. The eigenvalue corresponding to $\bmalpha$ is given by
\begin{equation*}
  \lambda_\alpha = \bmalpha^\intercal \bfY^\intercal \tilde{\calA} \bfY \bmalpha.
\end{equation*}
Let $\bfx = \bfY \bmalpha = \sum_{i = 1}^{N - r} \alpha_i \bfy_i$, then $\lambda_{\alpha} = \bfx^\intercal \tilde{\calA} \bfx$. Using the definition of $\tilde{\calA}$ from \eqref{eq:tilde_cal_A_def}, we get,
{
\small
\begin{equation}
  \label{eq:x_calA_x}
  \bfx^\intercal \tilde{\calA} \bfx = (q - s)\bfx^\intercal \bfR \bfx + s \bfx^\intercal \bmone \bmone^\intercal \bfx + [(p - q) - (r - s)]\sum_{k = 1}^K \bfx^\intercal \bfG_k \bfR \bfG_k \bfx + (r - s) \sum_{k = 1}^K \bfx^\intercal \bfG_k \bmone \bmone^\intercal \bfG_k \bfx.
\end{equation}
}%
We will consider each term in \eqref{eq:x_calA_x} separately. Before that, note that $\bfy_2, \dots, \bfy_{N - r} \in \nullspace{\bfR}$. This is because $\bfy_1 = \bmone / \sqrt{N}$ and $\bfy_2, \dots, \bfy_{N - r}$ are orthogonal to $\bfy_1$. Thus,
\begin{equation}
  \label{eq:null_space_F}
  \bfR (\bfI - \bmone \bmone^\intercal / N) \bfy_i = 0 \Rightarrow \bfR(\bfI - \bfy_1 \bfy_1^\intercal) \bfy_i = 0 \Rightarrow \bfR \bfy_i = 0, \,\,\,\,\, i = 2, 3, \dots, N - r.
\end{equation}
Now consider the first term in \eqref{eq:x_calA_x}.
\begin{equation*}
  \bfx^\intercal \bfR \bfx = \sum_{i = 1}^{N - r} \sum_{j = 1}^{N - r} \alpha_i \alpha_j \bfy_i^\intercal \bfR \bfy_j = \alpha_1^2 \bfy_1^\intercal \bfR \bfy_1 = 0.
\end{equation*}
Here, the second equality follows from \eqref{eq:null_space_F}, and the third equality follows as $\alpha_1 = 0$. Similarly, for the second term in \eqref{eq:x_calA_x},
\begin{equation*}
  \bfx^\intercal \bmone \bmone^\intercal \bfx = N \bfx^\intercal \bfy_1 \bfy_1^\intercal \bfx = N \sum_{i = 1}^{N - r} \sum_{j = 1}^{N - r} \alpha_i \alpha_j \bfy_i^\intercal \bfy_1 \bfy_1^\intercal \bfy_j = N \alpha_1^2 (\bfy_1^\intercal \bfy_1)^2 = 0.
\end{equation*}
Note that $\bfG_k = \bfG_k \bfG_k$ as $\bfG_k$ is a diagonal matrix with either $0$ or $1$ on its diagonal. For the third term in \eqref{eq:x_calA_x},
\begin{equation}
  \label{eq:x_Fk_x}
  \bfx^\intercal \bfG_k \bfR \bfG_k \bfx = \bfx^\intercal \bfG_k \bfG_k \bfR \bfG_k \bfG_k \bfx = \bfx_{[k]} \bfR_{[k]} \bfx_{[k]} \leq \frac{d}{K} \norm{\bfx_{[k]}}[2][2],
\end{equation}
where $\bfx_{[k]} \in \bbR^{N/K}$ contains elements of $\bfx$ corresponding to vertices in $\calC_k$. Similarly, $\bfR_{[k]} \in \bbR^{N/K \times N/K}$ contains the submatrix of $\bfR$ restricted to rows and columns corresponding to vertices in $\calC_k$. The last inequality holds because $\bfR_{[k]}$ is a $d/K$-regular graph by Assumption \ref{assumption:R_is_d_regular}, hence its maximum eigenvalue is $d/K$. Further, 
\begin{equation*}
  \sum_{k = 1}^K \bfx^\intercal \bfG_k \bfR \bfG_k \bfx \leq \frac{d}{K} \sum_{k = 1}^K \norm{\bfx_{[k]}}[2][2] = \frac{d}{K} \norm{\bfx}[2][2] = \frac{d}{K}.
\end{equation*}
Similarly, for the fourth term in \eqref{eq:x_calA_x},
\begin{equation*}
  \bfx^\intercal \bfG_k \bmone \bmone^\intercal \bfG_k \bfx = \bfx^\intercal \bfG_k \bfG_k \bmone \bmone^\intercal \bfG_k \bfG_k \bfx = \bfx_{[k]}^\intercal \bmone_{N/K} \bmone_{N/K}^\intercal \bfx_{[k]} \leq \frac{N}{K} \norm{\bfx_{[k]}}[2][2].
\end{equation*}
Here, $\bmone_{N/K} \in \bbR^{N/K}$ is an all-ones vector and the last inequality holds because $\bmone_{N/K} \bmone_{N/K}^\intercal$ is a $N/K$-regular graph. Because $\bfx_{[k]} \notin \spanof{\bfy_1, \dots, \bfy_K}$, there is at least one $k \in [K]$ for which $\bfx_{[k]}$ is not a constant vector (if this was not true, $\bfx_{[k]}$ will belong to span of $\bfy_1, \dots, \bfy_K$). Thus, at least for one $k \in [K]$, $\bfx^\intercal \bfG_k \bmone \bmone^\intercal \bfG_k \bfx < \frac{N}{K} \norm{\bfx_{[k]}}[2][2]$. Summing over $k \in [K]$, we get,
\begin{equation*}
  \sum_{k = 1}^K \bfx^\intercal \bfG_k \bmone \bmone^\intercal \bfG_k \bfx <  \frac{N}{K} \sum_{k = 1}^K \norm{\bfx_{[k]}}[2][2] = \frac{N}{K} \norm{\bfx}[2][2] = \frac{N}{K}.
\end{equation*}
Adding the four terms we get the following bound. For eigenvector $\bmalpha$ of $\bfY^\intercal \tilde{\calA} \bfY$ such that $\bmalpha \notin \spanof{\bfe_1, \dots, \bfe_K}$ and $\norm{\bmalpha}[2][2] = 1$,
\begin{equation}
  \label{eq:x_calA_x_bound}
  \lambda_\alpha = \bfx^\intercal \tilde{\calA} \bfx < [(p - q) - (r - s)] \frac{d}{K} + (r - s) \frac{N}{K} = \lambda_K.
\end{equation}
Thus, $\lambda_1, \dots, \lambda_K$ are the highest $K$ eigenvalues of $\bfY^\intercal \tilde{\calA} \bfY$ and hence $\bfe_1, \dots, \bfe_K$ are the top $K$ eigenvectors. Thus, the columns of $\bfY \calZ$ lie in the span of $\bfy_1, \dots, \bfy_K$.


\subsection{Proof of Lemma \ref{lemma:bound_on_D-calD}}
As $\bfD$ and $\calD$ are diagonal matrices, $\norm{\bfD - \calD} = \max_{i \in [N]} \abs{D_{ii} - \calD_{ii}}$. Applying union bound, we get,
\begin{equation*}
  \rmP(\max_{i \in [N]} \abs{D_{ii} - \calD_{ii}} \geq \epsilon) \leq \sum_{i = 1}^N \rmP(\abs{D_{ii} - \calD_{ii}} \geq \epsilon).
\end{equation*}
We consider an arbitrary term in this summation. For any $i \in [N]$, note that $D_{ii} = \sum_{j \neq i} A_{ij}$ is a sum of independent Bernoulli random variables such that $\rmE[D_{ii}] = \calD_{ii}$. We consider two cases depending on the value of $p$.

\paragraph*{Case 1: $p > \frac{1}{2}$} By Hoeffding's inequality,
\begin{equation*}
  \rmP(\abs{D_{ii} - \calD_{ii}} \geq \epsilon) \leq 2 \exp \left( -\frac{2\epsilon^2}{N} \right).
\end{equation*}
Setting $\epsilon = \sqrt{2(\alpha + 1)} \sqrt{p N \ln N}$, we get for any $\alpha > 0$,
\begin{equation*}
  \rmP(\abs{D_{ii} - \calD_{ii}} \geq \sqrt{2(\alpha + 1)} \sqrt{p N \ln N}) \leq 2 \exp \left( - \frac{4p(\alpha + 1)N \ln N}{N} \right) \leq N^{-(\alpha + 1)}.
\end{equation*}

\paragraph*{Case 2: $p \leq \frac{1}{2}$}  By Bernstein's inequality, as $\abs{A_{ij} - \calA_{ij}} \leq 1$ for all $i, j \in [N]$,
\begin{equation*}
  \rmP(\abs{D_{ii} - \calD_{ii}} \geq \epsilon) \leq 2 \exp \left( -\frac{\epsilon^2/2}{\sum_{j \neq i} \rmE[(A_{ij} - \calA_{ij})^2]  + \epsilon/3} \right).
\end{equation*}
Also note that,
\begin{equation*}
  \rmE[(A_{ij} - \calA_{ij})^2] \leq \calA_{ij}(1 - \calA_{ij})^2 + (1 - \calA_{ij})(-\calA_{ij})^2 = \calA_{ij} (1 - \calA_{ij}) \leq \calA_{ij} \leq p.
\end{equation*}
Thus,
\begin{equation*}
  \rmP(\abs{D_{ii} - \calD_{ii}} \geq \epsilon) \leq 2 \exp \left( -\frac{\epsilon^2/2}{Np  + \epsilon/3} \right).
\end{equation*}
Let $\epsilon = c \sqrt{p N \ln N}$ for some constant $c > 0$ and assume that $p \geq C\frac{\ln N}{N}$ for some $C > 0$. We get,
\begin{align*}
  2 \exp \left( -\frac{\epsilon^2/2}{Np  + \epsilon/3} \right) = 2 \exp \left(- \frac{c^2 p N \ln N}{2(Np + c \sqrt{p N \ln N}/3)} \right) &= 2 \exp \Big(- \frac{c^2 \ln N}{2(1 + \frac{c}{3} \sqrt{\frac{\ln N}{p N}})} \Big) \\
  &\leq 2 \exp \Big(- \frac{c^2 \ln N}{2(1 + \frac{c}{3 \sqrt{C}})} \Big).
\end{align*}
Let $c$ be such that $\frac{c^2}{2(1 + c/3\sqrt{C})} \geq 2 (\alpha + 1)$. Such a $c$ can always be chosen as $\underset{c \rightarrow \infty}{\lim} \frac{c^2}{2(1 + c/3\sqrt{C})} = \infty$. Then,
\begin{equation*}
  \rmP(\abs{D_{ii} - \calD_{ii}} \geq \epsilon) \leq N^{-(\alpha + 1)}.
\end{equation*}

Thus, there always exists a constant $\const_1(C, \alpha)$ that depends only on $C$ and $\alpha$ such that for all $\alpha > 0$ and for all values of $p \geq C \ln N / N$,
\begin{equation*}
  \rmP(\abs{D_{ii} - \calD_{ii}} \geq \const_1(C, \alpha) \sqrt{p N \ln N}) \leq N^{-(\alpha + 1)}.
\end{equation*}
Applying the union bound over all $i \in [N]$ yields the desired result.


\subsection{Proof of Lemma \ref{lemma:bound_on_A-calA}}

Note that $\max_{i, j \in [N]} \calA_{ij} = p$. Define $g = p N = N \max_{i, j \in [N]} \calA_{ij}$. Note that, $g \geq C \ln N$ as $p \geq C\frac{\ln N}{N}$. By Theorem 5.2 from \cite{LeiEtAl:2015:ConsistencyOfSpectralClusteringInSBM}, for any $\alpha > 0$, there exists a constant $\const_4(C, \alpha)$ such that, $$\norm{\bfA - \calA} \leq \const_4(C, \alpha) \sqrt{g} = \const_4(C, \alpha) \sqrt{pN}$$ with probability at least $1 - N^{-\alpha}$.


\subsection{Proof of Lemma \ref{lemma:bound_on_eigenvector_diff}}

Because $\bfY^\intercal \bfY = \bfI$, for any orthonormal matrix $\bfU \in \bbR^{K \times K}$ such that $\bfU \bfU^\intercal = \bfU^\intercal \bfU = \bfI$, $$\norm{\bfY \calZ - \bfY \bfZ \bfU}[F][2] = \norm{\bfY(\calZ - \bfZ \bfU)}[F][2] = \trace{(\calZ - \bfZ \bfU)^\intercal \bfY^\intercal \bfY (\calZ - \bfZ \bfU)} = \norm{\calZ - \bfZ \bfU}[F][2].$$ Thus, it is enough to show an upper bound on $\norm{\calZ - \bfZ \bfU}[F]$, where recall that columns of $\calZ \in \bbR^{N - r \times K}$ and $\bfZ \in \bbR^{N - r \times K}$ contain the leading $K$ eigenvectors of $\bfY^\intercal \calL \bfY$ and $\bfY^\intercal \bfL \bfY$ respectively. Thus, $$\inf_{\bfU \in \bbR^{K \times K} : \bfU \bfU^\intercal = \bfU^\intercal \bfU = \bfI} \norm{\bfY \calZ - \bfY \bfZ \bfU}[F] = \inf_{\bfU \in \bbR^{K \times K} : \bfU \bfU^\intercal = \bfU^\intercal \bfU = \bfI} \norm{\calZ - \bfZ \bfU}[F].$$ By equation (2.6) and Proposition 2.2 in \cite{VuLei:2013:MinimaxSparsePrincipalSubspaceEstimationInHighDimensions}, $$\inf_{\bfU \in \bbR^{K \times K} : \bfU \bfU^\intercal = \bfU^\intercal \bfU = \bfI}\norm{\calZ - \bfZ\bfU}[F] \leq \sqrt{2} \norm{\calZ\calZ^\intercal (\bfI - \bfZ \bfZ^\intercal)}[F].$$
Moreover, $\norm{\calZ\calZ^\intercal (\bfI - \bfZ \bfZ^\intercal)}[F] \leq \sqrt{K} \norm{\calZ\calZ^\intercal (\bfI - \bfZ \bfZ^\intercal)}$ as $\rank{\calZ\calZ^\intercal (\bfI - \bfZ \bfZ^\intercal)} \leq K$. Thus, we get, 
\begin{equation}
  \label{eq:Z-ZU_leq_ZZ_I-ZZ}
  \inf_{\bfU \in \bbR^{K \times K} : \bfU\bfU^\intercal = \bfU^\intercal \bfU = \bfI}\norm{\calZ - \bfZ\bfU}[F] \leq \sqrt{2K} \norm{\calZ\calZ^\intercal (\bfI - \bfZ \bfZ^\intercal)}.
\end{equation}

Let $\mu_1 \leq \mu_2 \leq \dots \leq \mu_{N - r}$ be eigenvalues of $\bfY^\intercal \calL \bfY$ and $\alpha_1 \leq \alpha_2 \leq \dots \leq \alpha_{N - r}$ be eigenvalues of $\bfY^\intercal \bfL \bfY$. By Weyl's perturbation theorem, $$\abs{\mu_i - \alpha_i} \leq \norm{\bfY^\intercal \calL \bfY - \bfY^\intercal \bfL \bfY}, \,\,\,\, \forall i \in [N - r].$$ Define $\gamma = \mu_{K + 1} - \mu_{K}$ to be the eigen-gap between the $K^{th}$ and $(K + 1)^{th}$ eigenvalues of $\bfY^\intercal \calL \bfY$. 

\paragraph*{Case 1: $\norm{\bfY^\intercal \calL \bfY - \bfY^\intercal \bfL \bfY} \leq \frac{\gamma}{4}$} If $\norm{\bfY^\intercal \calL \bfY - \bfY^\intercal \bfL \bfY} \leq \frac{\gamma}{4}$, then $\abs{\mu_i - \alpha_i} \leq \frac{\gamma}{4}$ for all $i \in [N - r]$ by the inequality given above. Thus, $\alpha_1, \alpha_2, \dots, \alpha_K \in [0, \mu_K + \frac{\gamma}{4}]$ and $\alpha_{K + 1}, \alpha_{K + 2}, \dots, \alpha_{N - r} \in [\mu_{K + 1} - \frac{\gamma}{4}, \infty)$. Let $S = [0, \mu_K + \frac{\gamma}{4}]$, then $\mu_1, \dots, \mu_K \in S$ and $\alpha_{K + 1}, \dots, \alpha_{N - r} \notin S$. Define $\delta$ as, $$\delta = \min\{ \abs{\alpha_i - s}, \alpha_i \notin S, s \in S\}.$$ Then, $\delta \geq [\mu_{K + 1} - \gamma/4] - [\mu_{K} + \gamma/4] = \gamma/2$. By Davis-Kahan $\sin \Theta$ theorem, $$\norm{\calZ\calZ^\intercal (\bfI - \bfZ \bfZ^\intercal)} \leq \frac{1}{\delta} \norm{\bfY^\intercal \calL \bfY - \bfY^\intercal \bfL \bfY} = \frac{2}{\gamma} \norm{\bfY^\intercal \calL \bfY - \bfY^\intercal \bfL \bfY}.$$

\paragraph*{Case 2: $\norm{\bfY^\intercal \calL \bfY - \bfY^\intercal \bfL \bfY} > \frac{\gamma}{4}$} Note that $\norm{\calZ\calZ^\intercal (\bfI - \bfZ \bfZ^\intercal)} \leq 1$ as, $$\norm{\calZ\calZ^\intercal (\bfI - \bfZ \bfZ^\intercal)} \leq \norm{\calZ\calZ^\intercal}\,\,\norm{\bfI - \bfZ \bfZ^\intercal} = 1 . 1 = 1.$$ Thus, if $\norm{\bfY^\intercal \calL \bfY - \bfY^\intercal \bfL \bfY} > \frac{\gamma}{4}$, then, $$\norm{\calZ\calZ^\intercal (\bfI - \bfZ \bfZ^\intercal)} \leq \frac{4}{\gamma} \norm{\bfY^\intercal \calL \bfY - \bfY^\intercal \bfL \bfY}.$$

In both cases, $\norm{\calZ\calZ^\intercal (\bfI - \bfZ \bfZ^\intercal)} \leq \frac{4}{\gamma} \norm{\bfY^\intercal \calL \bfY - \bfY^\intercal \bfL \bfY}$. Using \eqref{eq:L-calL_bound} and \eqref{eq:Z-ZU_leq_ZZ_I-ZZ}, we get with probability at least $1 - 2N^{-\alpha}$, $$\inf_{\bfU \in \bbR^{K \times K} : \bfU\bfU^\intercal = \bfU^\intercal \bfU = \bfI}\norm{\calZ - \bfZ\bfU}[F] \leq \const_5(C, \alpha) \frac{4\sqrt{2K}}{\gamma} \sqrt{p N \ln N}.$$


\subsection{Proof of Lemma \ref{lemma:k_means_error}}

Equation \eqref{eq:num_mistakes_bound} directly follows from Lemma 5.3 in \cite{LeiEtAl:2015:ConsistencyOfSpectralClusteringInSBM}. We only need to show that $$\frac{8 (2 + \epsilon)}{\delta^2} \norm{\bfX \bfU - \bfY \calZ}[F][2] < \frac{N}{K}.$$ Equation \eqref{eq:correct_solution_on_non-mistakes} then follows from Lemma 5.3 in \cite{LeiEtAl:2015:ConsistencyOfSpectralClusteringInSBM}. Recall that $\delta = \sqrt{\frac{2K}{N}}$. Using Lemma \ref{lemma:bound_on_eigenvector_diff}, we get
\begin{equation*}
  \frac{8 (2 + \epsilon)}{\delta^2} \norm{\bfX \bfU^\intercal - \bfY \calZ}[F][2] \leq \const_5(C, \alpha)^2 \frac{128(2 + \epsilon)}{\gamma^2} p N^2 \ln N < \frac{N}{K}.
\end{equation*}
Here, the last inequality follows from the assumption that $\gamma^2 > \const_5(C, \alpha)^2 . 128 (2 + \epsilon) p NK \ln N$.


\subsection{Proof of Lemma \ref{lemma:calQ-Q_bound}}

We begin by showing a simple result. Let $a, b > 0$. Then,
\begin{equation*}
    \abs{\sqrt{a} - \sqrt{b}} = \frac{\abs{(\sqrt{a} - \sqrt{b})(\sqrt{a} + \sqrt{b})}}{\sqrt{a} + \sqrt{b}} = \frac{\abs{a - b}}{\sqrt{a} + \sqrt{b}} \leq \frac{\abs{a - b}}{\sqrt{b}}.
\end{equation*}
Further,
\begin{equation*}
    \Big\vert \frac{1}{\sqrt{a}} - \frac{1}{\sqrt{b}}  \Big\vert = \frac{\abs{\sqrt{a} - \sqrt{b}}}{\sqrt{ab}} \leq \frac{\abs{a - b}}{b\sqrt{a}}.
\end{equation*}

Coming back to the bound on $\norm{\calQ^{-1} - \bfQ^{-1}}$, note that, as $\calQ^{-1} = (\lambda_1 - p)^{-1/2} \bfI$, we have that
\begin{equation*}
    \norm{\calQ^{-1} - \bfQ^{-1}} = \max \left\{ \Big\vert \nu_i - \frac{1}{\sqrt{\lambda_1 - p}} \Big\vert \;:\; \nu_i \text{ is an eigenvalue of } \bfQ^{-1} \right\}.
\end{equation*}
As $\bfQ = \sqrt{\bfY^\intercal \bfD \bfY}$, the eigenvalues of $\bfQ^{-1}$ are given by $1/\sqrt{\mu{'}_1}$, \dots, $1/\sqrt{\mu{'}_{N - r}}$, where, $\mu{'}_1$, \dots, $\mu{'}_{N - r}$ are the eigenvalues of $\bfY^\intercal \bfD \bfY$. Moreover, by substituting $a = \mu{'}_i$ and $b = \lambda_1 - p$ in the inequality derived above, we get
\begin{equation}
    \label{eq:Q_calQ_eigenvalue_diff}
    \Big\vert \frac{1}{\sqrt{\mu{'}_i}} - \frac{1}{\sqrt{\lambda_1 - p}} \Big\vert \leq \frac{\abs{\mu{'}_i - (\lambda_1 - p)}}{(\lambda_1 - p) \sqrt{\mu{'}_i}}, \;\;\; \forall\; i \in [N - r].
\end{equation}
By Weyl's perturbation theorem, for any $i \in [N - r]$,
\begin{equation*}
    \abs{\mu{'}_i - (\lambda_1 - p)} \leq \norm{\bfY^\intercal \bfD \bfY - \bfY^\intercal \calD \bfY} \leq \norm{\bfD - \calD},
\end{equation*}
where the last inequality follows as $\bfY^\intercal \bfY = \bfI$. Let us assume for now that $\norm{\bfD - \calD} \leq \frac{\lambda_1 - p}{2}$ (we prove this below). Then, $\abs{\mu{'}_i - (\lambda_1 - p)} \leq \frac{\lambda_1 - p}{2}$ for all $i \in [N - r]$. Hence,
\begin{equation*}
    \mu{'}_i \geq \frac{\lambda_1 - p}{2}, \;\;\; \forall \; i \in [N - r].
\end{equation*}
Using this in \eqref{eq:Q_calQ_eigenvalue_diff} results in
\begin{equation*}
    \Big\vert \frac{1}{\sqrt{\mu{'}_i}} - \frac{1}{\sqrt{\lambda_1 - p}} \Big\vert \leq \frac{\sqrt{2}}{\sqrt{(\lambda_1 - p)^3}} \norm{\bfD - \calD}, \;\;\; \forall\; i \in [N - r].
\end{equation*}
Taking the maximum over all $i \in [N - r]$ yields the desired result. Next, we prove that $\norm{\bfD - \calD} \leq \frac{\lambda_1 - p}{2}$.

Recall from Lemma \ref{lemma:bound_on_D-calD} that $\norm{\bfD - \calD} \leq \const_1(C, \alpha) \sqrt{p N \ln N}$, where the proof of Lemma \ref{lemma:bound_on_D-calD} requires that $\const_1(C, \alpha)$ satisfies the following condition:
\begin{equation*}
    \frac{\const_1(C, \alpha)^2}{2\left(1 + \frac{\const_1(C, \alpha)}{3\sqrt{C}}\right)} \geq 2(\alpha + 1).
\end{equation*}
Additionally, to show that $\norm{\bfD - \calD} \leq \frac{\lambda_1 - p}{2}$, we need to ensure that $\const_1(C, \alpha) \leq \frac{\lambda_1 - p}{2 \sqrt{p N \ln N}}$. A constant $\const_1(C, \alpha)$ that satisfies both these conditions exists if:
\begin{equation*}
    \frac{(\lambda_1 - p)^2 / 4pN \ln N}{2\left(1 + \frac{(\lambda_1 - p) / 2\sqrt{pN \ln N}}{3\sqrt{C}} \right)} \geq 2(\alpha + 1).
\end{equation*}
Simplifying the expression above results in
\begin{equation*}
    \frac{1}{\left(\frac{\sqrt{pN \ln N}}{\lambda_1 - p}\right)\left(\frac{\sqrt{pN \ln N}}{\lambda_1 - p} + \frac{1}{6\sqrt{C}} \right)} \geq 16(\alpha + 1).
\end{equation*}
The assumption made in the lemma guarantees that such a condition is satisfied. Hence, $\const_1(C, \alpha)$ can be set such that $\norm{\bfD - \calD} \leq \frac{\lambda_1 - p}{2}$.


\subsection{Proof of Lemma \ref{lemma:eigenvector_diff_bound_normalized}}

As in the proof of Lemma \ref{lemma:bound_on_eigenvector_diff}, because $\bfY^\intercal \bfY = \bfI$, for any orthonormal matrix $\bfU \in \bbR^{K \times K}$ such that $\bfU^\intercal \bfU = \bfU \bfU^\intercal = \bfI$,
\begin{equation*}
    \norm{\bfY \calQ^{-1} \calZ - \bfY \bfQ^{-1} \bfZ \bfU}[F] =  \norm{\calQ^{-1} \calZ - \bfQ^{-1} \bfZ \bfU}[F].
\end{equation*}
As $\calQ, \bfQ \in \bbR^{N - r \times N - r}$ and $\calZ, \bfZ \in \bbR^{N - r \times K}$, we have that $\rank{\calQ^{-1} \calZ} \leq K$ and $\rank{\bfQ^{-1} \bfZ \bfU} \leq K$, and hence $\rank{\calQ^{-1} \calZ - \bfQ^{-1} \bfZ \bfU} \leq 2K$. Therefore,
\begin{equation*}
    \norm{\calQ^{-1} \calZ - \bfQ^{-1} \bfZ \bfU}[F] \leq \sqrt{2K} \norm{\calQ^{-1} \calZ - \bfQ^{-1} \bfZ \bfU}.
\end{equation*}
Moreover, using $\calQ^{-1} = (\sqrt{\lambda_1 - p})^{-1} \bfI$ and Lemma \ref{lemma:calQ-Q_bound},
\begin{align*}
    \norm{\calQ^{-1} \calZ - \bfQ^{-1} \bfZ \bfU} &\leq \norm{\calQ^{-1}} \cdot \norm{\calZ - \bfZ \bfU} + \norm{\calQ^{-1} - \bfQ^{-1}} \cdot \norm{\bfZ \bfU} \\
    &\leq \frac{1}{\sqrt{\lambda_1 - p}} \norm{\calZ - \bfZ \bfU} + \sqrt{\frac{2}{(\lambda_1 - p)^3}} \norm{\bfD - \calD} \cdot \norm{\bfZ \bfU}.
\end{align*}
Note that $\norm{\bfZ \bfU} = \sqrt{\lambdamax{\bfU^\intercal \bfZ^\intercal \bfZ \bfU}} = \sqrt{\lambdamax{\bfU^\intercal \bfU}} = \sqrt{\lambdamax{\bfI}} = 1$. Also note that $\norm{\calZ - \bfZ \bfU} \leq \norm{\calZ - \bfZ \bfU}[F]$. Combining all of this information, we get,
{
\small
\begin{eqnarray*}
    \inf_{\bfU : \bfU^\intercal \bfU = \bfU\bfU^\intercal = \bfI} \norm{\bfY \calQ^{-1} \calZ - &\bfY& \bfQ^{-1} \bfZ \bfU}[F] \leq \\ &&\sqrt{\frac{2K}{\lambda_1 - p}} \inf_{\bfU : \bfU^\intercal \bfU = \bfU\bfU^\intercal = \bfI} \norm{\calZ - \bfZ \bfU}[F] + \sqrt{\frac{4K}{(\lambda_1 - p)^3}} \norm{\bfD - \calD}.
\end{eqnarray*}
}%
Let $\mu_1 \leq \mu_2 \leq \dots \leq \mu_{N - r}$ be eigenvalues of $\calQ^{-1} \bfY^\intercal \calL \bfY \calQ^{-1}$ and $\alpha_1 \leq \alpha_2 \leq \dots \leq \alpha_{N - r}$ be eigenvalues of $\bfQ^{-1} \bfY^\intercal \bfL \bfY \bfQ^{-1}$. Define $\gamma = \mu_{K + 1} - \mu_K$. Using a strategy similar to the one used in the proof of Lemma \ref{lemma:bound_on_eigenvector_diff}, we get:
\begin{equation*}
    \inf_{\bfU : \bfU^\intercal \bfU = \bfU\bfU^\intercal = \bfI} \norm{\calZ - \bfZ \bfU}[F] \leq \frac{4\sqrt{2K}}{\gamma} \norm{\calQ^{-1} \bfY^\intercal \calL \bfY \calQ^{-1} - \bfQ^{-1} \bfY^\intercal \bfL \bfY \bfQ^{-1}}.
\end{equation*}
Using \eqref{eq:calQYcalLYcalQ-QYLYQ_bound} and Lemma \ref{lemma:bound_on_D-calD} results in
{
\small
\begin{align*}
    \inf_{\bfU : \bfU^\intercal \bfU = \bfU\bfU^\intercal = \bfI}  &\norm{\calZ - \bfZ \bfU}[F] \\
    \leq& \frac{8\sqrt{2K}}{\gamma (\lambda_1 - p)} \Bigg[ \left( \frac{(\lambda_1 - \bar{\lambda}) \const_1(C, \alpha) \sqrt{2}}{\lambda_1 - p} + \frac{\const_5(C, \alpha)}{2}\right) \sqrt{pN \ln N} + \\
    &\left( \frac{(\lambda_1 - \bar{\lambda}) \const_1(C, \alpha)^2}{\lambda_1 - p} + \const_1(C, \alpha) \const_5(C, \alpha) \sqrt{2} \right) \frac{p N \ln N}{\lambda_1 - p} + \\
    &\const_1(C, \alpha)^2 \const_5(C, \alpha) \frac{(pN \ln N)^{3/2}}{(\lambda_1 - p)^2} \Bigg] \\
    \leq & \frac{8\sqrt{2K}}{\gamma (\lambda_1 - p)} \Bigg[ \frac{(\lambda_1 - \bar{\lambda}) \const_1(C, \alpha) \sqrt{2}}{\lambda_1 - p} + \frac{\const_5(C, \alpha)}{2} + \\
    & \left( \frac{(\lambda_1 - \bar{\lambda}) \const_1(C, \alpha)^2}{\lambda_1 - p} + \const_1(C, \alpha) \const_5(C, \alpha) \sqrt{2} \right) \const_2(C, \alpha)  + \\
    & \const_1(C, \alpha)^2 \const_5(C, \alpha)\const_2(C, \alpha)^2 \Bigg] \sqrt{p N \ln N} \\
    \leq & \frac{8\sqrt{2K} \const_3(C, \alpha)}{\gamma (\lambda_1 - p)} \sqrt{p N \ln N}.
\end{align*}
}%
Here, the second inequality follows from the assumption that there is a constant $\const_2(C, \alpha)$ that satisfies $\frac{\sqrt{p N \ln N}}{\lambda_1 - p} \leq \const_2(C, \alpha)$. The last inequality follows by choosing $\const_3(C, \alpha)$ such that the expression between the square brackets in the second inequality is less than $\const_3(C, \alpha)$. Using the expression above, we get:
{
\small
\begin{equation*}
    \inf_{\bfU : \bfU^\intercal \bfU = \bfU\bfU^\intercal = \bfI} \norm{\bfY \calQ^{-1} \calZ - \bfY \bfQ^{-1} \bfZ \bfU}[F] \leq \left[\frac{16 K \const_3(C, \alpha)}{\gamma (\lambda_1 - p)^{3/2}} + \frac{2 \const_1(C, \alpha) \sqrt{K}}{(\lambda_1 - p)^{3/2}} \right] \sqrt{p N \ln N}.
\end{equation*}
}

%% file: appendix_additional_experiments.tex
\section{Additional experiments}
\label{appendix:additional_experiments}

\rebuttal{In this section, we present experimental results to demonstrate the performance of \textsc{NRepSC}. We also experiment with another real-world network and present a few additional plots for \textsc{URepSC} that were left out of \Cref{section:numerical_results} due to space constraints. This section ends with a numerical validation of the time complexity of our algorithms.}


\rebuttal{\subsection{Experiments with \textsc{NRepSC}}}
\label{appendix:experiments_with_nrepsc}

\begin{figure}[t]
    \centering
    \subfloat[][Accuracy vs no. of nodes]{\includegraphics[width=0.33\textwidth]{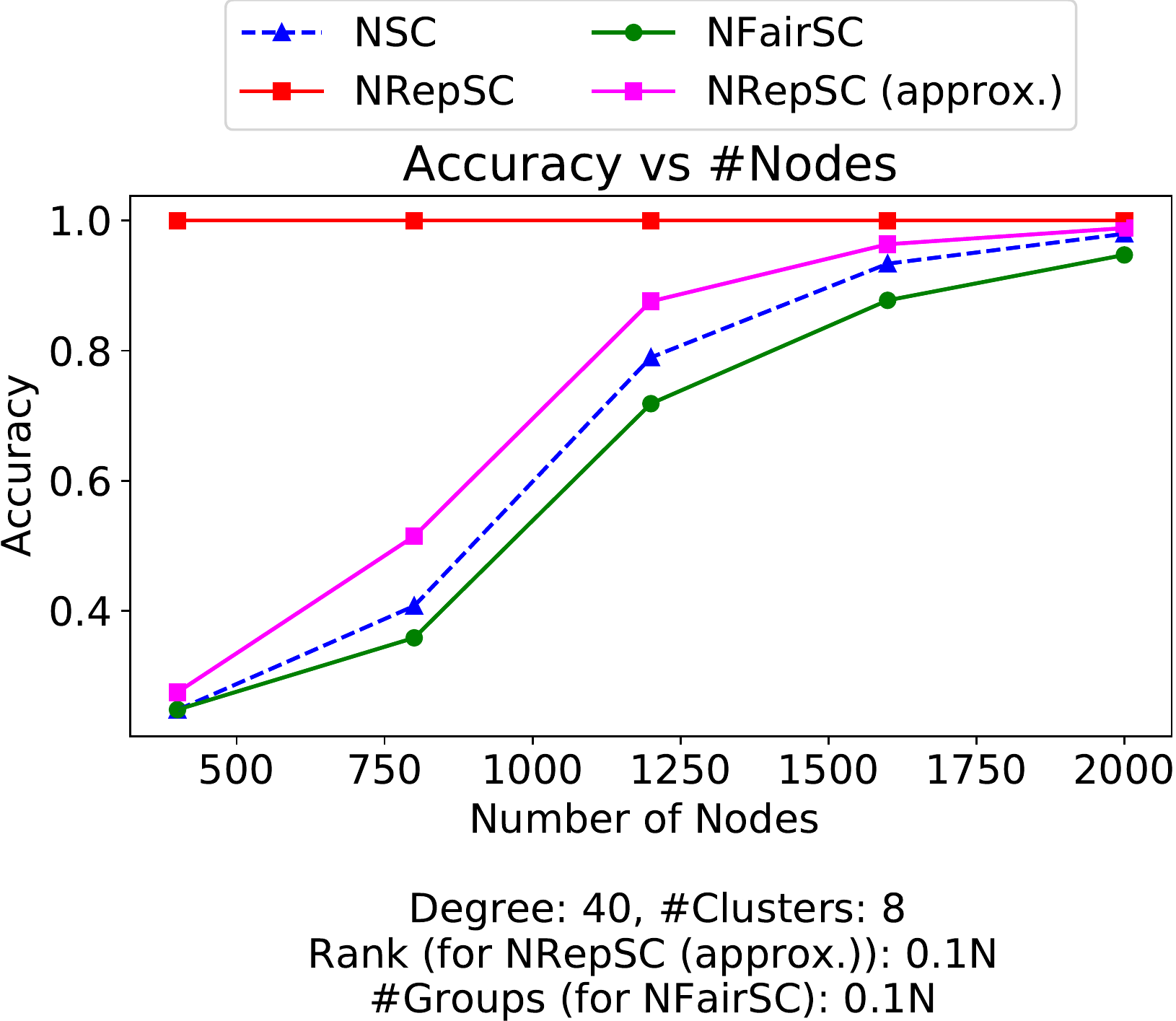}\label{fig:d_reg_norm:vs_N}}%
    \subfloat[][Accuracy vs no. of clusters]{\includegraphics[width=0.33\textwidth]{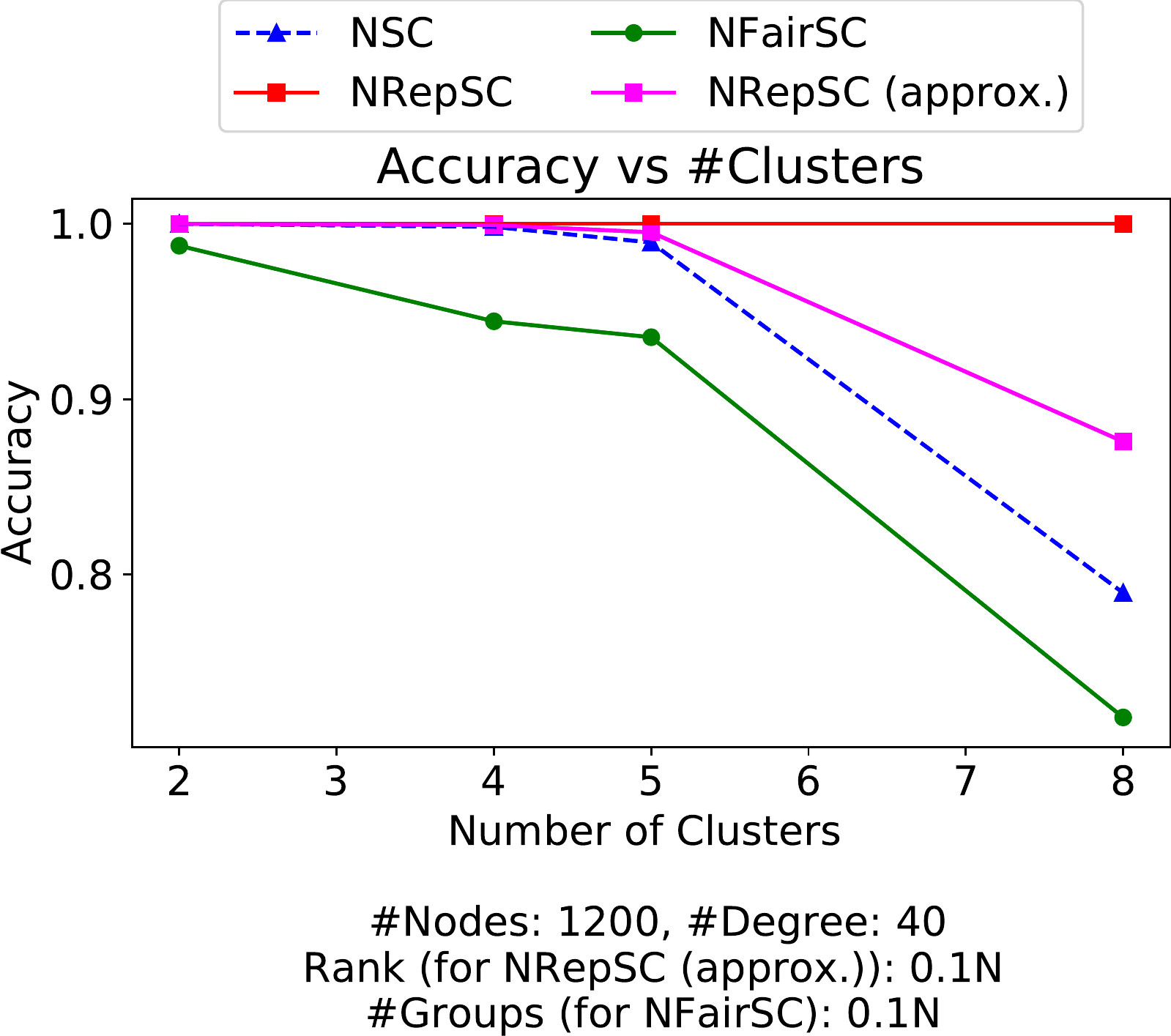}\label{fig:d_reg_norm:vs_K}}%
    \subfloat[][Accuracy vs degree of $\calR$]{\includegraphics[width=0.33\textwidth]{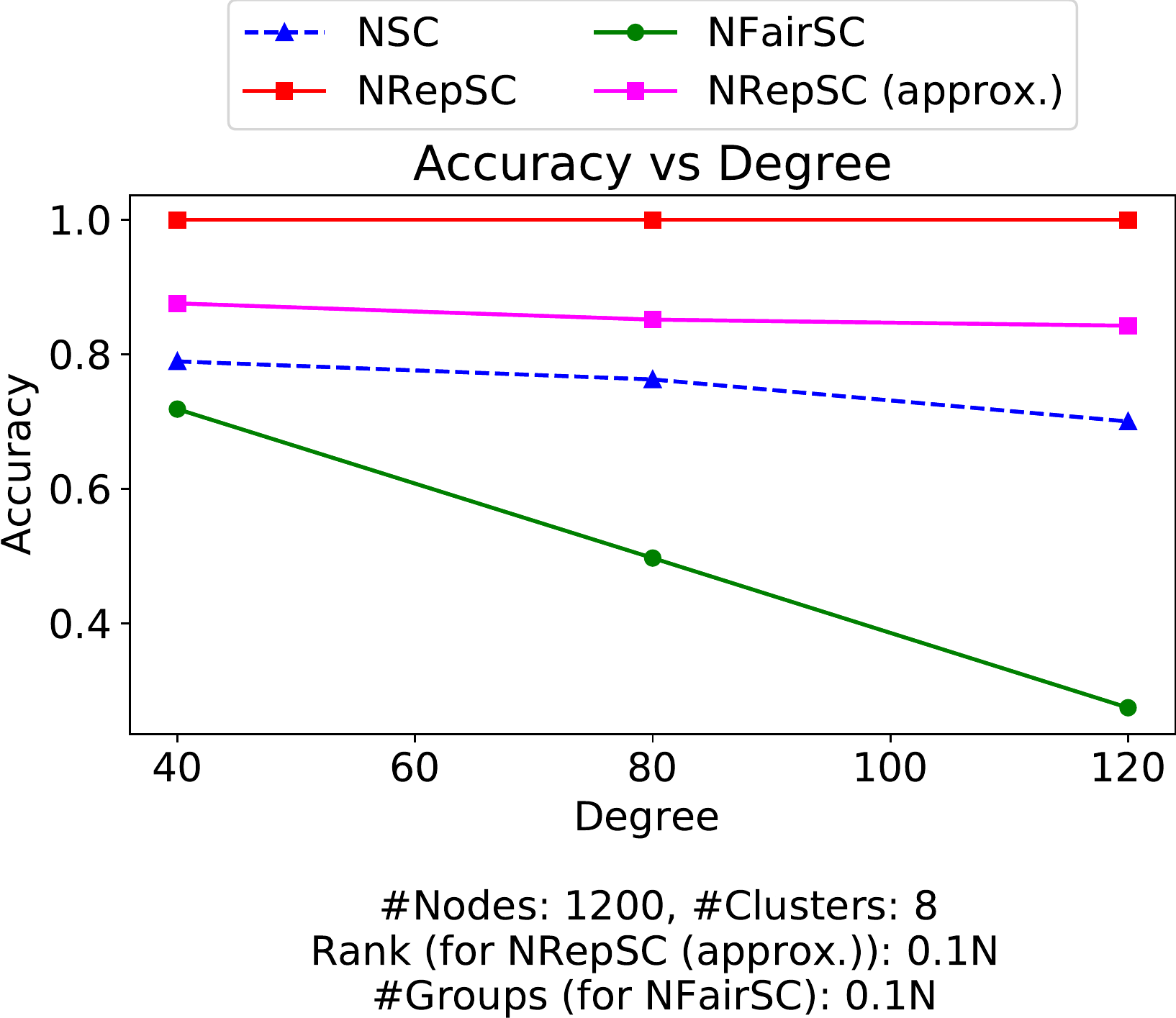}\label{fig:d_reg_norm:vs_d}}
    \caption{Comparing \textsc{NRepSC} with other ``normalized'' algorithms using synthetically generated $d$-regular representation graphs.}
    \label{fig:d_reg_norm}
\end{figure}

\begin{figure}[t]
    \centering
    \subfloat[][$N = 1000$, $K = 4$]{\includegraphics[width=0.48\textwidth]{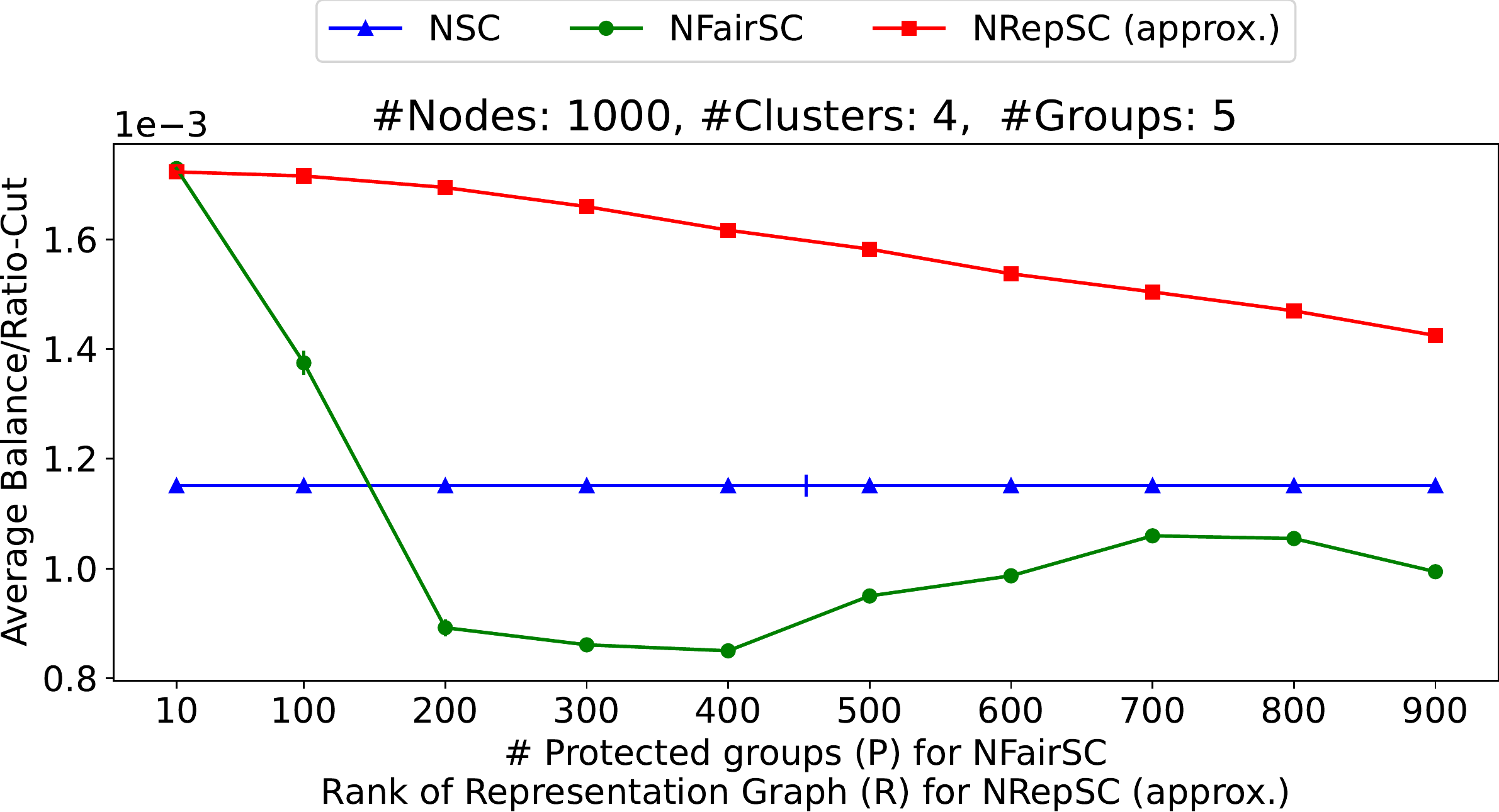}}%
    \hspace{0.5cm}\subfloat[][$N = 3000$, $K = 4$]{\includegraphics[width=0.48\textwidth]{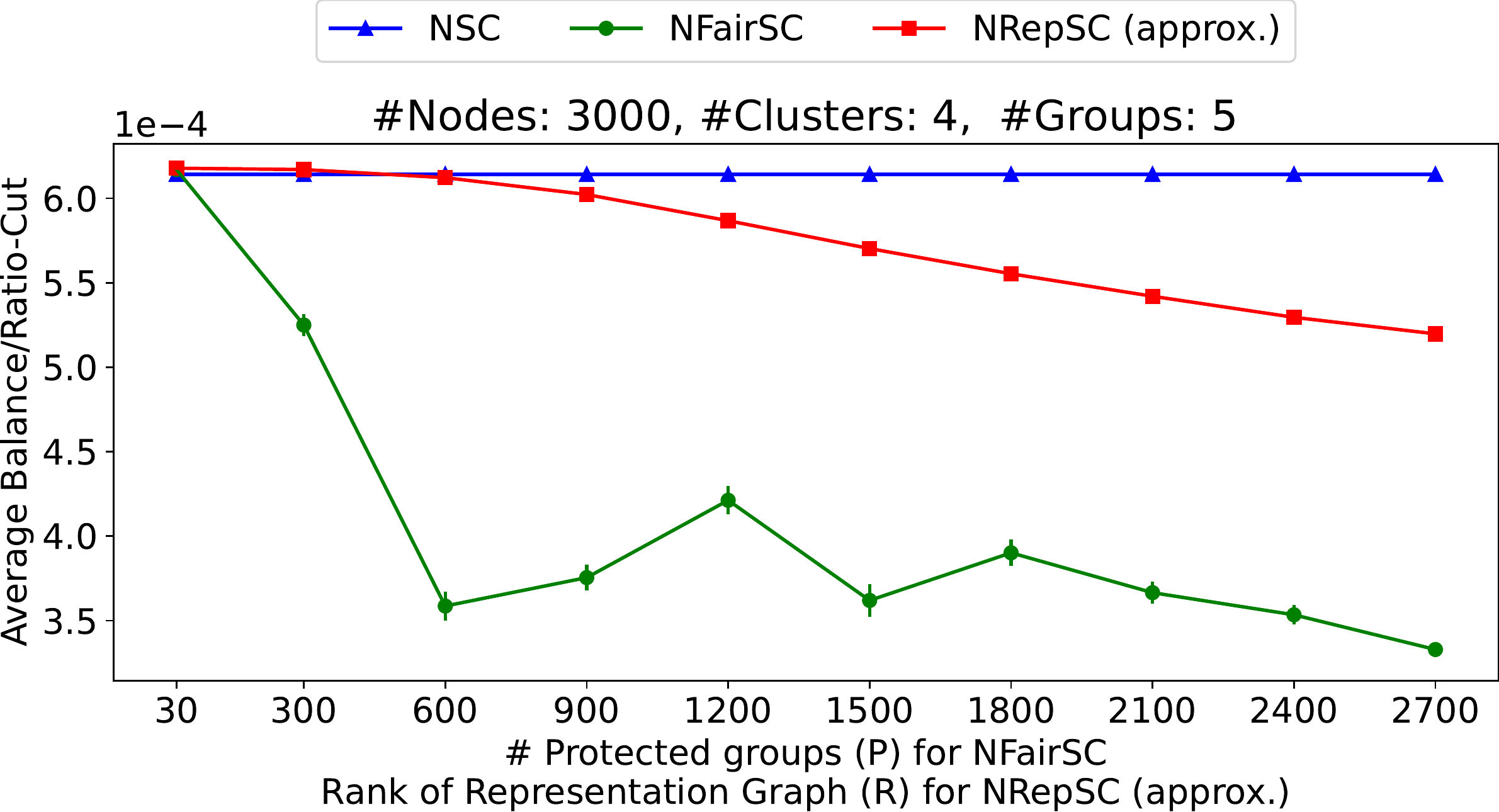}}

    \subfloat[][$N = 1000$, $K = 8$]{\includegraphics[width=0.48\textwidth]{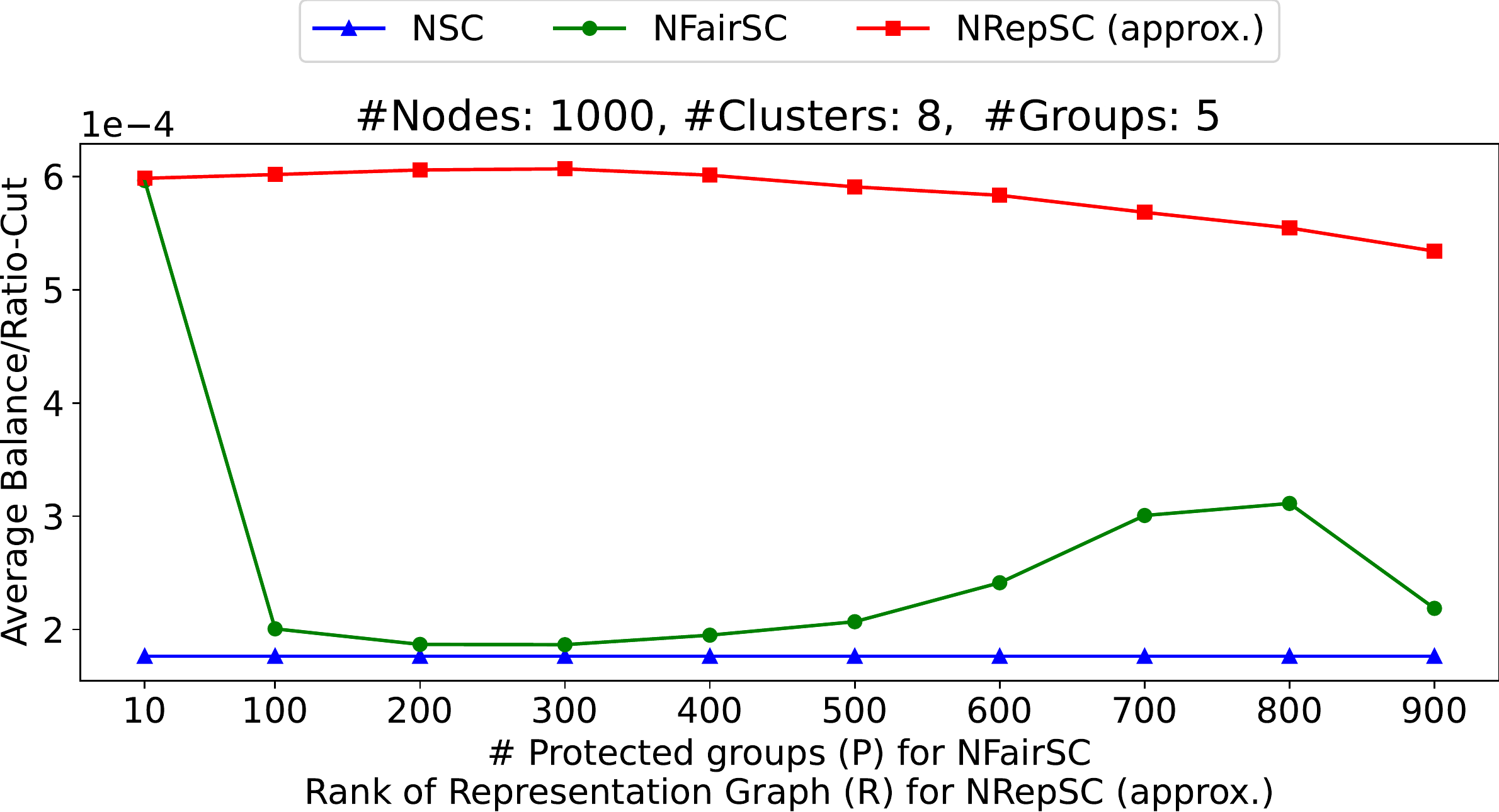}}%
    \hspace{0.5cm}\subfloat[][$N = 3000$, $K = 8$]{\includegraphics[width=0.48\textwidth]{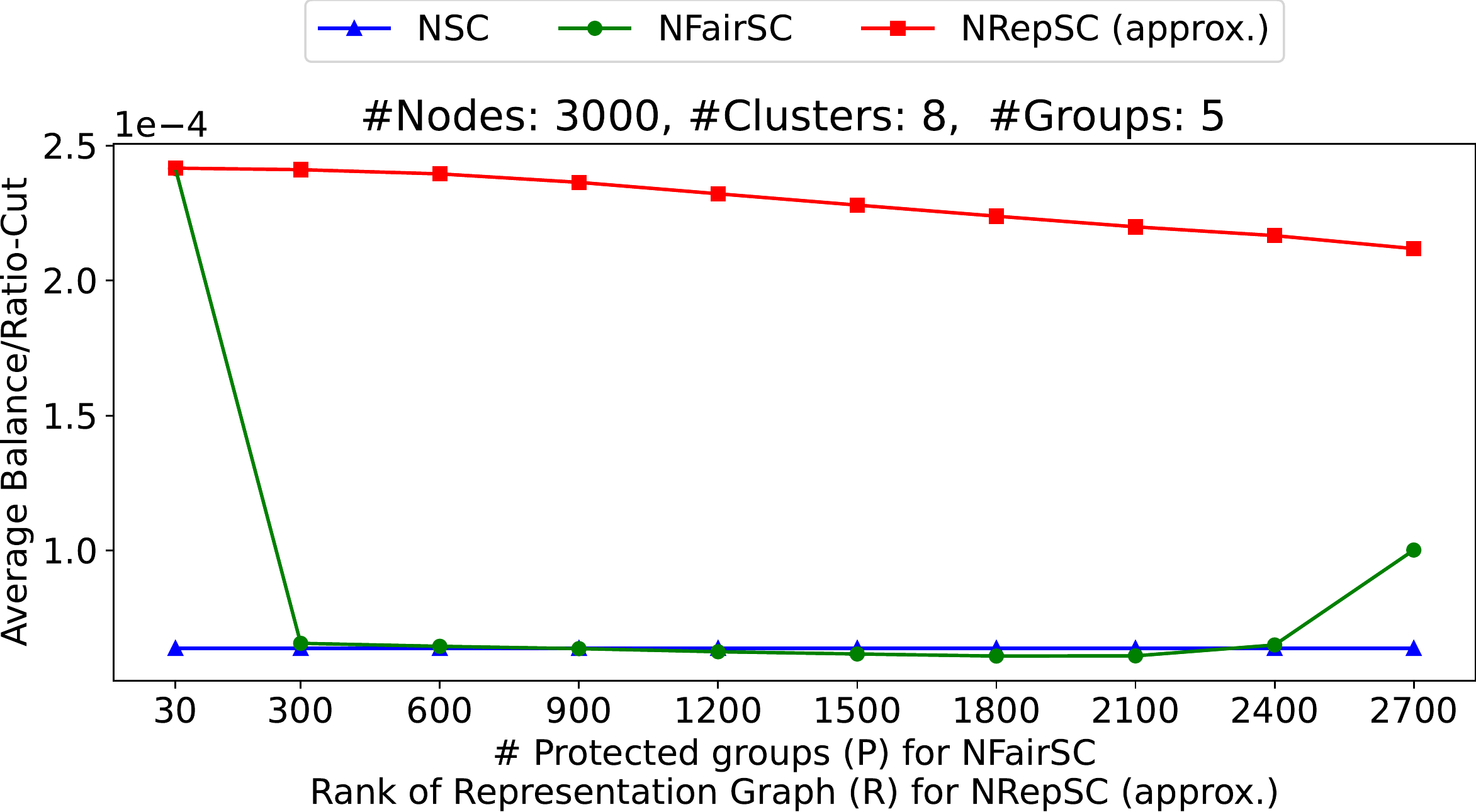}}
    \caption{Comparing \textsc{NRepSC (approx.)} with \textsc{NFairSC} using synthetically generated representation graphs sampled from an SBM.}
    \label{fig:sbm_comparison_norm}
\end{figure}

\begin{figure}[t]
    \centering
    \subfloat[][$K = 2$]{\includegraphics[width=0.48\textwidth]{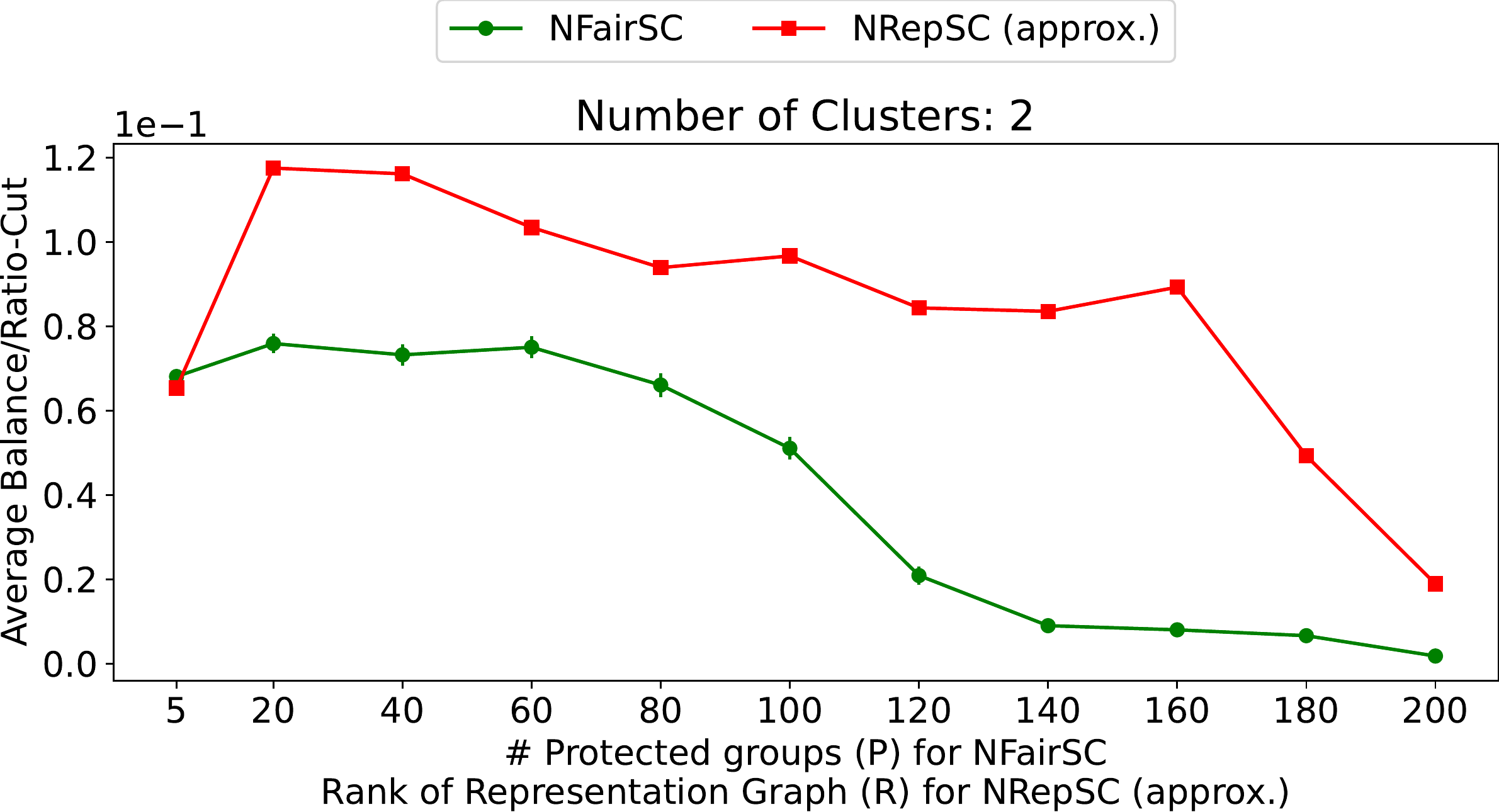}}%
    \hspace{0.5cm}\subfloat[][$K = 4$]{\includegraphics[width=0.48\textwidth]{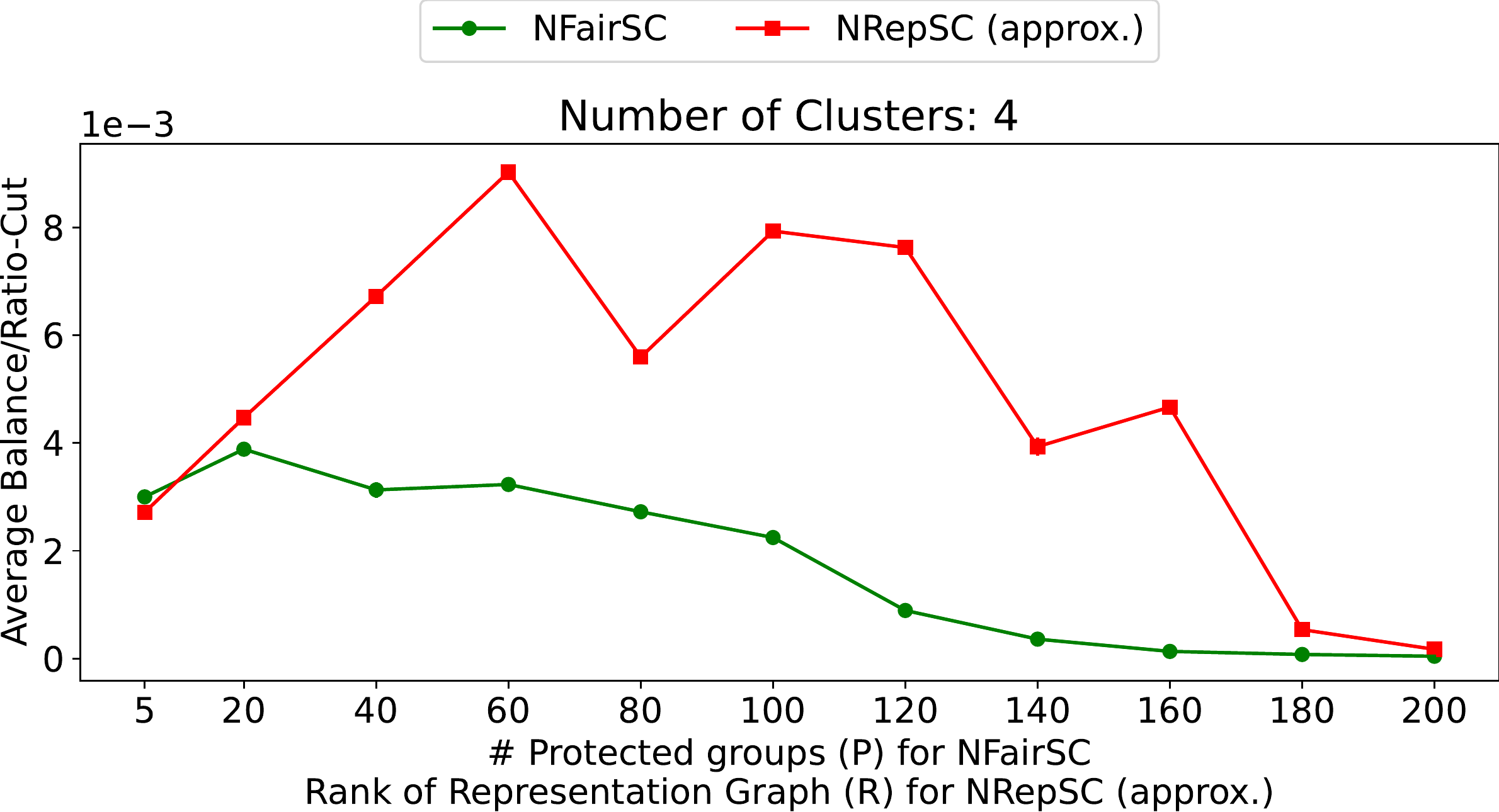}}

    \subfloat[][$K = 6$]{\includegraphics[width=0.48\textwidth]{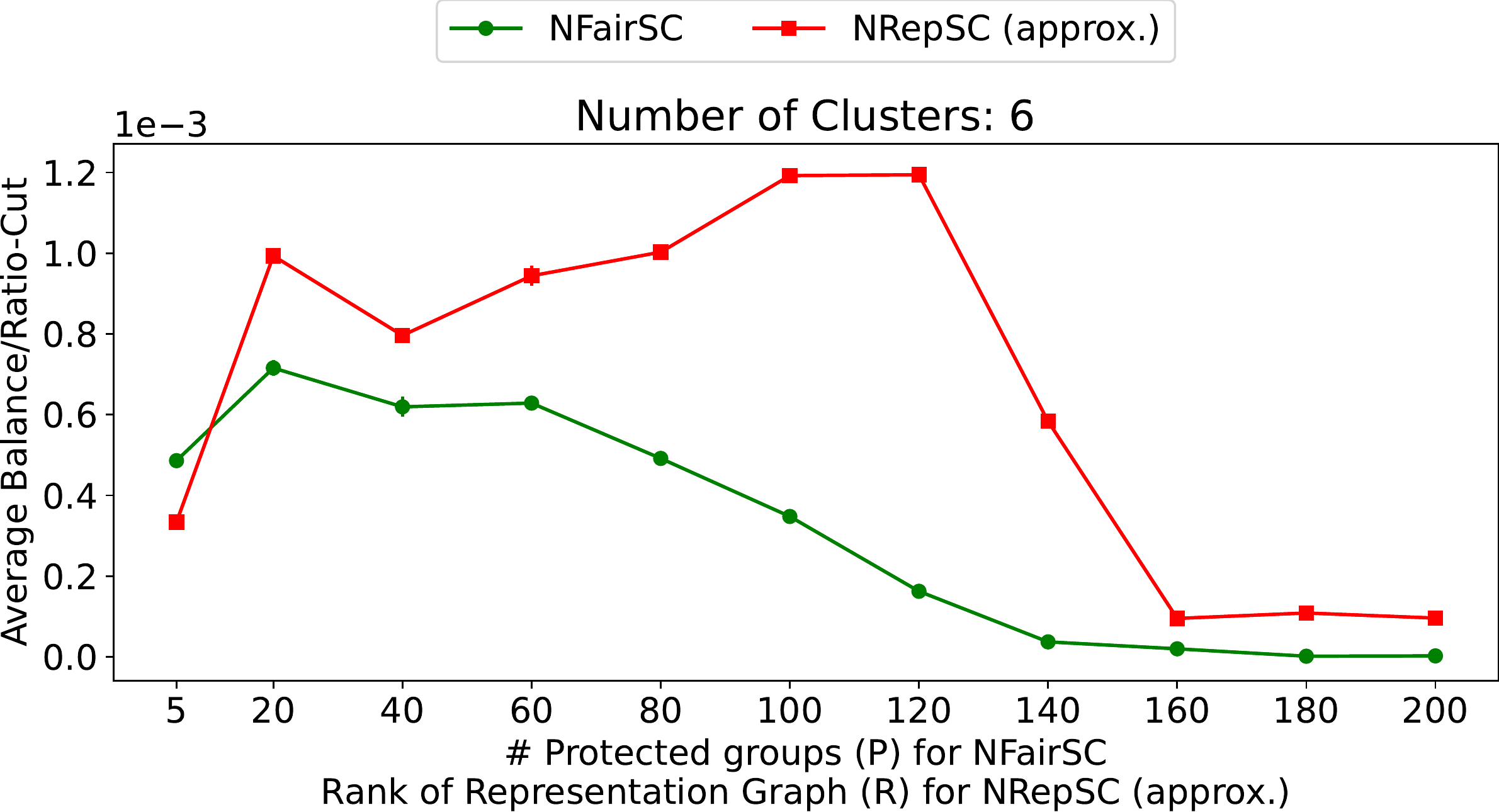}}%
    \hspace{0.5cm}\subfloat[][$K = 8$]{\includegraphics[width=0.48\textwidth]{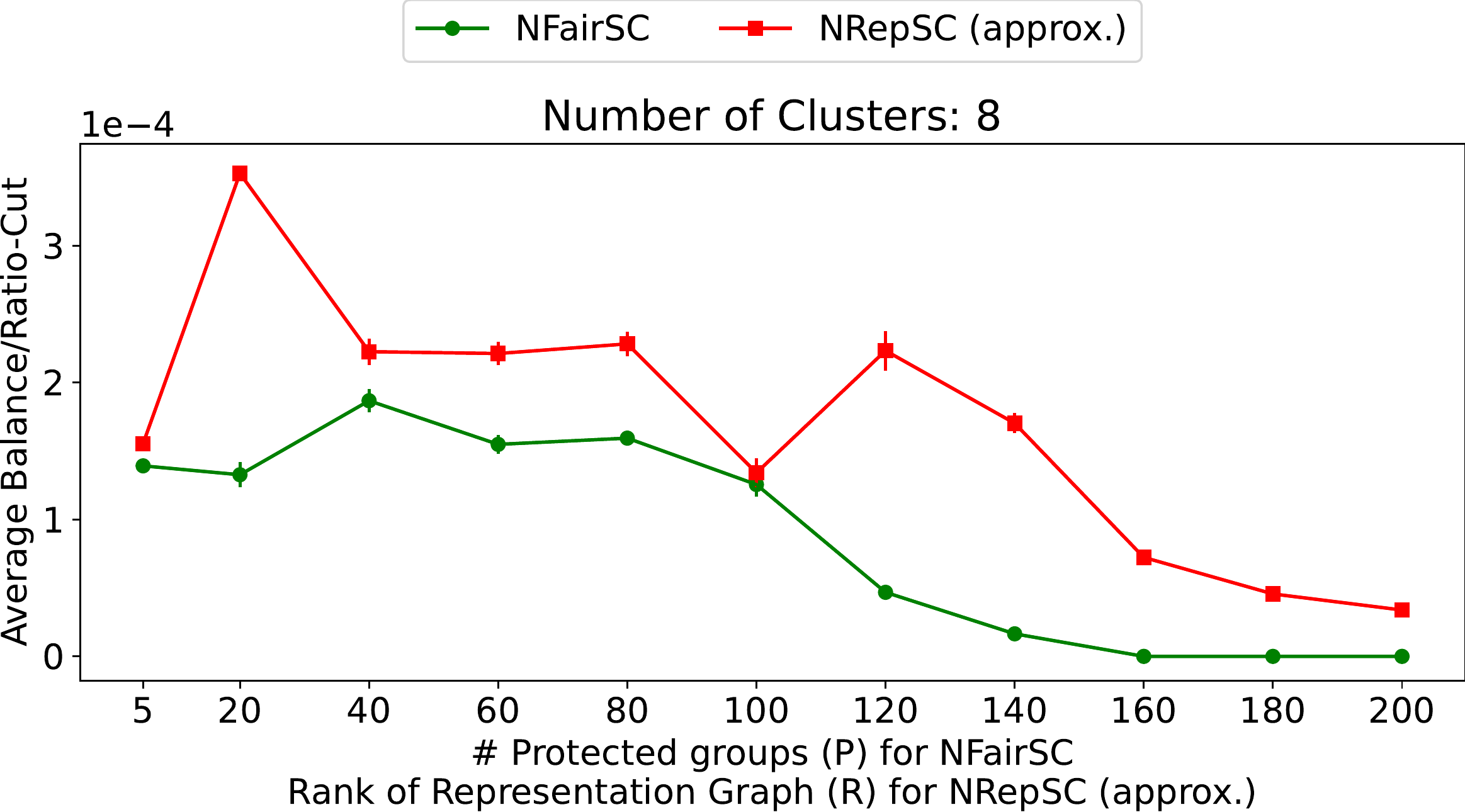}}
    \caption{Comparing \textsc{NRepSC (approx.)} with \textsc{NFairSC} on FAO trade network.}
    \label{fig:real_data_comparison_norm}
\end{figure}

\Cref{fig:d_reg_norm} compares the performance of \textsc{NRepSC} with \textsc{NFairSC} \citep{KleindessnerEtAl:2019:GuaranteesForSpectralClusteringWithFairnessConstraints} and normalized spectral clustering (\textsc{NSC}) on synthetic $d$-regular representation graphs sampled from $\calR$-PP, as described in \Cref{section:numerical_results}. \Cref{fig:sbm_comparison_norm} has the same semantics as \Cref{fig:d_reg_norm}, but uses representation graphs sampled from the traditional planted partition model, as is the case with the second type of experiments in \Cref{section:numerical_results}. Finally, \Cref{fig:real_data_comparison_norm} uses the same FAO trade network as \Cref{section:numerical_results}. All the results follow the same trends as \textsc{URepSC} in \Cref{section:numerical_results}.

\begin{figure}[t]
    \centering
    \subfloat[][Unnormalized case]{\includegraphics[width=0.4\textwidth]{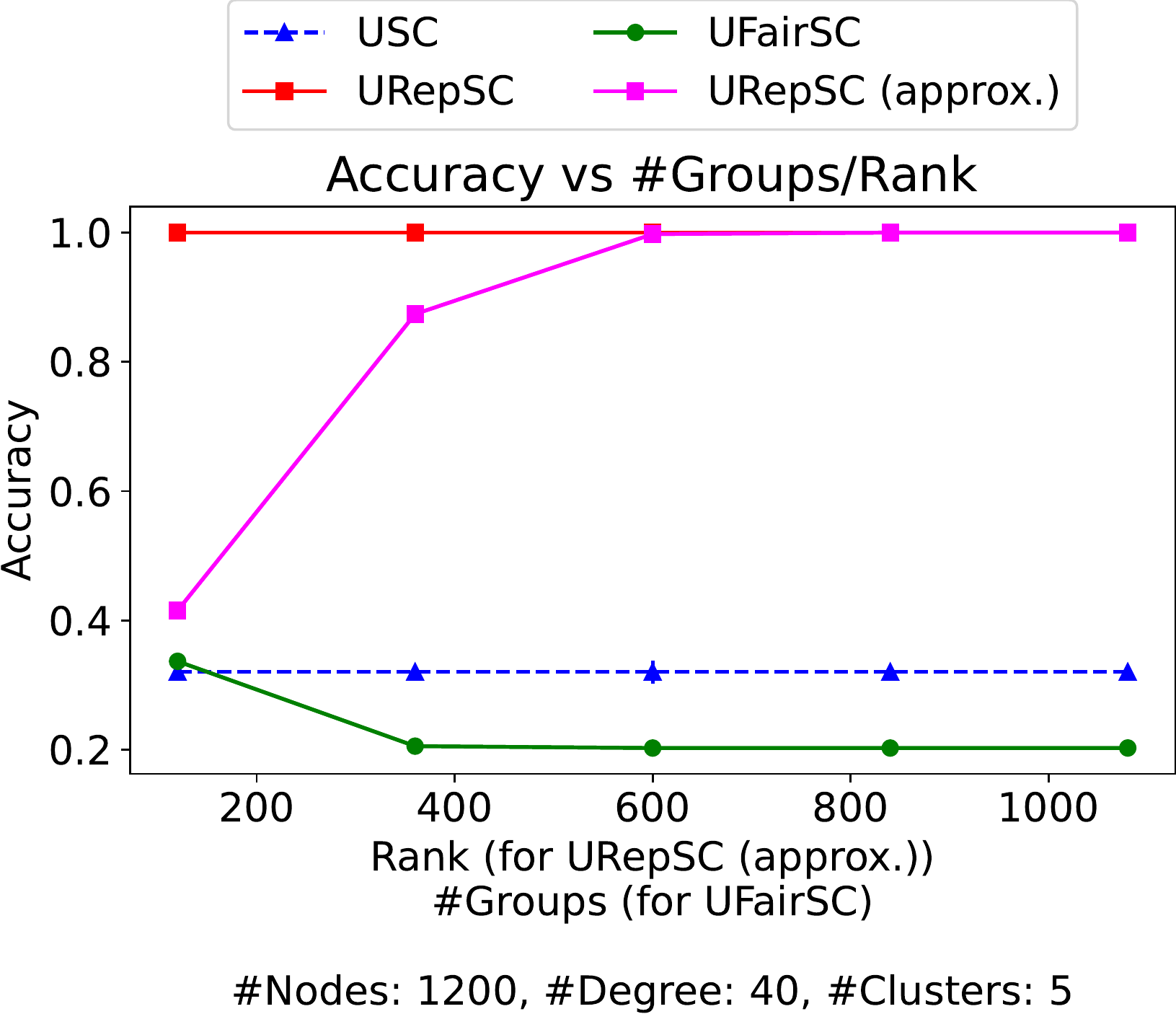}\label{fig:d_reg_unnorm:rank_groups}}%
    \hspace{1cm}\subfloat[][Normalized case]{\includegraphics[width=0.4\textwidth]{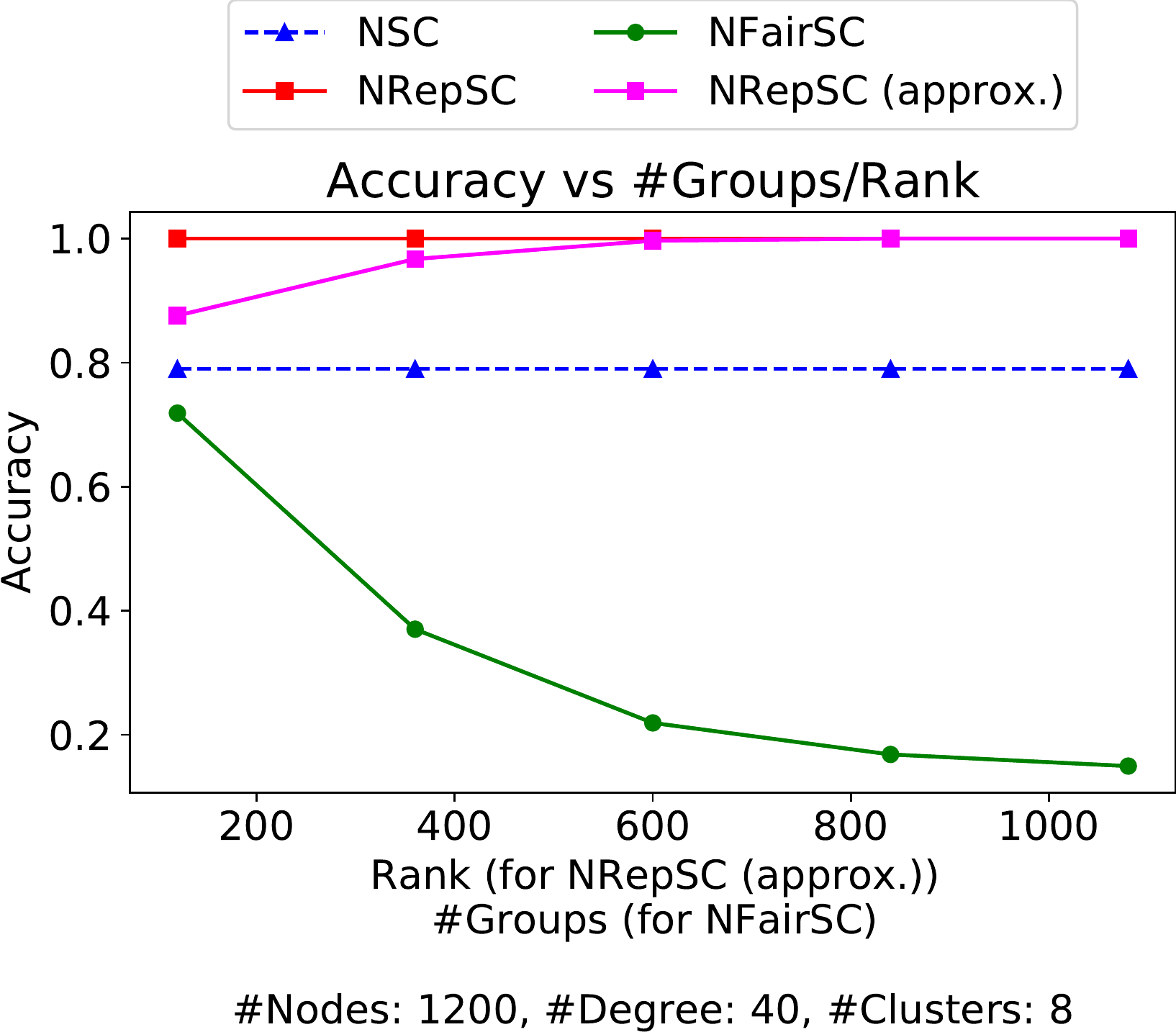}\label{fig:d_reg_norm:rank_groups}}%
    \caption{Accuracy vs the values of $P$ and $R$ used by \textsc{U/NFairSC} and \textsc{U/NRepSC}, respectively, for $d$-regular representation graphs.}
\end{figure}

One may be tempted to think that \textsc{UFairSC} and \textsc{NFairSC} may perform well with a more carefully chosen value of $P$, the number of protected groups. However, \Cref{fig:d_reg_unnorm:rank_groups,fig:d_reg_norm:rank_groups} show that this is not true. These figures plot the performance of \textsc{UFairSC} and \textsc{NFairSC} as a function of the number of protected groups $P$. Also shown is the performance of the approximate variants of our algorithms for various values of rank $R$. As expected, the accuracy increases with $R$ as the approximation of $\bfR$ becomes better but no similar gains are observed for \textsc{UFairSC} and \textsc{NFairSC} for various values of $P$.

\rebuttal{\subsection{Experiments with the Air-Transportation Network}}
\label{appendix:experiments_air_transportation}

\rebuttal{This section demonstrates the performance of \textsc{URepSC (approx.)} and \textsc{NRepSC (approx.)} on another real-world network called the Air-Transportation Network \citep{CardilloEtAl:2013:EmergenceOfNetworkFeaturesFromMultiplexity}. In this network, nodes correspond to airports, edges correspond to direct connections, and layers correspond to airlines. We took three designated ``major'' airlines (Air-France, British, and Lufthansa) and constructed a similarity graph by taking the union of the edges in these layers. Similarly, we constructed the representation graph by considering three “lowcost” airlines (Air-Berlin, EasyJet, and RyanAir). Nodes that were isolated in either of the similarity/representation graphs were dropped. The resulting graphs have $106$ nodes.}

\begin{figure}[t]
    \centering
    \subfloat[][$K = 2$]{\includegraphics[width=0.48\textwidth]{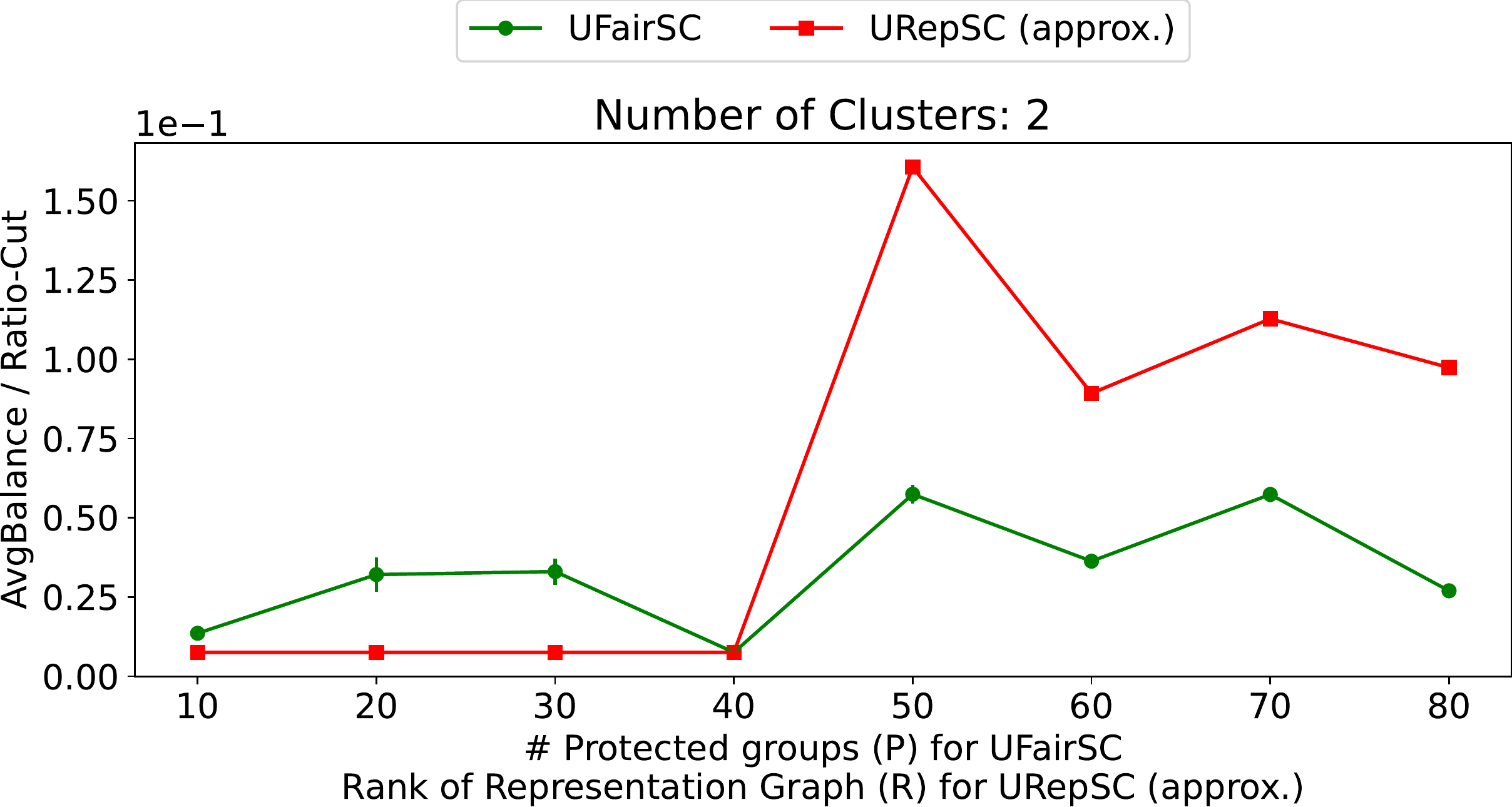}}%
    \hspace{0.5cm}\subfloat[][$K = 4$]{\includegraphics[width=0.48\textwidth]{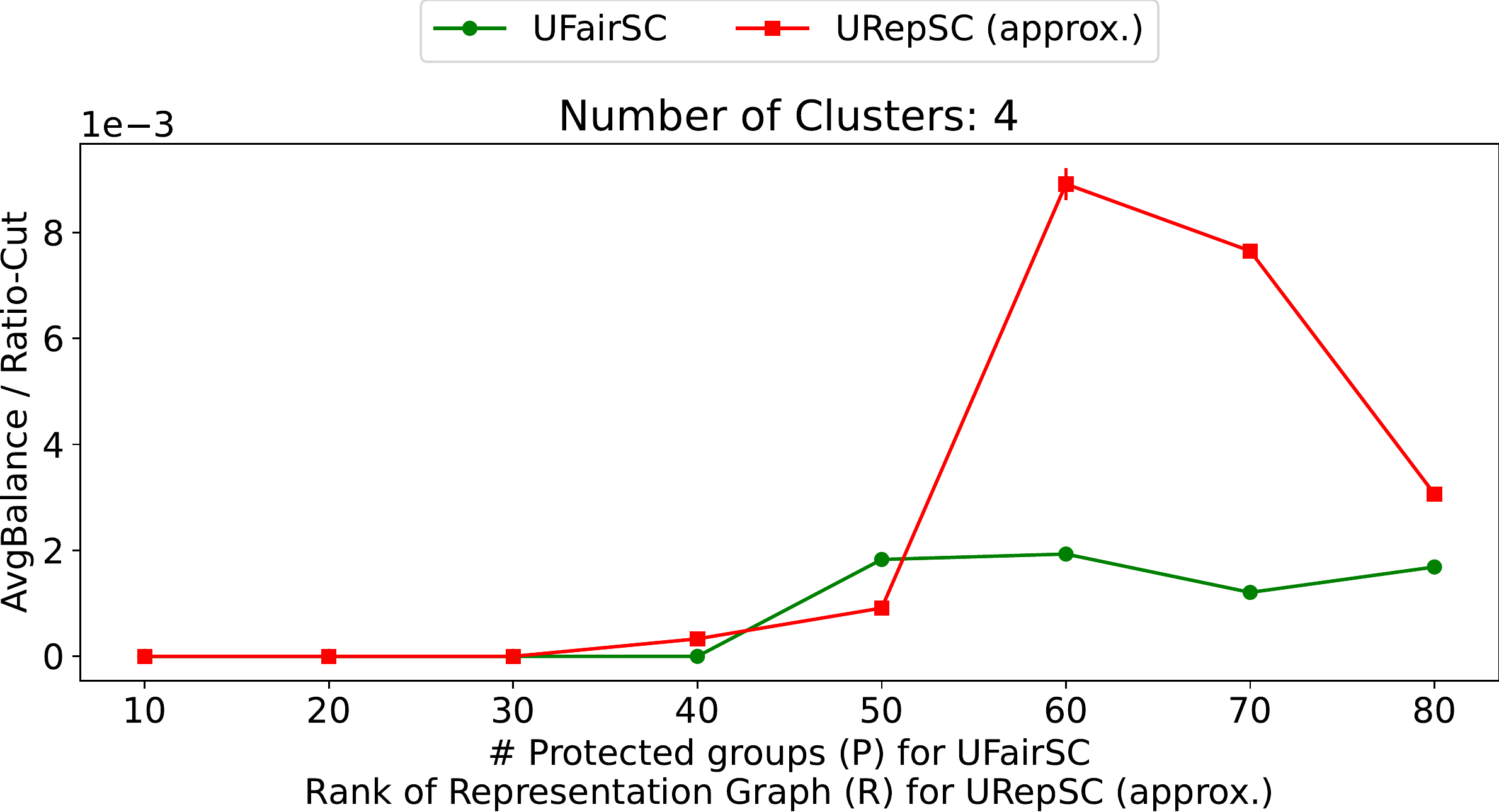}}

    \subfloat[][$K = 6$]{\includegraphics[width=0.48\textwidth]{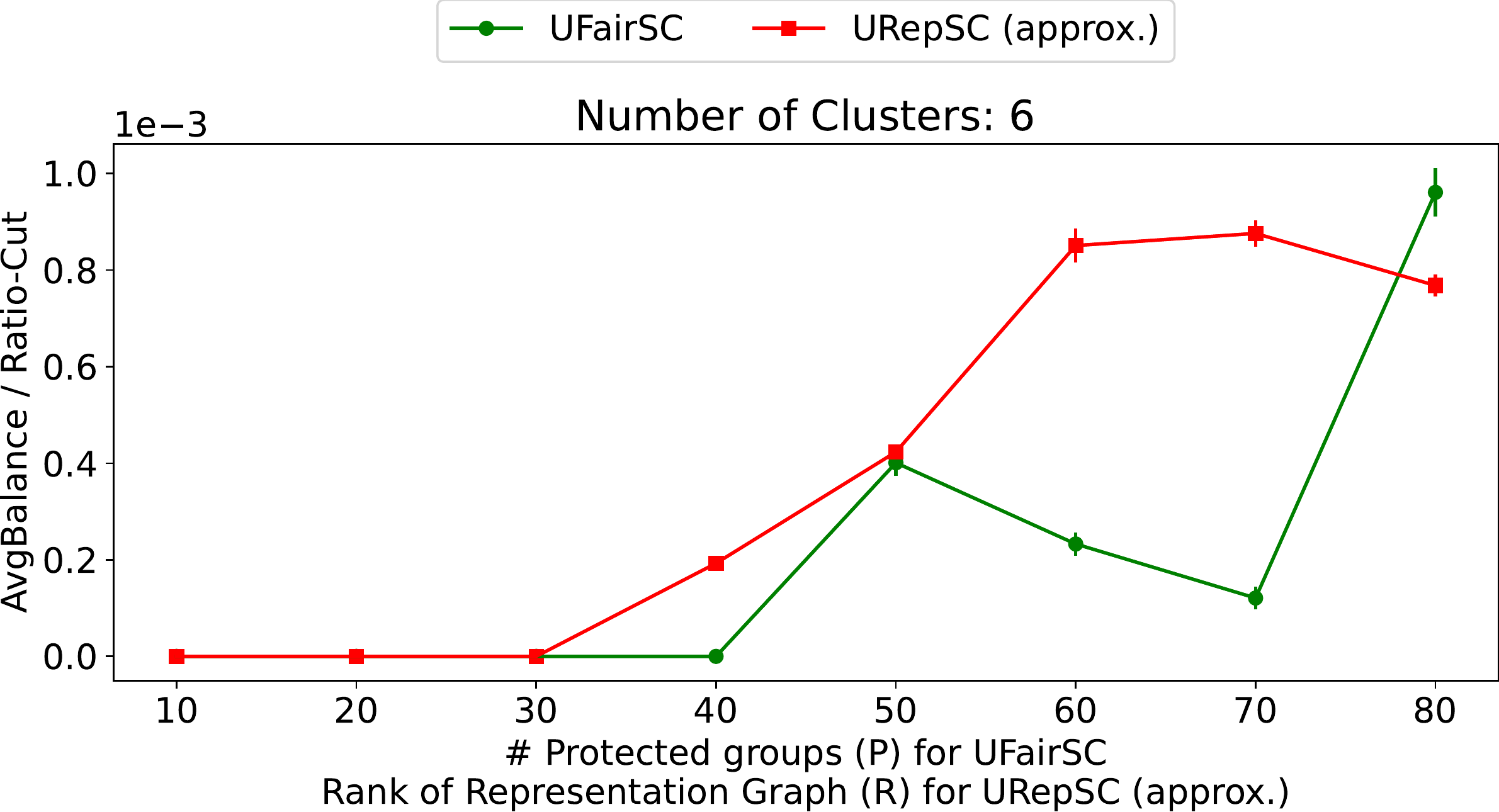}}%
    \hspace{0.5cm}\subfloat[][$K = 8$]{\includegraphics[width=0.48\textwidth]{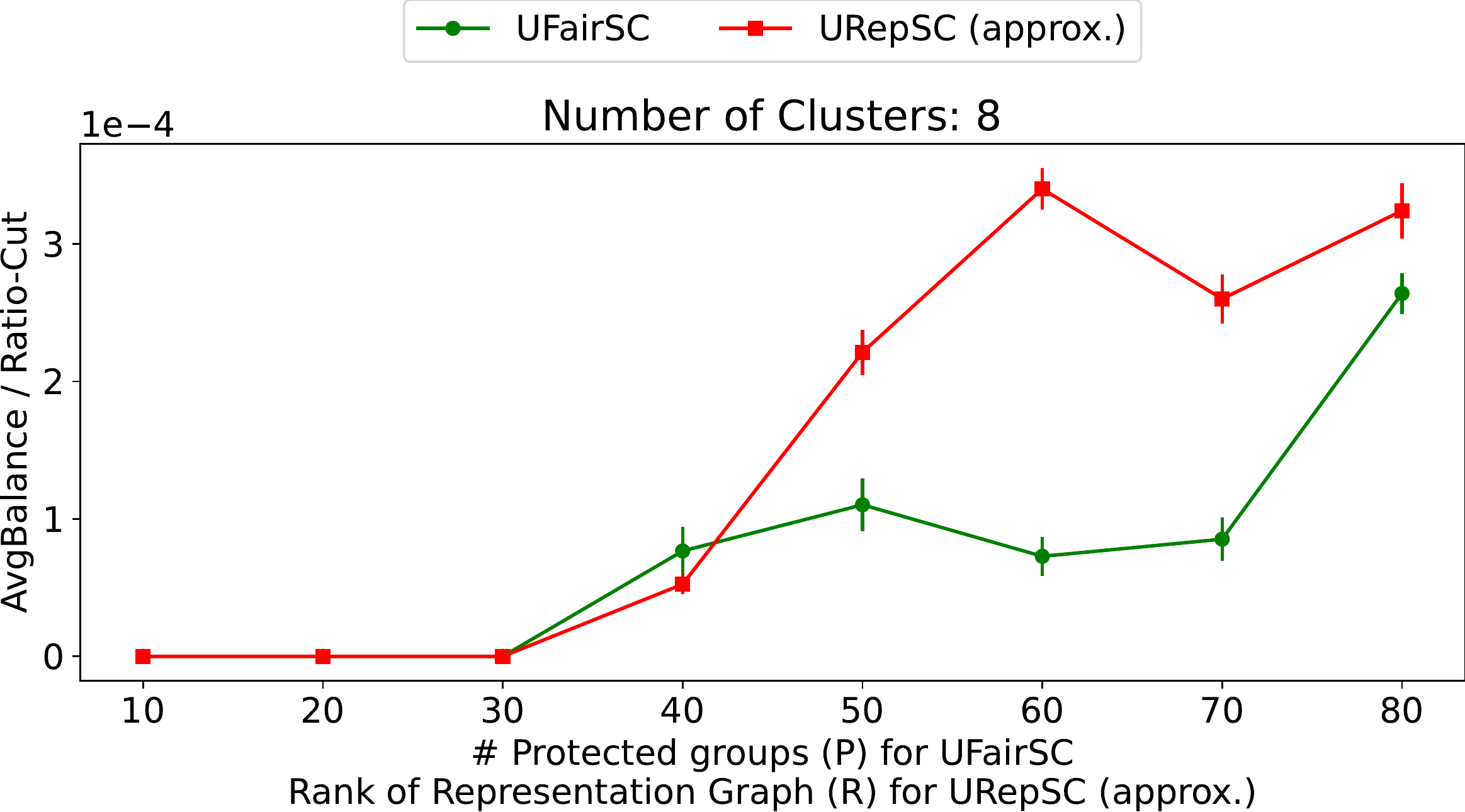}}
    \caption{\rebuttal{Comparing \textsc{URepSC (approx.)} with \textsc{UFairSC} on the air-transportation network.}}
    \label{fig:atn_comparison_unnorm}
\end{figure}

\begin{figure}[t]
    \centering
    \subfloat[][$K = 2$]{\includegraphics[width=0.48\textwidth]{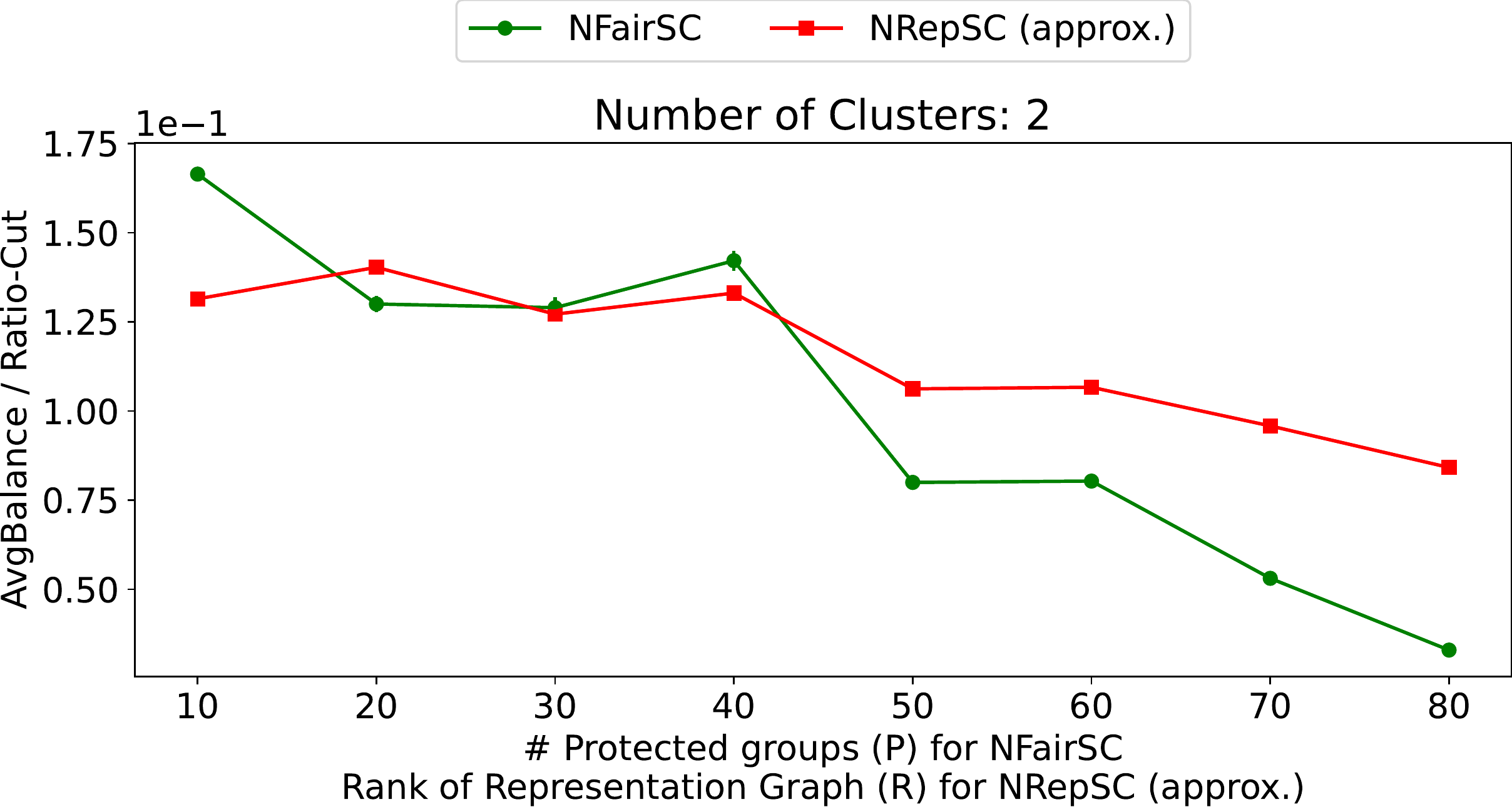}}%
    \hspace{0.5cm}\subfloat[][$K = 4$]{\includegraphics[width=0.48\textwidth]{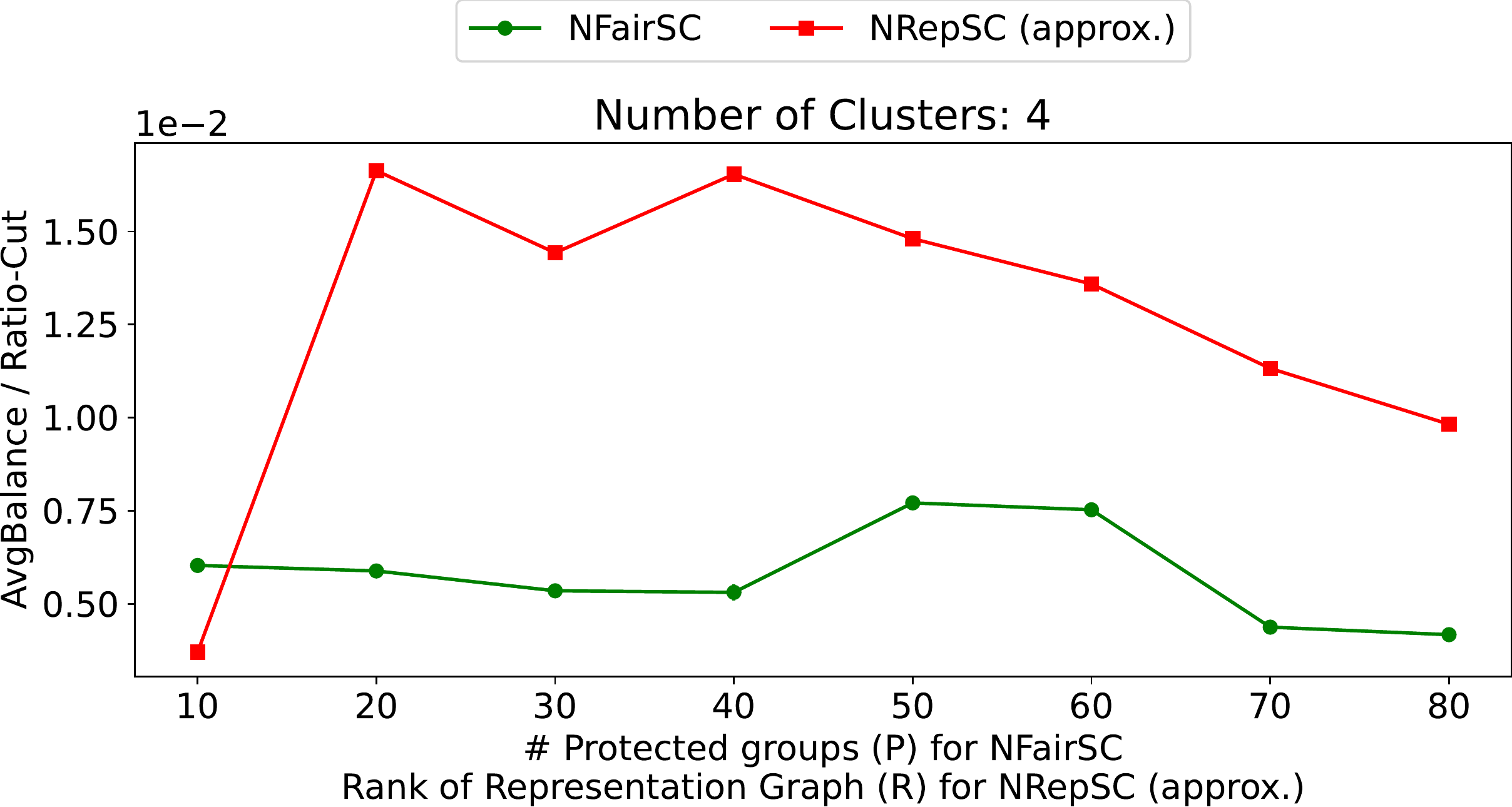}}

    \subfloat[][$K = 6$]{\includegraphics[width=0.48\textwidth]{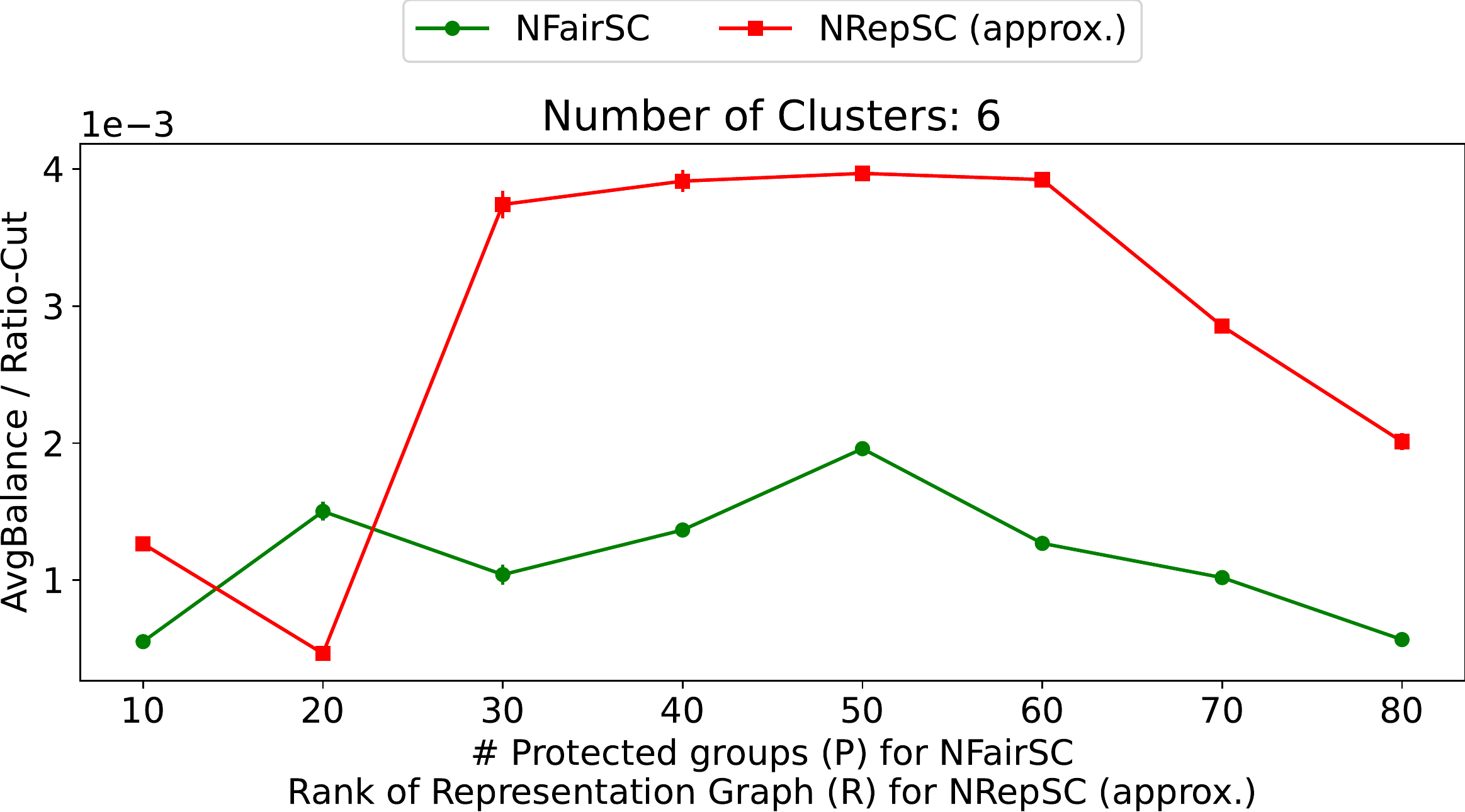}}%
    \hspace{0.5cm}\subfloat[][$K = 8$]{\includegraphics[width=0.48\textwidth]{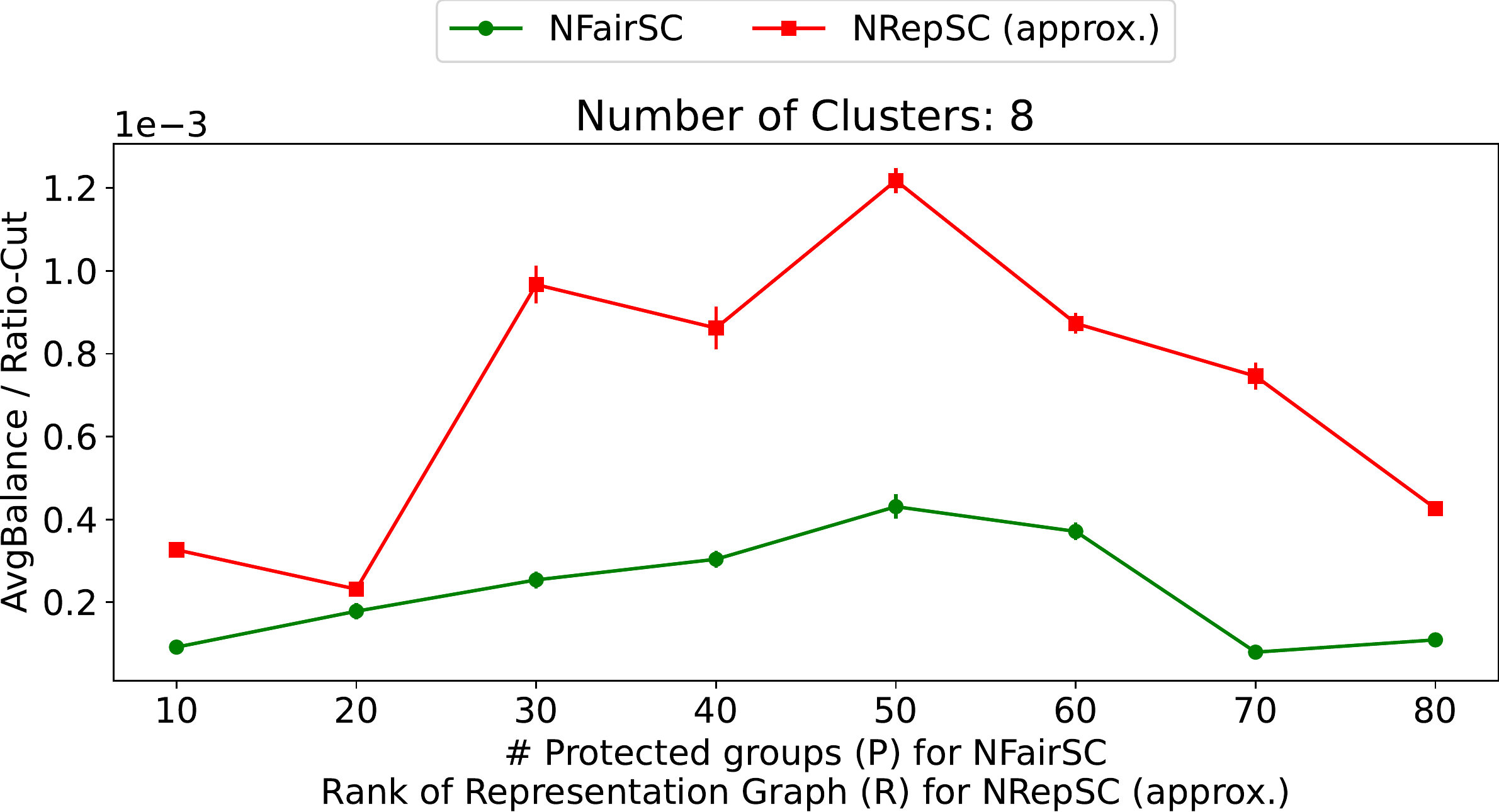}}
    \caption{\rebuttal{Comparing \textsc{NRepSC (approx.)} with \textsc{NFairSC} on the air-transportation network.}}
    \label{fig:atn_comparison_norm}
\end{figure}

\rebuttal{\Cref{fig:atn_comparison_unnorm,fig:atn_comparison_norm} compare the performance of \textsc{URepSC (approx.)} and \textsc{NRepSC (approx.)} on this network with that of \textsc{UFairSC} and \textsc{NFairSC} respectively. The semantics of these plots are identical to the corresponding plots for the FAO trade network (\Cref{fig:real_data_comparison_unnorm,fig:real_data_comparison_norm}). As before, the value of $R$ can be chosen to get a high balance at a competitive ratio-cut.}

\rebuttal{\subsection{A few additional plots for \textsc{URepSC}}}
\label[]{appendix:a_few_additional_plots_for_urepsc}

\begin{figure}[t]
    \centering
    \subfloat[][$N = 1000$, $K = 4$]{\includegraphics[width=0.48\textwidth]{Images/U_AccVsGroupRankSBM_1000_4.pdf}}%
    \hspace{0.5cm}\subfloat[][$N = 3000$, $K = 4$]{\includegraphics[width=0.48\textwidth]{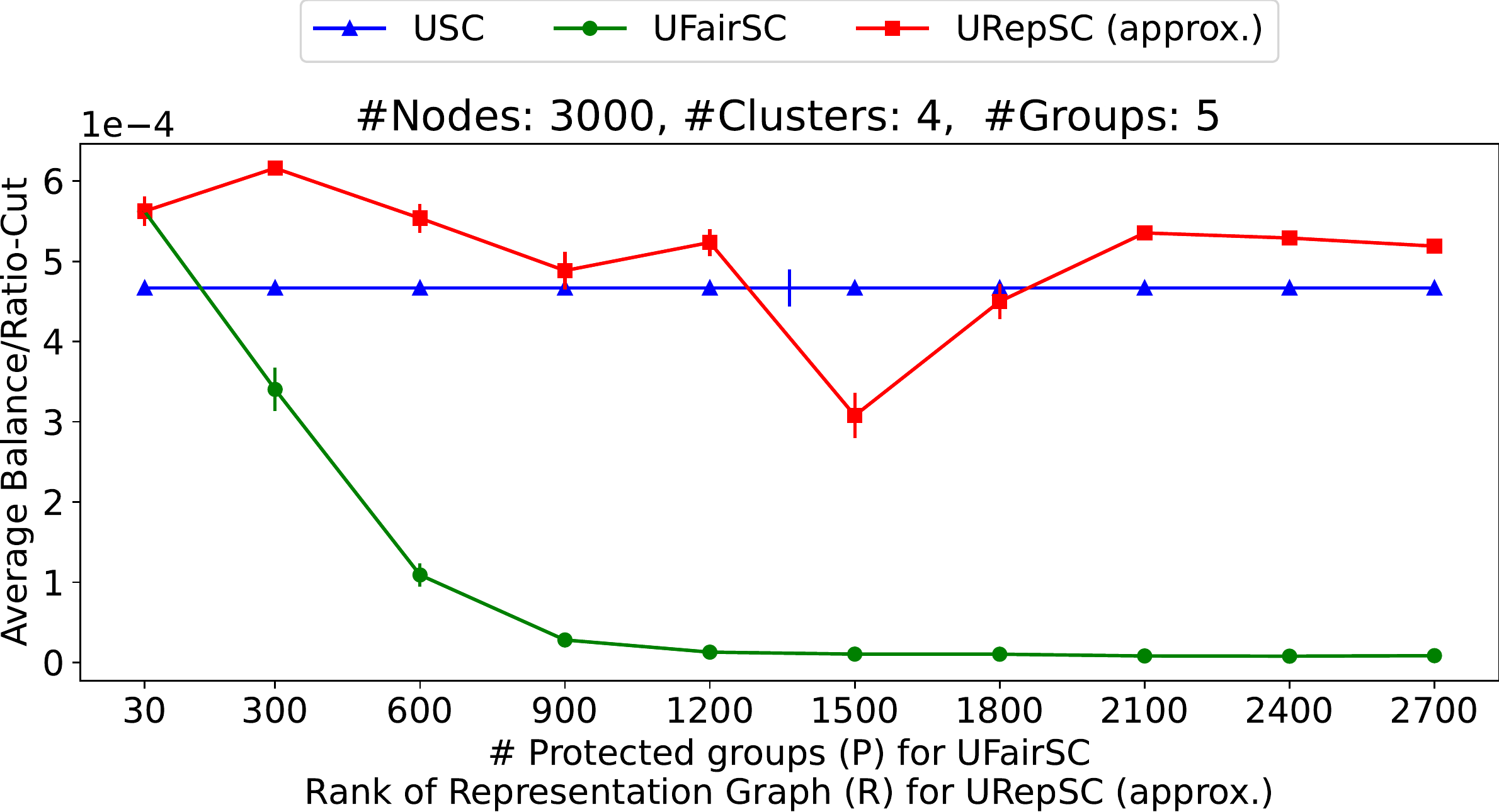}}

    \subfloat[][$N = 1000$, $K = 8$]{\includegraphics[width=0.48\textwidth]{Images/U_AccVsGroupRankSBM_1000_8}}%
    \hspace{0.5cm}\subfloat[][$N = 3000$, $K = 8$]{\includegraphics[width=0.48\textwidth]{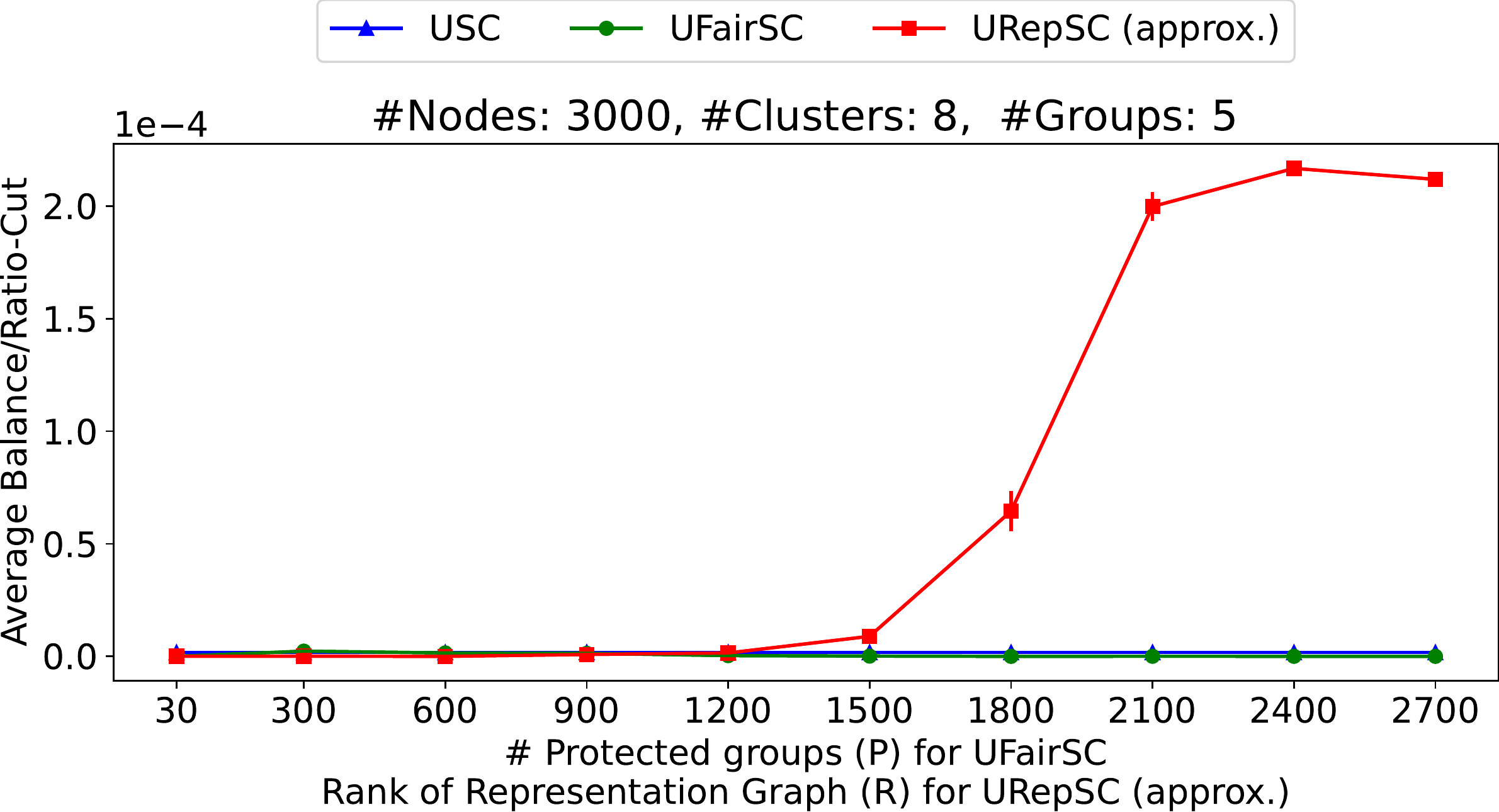}}
    \caption{Comparing \textsc{URepSC (approx.)} with \textsc{UFairSC} using synthetically generated representation graphs sampled from an SBM.}
    \label{fig:sbm_comparison_unnorm_additional}
\end{figure}

\begin{figure}[t]
    \centering
    \subfloat[][$K = 2$]{\includegraphics[width=0.48\textwidth]{Images/U_Trade_2}}%
    \hspace{0.5cm}\subfloat[][$K = 4$]{\includegraphics[width=0.48\textwidth]{Images/U_Trade_4}}

    \subfloat[][$K = 6$]{\includegraphics[width=0.48\textwidth]{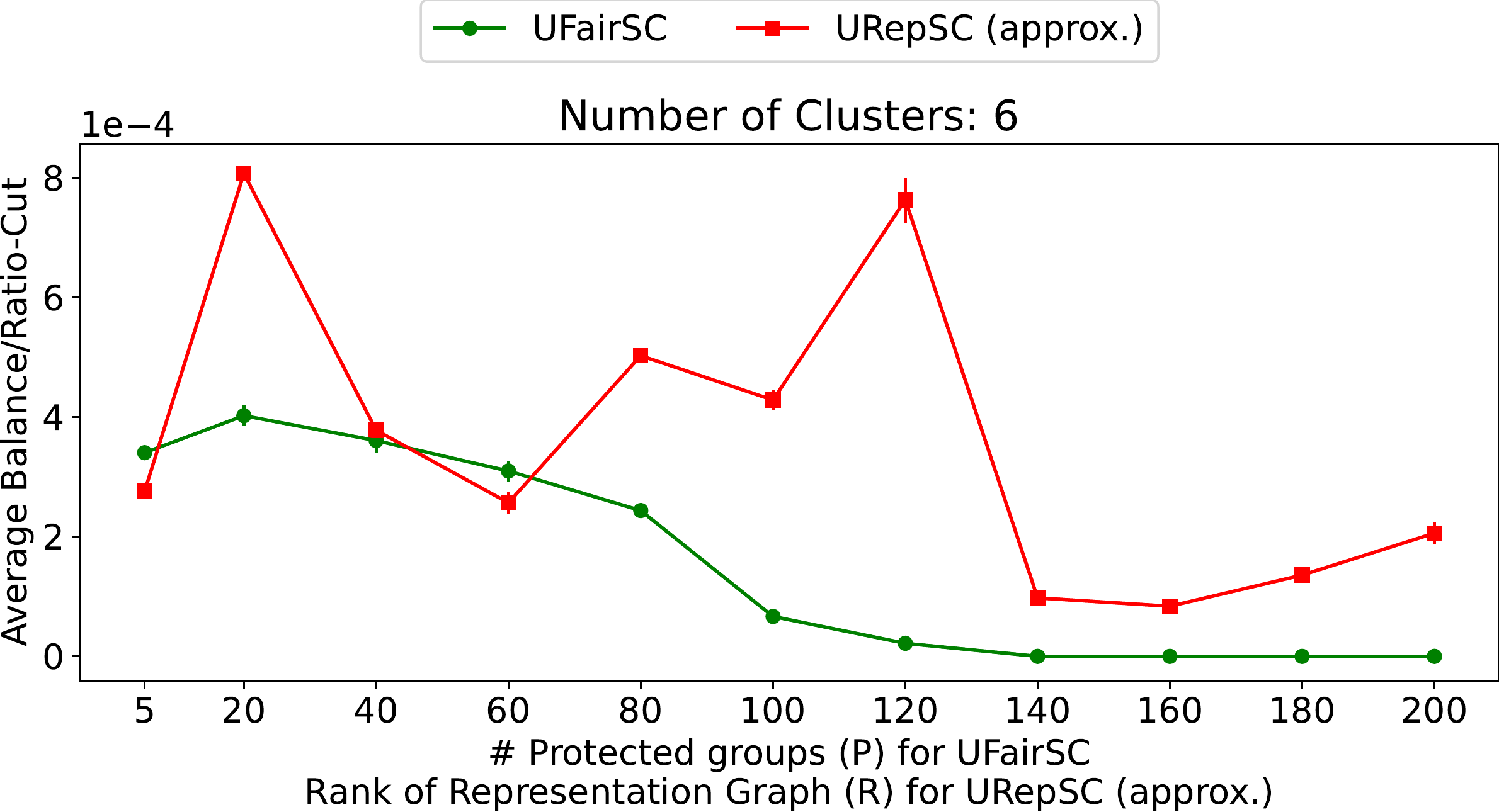}}%
    \hspace{0.5cm}\subfloat[][$K = 8$]{\includegraphics[width=0.48\textwidth]{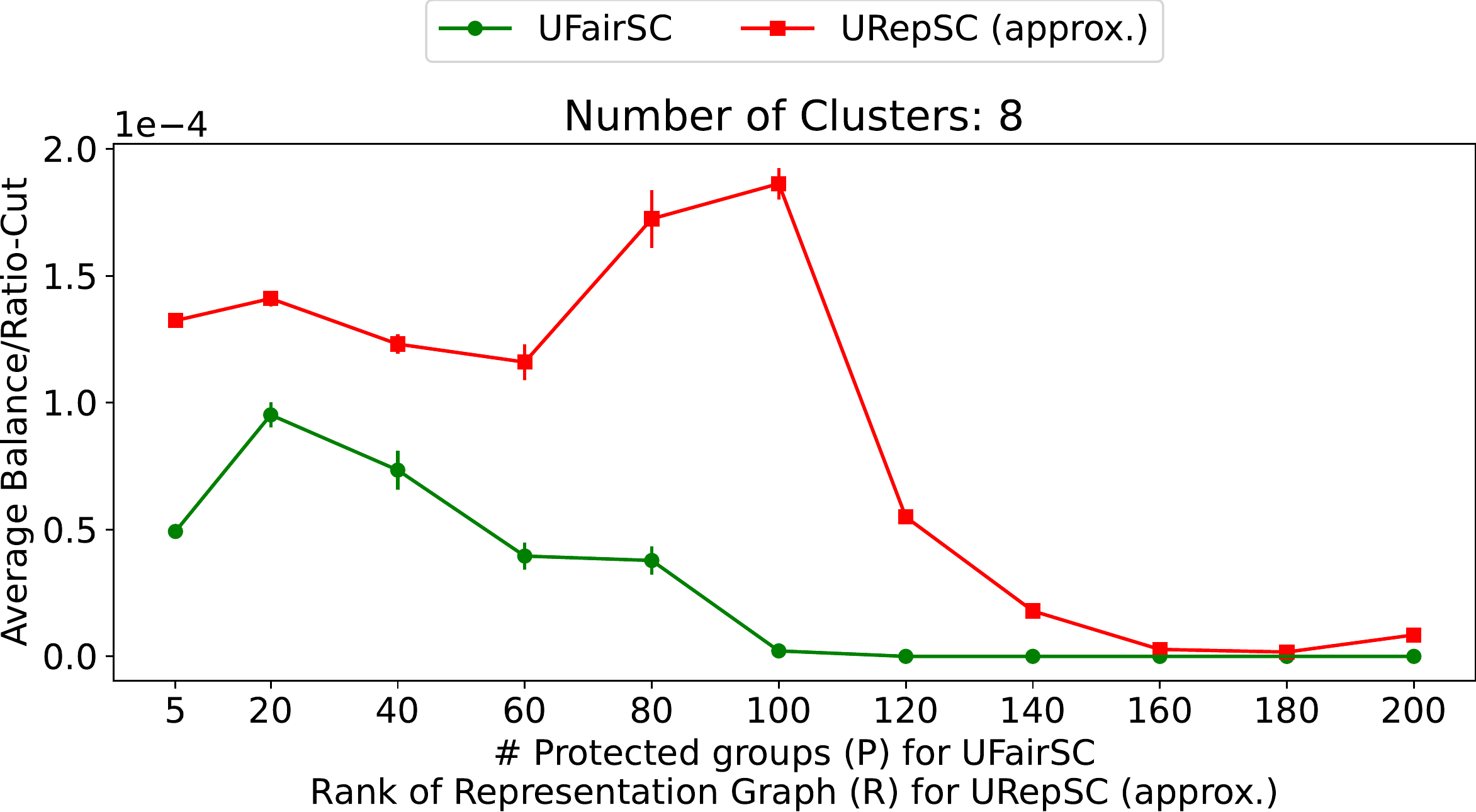}}
    \caption{Comparing \textsc{URepSC (approx.)} with \textsc{UFairSC} on FAO trade network.}
    \label{fig:real_data_comparison_unnorm_additional}
\end{figure}

\Cref{fig:sbm_comparison_unnorm_additional,fig:real_data_comparison_unnorm_additional} provide a few more configurations of $N$ and $K$ pairs for \textsc{URepSC} and have the same semantics as \Cref{fig:sbm_comparison_unnorm,fig:real_data_comparison_unnorm} respectively.

\rebuttal{\subsection{Numerical validation of time complexity}}
\label{appendix:numerical_validation_of_time_complexity}

\begin{figure}[t]
    \centering
    \includegraphics[width=0.5\textwidth]{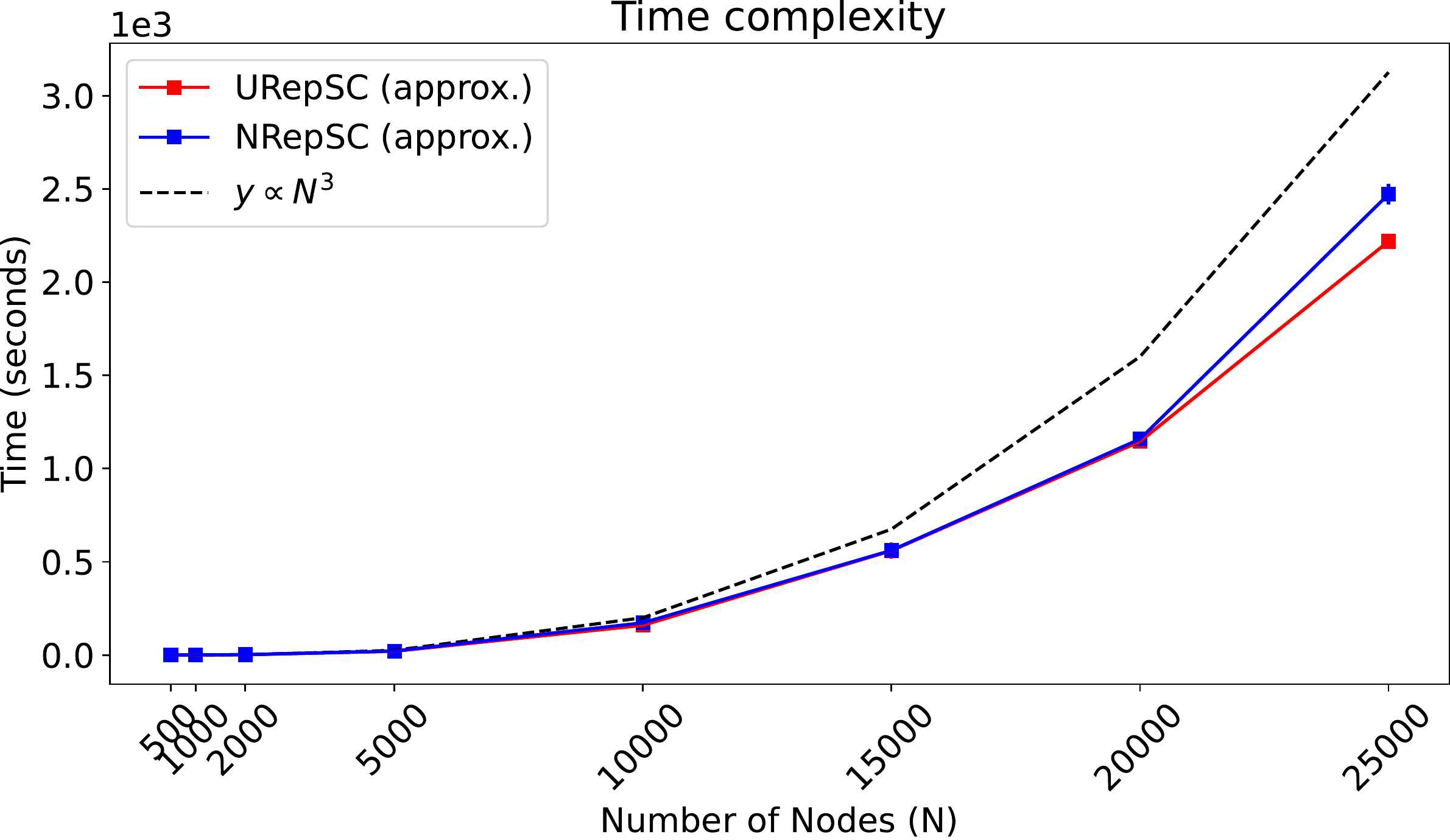}
    \caption{\rebuttal{Time taken by \textsc{U/NRepSC (approx.)} as a function of the number of nodes in the graph.}}
    \label{fig:time_complexity}
\end{figure}

\rebuttal{We close this section with a numerical validation of the time complexity of our algorithms. Recall from \Cref{remark:computational_complexity} that \textsc{URepSC} has a time complexity of $O(N^3)$. A similar analysis holds for \textsc{NRepSC} as well. \textsc{URepSC (approx.)} and \textsc{NRepSC (approx.)} involve an additional low-rank approximation step, but still have $O(N^3)$ time complexity. \Cref{fig:time_complexity} plots the time in seconds taken by \textsc{URepSC (approx.)} and \textsc{NRepSC (approx.)} as a function of the number of nodes in the graph. We used representation graphs sampled from a planted partition model for these experiments with $K=4$, $R=0.5N$, and the remaining configurations same as in \Cref{section:numerical_results}. The dotted line indicates the $O(N^3)$ growth.}

%% file: appendix_ratio-cut_balance_separated.tex
\section{\rebuttal{Plots with ratio-cut and average balance separated out}}
\label{appendix:separated_plots}

\rebuttal{Up to this point, the plots either show accuracy (for $d$-regular graphs) or the ratio between average balance and ratio-cut (for planted partition based representation graphs and real-world networks) on $y$-axis as these quantities adequately convey the idea that our algorithms produce high-quality fair clusters. We now show the corresponding plots for each case with average balance and ratio-cut separated out.}

\begin{figure}[t]
    \centering
    \subfloat[][Average balance vs no. of nodes]{\includegraphics[width=0.45\textwidth]{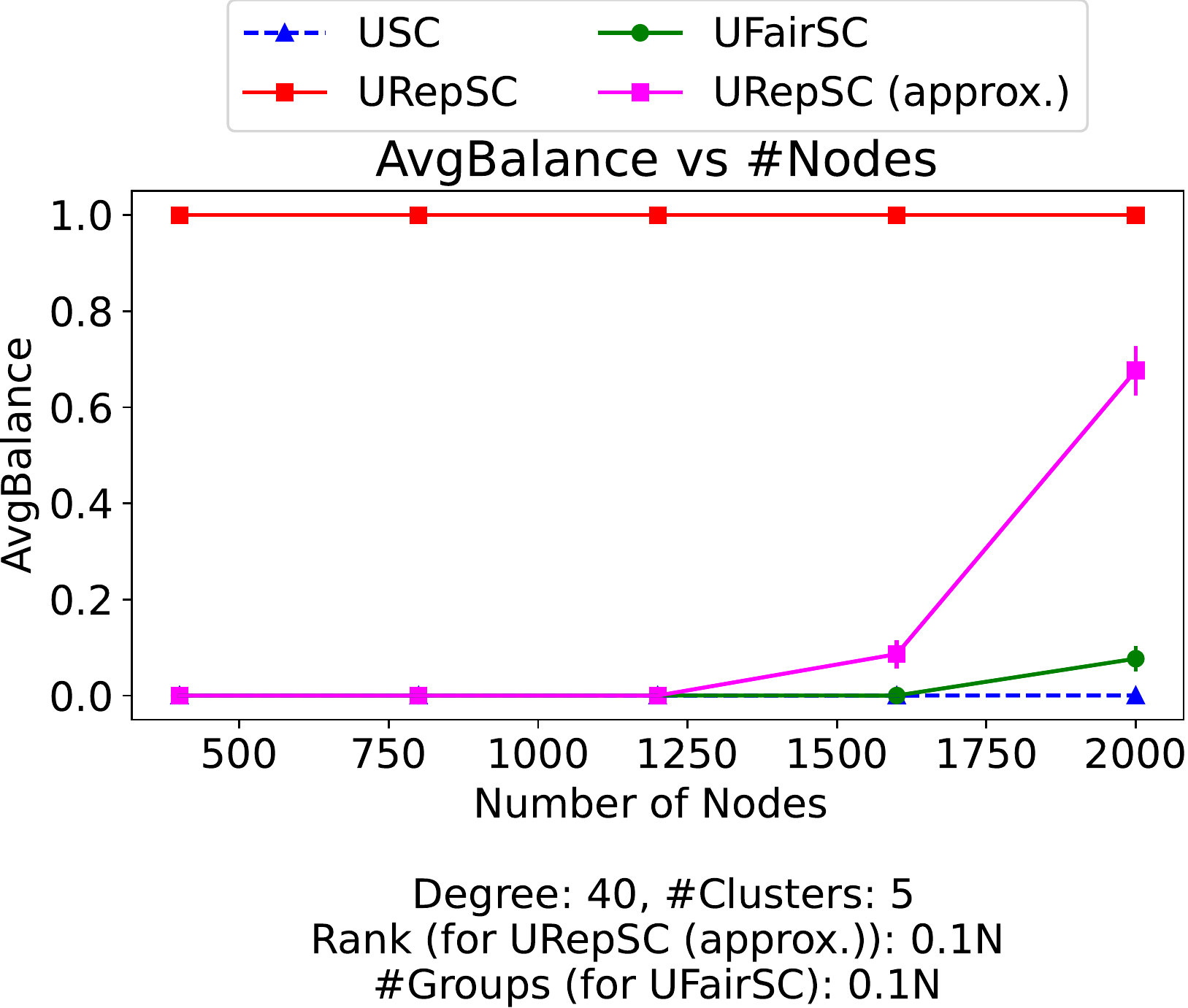}}%
    \hspace{1cm}
    \subfloat[][Ratio-cut vs no. of nodes]{\includegraphics[width=0.45\textwidth]{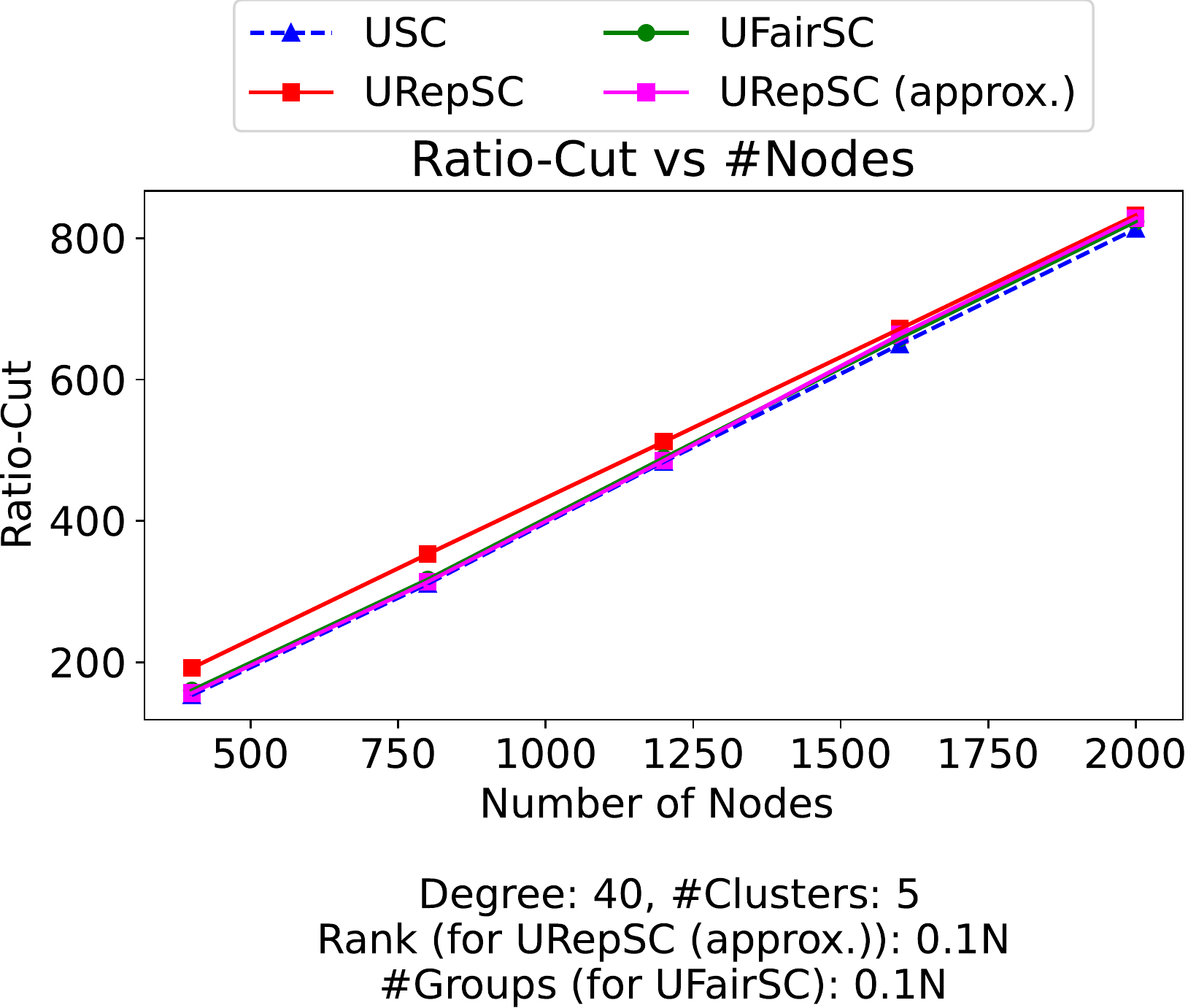}}

    \subfloat[][Average balance vs no. of clusters]{\includegraphics[width=0.45\textwidth]{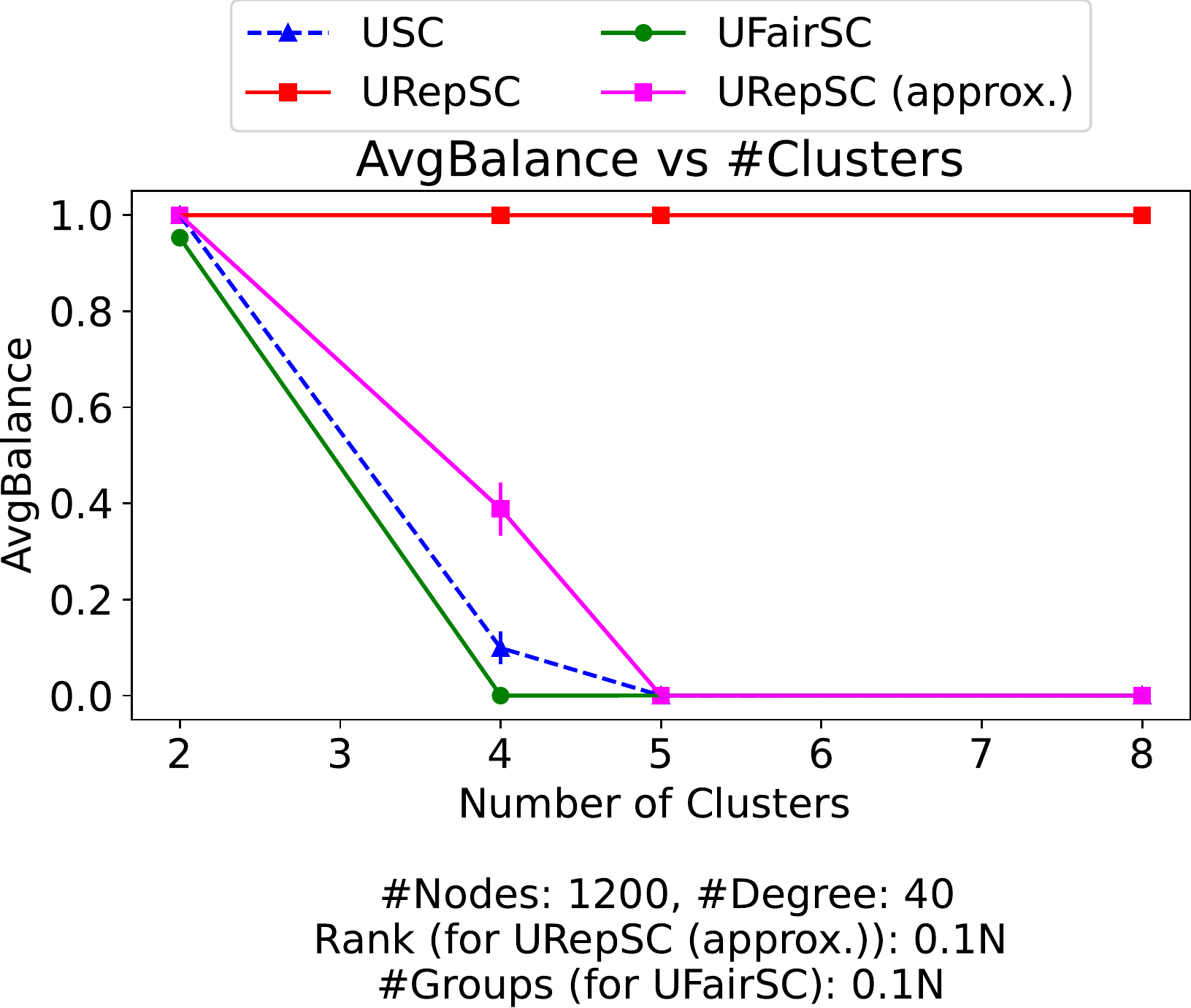}}%
    \hspace{1cm}
    \subfloat[][Ratio-cut vs no. of clusters]{\includegraphics[width=0.45\textwidth]{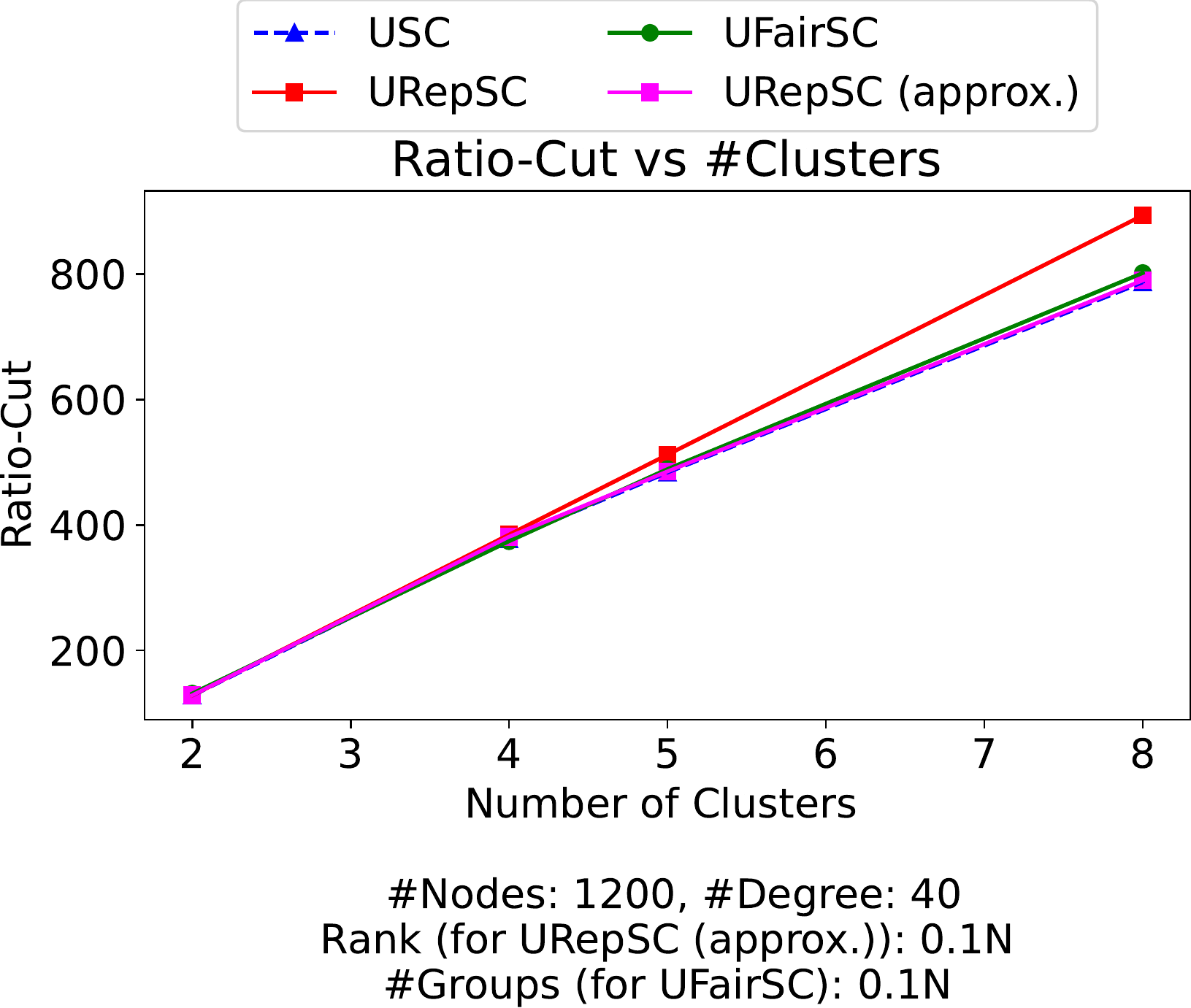}}

    \subfloat[][Average balance vs degree of $\calR$]{\includegraphics[width=0.45\textwidth]{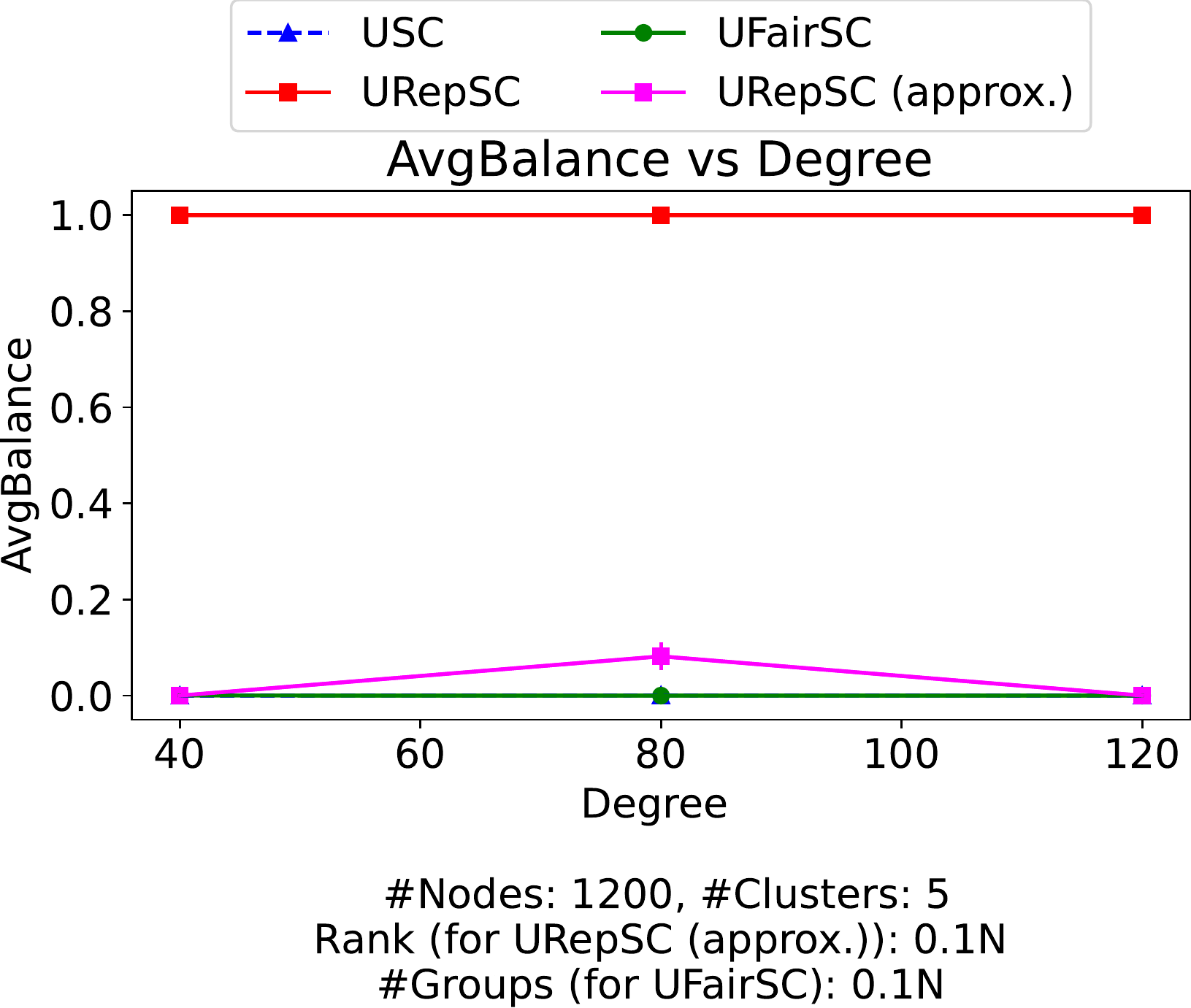}}%
    \hspace{1cm}
    \subfloat[][Ratio-cut vs degree of $\calR$]{\includegraphics[width=0.45\textwidth]{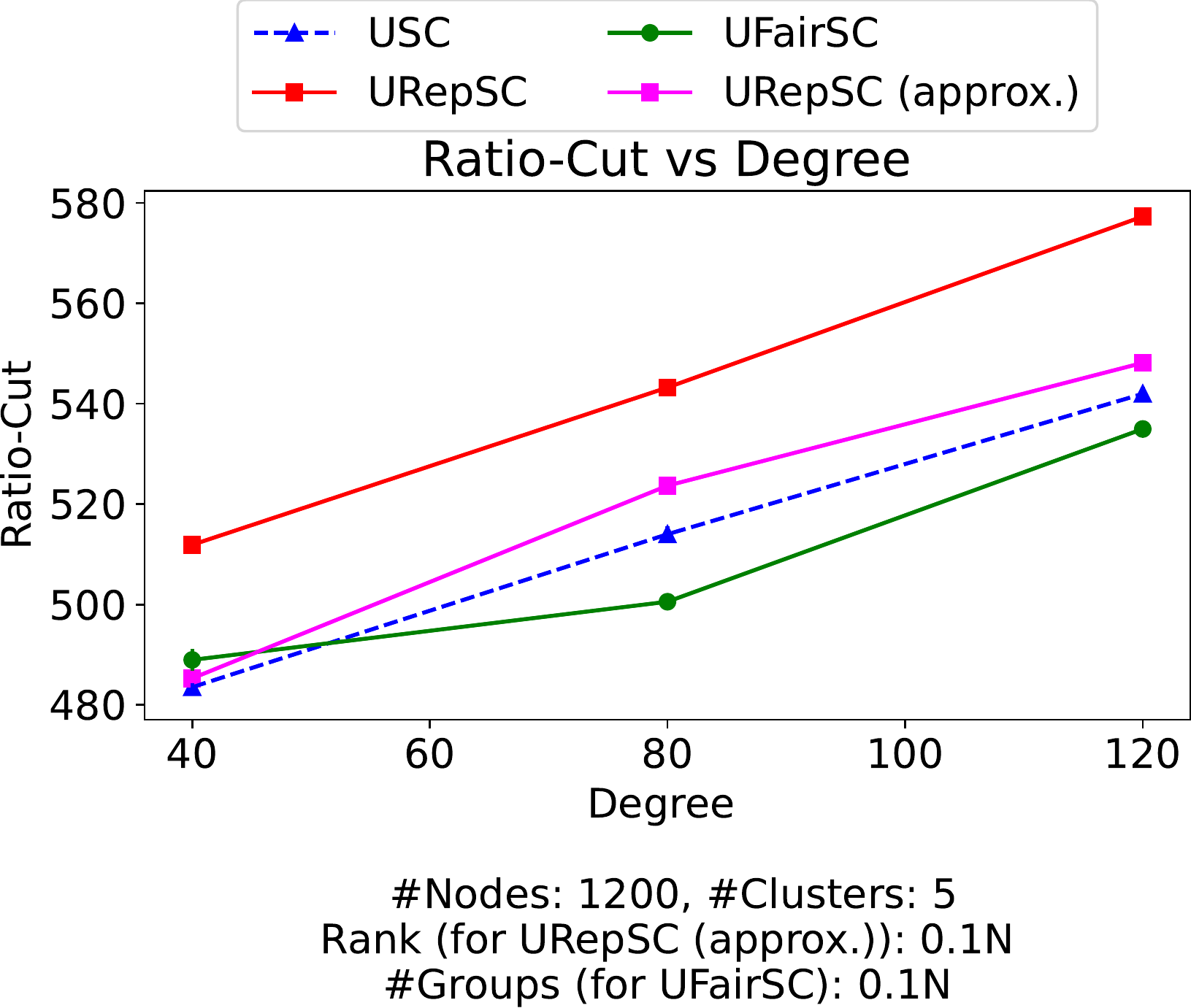}}

    \caption{\rebuttal{Comparing \textsc{URepSC} with other ``unnormalized'' algorithms using synthetically generated $d$-regular representation graphs.}}
    \label{fig:d_reg_unnorm_separated}
\end{figure}

\begin{figure}[t]
    \centering
    \subfloat[][Average balance vs no. of nodes]{\includegraphics[width=0.45\textwidth]{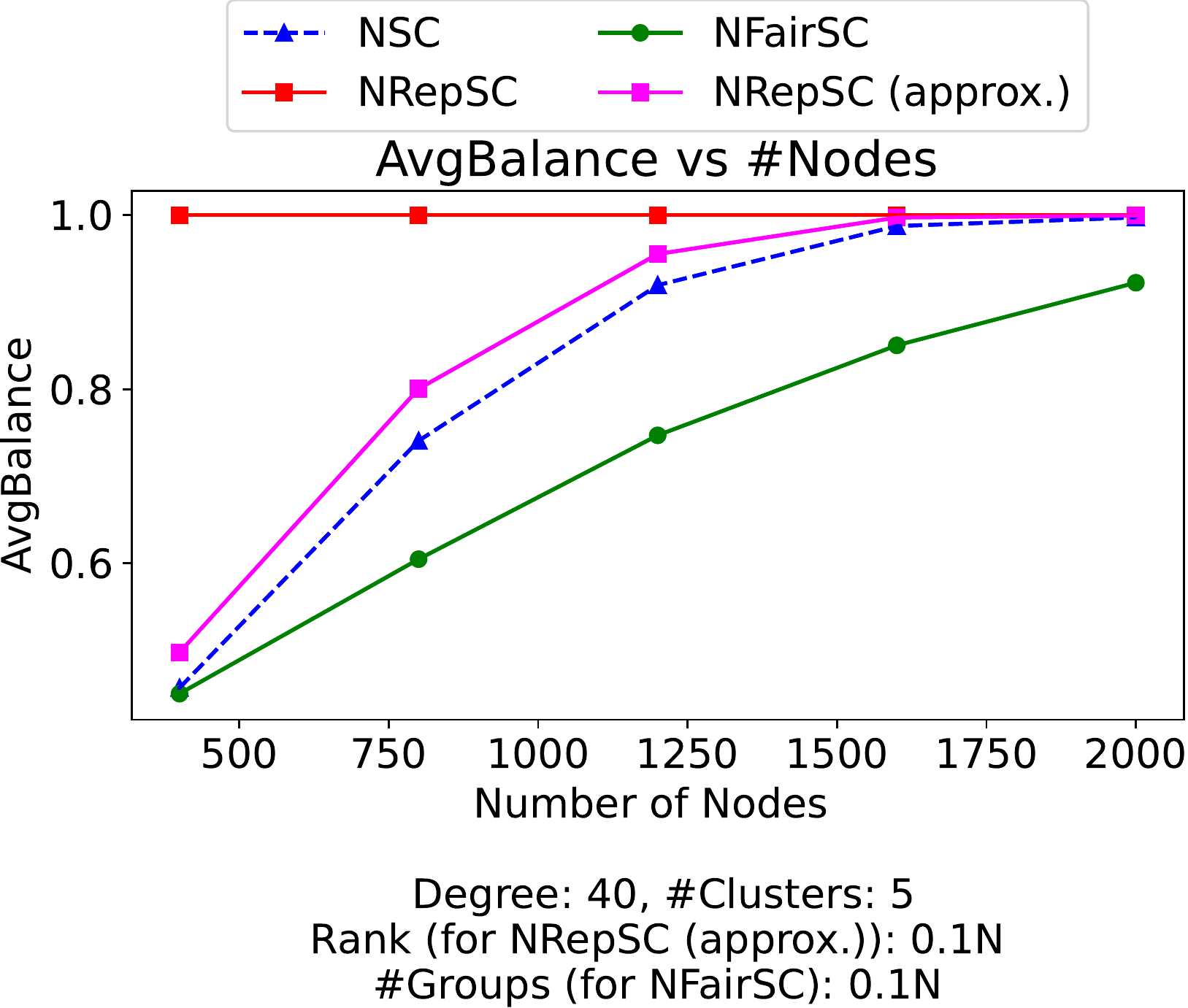}}%
    \hspace{1cm}
    \subfloat[][Ratio-cut vs no. of nodes]{\includegraphics[width=0.45\textwidth]{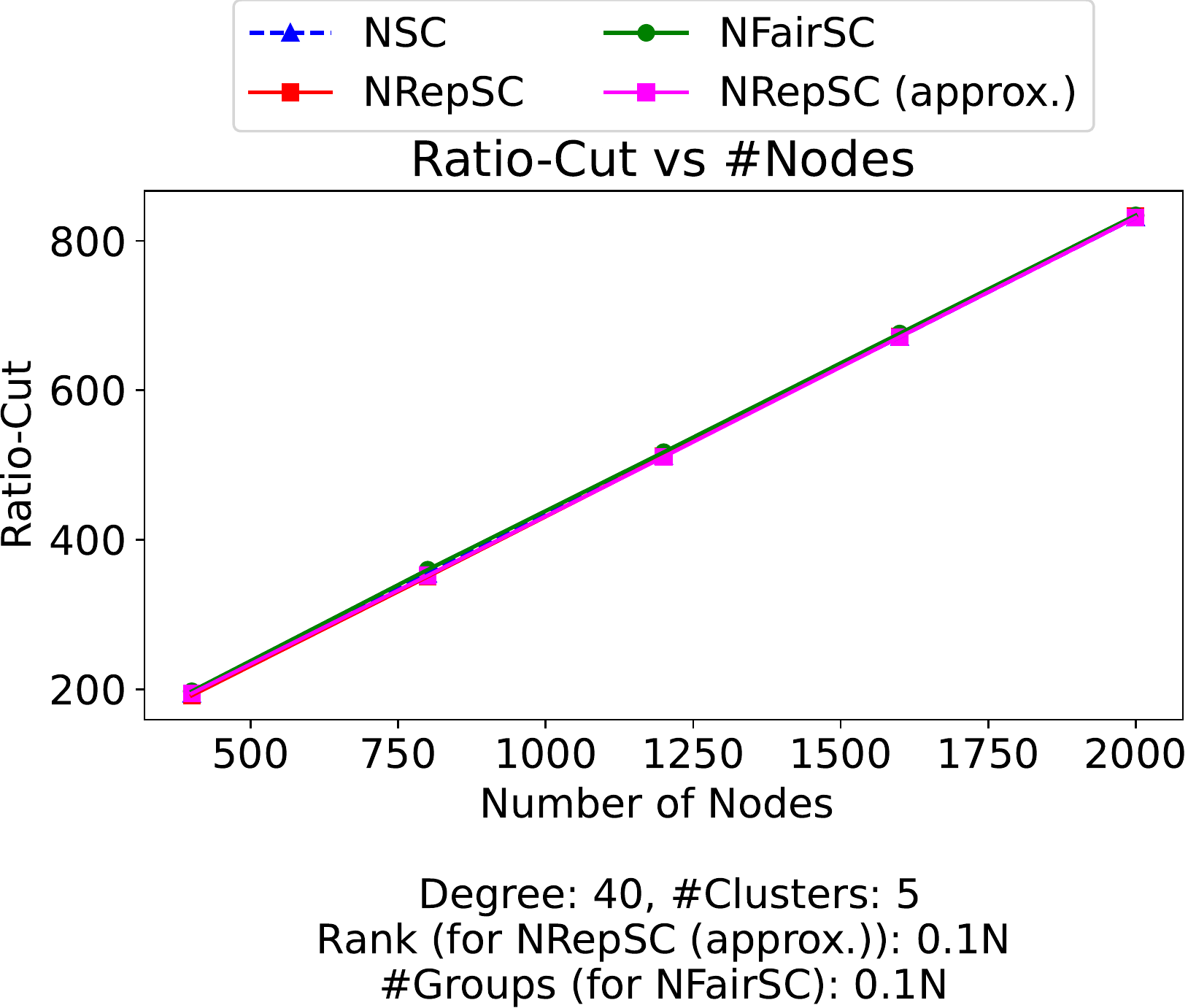}}

    \subfloat[][Average balance vs no. of clusters]{\includegraphics[width=0.45\textwidth]{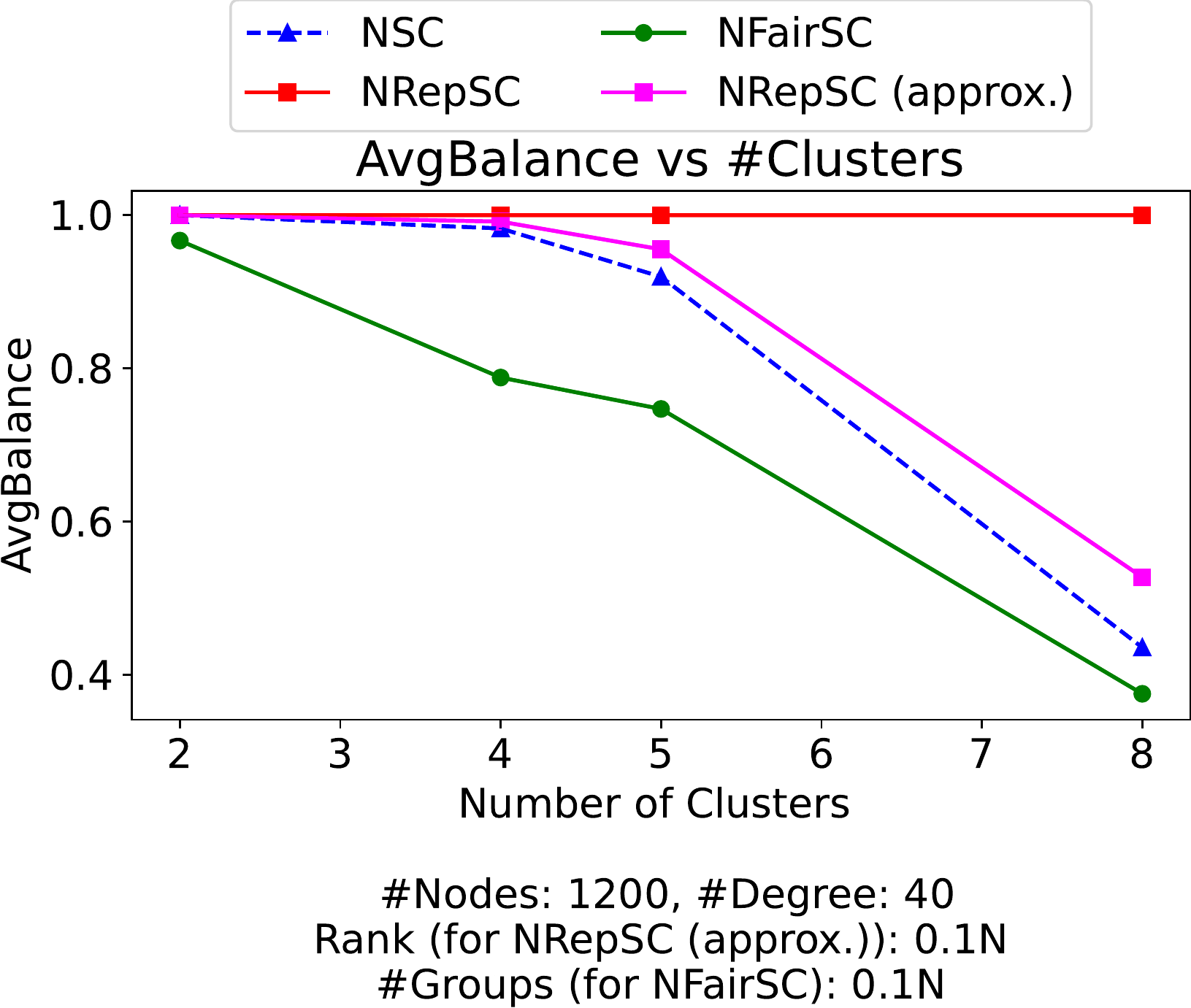}}%
    \hspace{1cm}
    \subfloat[][Ratio-cut vs no. of clusters]{\includegraphics[width=0.45\textwidth]{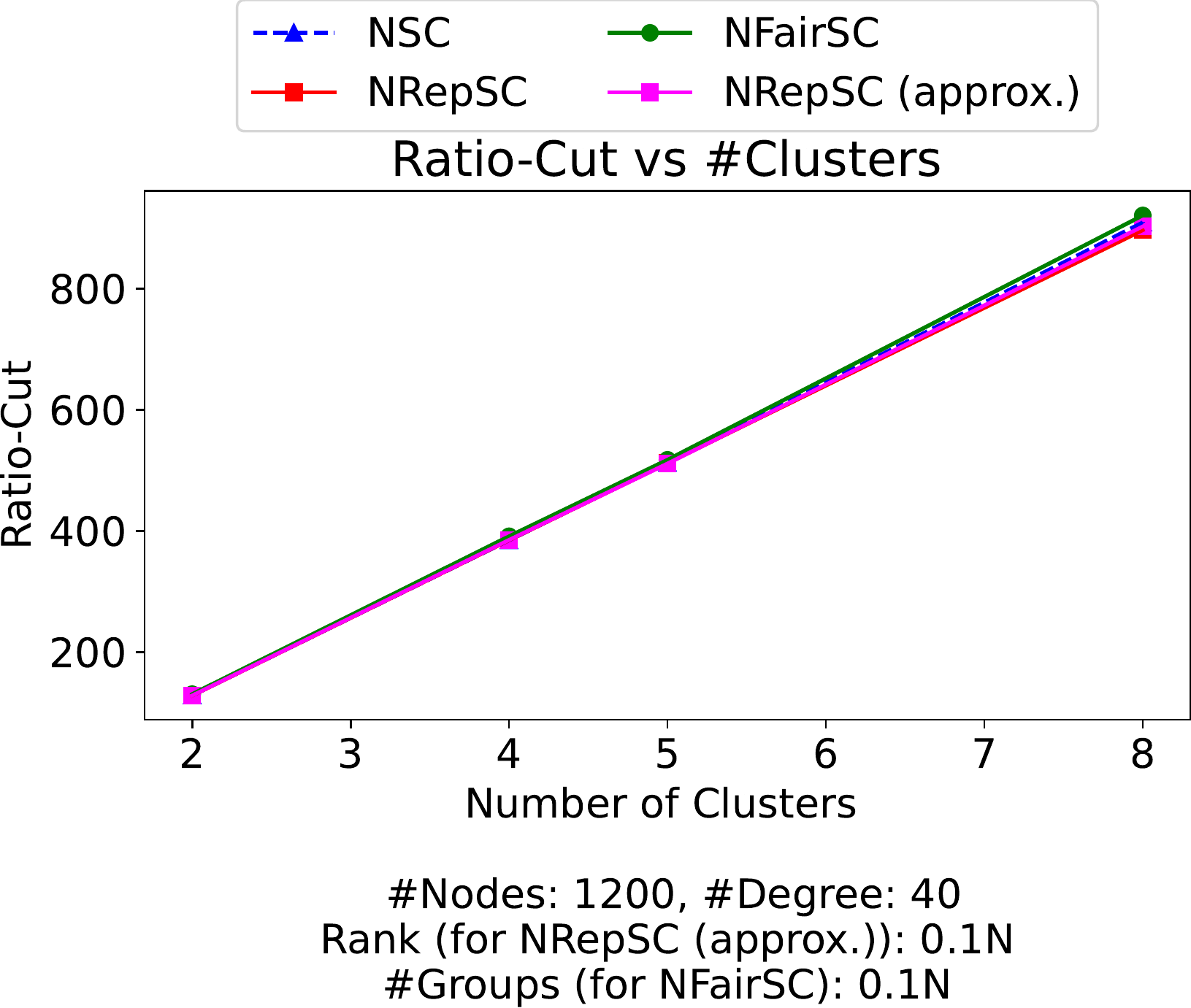}}

    \subfloat[][Average balance vs degree of $\calR$]{\includegraphics[width=0.45\textwidth]{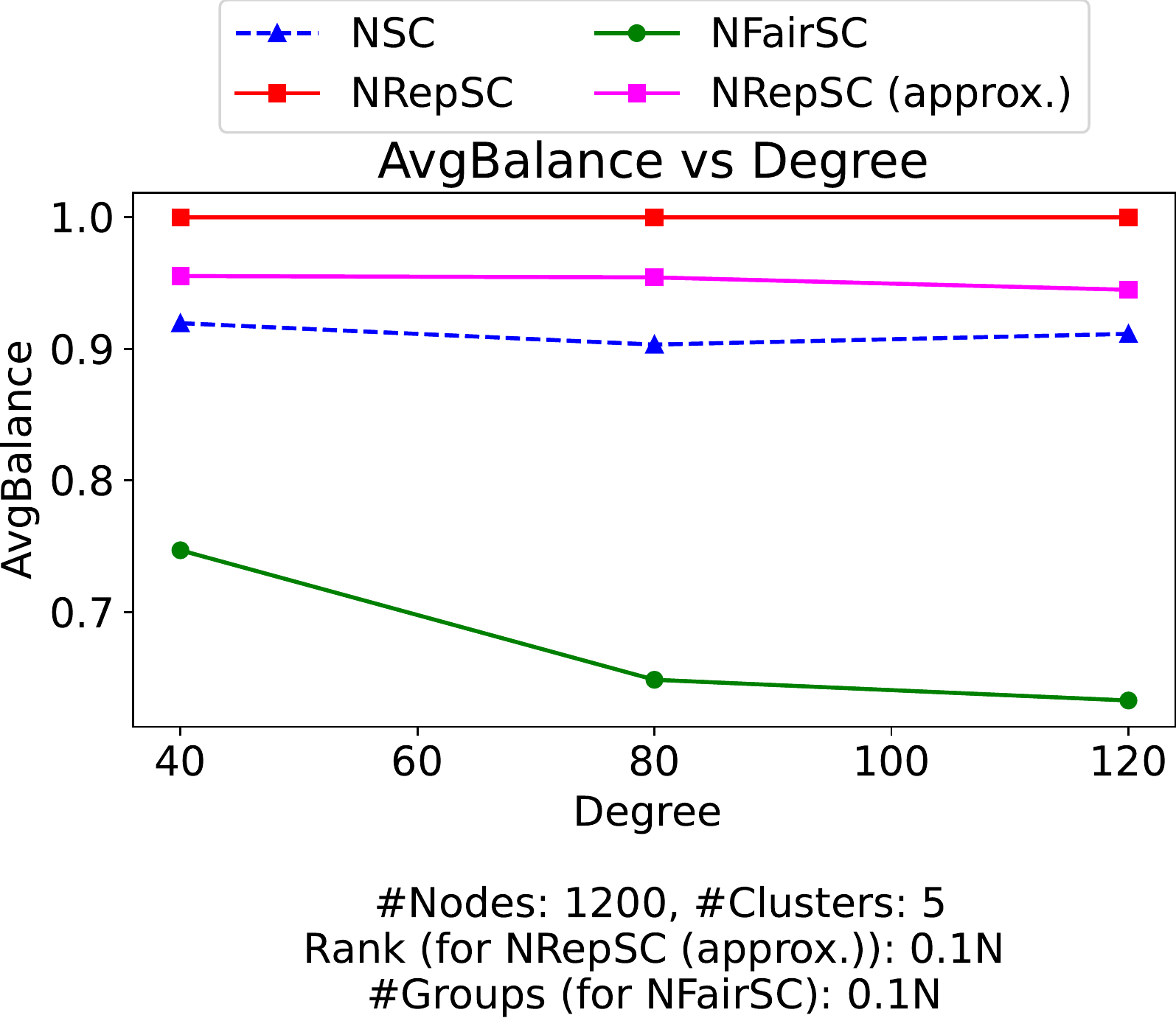}}%
    \hspace{1cm}
    \subfloat[][Ratio-cut vs degree of $\calR$]{\includegraphics[width=0.45\textwidth]{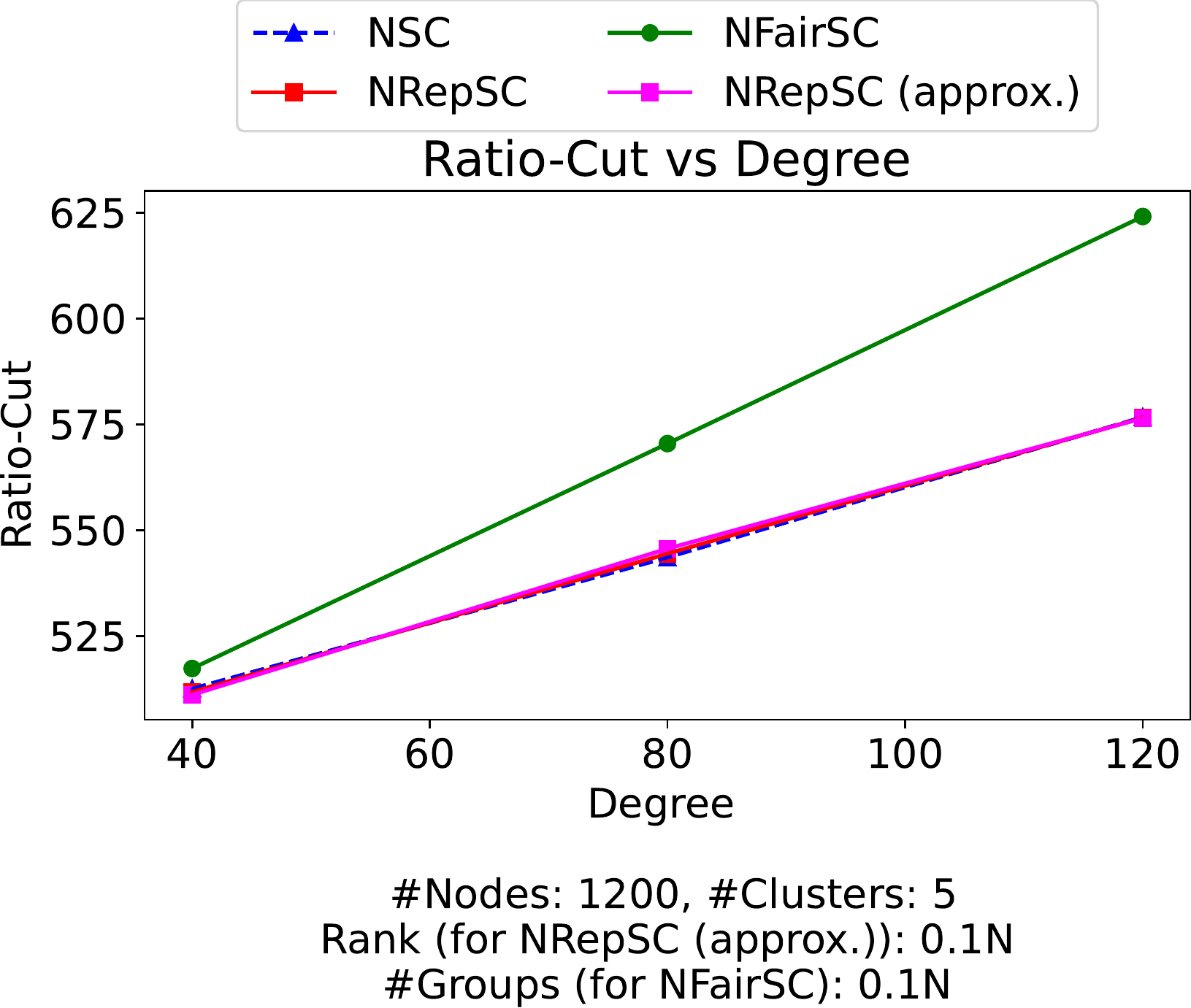}}

    \caption{\rebuttal{Comparing \textsc{NRepSC} with other ``normalized'' algorithms using synthetically generated $d$-regular representation graphs.}}
    \label{fig:d_reg_norm_separated}
\end{figure}

\begin{figure}[t]
    \centering
    \subfloat[][Average balance, $N=1000$, $K=4$]{\includegraphics[width=0.45\textwidth]{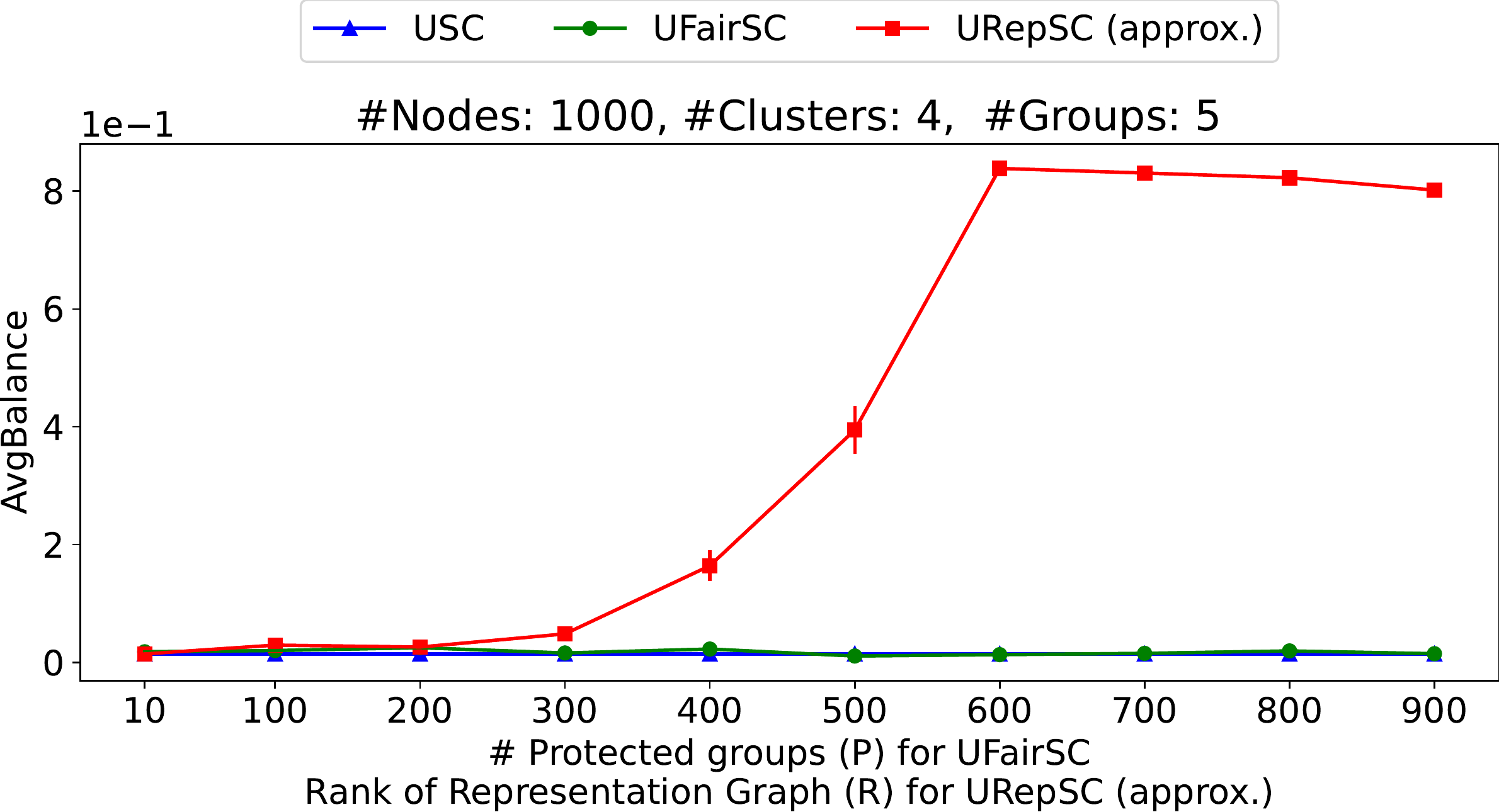}}%
    \hspace{1cm}
    \subfloat[][Ratio-cut, $N=1000$, $K=4$]{\includegraphics[width=0.45\textwidth]{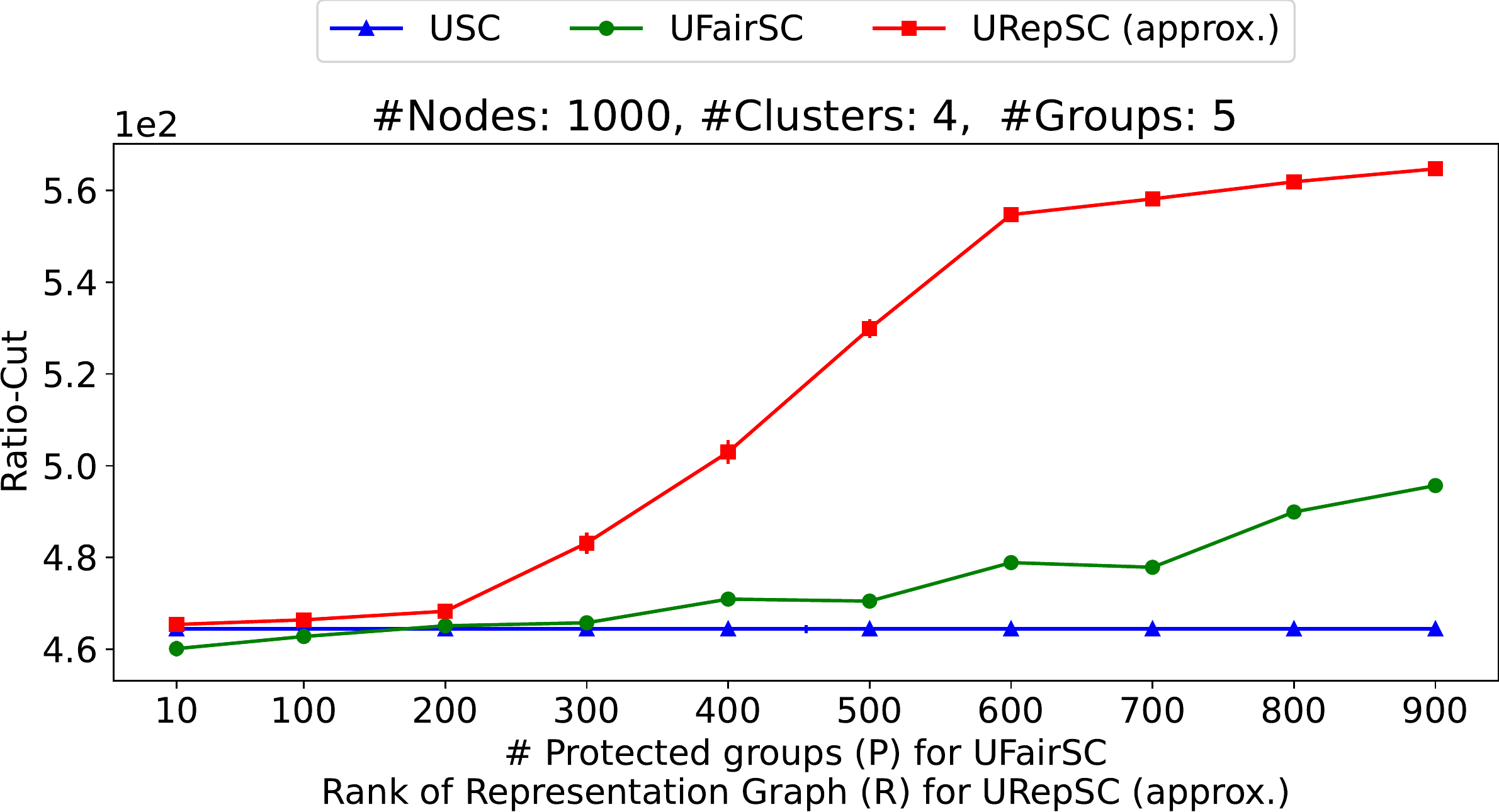}}

    \subfloat[][Average balance, $N=1000$, $K=8$]{\includegraphics[width=0.45\textwidth]{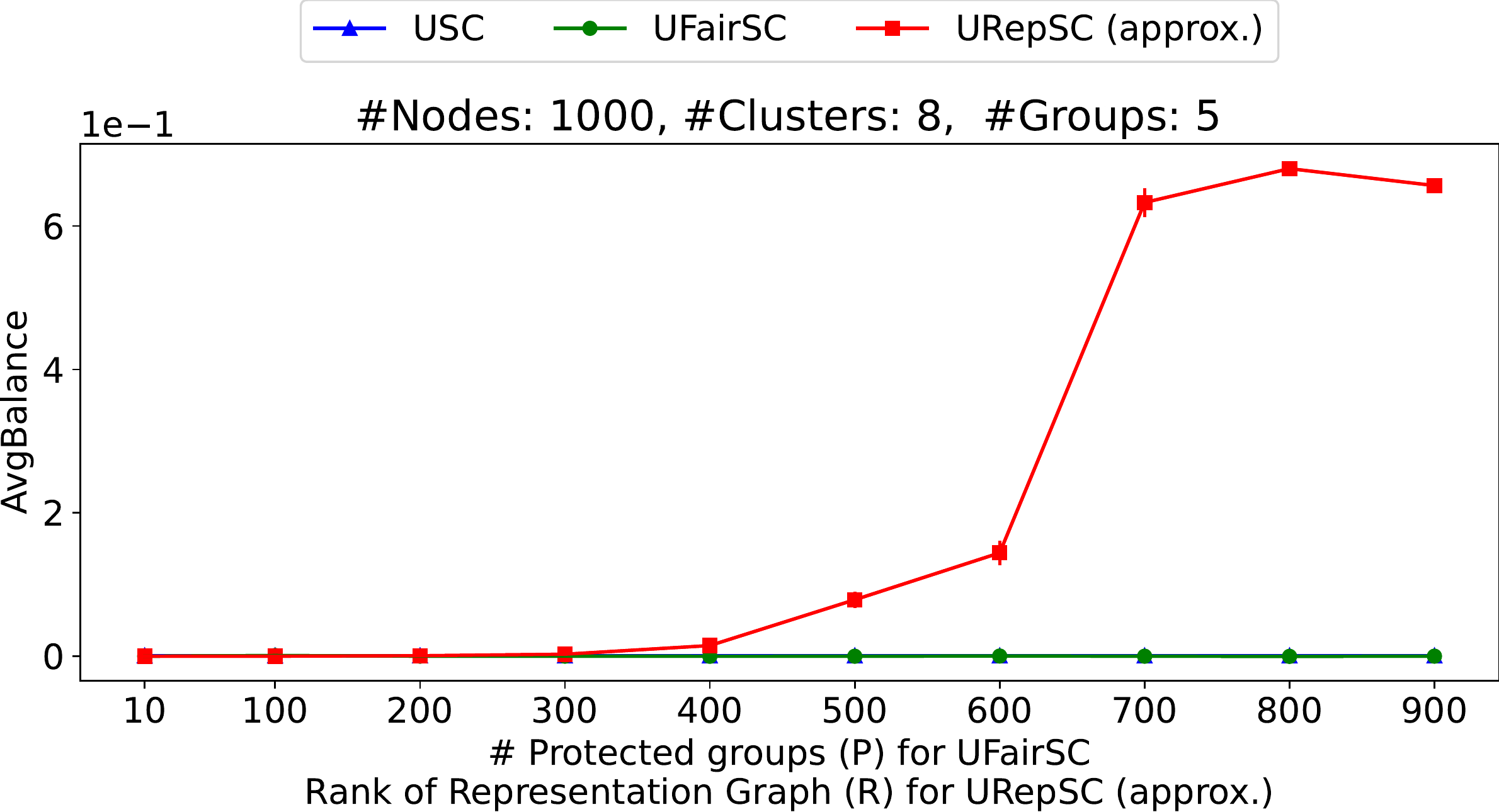}}%
    \hspace{1cm}
    \subfloat[][Ratio-cut, $N=1000$, $K=8$]{\includegraphics[width=0.45\textwidth]{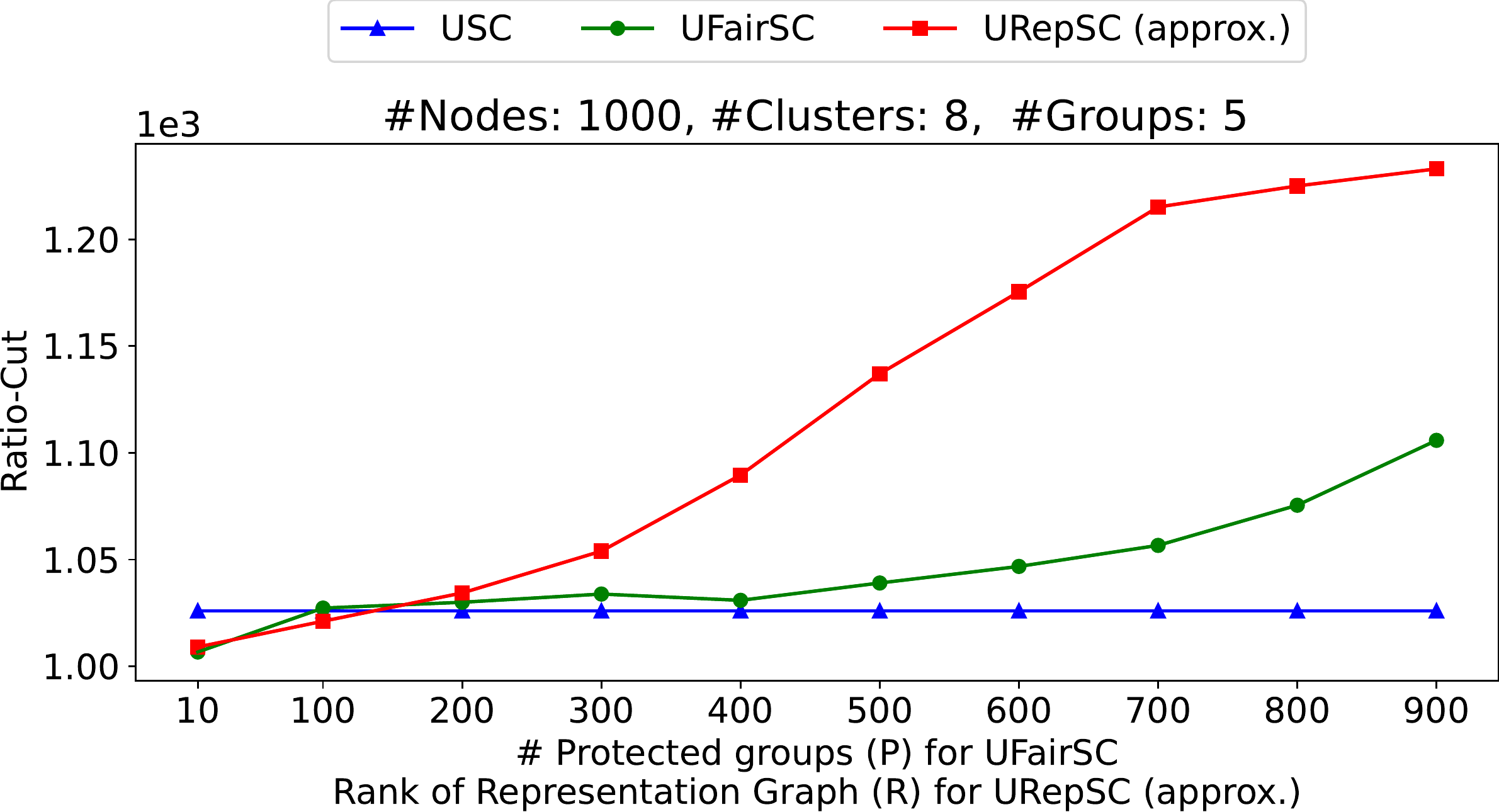}}

    \subfloat[][Average balance, $N=3000$, $K=4$]{\includegraphics[width=0.45\textwidth]{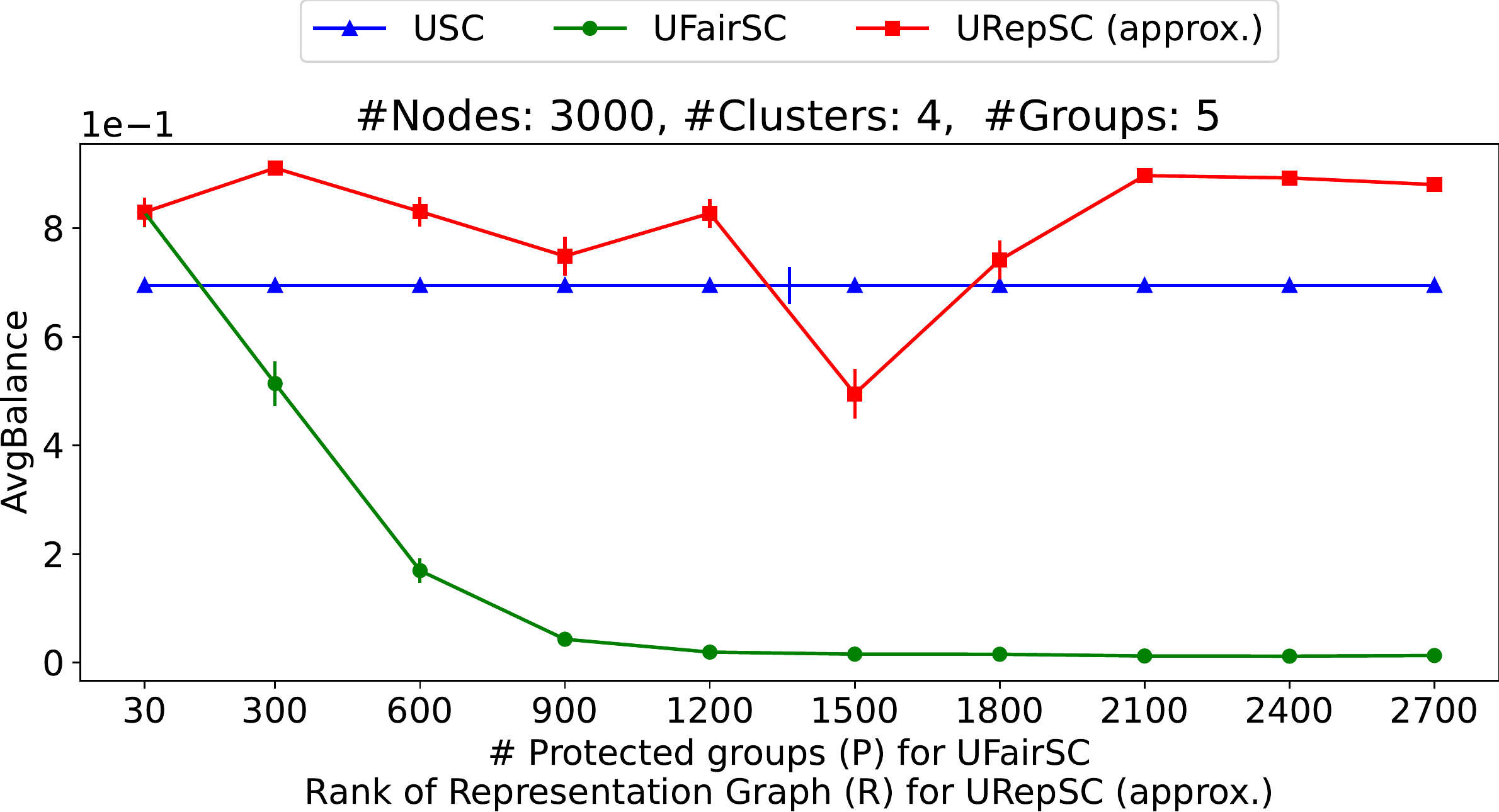}}%
    \hspace{1cm}
    \subfloat[][Ratio-cut, $N=3000$, $K=4$]{\includegraphics[width=0.45\textwidth]{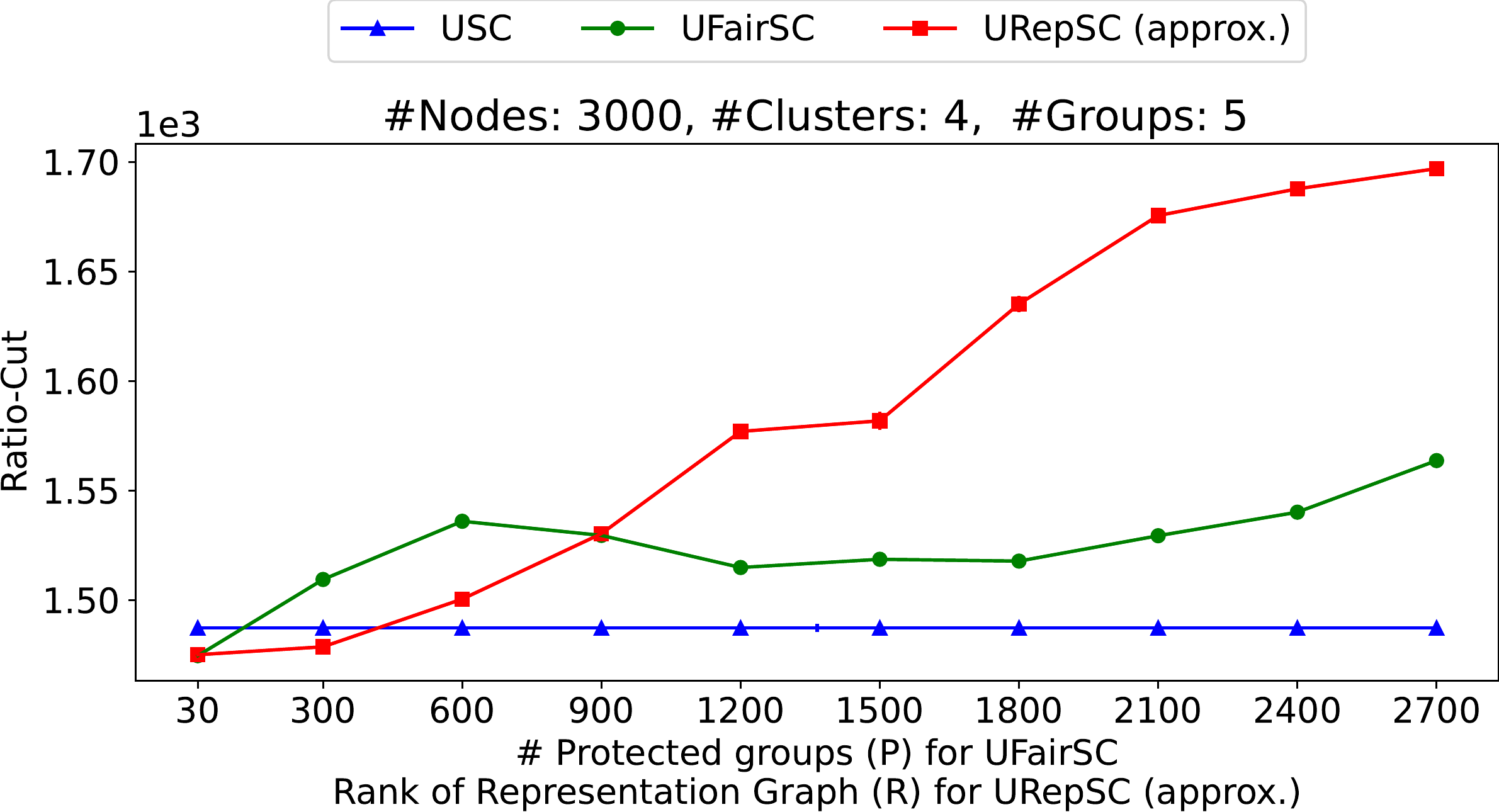}}

    \subfloat[][Average balance, $N=3000$, $K=8$]{\includegraphics[width=0.45\textwidth]{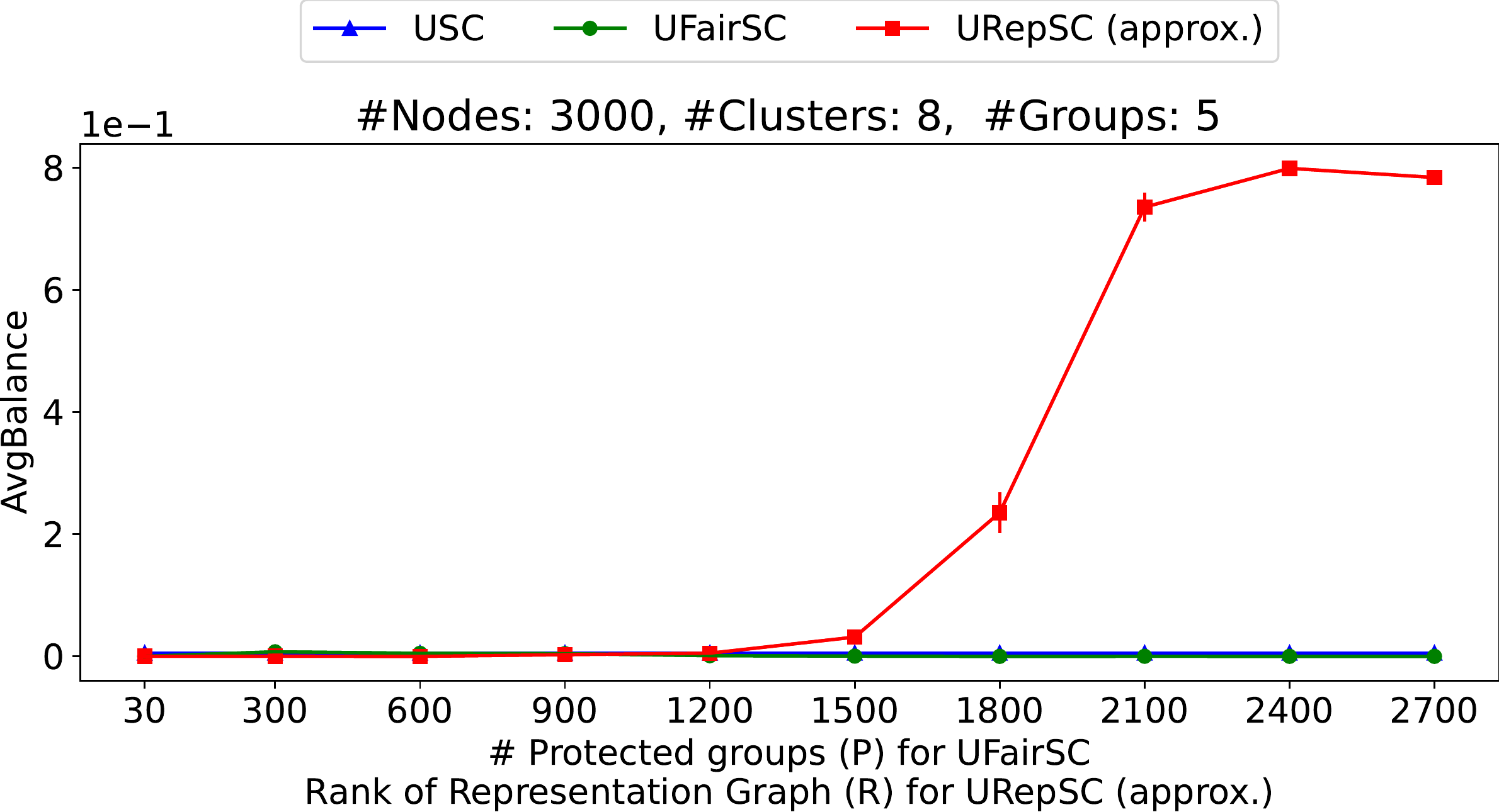}}%
    \hspace{1cm}
    \subfloat[][Ratio-cut, $N=3000$, $K=8$]{\includegraphics[width=0.45\textwidth]{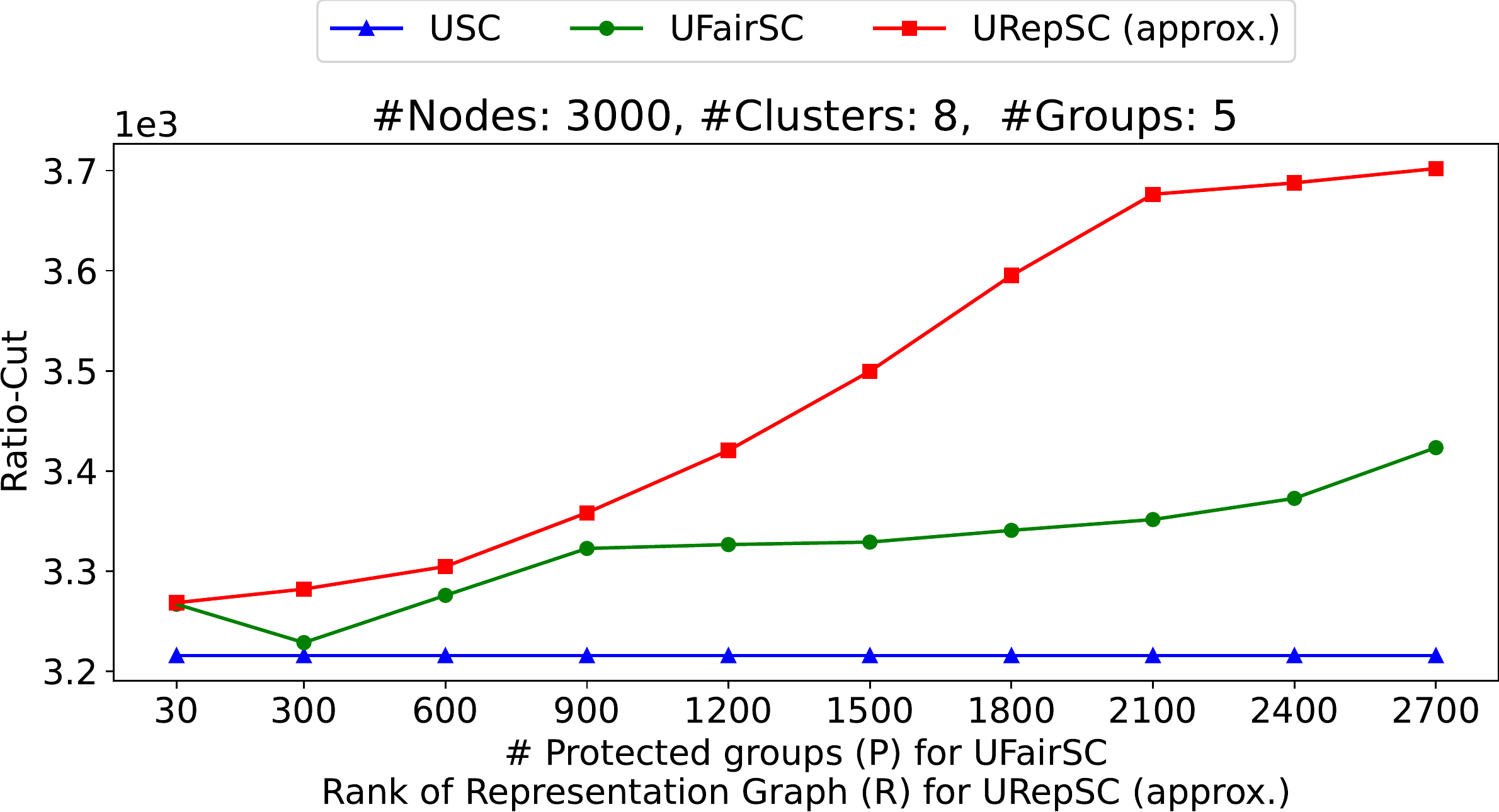}}

    \caption{\rebuttal{Comparing \textsc{URepSC} with other ``unnormalized'' algorithms using representation graphs sampled from a planted partition model.}}
    \label{fig:sbm_comparison_unnorm_separated}
\end{figure}

\begin{figure}[t]
    \centering
    \subfloat[][Average balance, $N=1000$, $K=4$]{\includegraphics[width=0.45\textwidth]{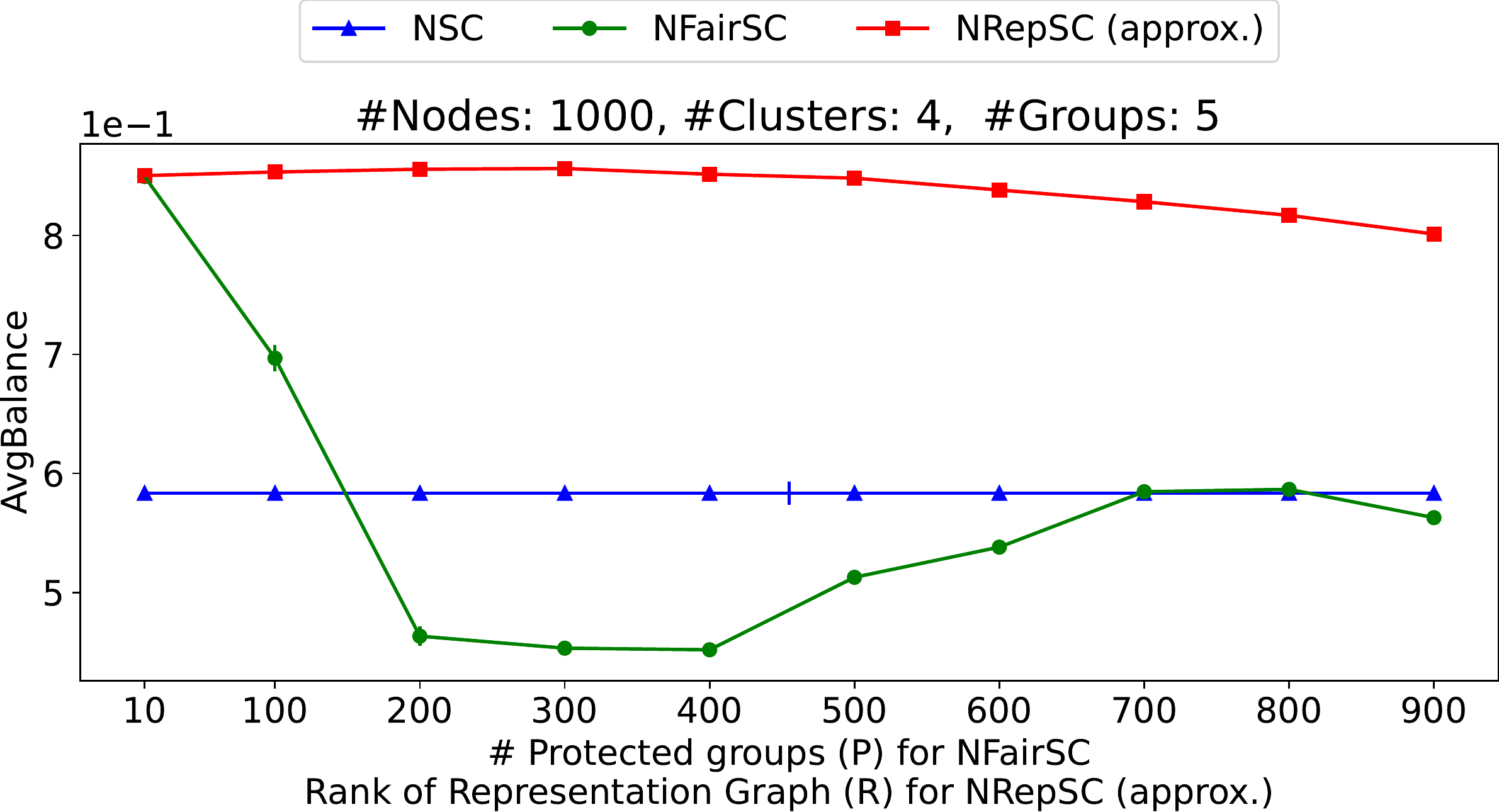}}%
    \hspace{1cm}
    \subfloat[][Ratio-cut, $N=1000$, $K=4$]{\includegraphics[width=0.45\textwidth]{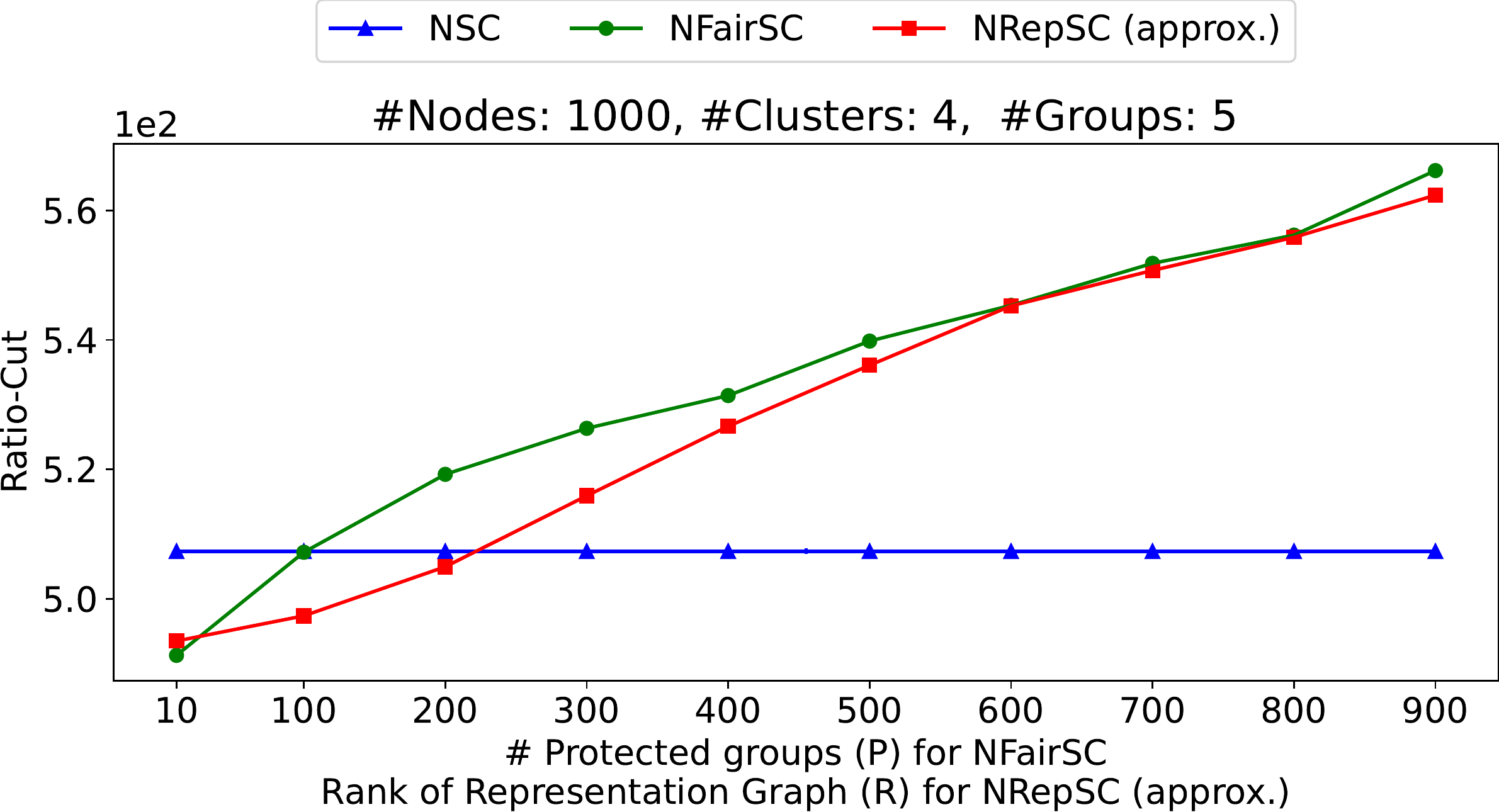}}

    \subfloat[][Average balance, $N=1000$, $K=8$]{\includegraphics[width=0.45\textwidth]{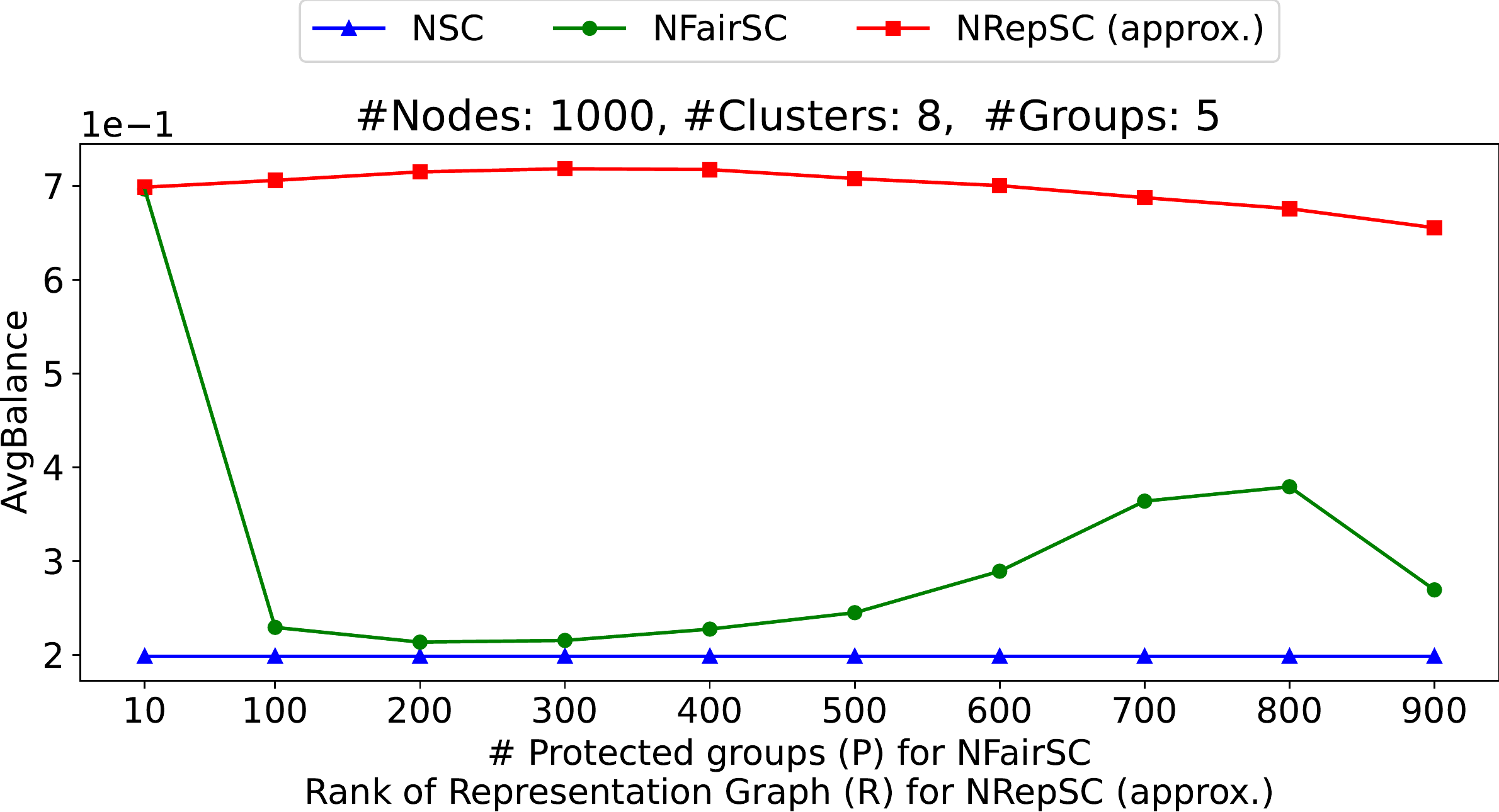}}%
    \hspace{1cm}
    \subfloat[][Ratio-cut, $N=1000$, $K=8$]{\includegraphics[width=0.45\textwidth]{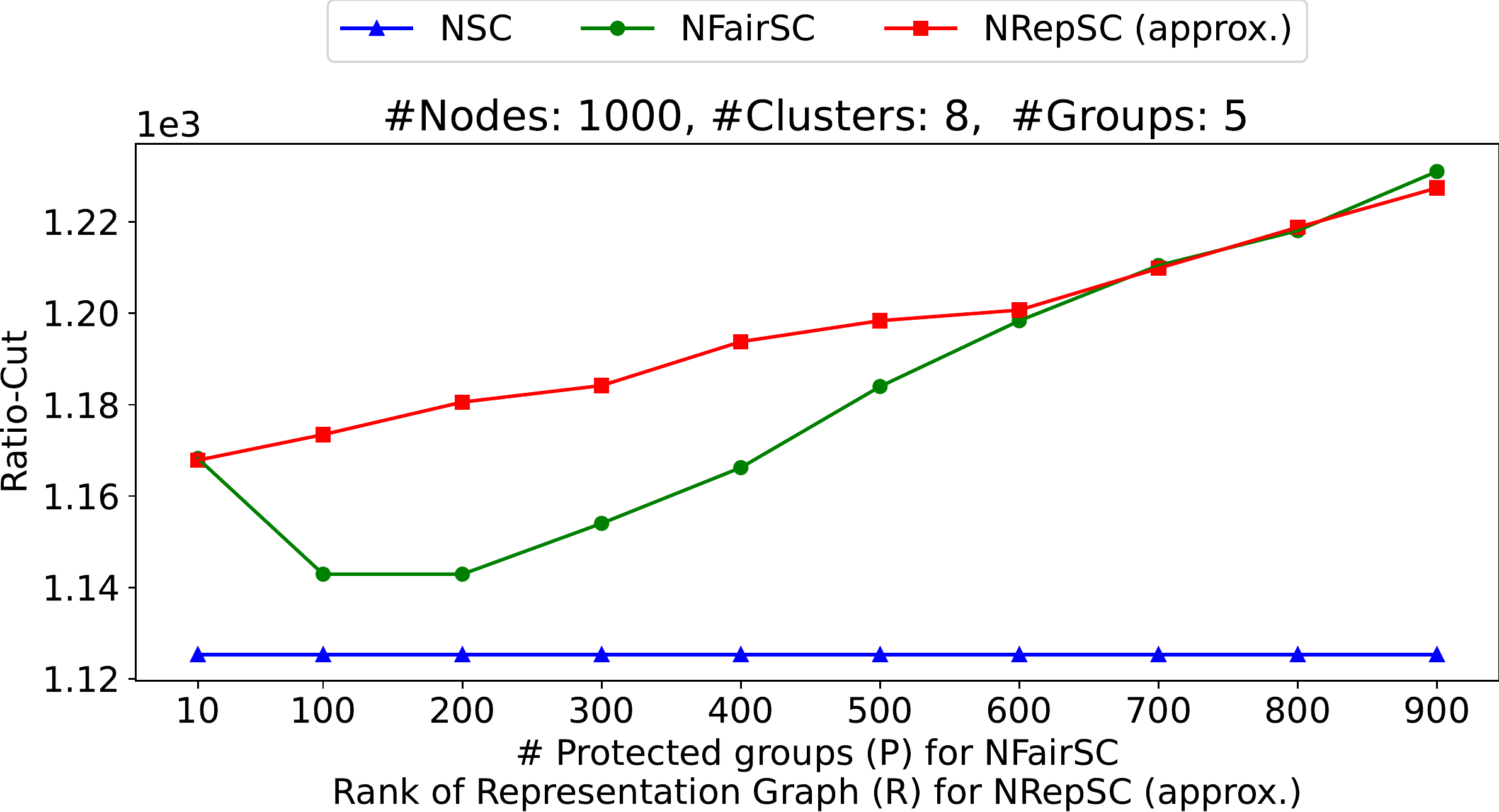}}

    \subfloat[][Average balance, $N=3000$, $K=4$]{\includegraphics[width=0.45\textwidth]{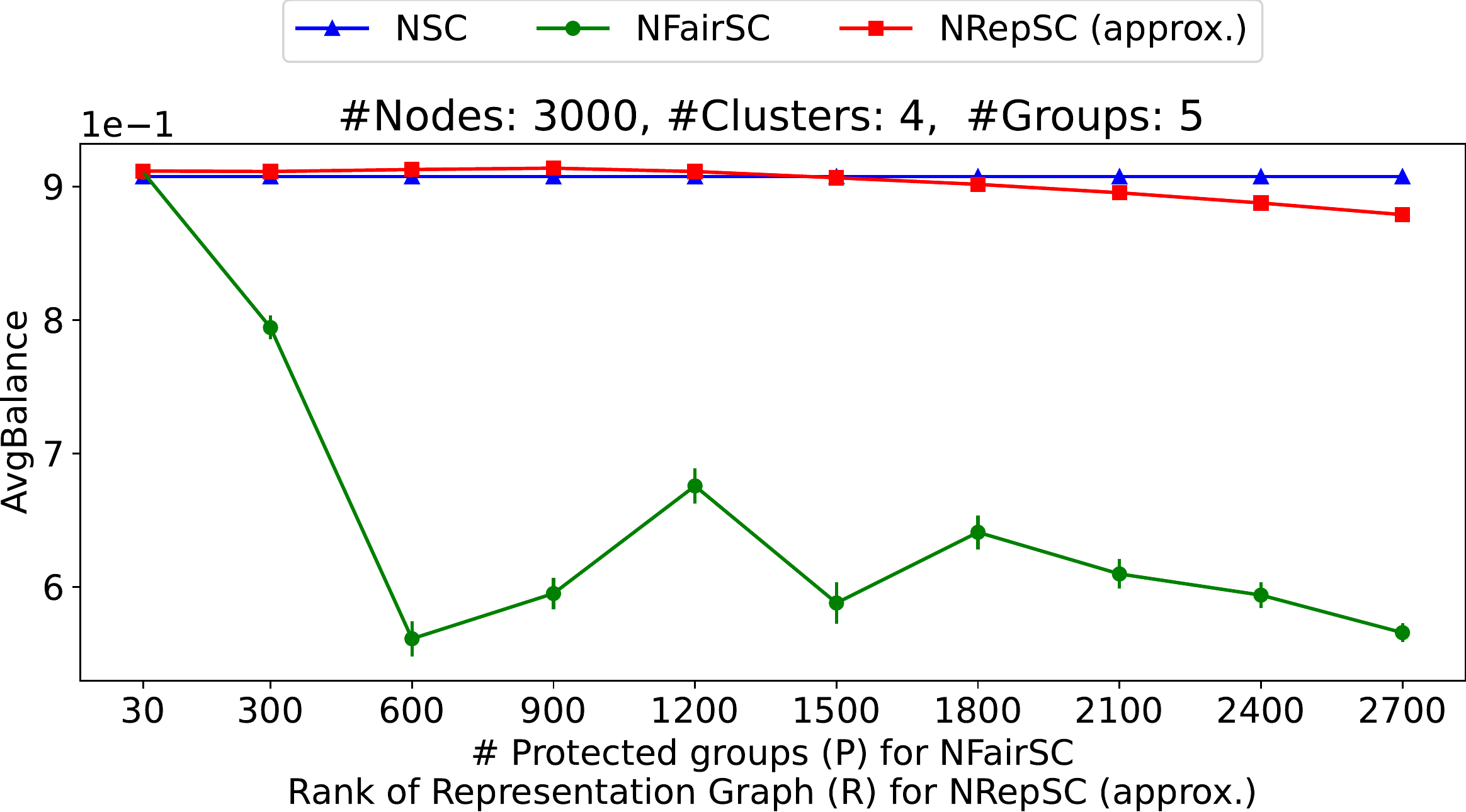}}%
    \hspace{1cm}
    \subfloat[][Ratio-cut, $N=3000$, $K=4$]{\includegraphics[width=0.45\textwidth]{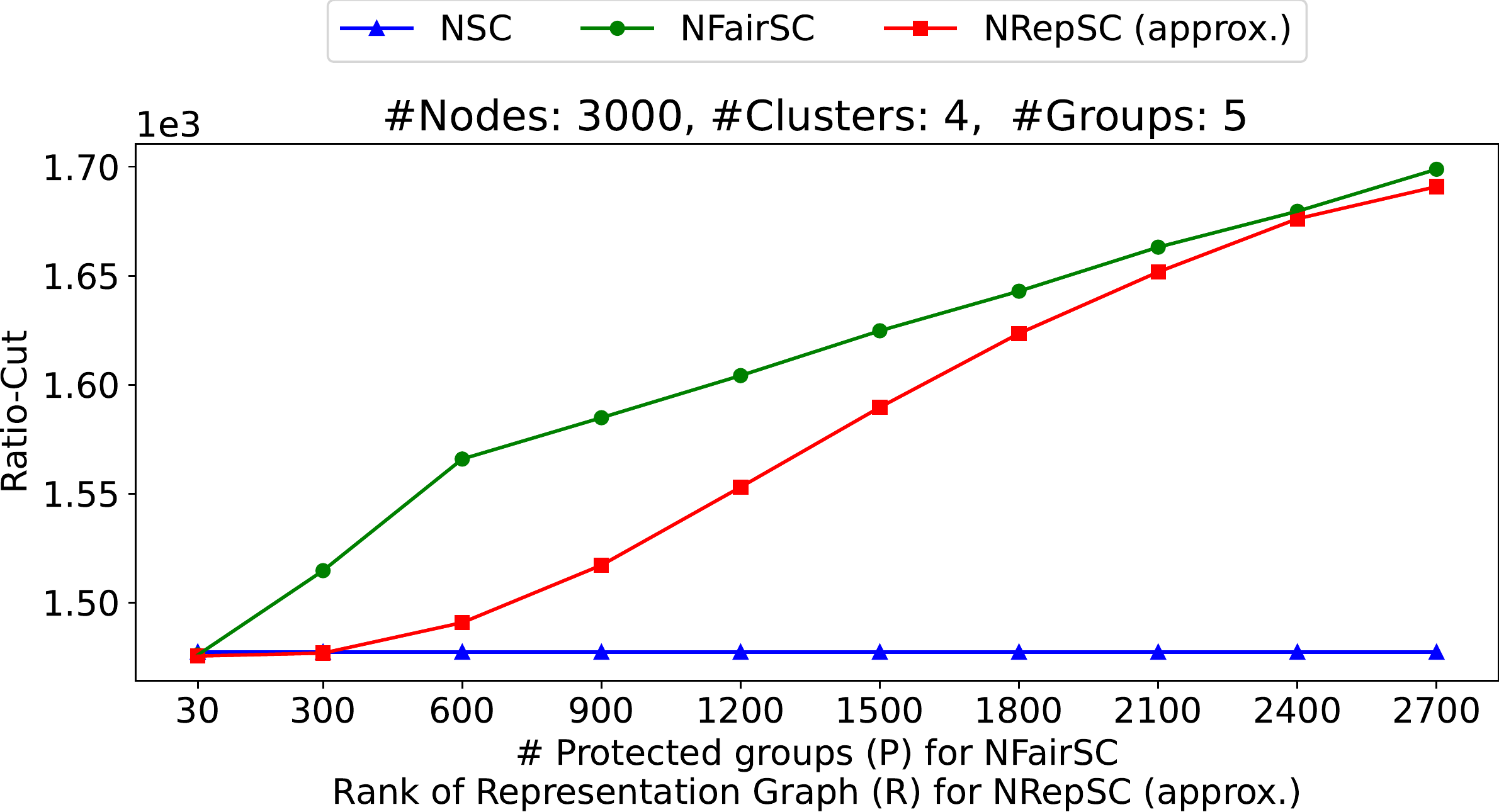}}

    \subfloat[][Average balance, $N=3000$, $K=8$]{\includegraphics[width=0.45\textwidth]{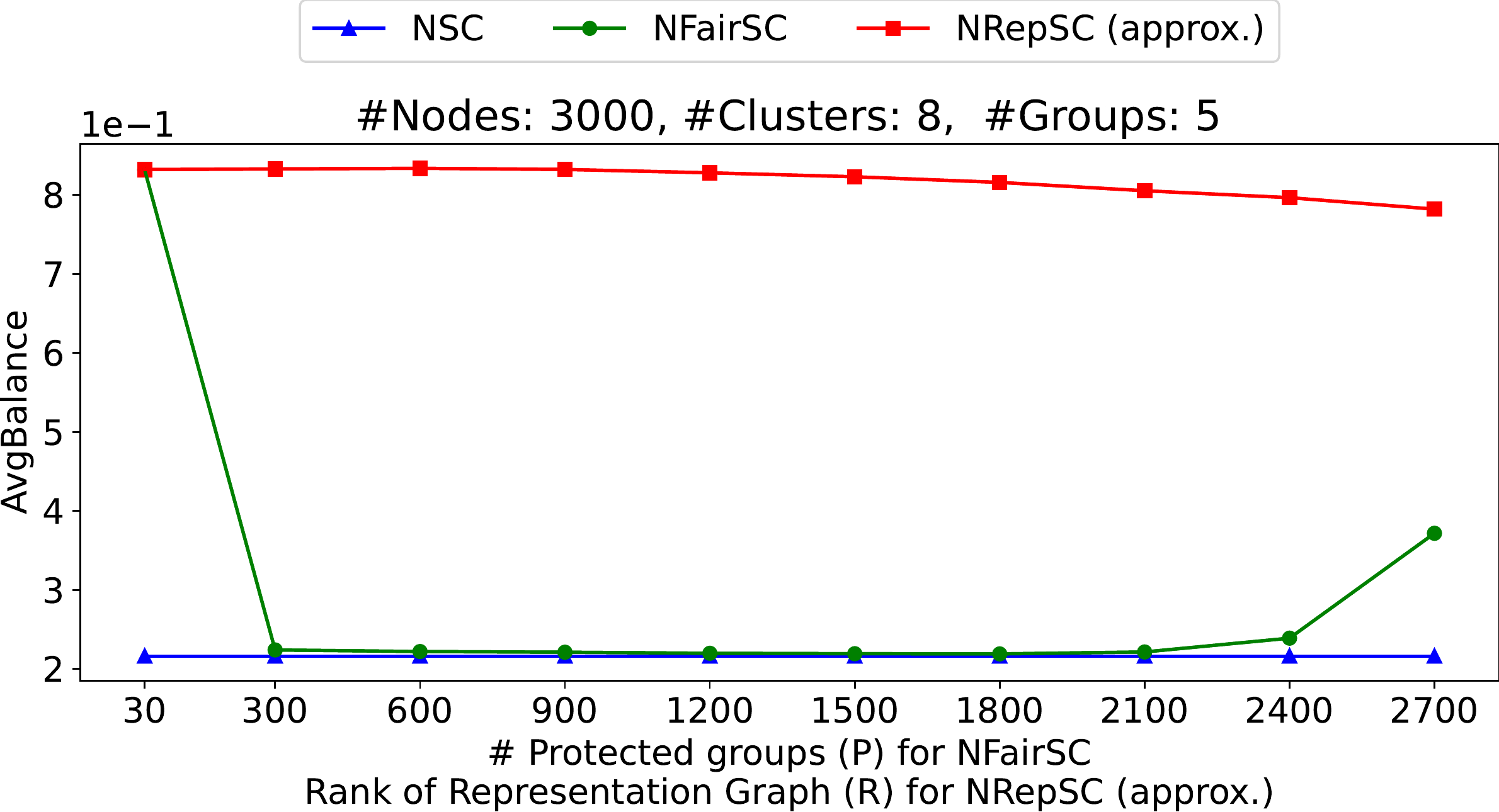}}%
    \hspace{1cm}
    \subfloat[][Ratio-cut, $N=3000$, $K=8$]{\includegraphics[width=0.45\textwidth]{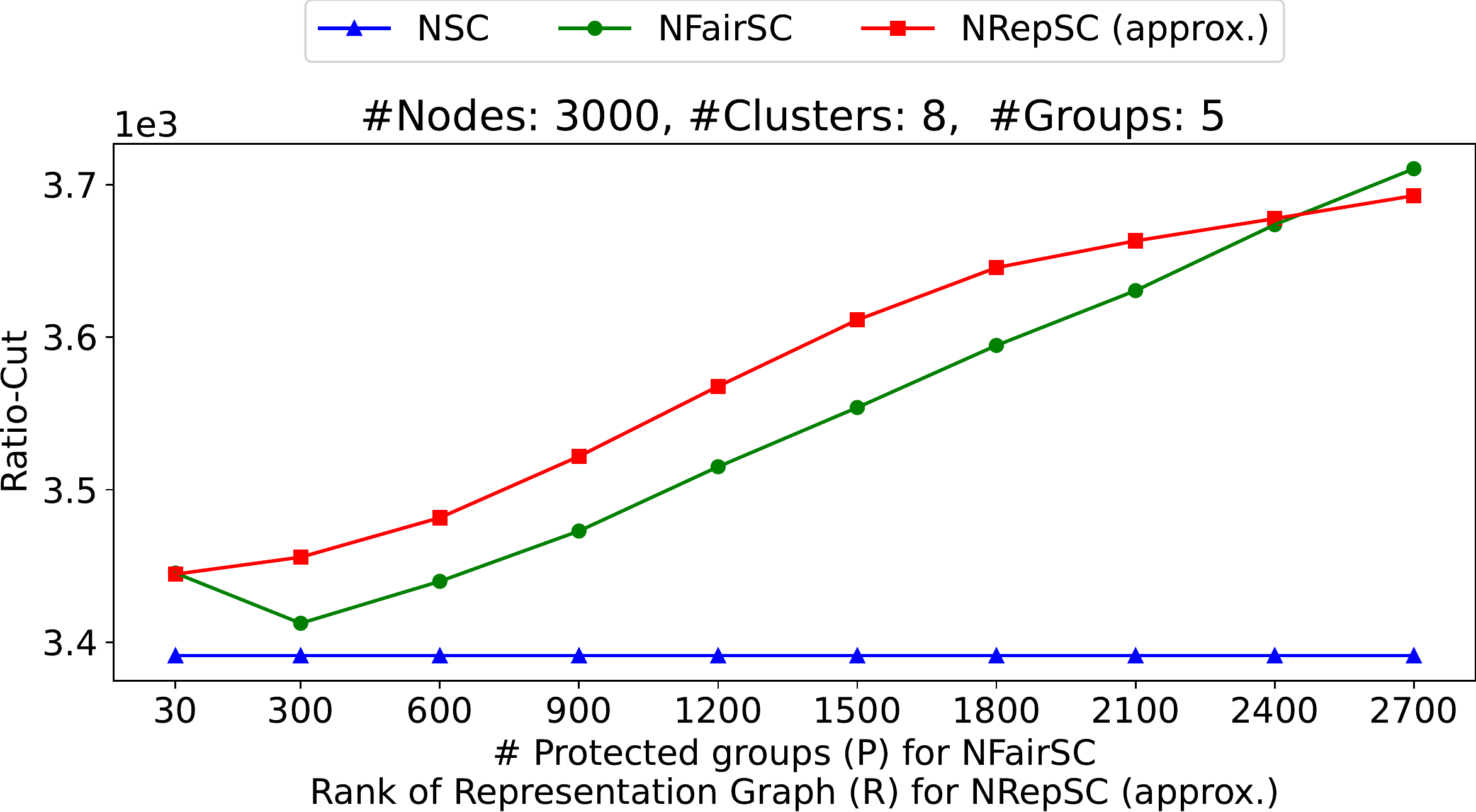}}

    \caption{\rebuttal{Comparing \textsc{NRepSC} with other ``normalized'' algorithms using representation graphs sampled from a planted partition model.}}
    \label{fig:sbm_comparison_norm_separated}
\end{figure}

\begin{figure}[t]
    \centering
    \subfloat[][Average balance, $K=2$]{\includegraphics[width=0.45\textwidth]{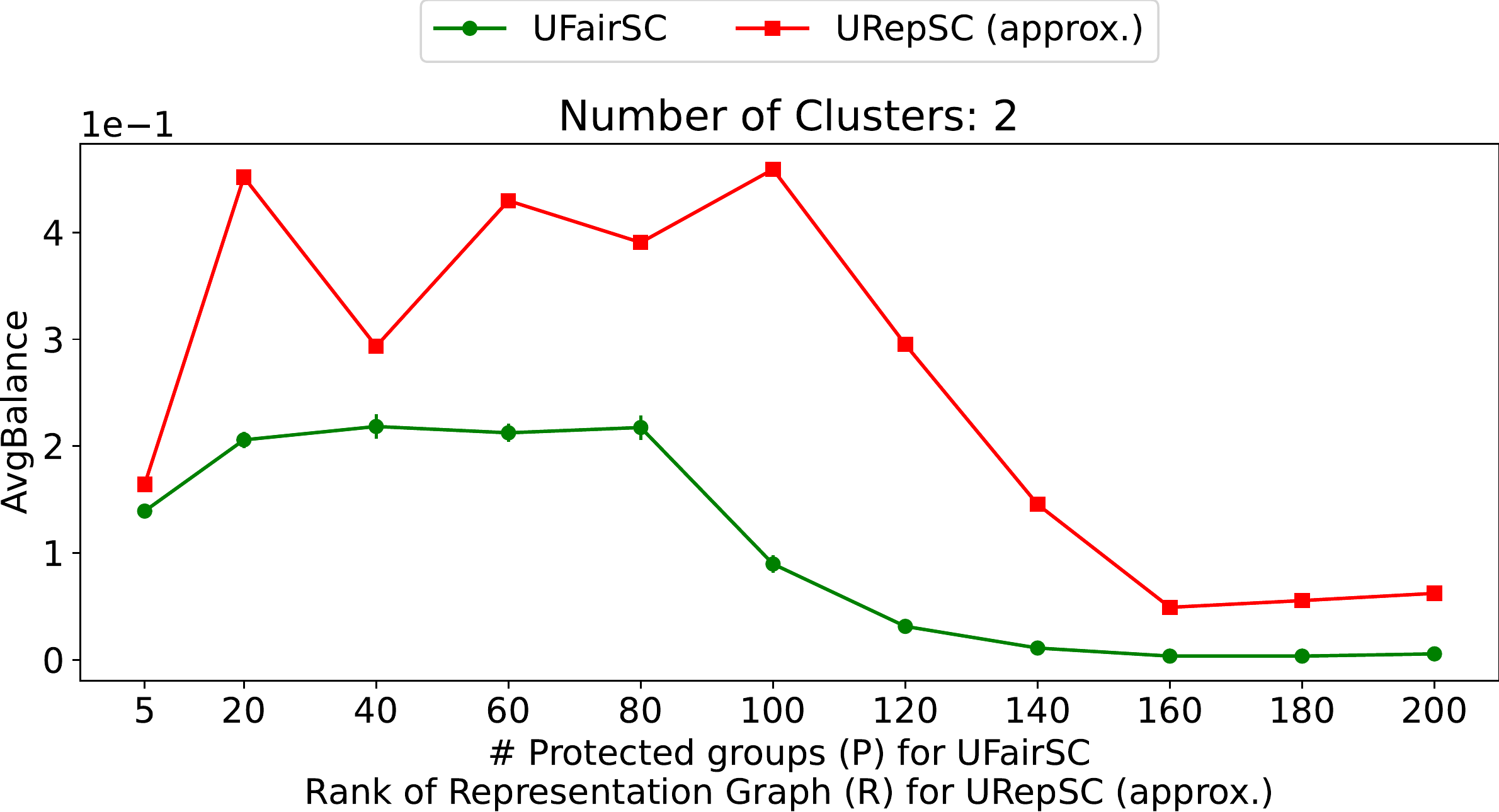}}%
    \hspace{1cm}
    \subfloat[][Ratio-cut, $K=2$]{\includegraphics[width=0.45\textwidth]{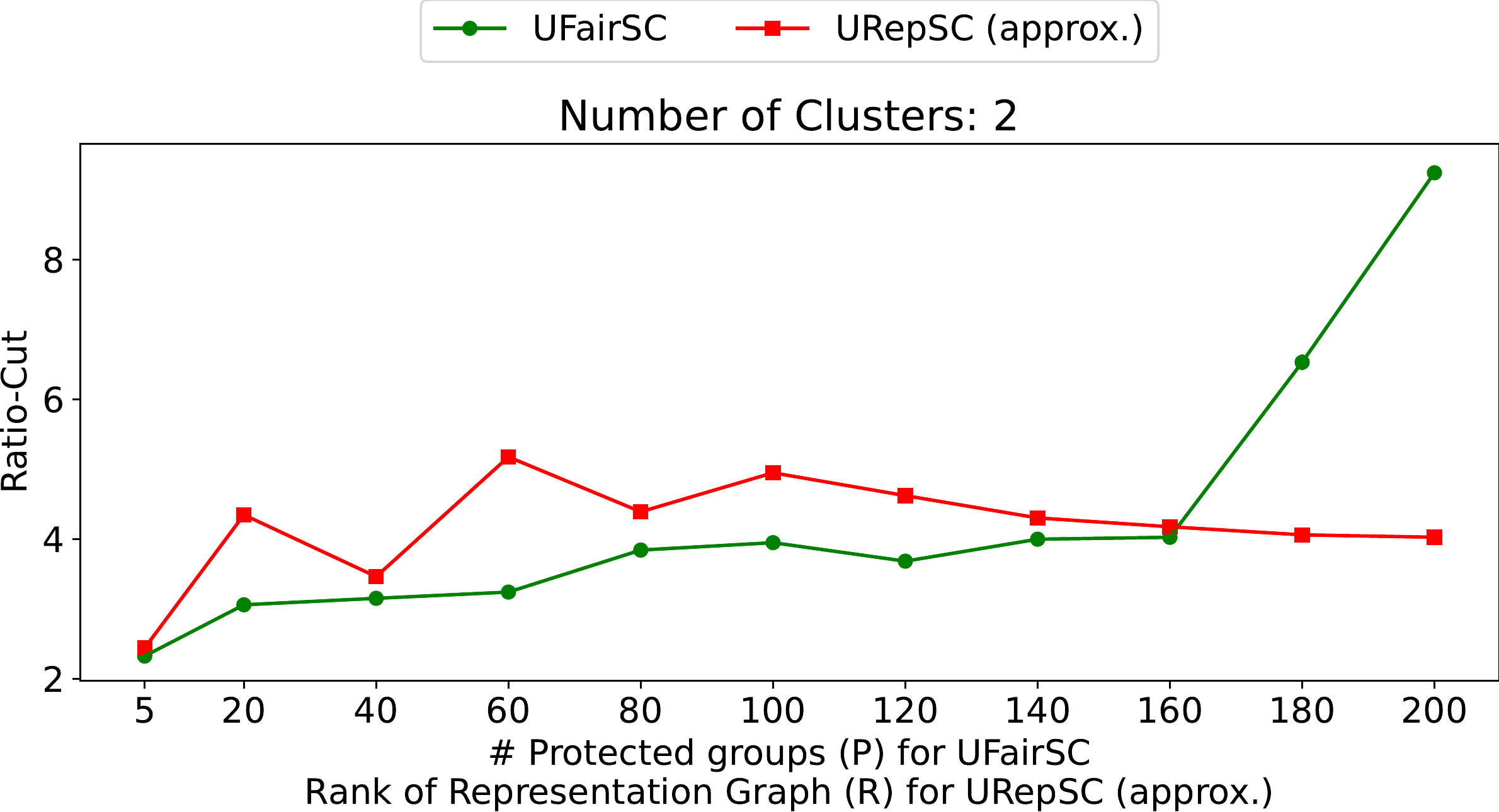}}

    \subfloat[][Average balance, $K=4$]{\includegraphics[width=0.45\textwidth]{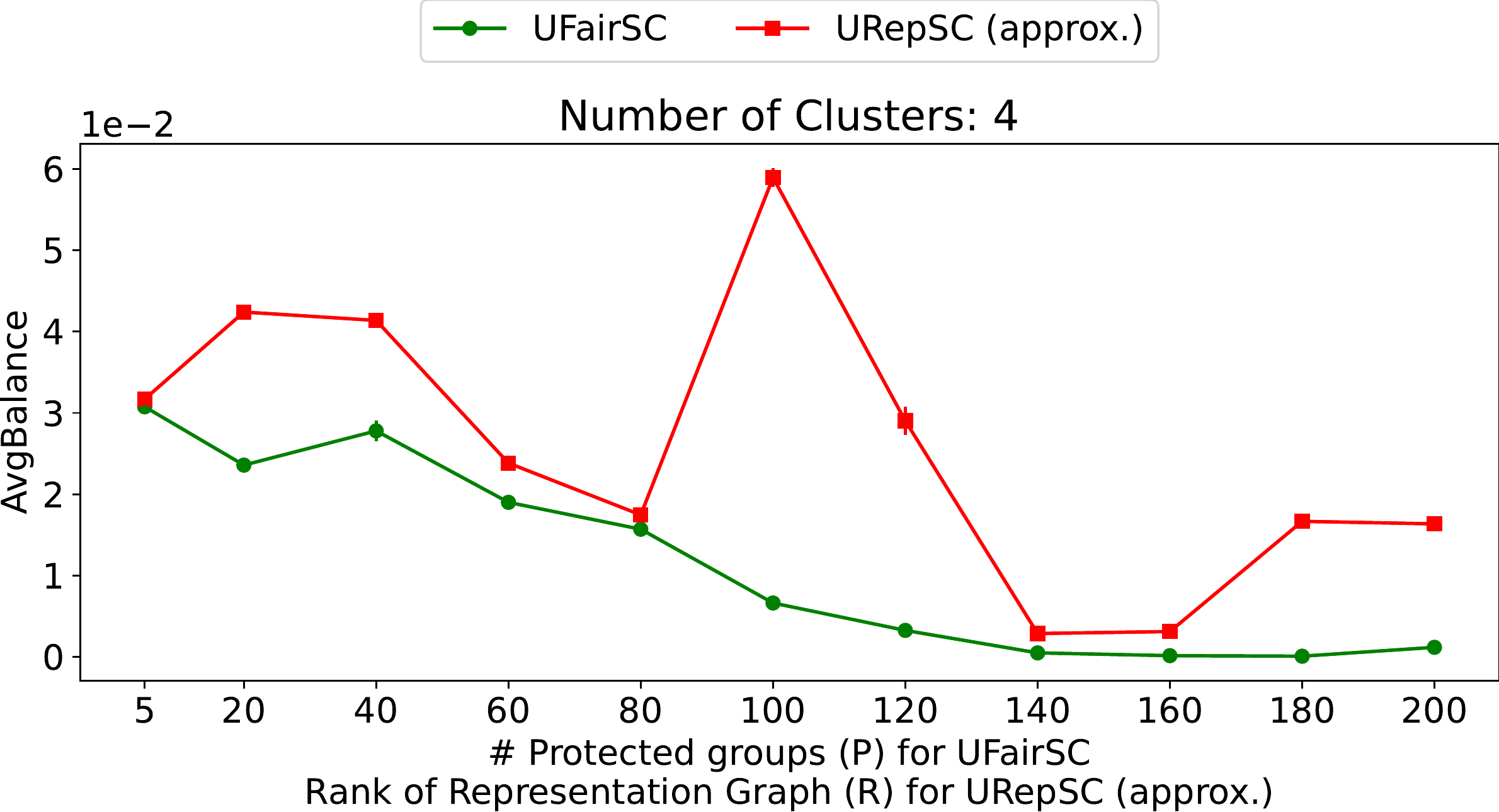}}%
    \hspace{1cm}
    \subfloat[][Ratio-cut, $K=4$]{\includegraphics[width=0.45\textwidth]{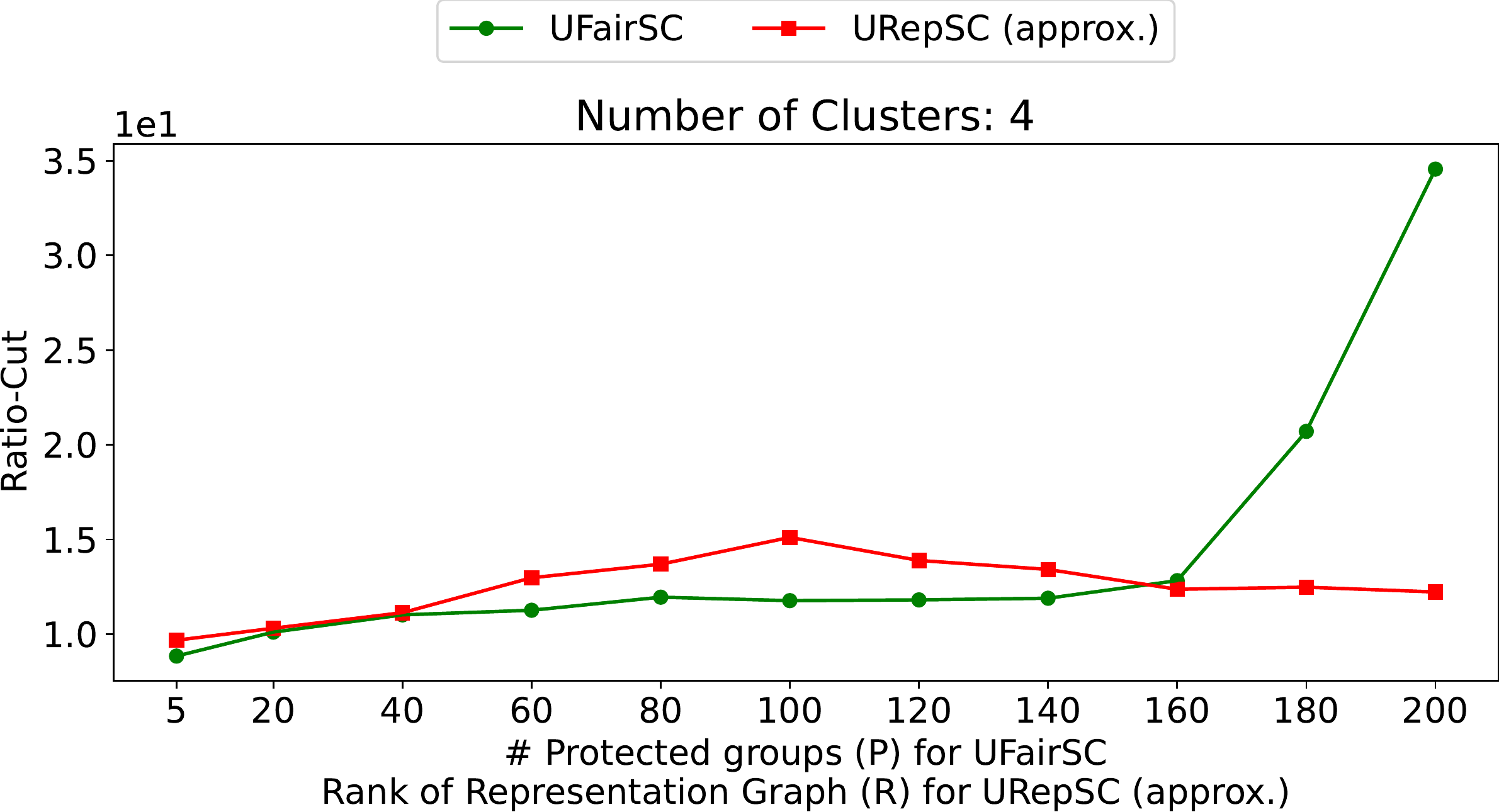}}

    \subfloat[][Average balance, $K=6$]{\includegraphics[width=0.45\textwidth]{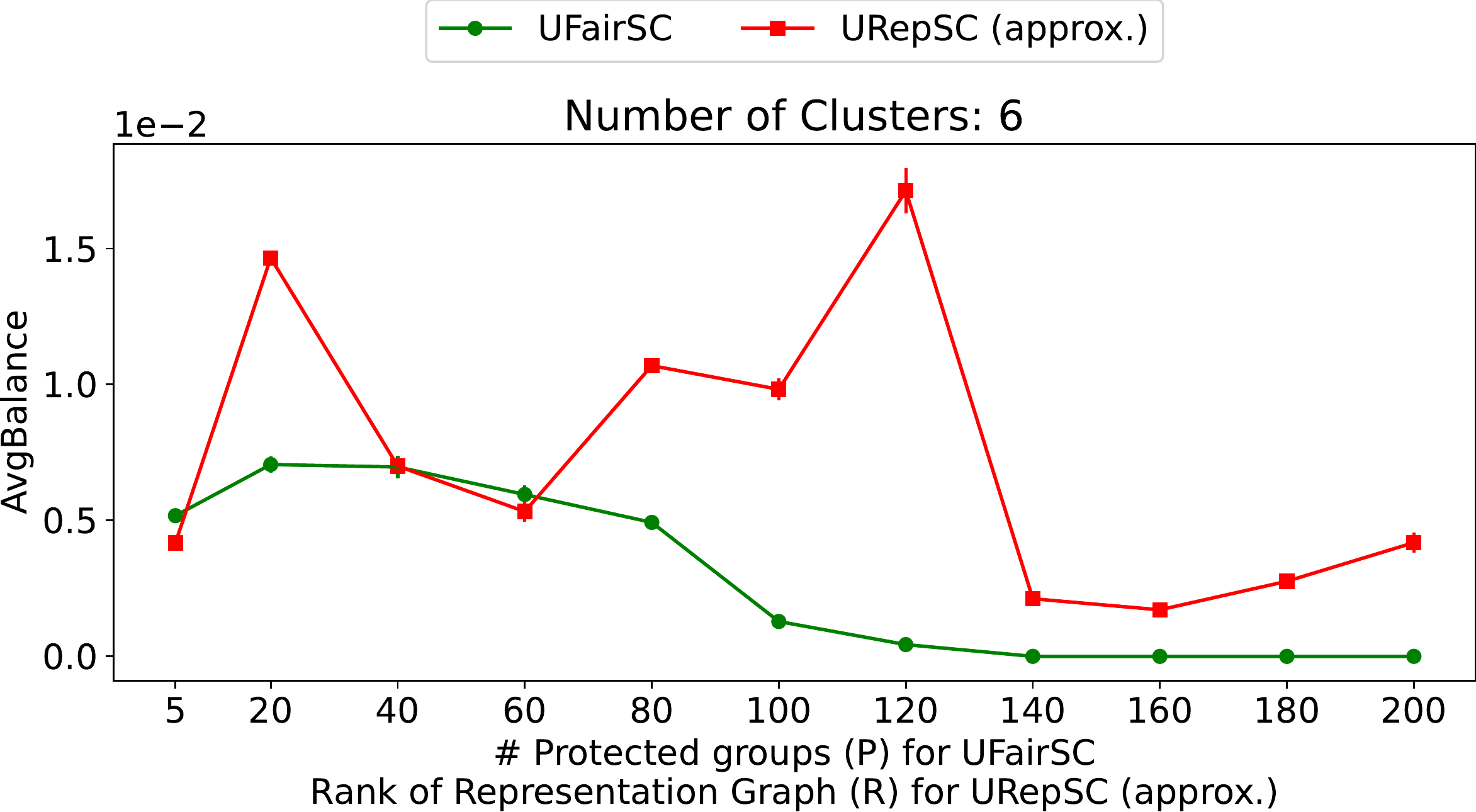}}%
    \hspace{1cm}
    \subfloat[][Ratio-cut, $K=6$]{\includegraphics[width=0.45\textwidth]{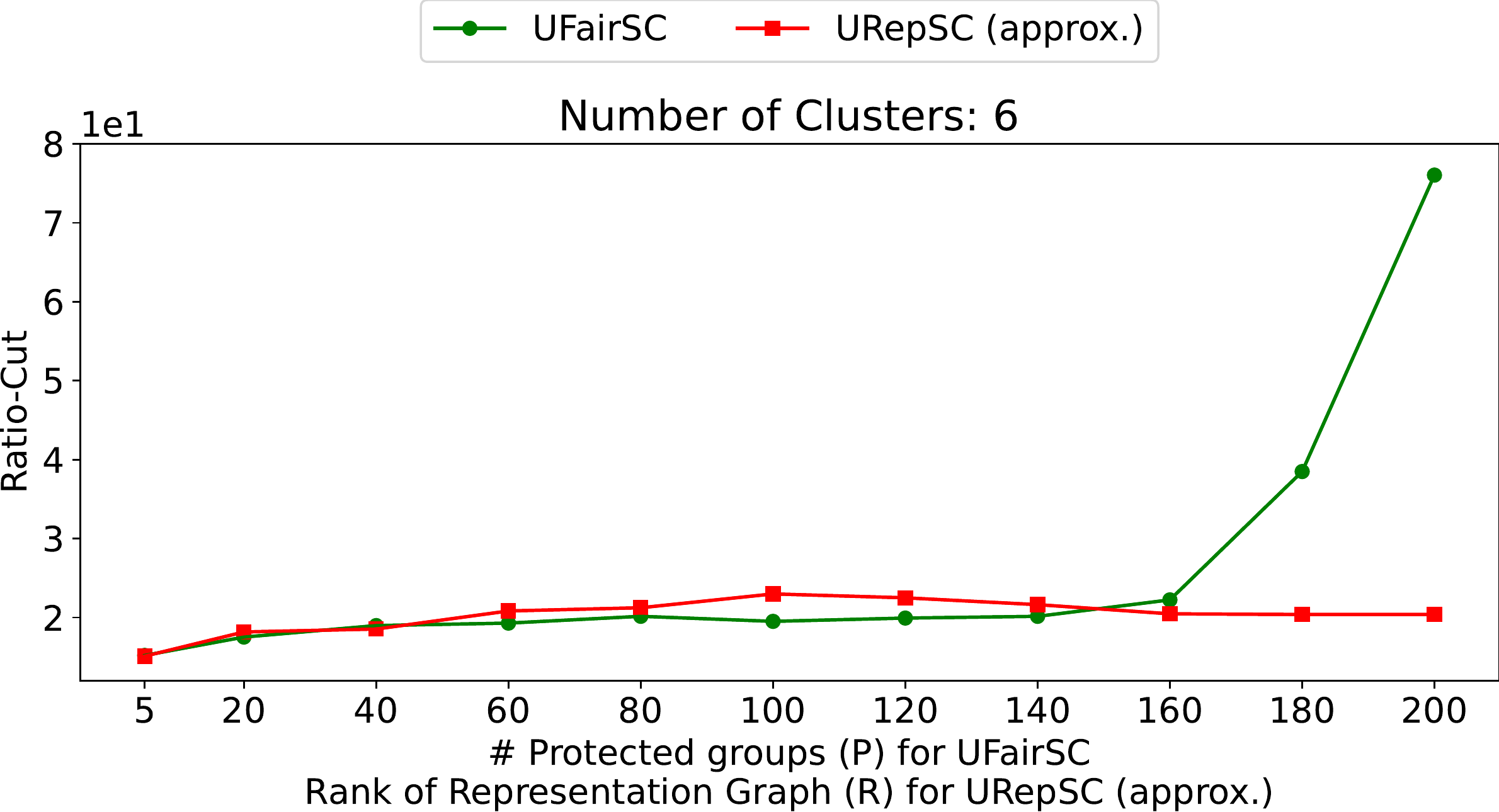}}

    \subfloat[][Average balance, $K=8$]{\includegraphics[width=0.45\textwidth]{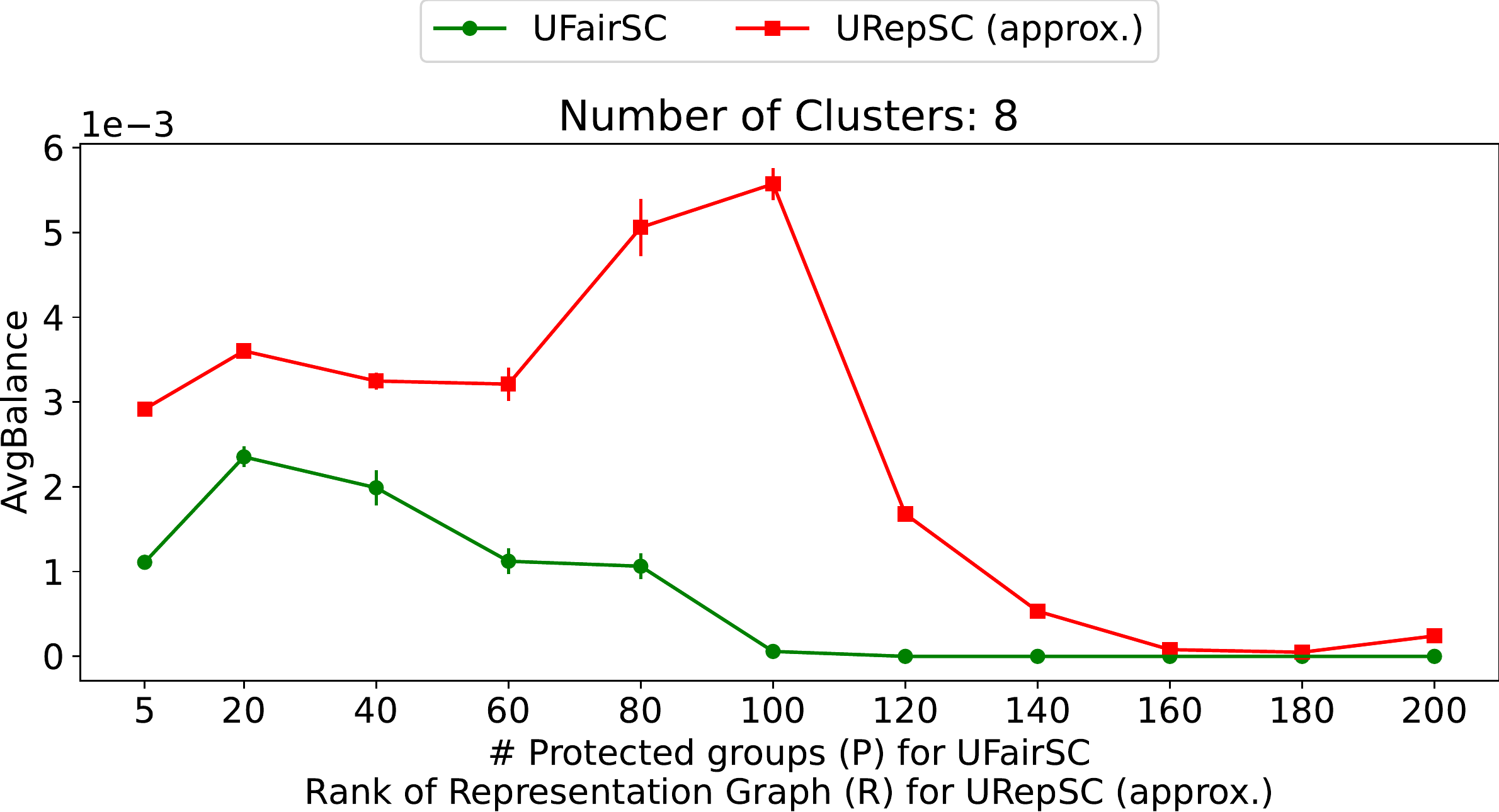}}%
    \hspace{1cm}
    \subfloat[][Ratio-cut, $K=8$]{\includegraphics[width=0.45\textwidth]{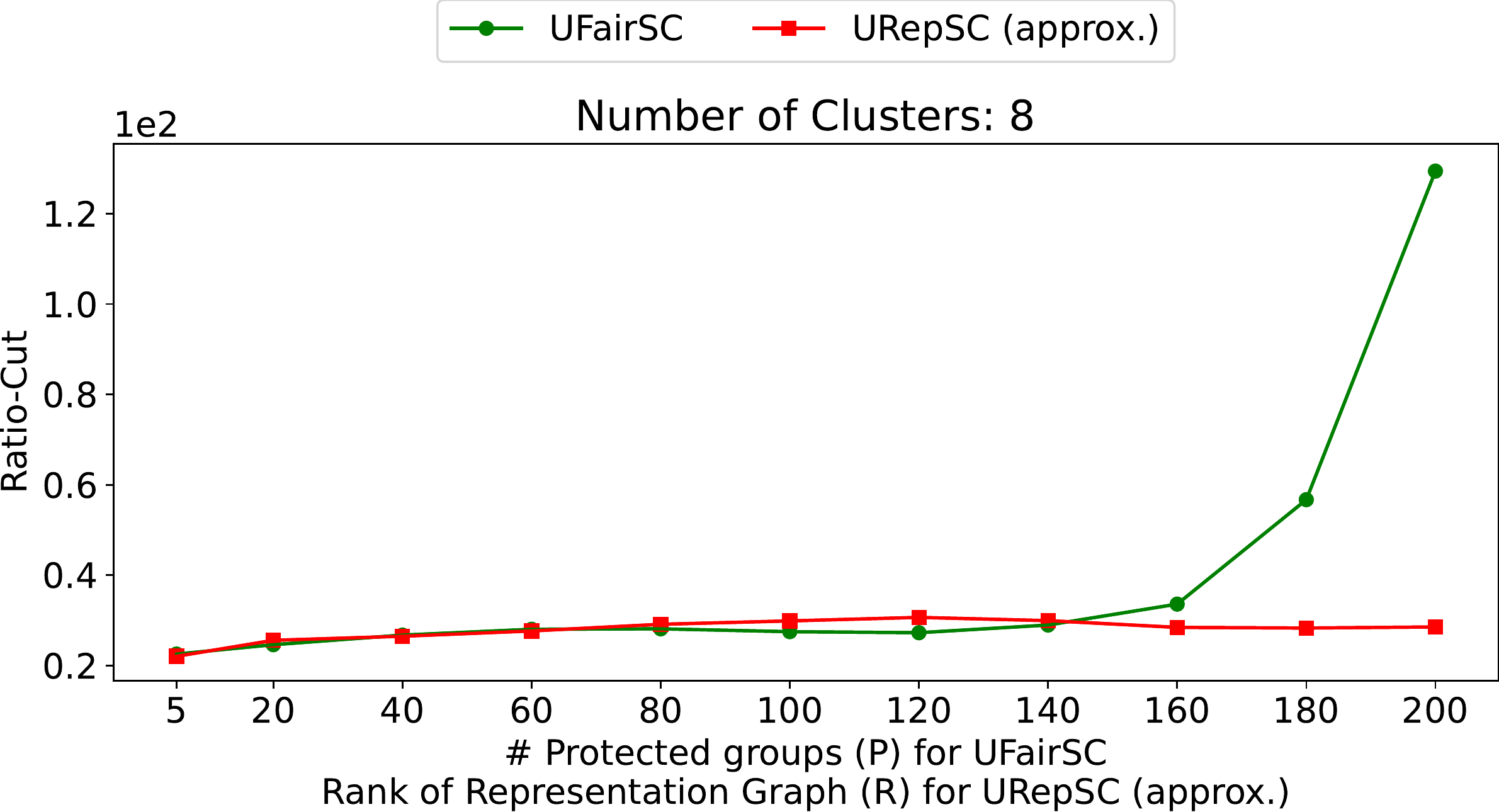}}

    \caption{\rebuttal{Comparing \textsc{URepSC} with other ``unnormalized'' algorithms on the FAO trade network.}}
    \label{fig:real_data_comparison_unnorm_separated}
\end{figure}

\begin{figure}[t]
    \centering
    \subfloat[][Average balance, $K=2$]{\includegraphics[width=0.45\textwidth]{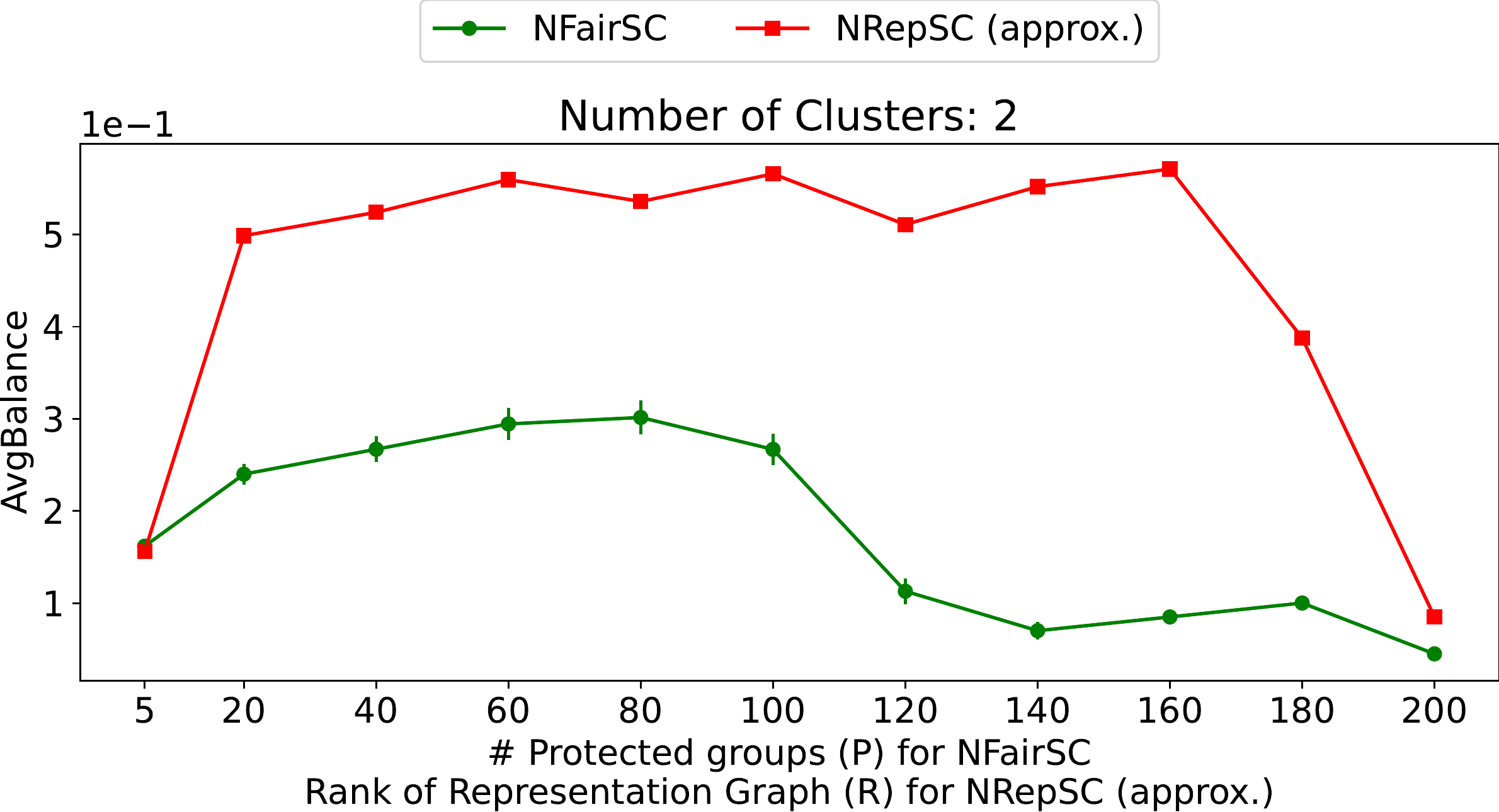}}%
    \hspace{1cm}
    \subfloat[][Ratio-cut, $K=2$]{\includegraphics[width=0.45\textwidth]{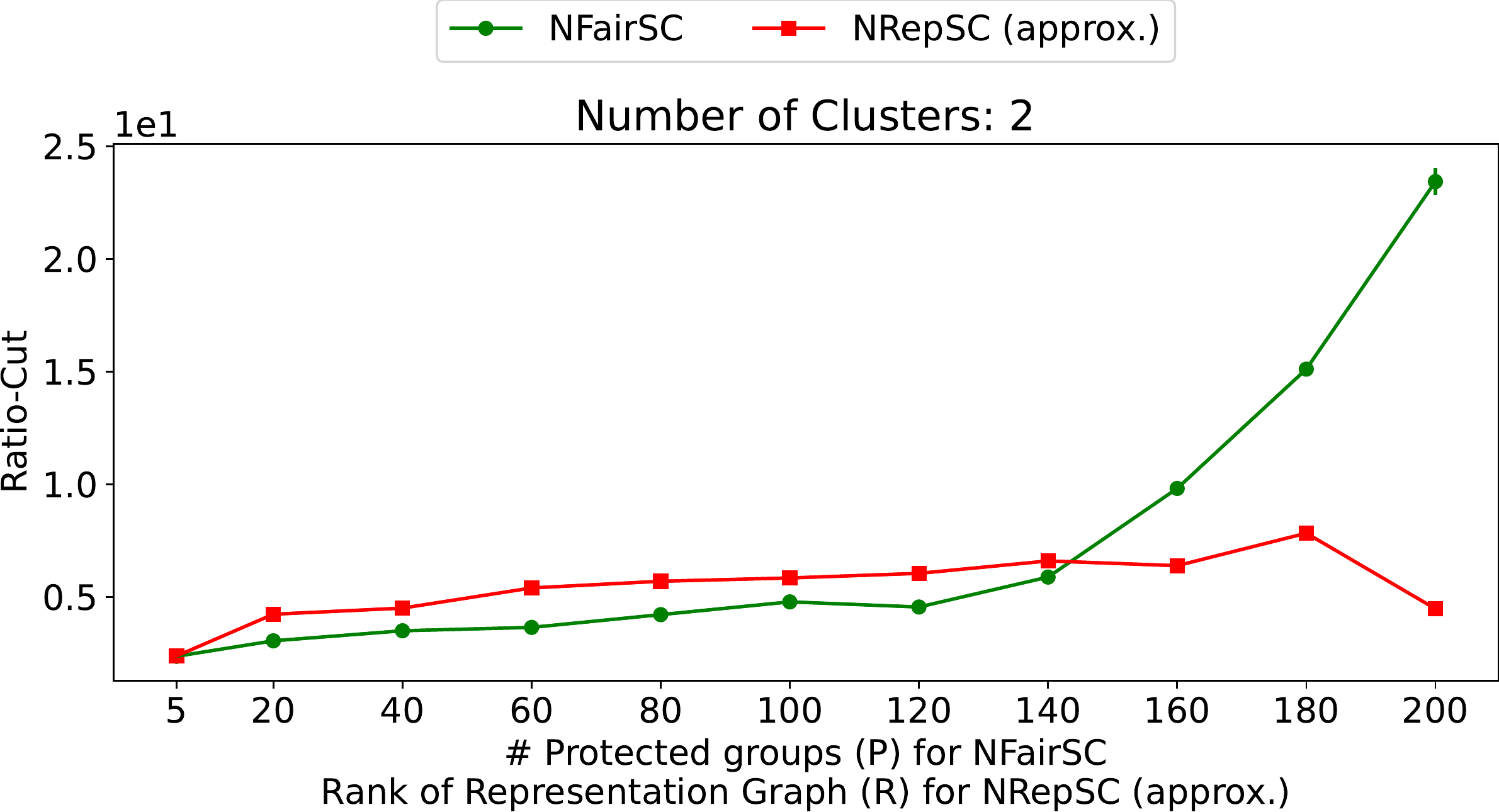}}

    \subfloat[][Average balance, $K=4$]{\includegraphics[width=0.45\textwidth]{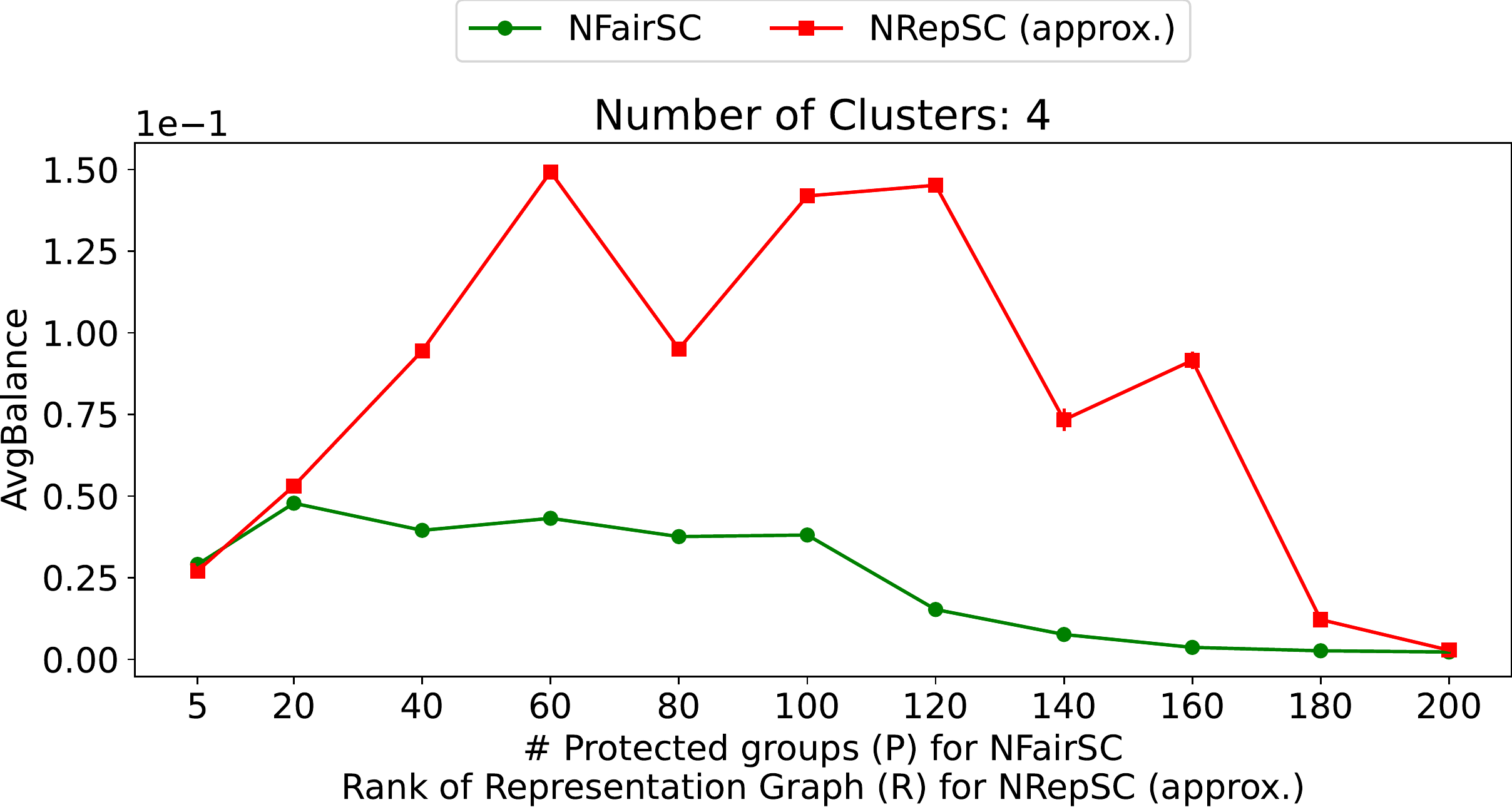}}%
    \hspace{1cm}
    \subfloat[][Ratio-cut, $K=4$]{\includegraphics[width=0.45\textwidth]{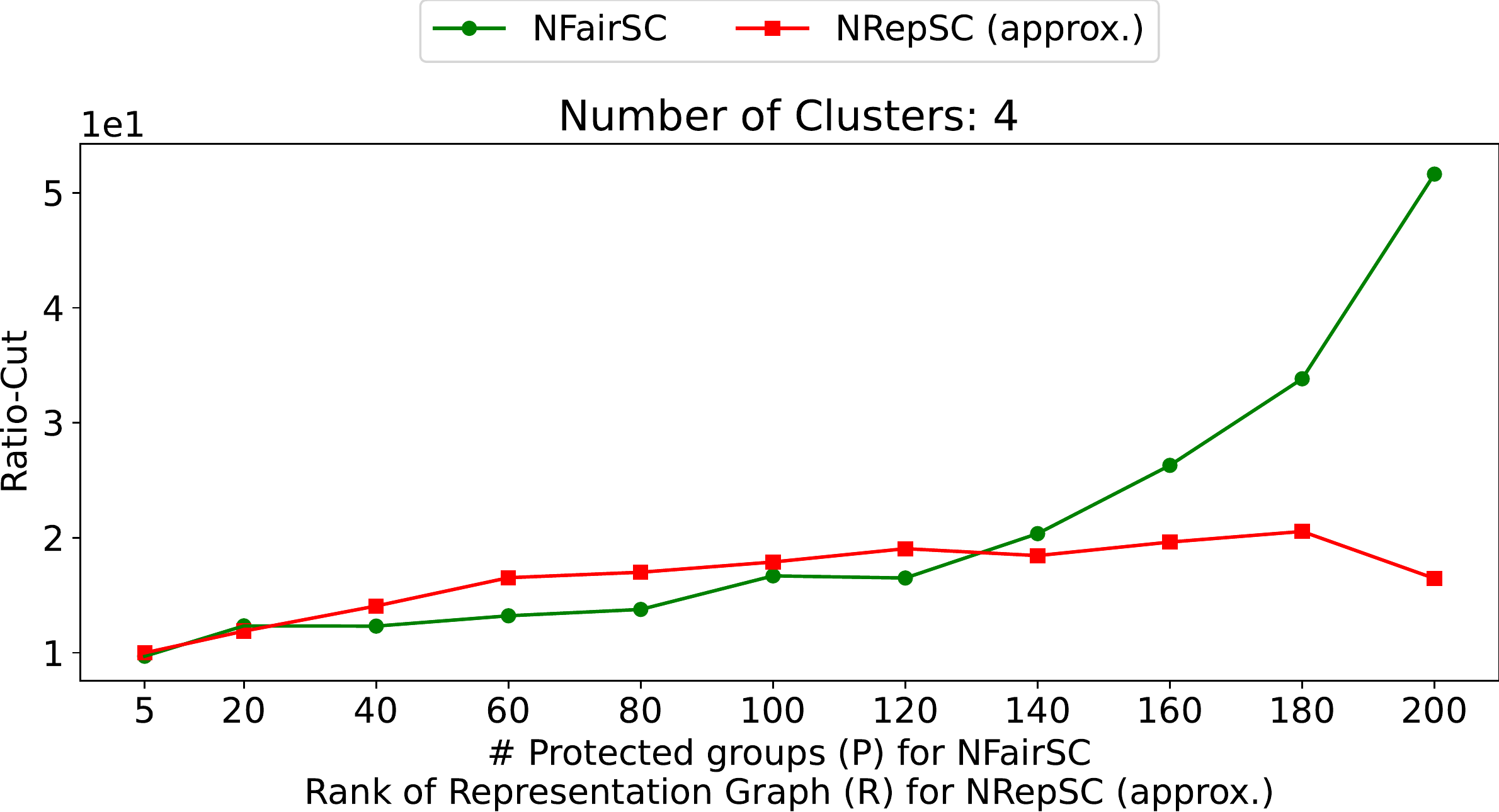}}

    \subfloat[][Average balance, $K=6$]{\includegraphics[width=0.45\textwidth]{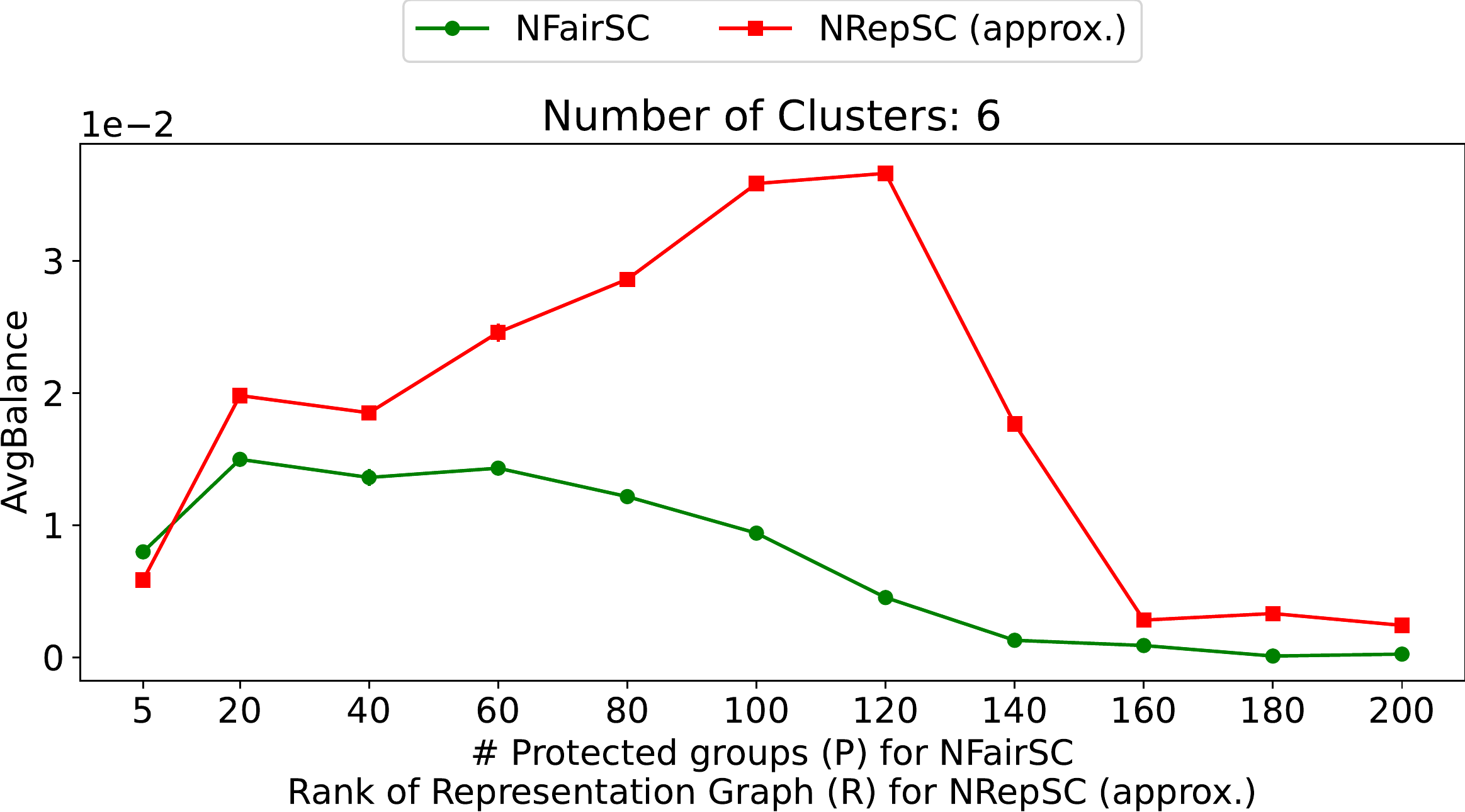}}%
    \hspace{1cm}
    \subfloat[][Ratio-cut, $K=6$]{\includegraphics[width=0.45\textwidth]{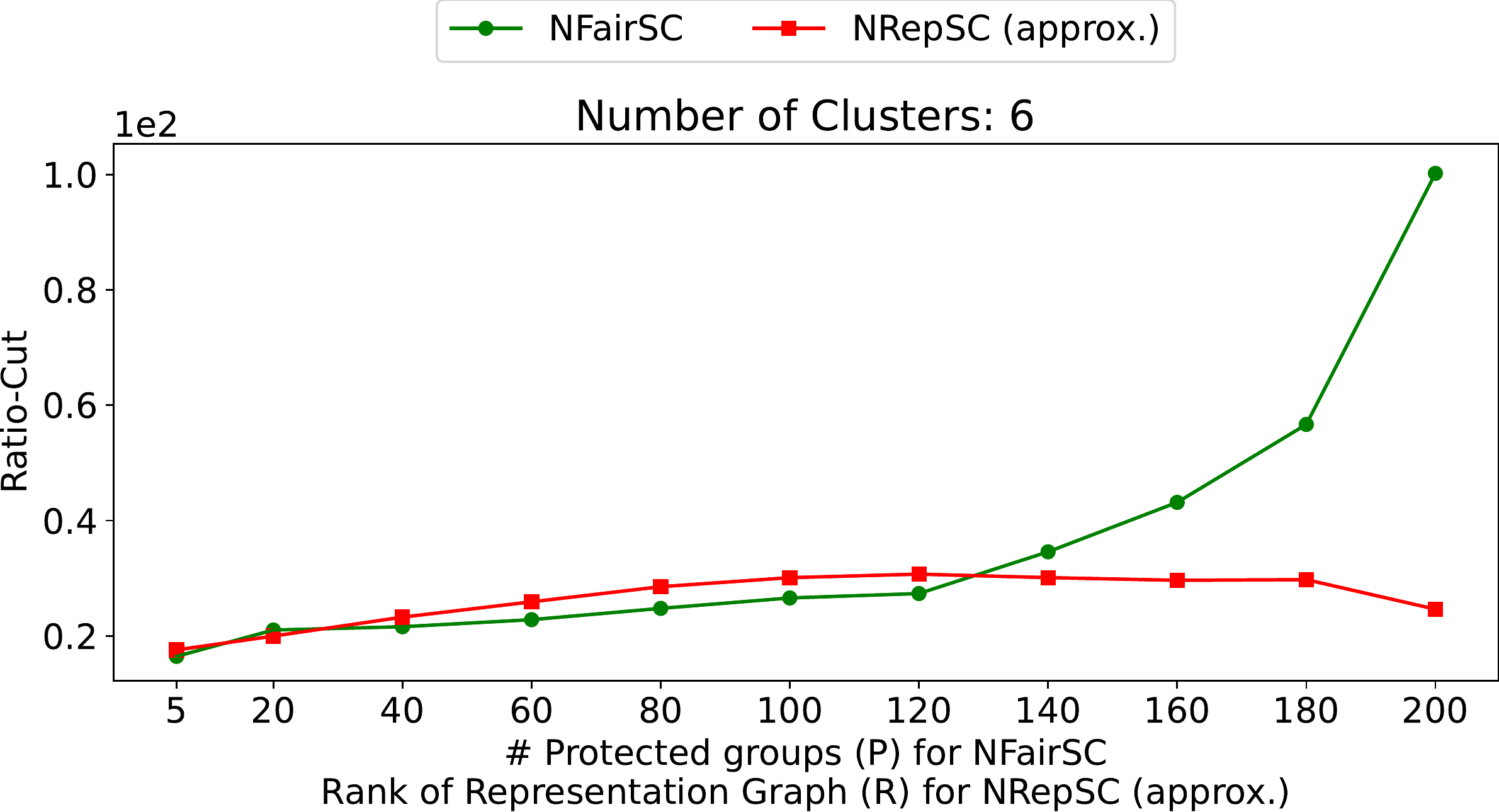}}

    \subfloat[][Average balance, $K=8$]{\includegraphics[width=0.45\textwidth]{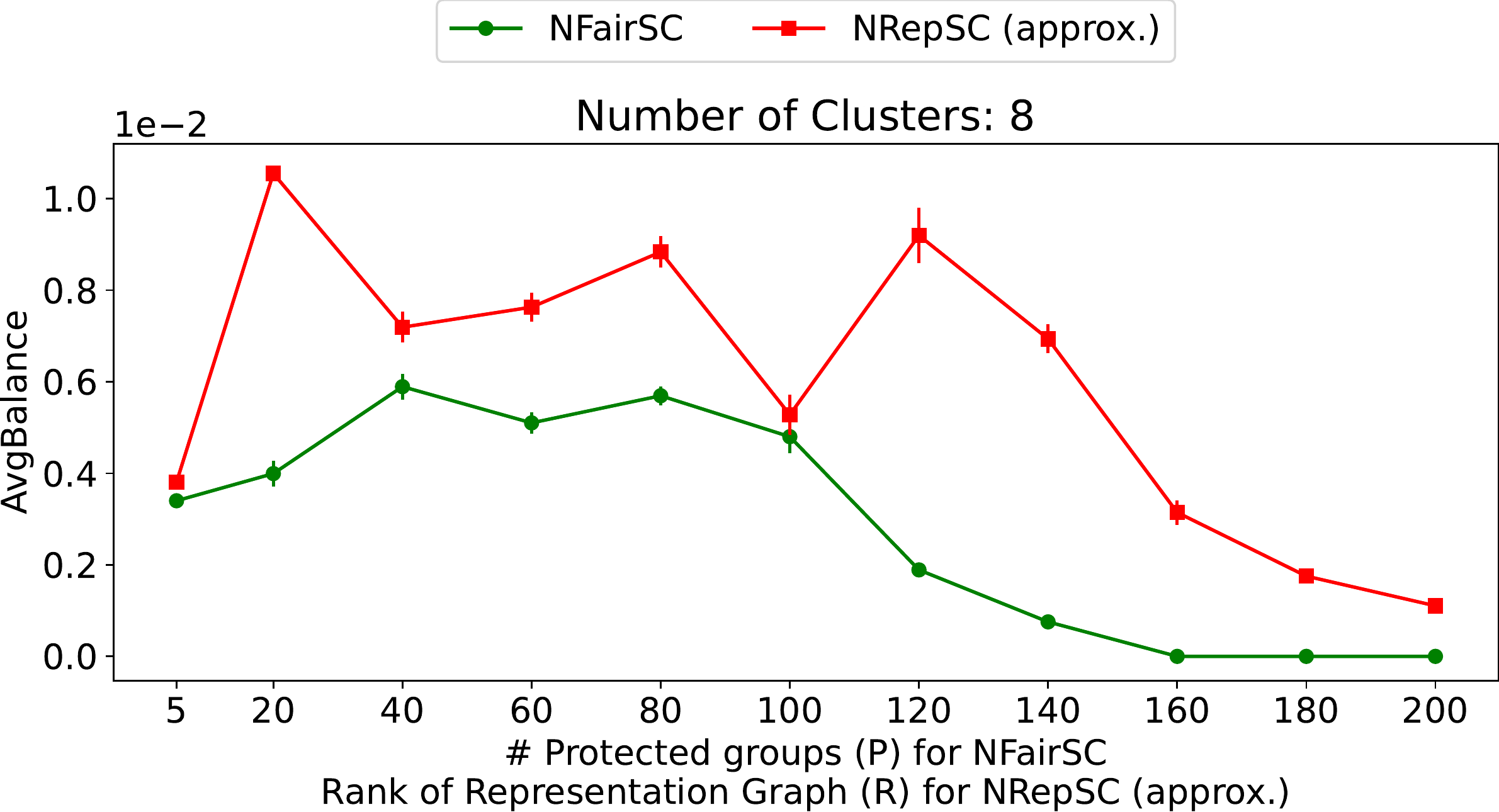}}%
    \hspace{1cm}
    \subfloat[][Ratio-cut, $K=8$]{\includegraphics[width=0.45\textwidth]{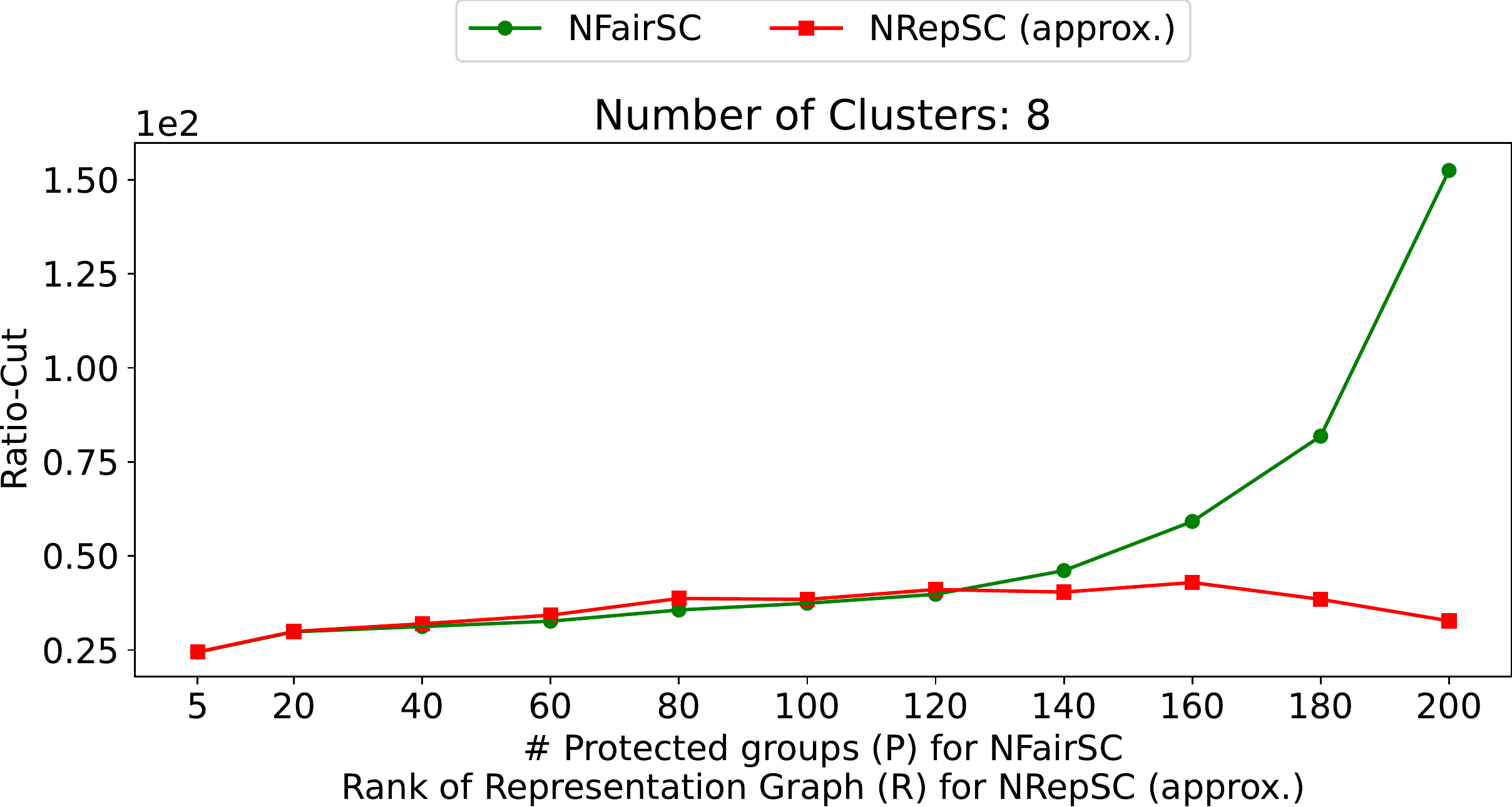}}

    \caption{\rebuttal{Comparing \textsc{NRepSC} with other ``normalized'' algorithms on the FAO trade network.}}
    \label{fig:real_data_comparison_norm_separated}
\end{figure}

\begin{figure}[t]
    \centering
    \subfloat[][Average balance, $K=2$]{\includegraphics[width=0.45\textwidth]{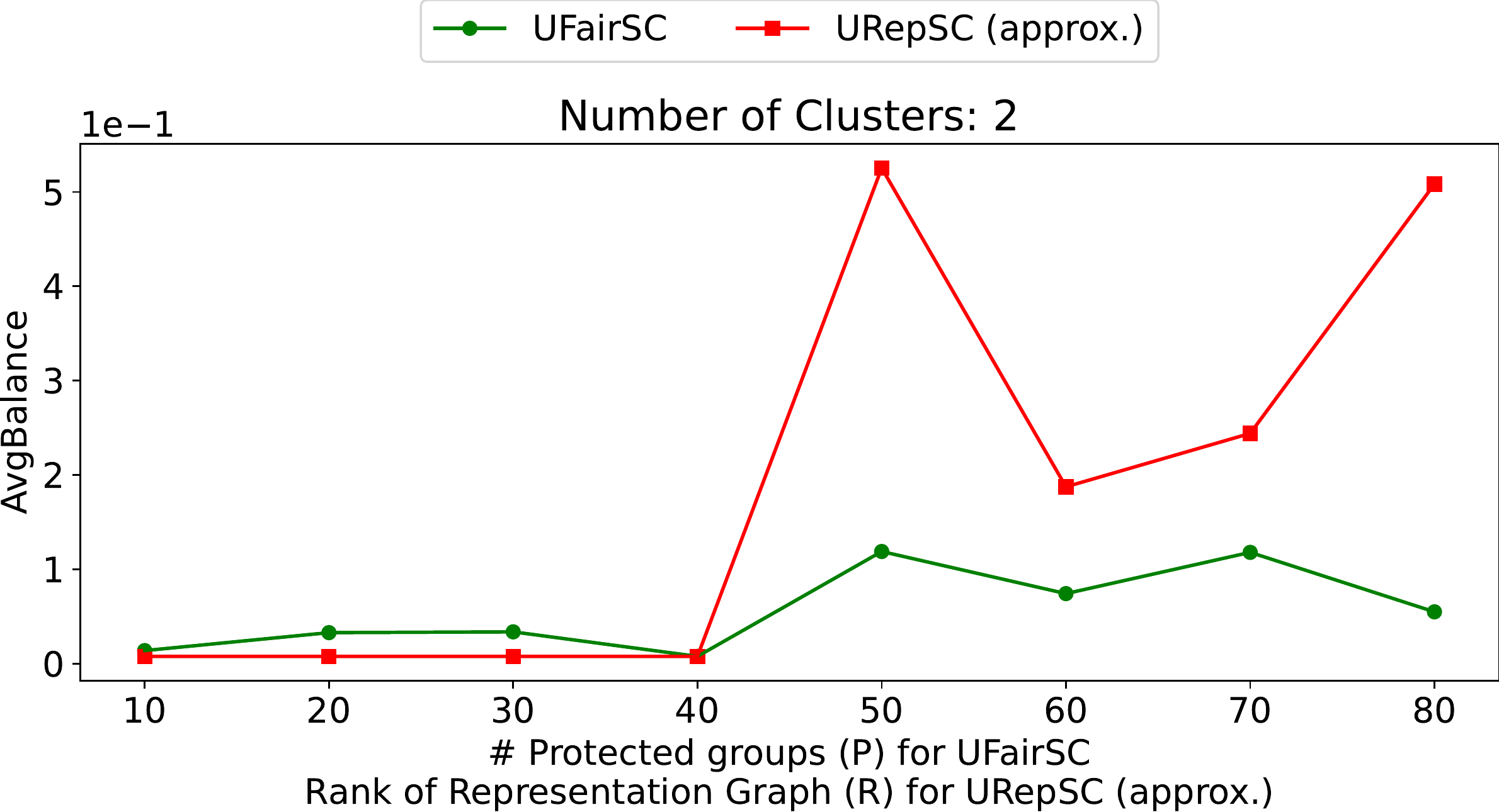}}%
    \hspace{1cm}
    \subfloat[][Ratio-cut, $K=2$]{\includegraphics[width=0.45\textwidth]{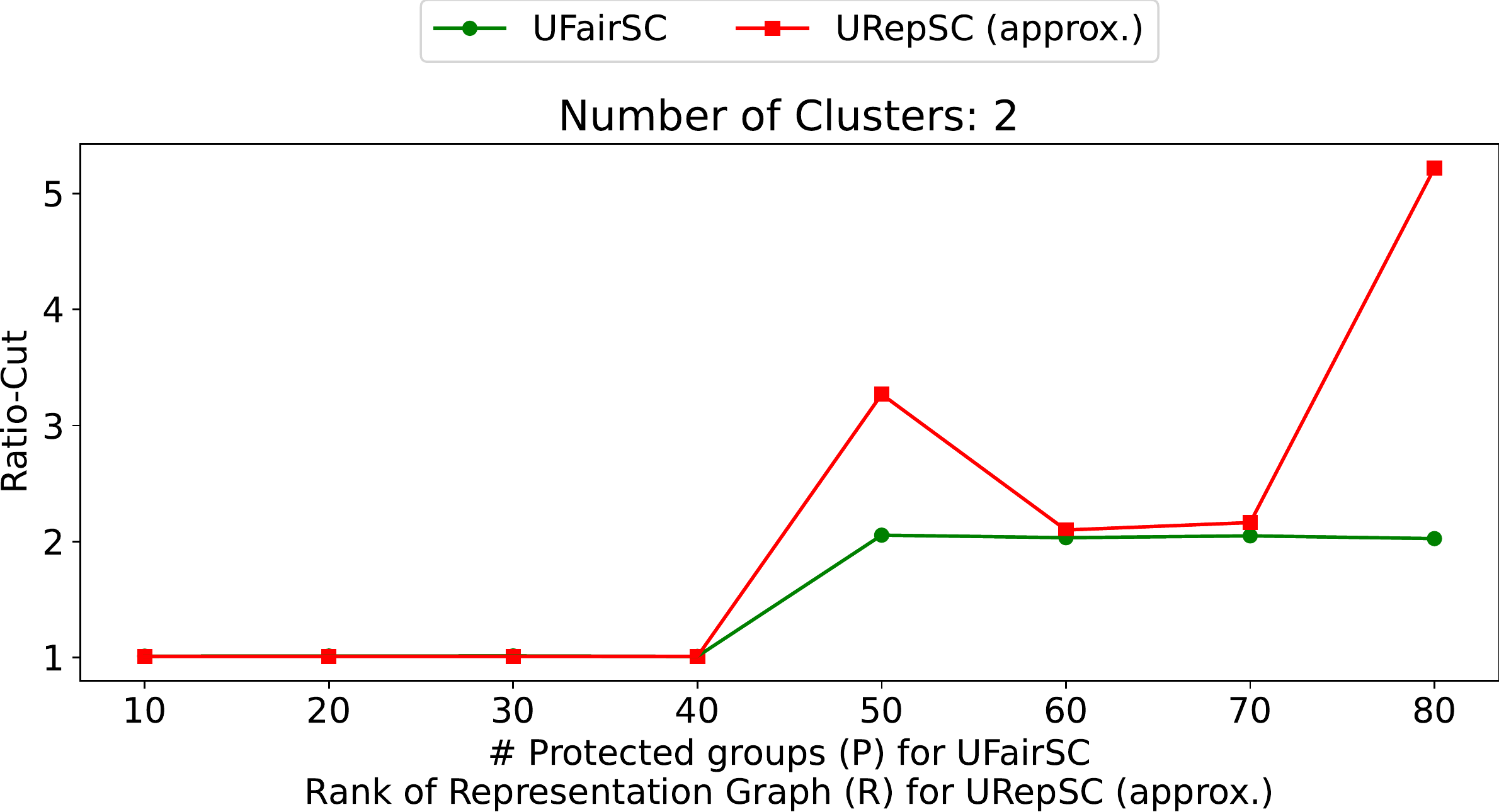}}

    \subfloat[][Average balance, $K=4$]{\includegraphics[width=0.45\textwidth]{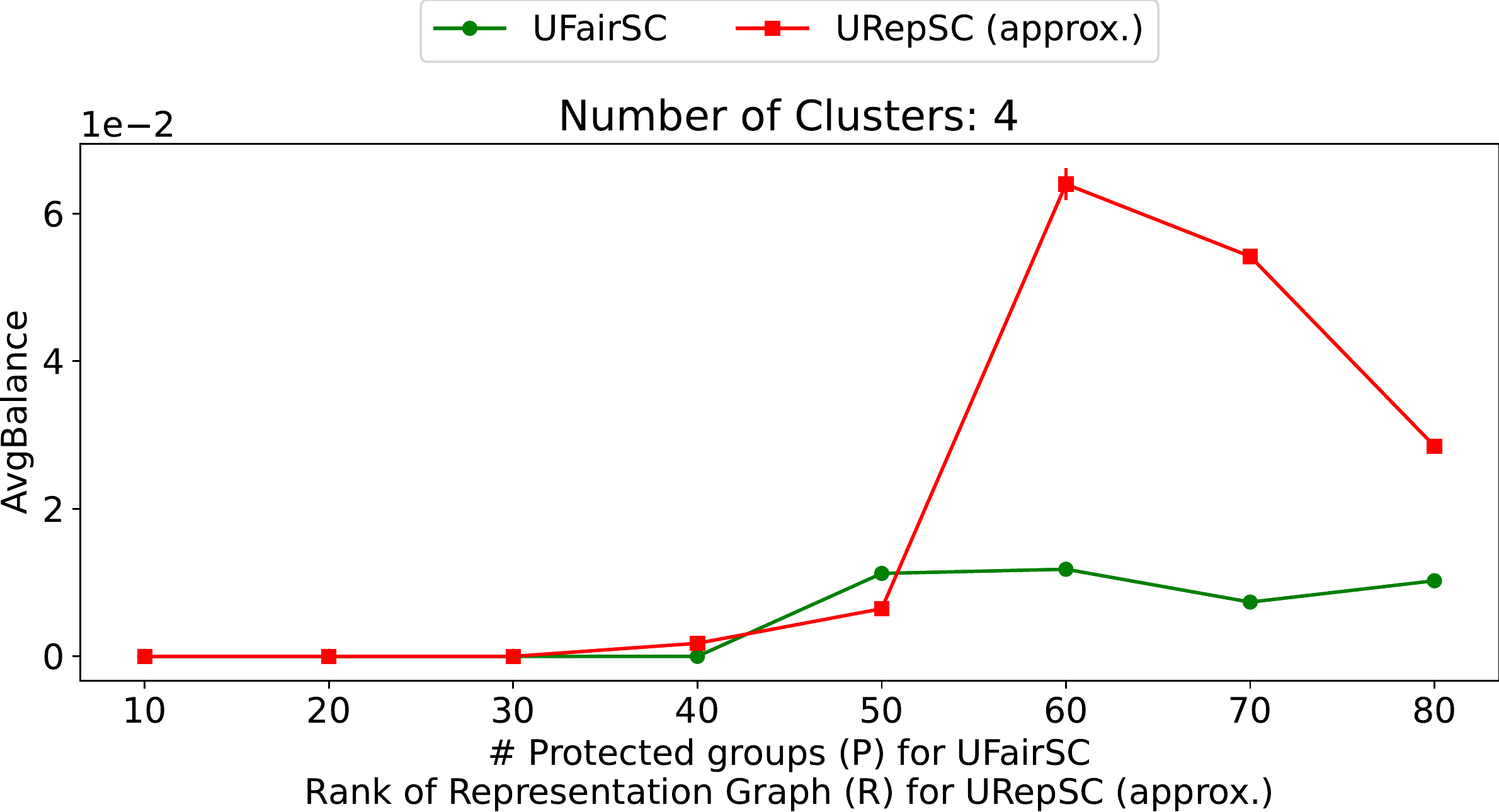}}%
    \hspace{1cm}
    \subfloat[][Ratio-cut, $K=4$]{\includegraphics[width=0.45\textwidth]{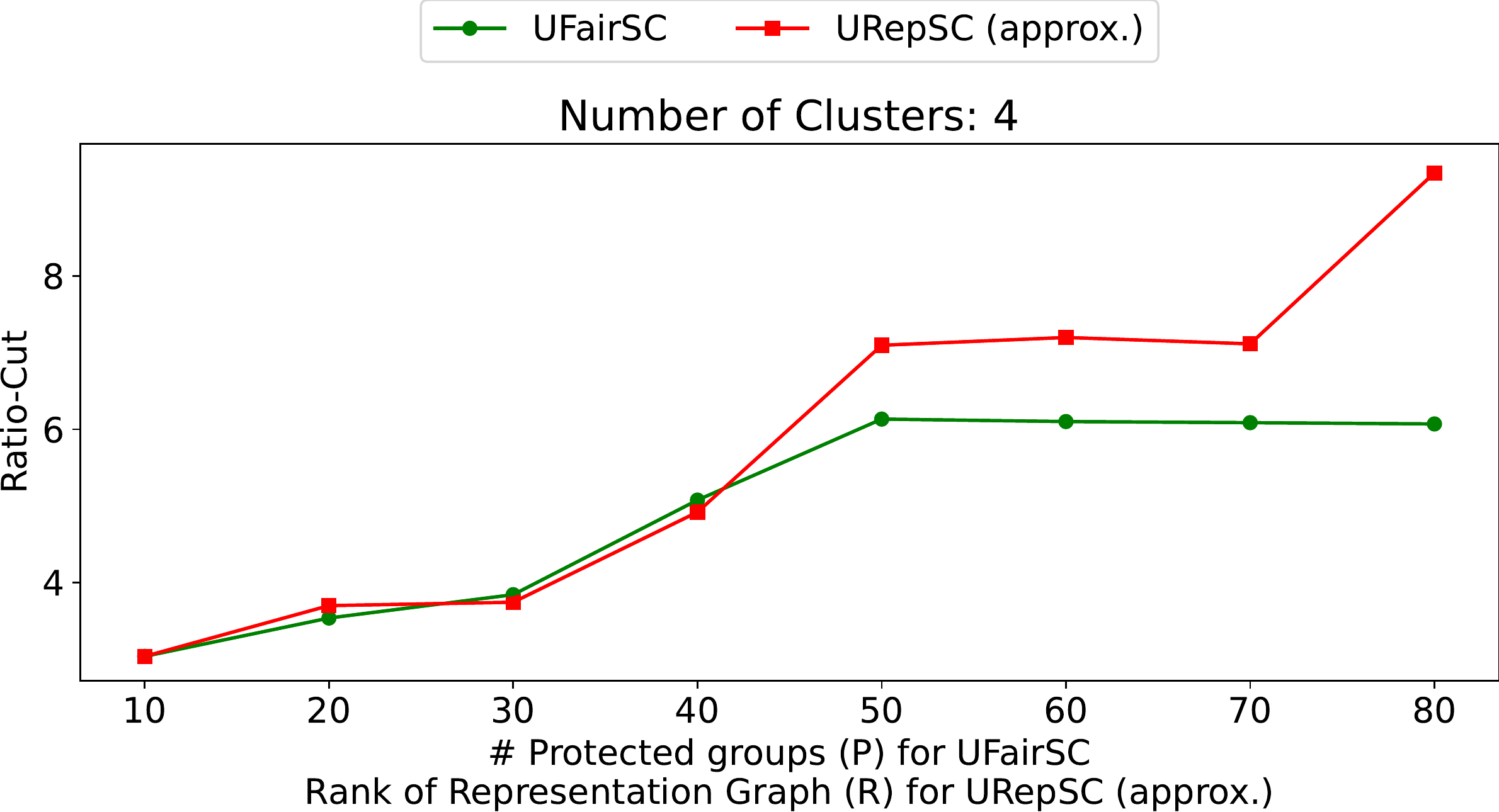}}

    \subfloat[][Average balance, $K=6$]{\includegraphics[width=0.45\textwidth]{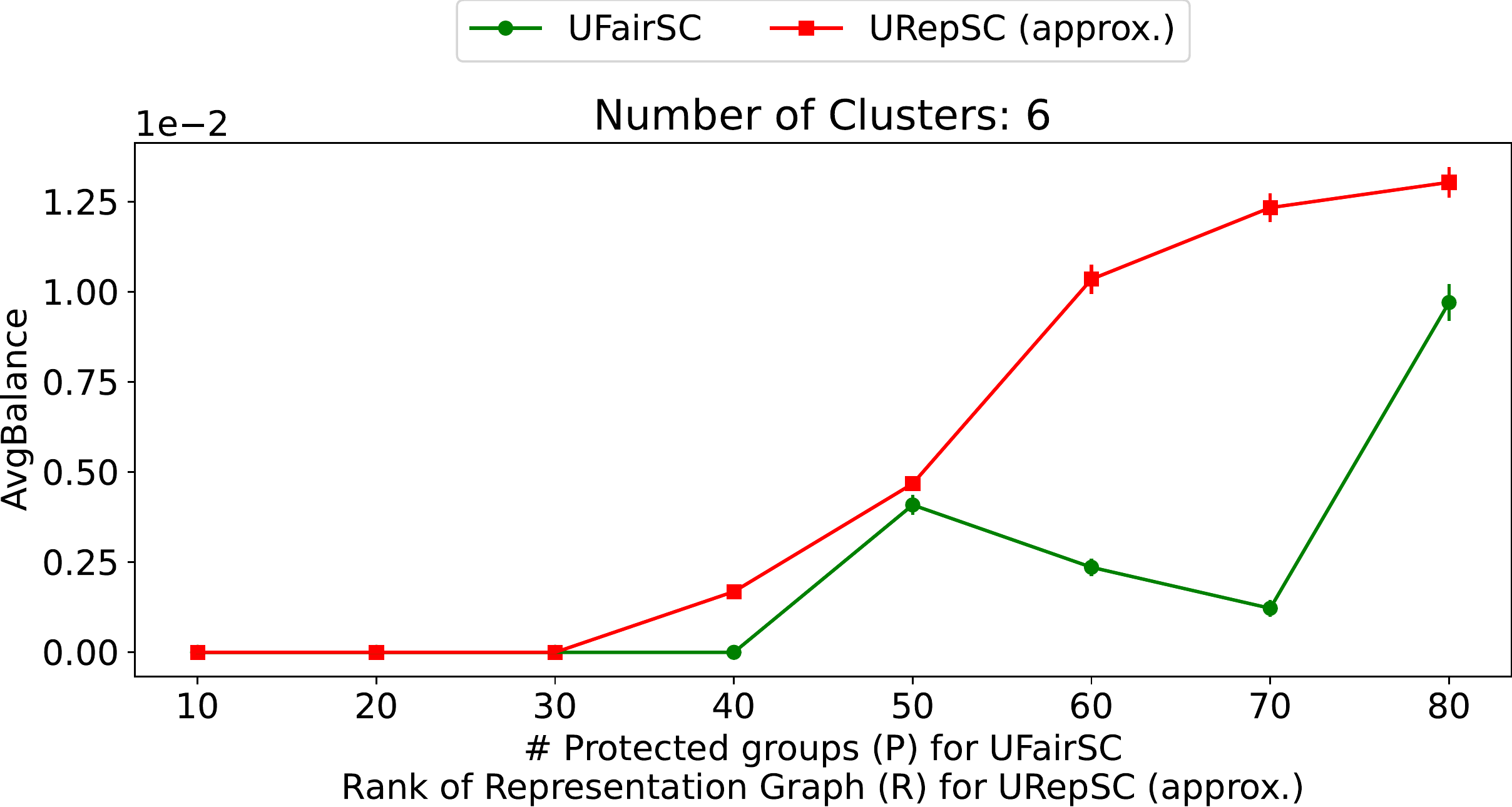}}%
    \hspace{1cm}
    \subfloat[][Ratio-cut, $K=6$]{\includegraphics[width=0.45\textwidth]{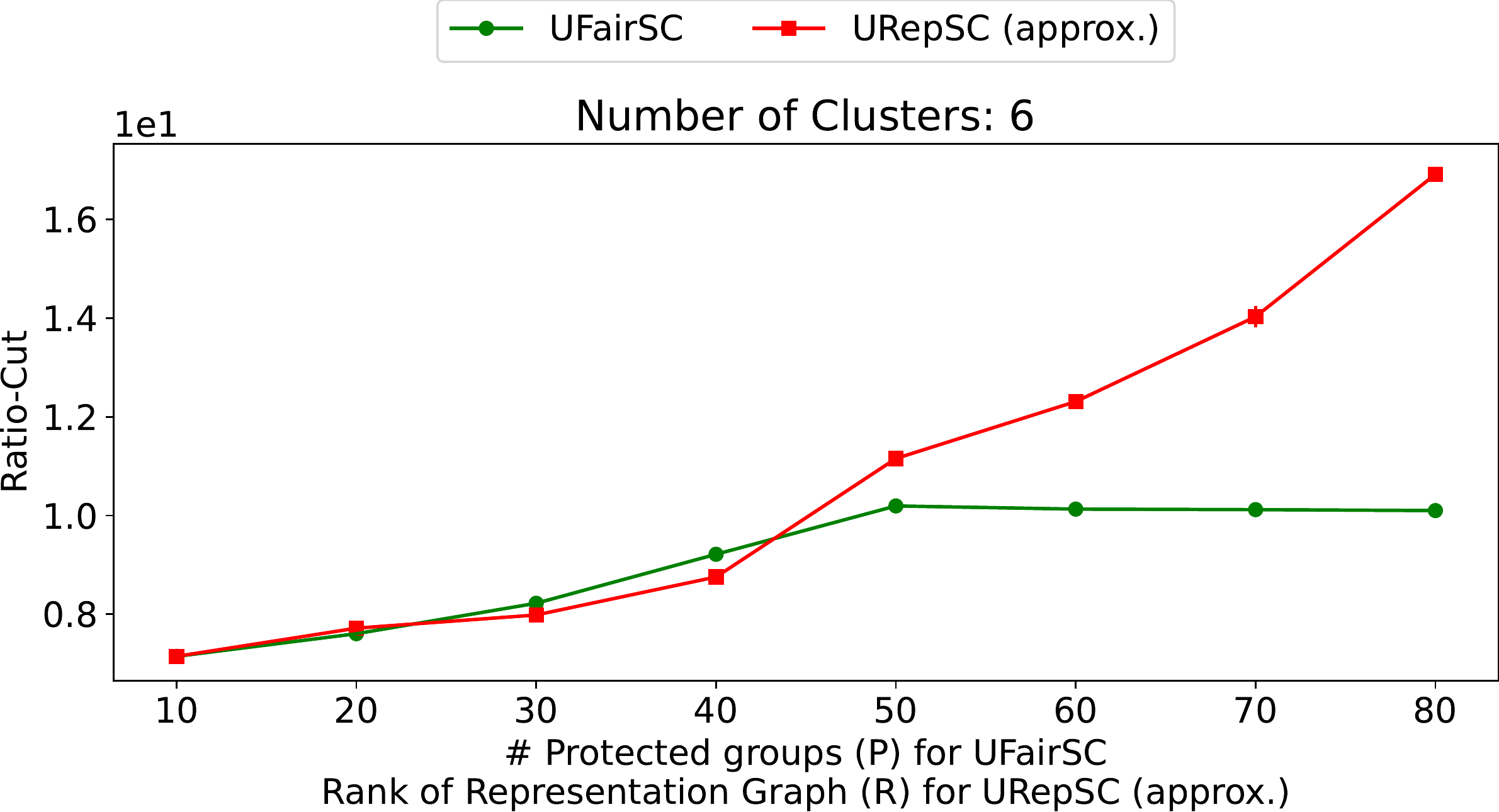}}

    \subfloat[][Average balance, $K=8$]{\includegraphics[width=0.45\textwidth]{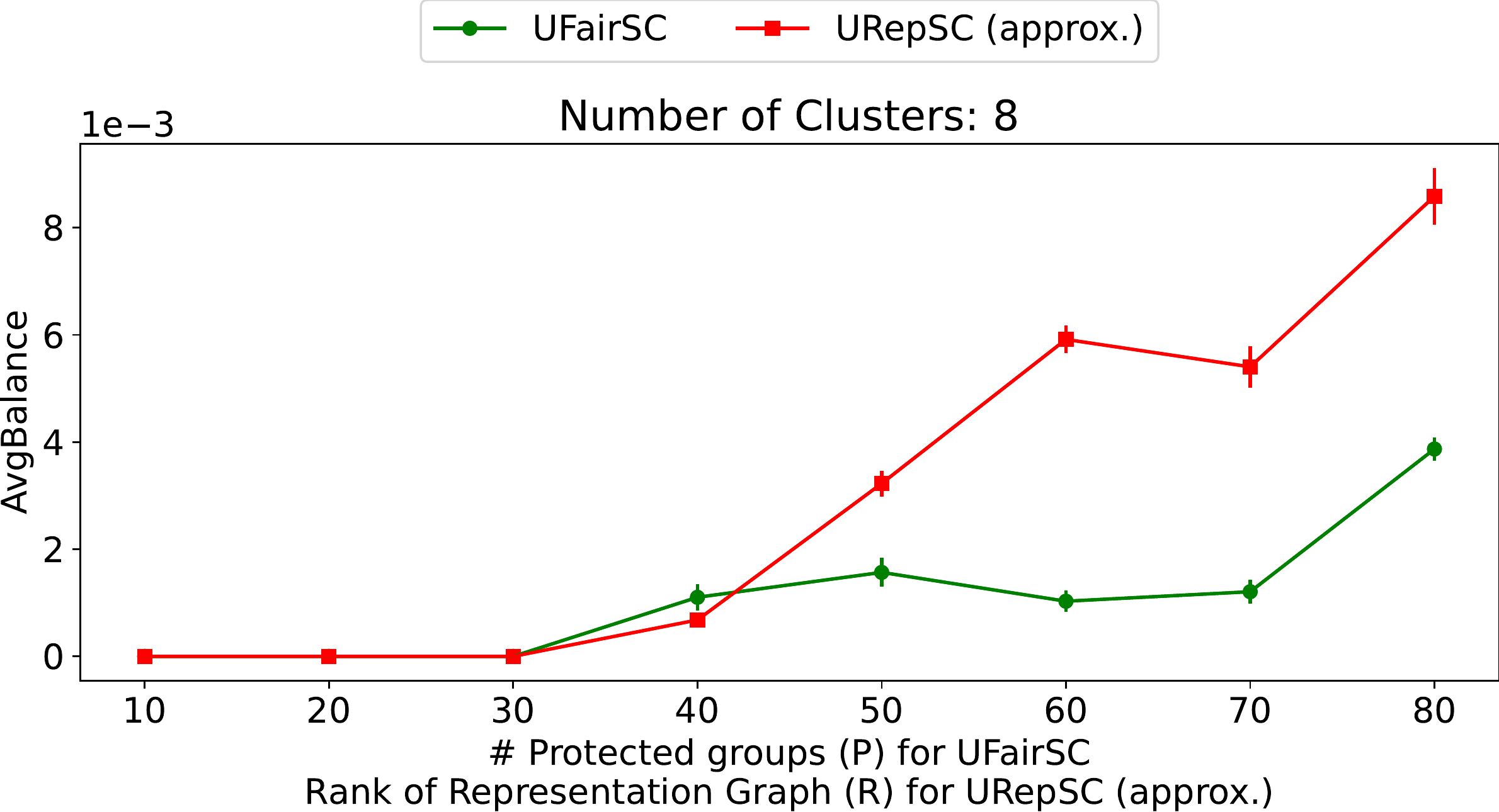}}%
    \hspace{1cm}
    \subfloat[][Ratio-cut, $K=8$]{\includegraphics[width=0.45\textwidth]{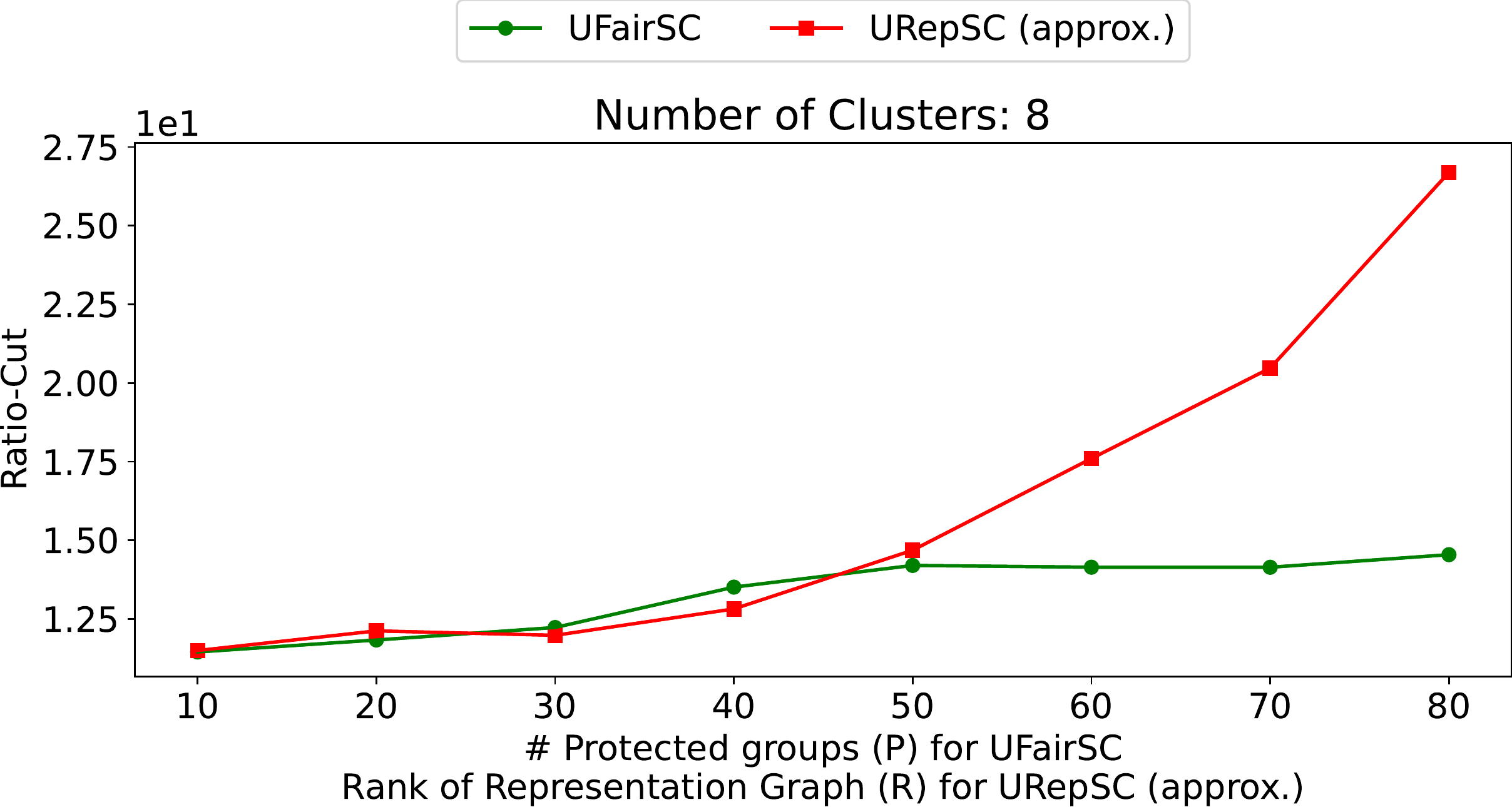}}

    \caption{\rebuttal{Comparing \textsc{URepSC} with other ``unnormalized'' algorithms on the air transportation network.}}
    \label{fig:atn_comparison_unnorm_separated}
\end{figure}

\begin{figure}[t]
    \centering
    \subfloat[][Average balance, $K=2$]{\includegraphics[width=0.45\textwidth]{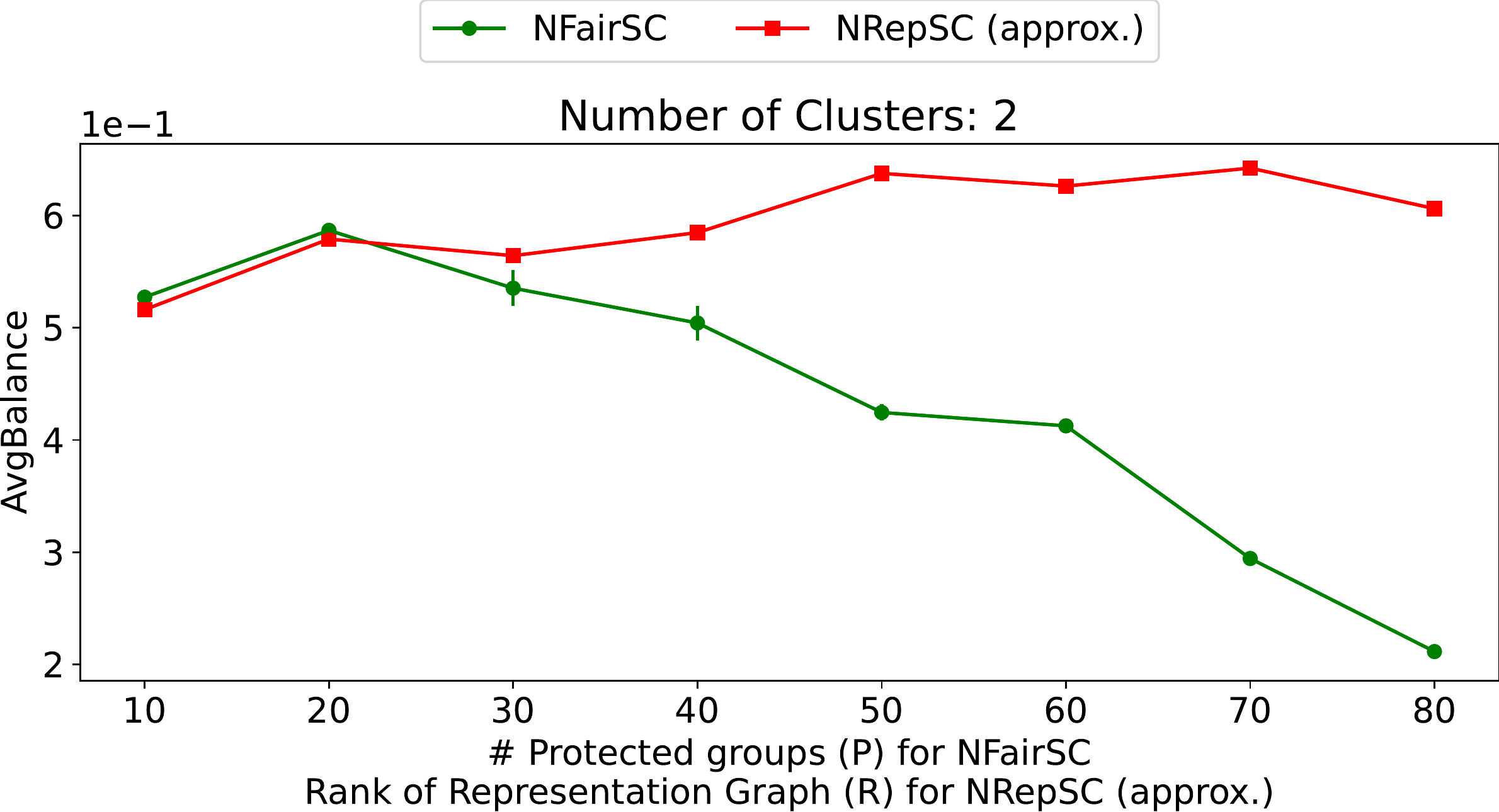}}%
    \hspace{1cm}
    \subfloat[][Ratio-cut, $K=2$]{\includegraphics[width=0.45\textwidth]{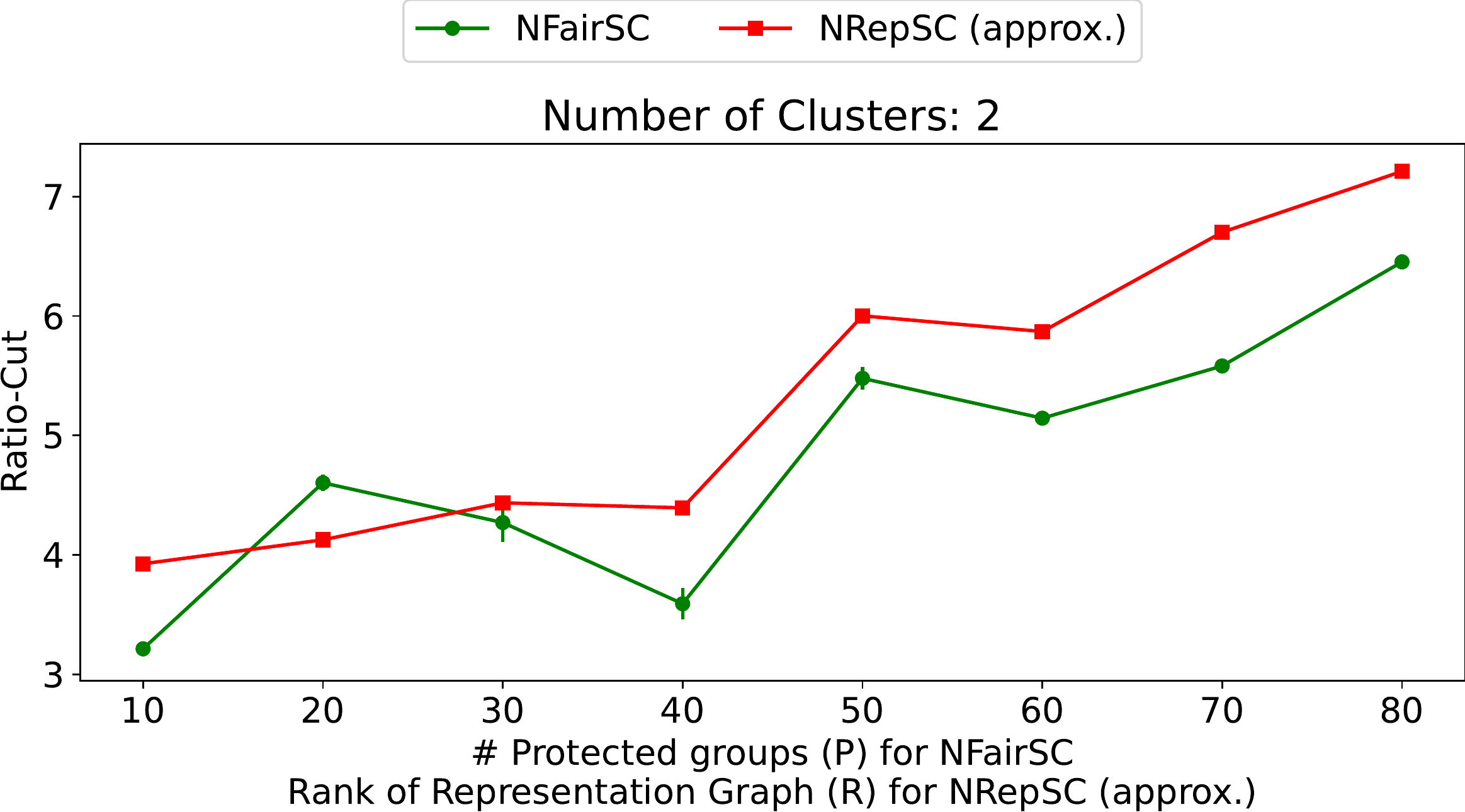}}

    \subfloat[][Average balance, $K=4$]{\includegraphics[width=0.45\textwidth]{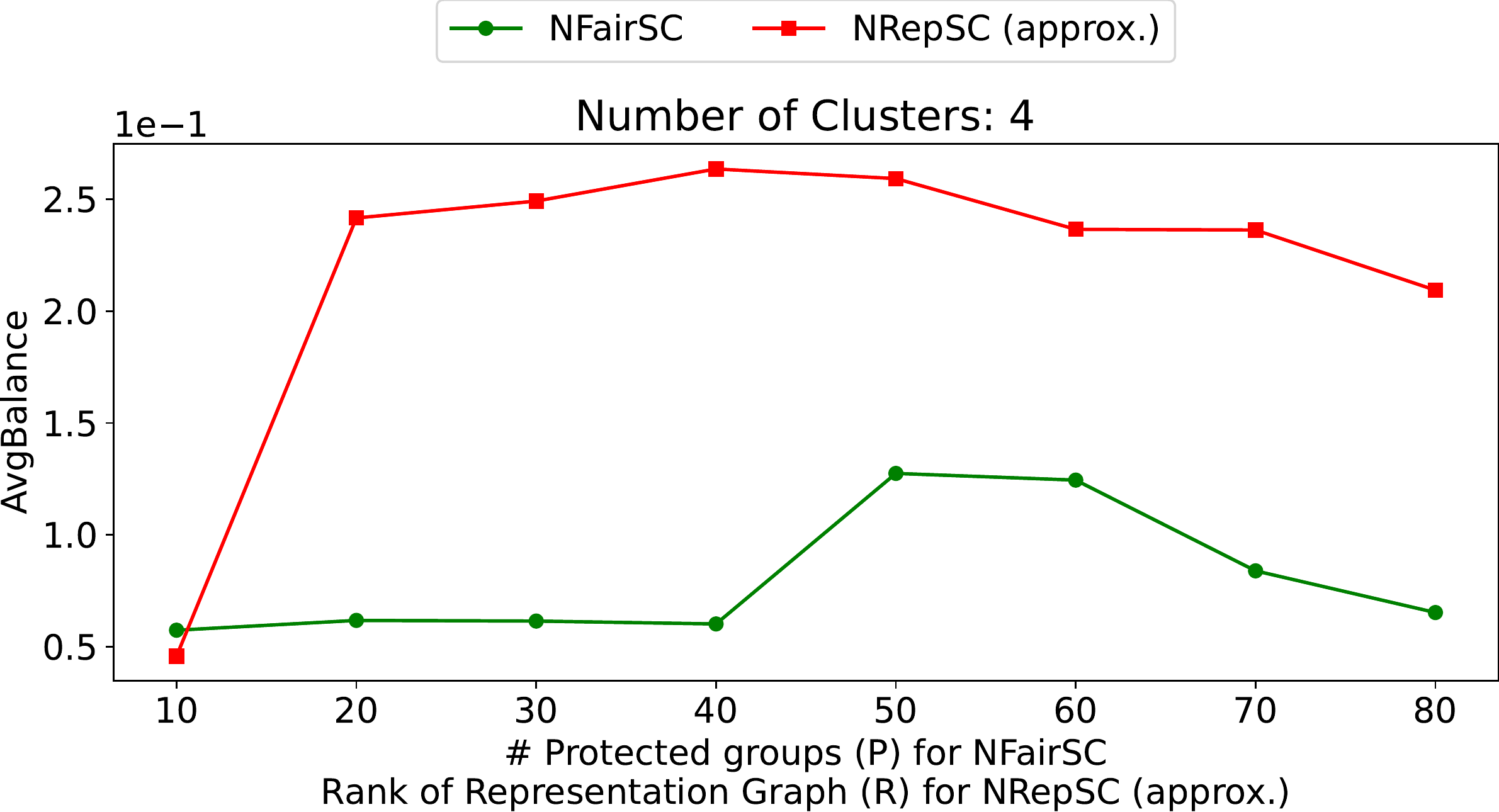}}%
    \hspace{1cm}
    \subfloat[][Ratio-cut, $K=4$]{\includegraphics[width=0.45\textwidth]{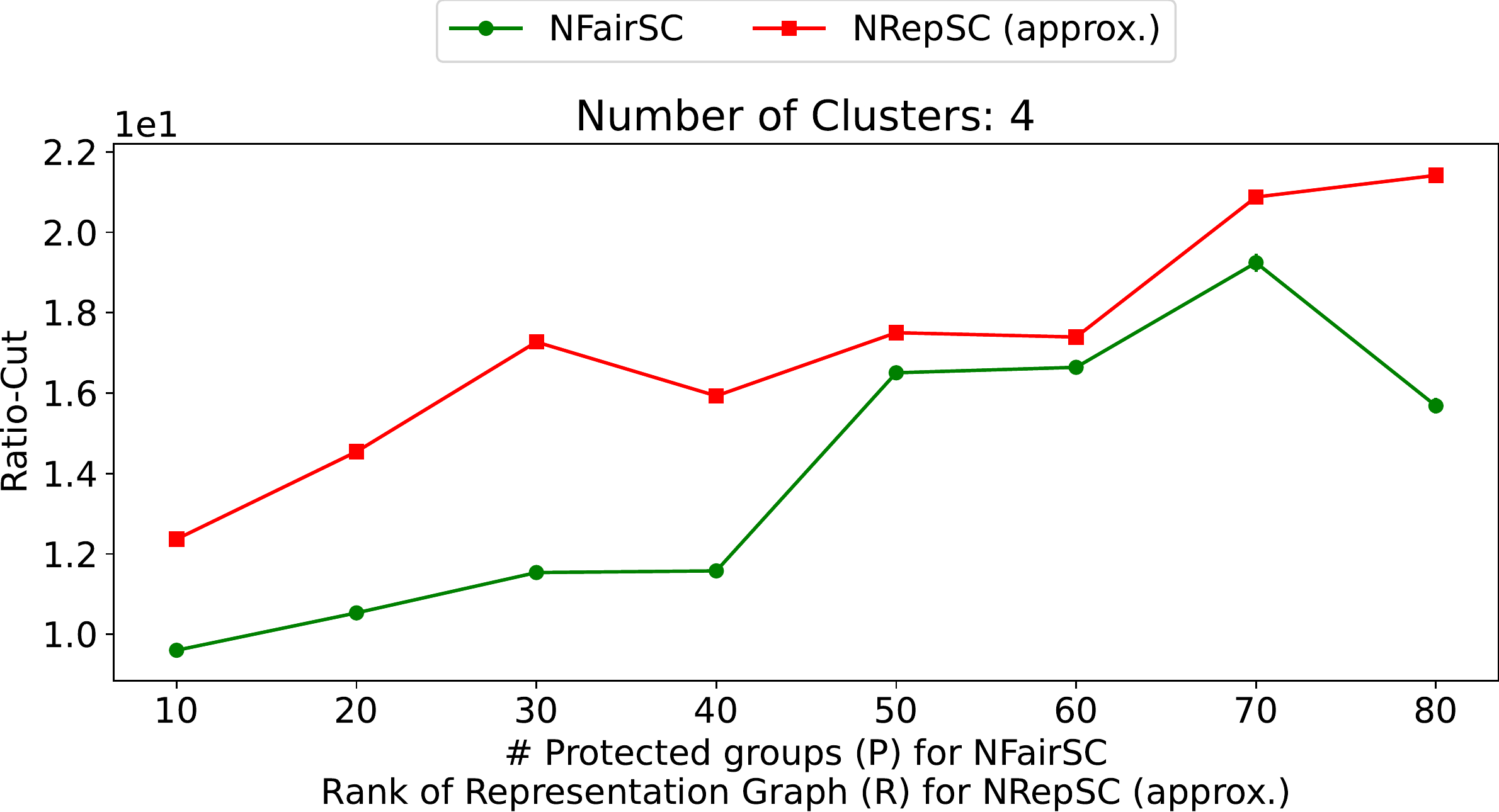}}

    \subfloat[][Average balance, $K=6$]{\includegraphics[width=0.45\textwidth]{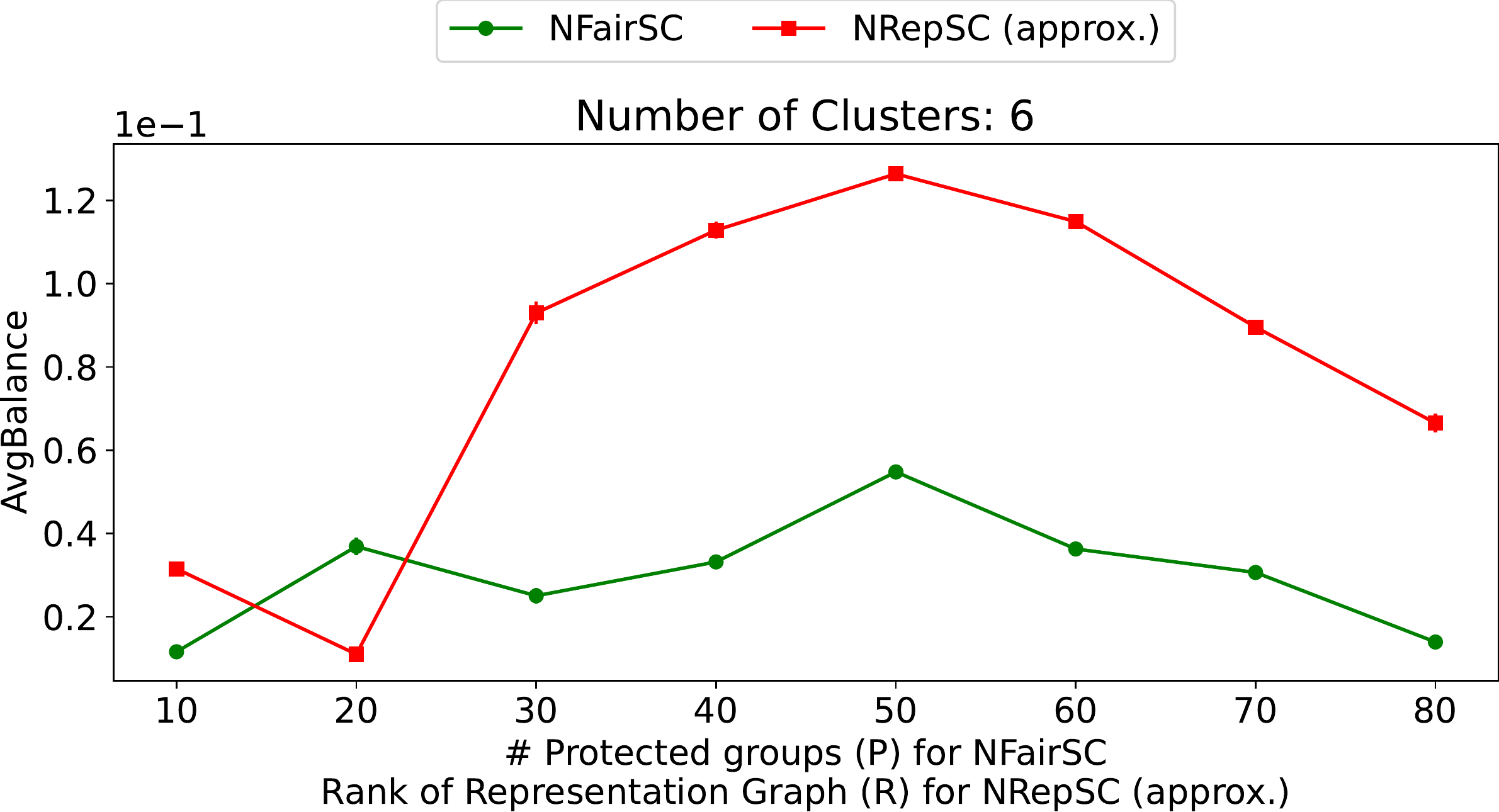}}%
    \hspace{1cm}
    \subfloat[][Ratio-cut, $K=6$]{\includegraphics[width=0.45\textwidth]{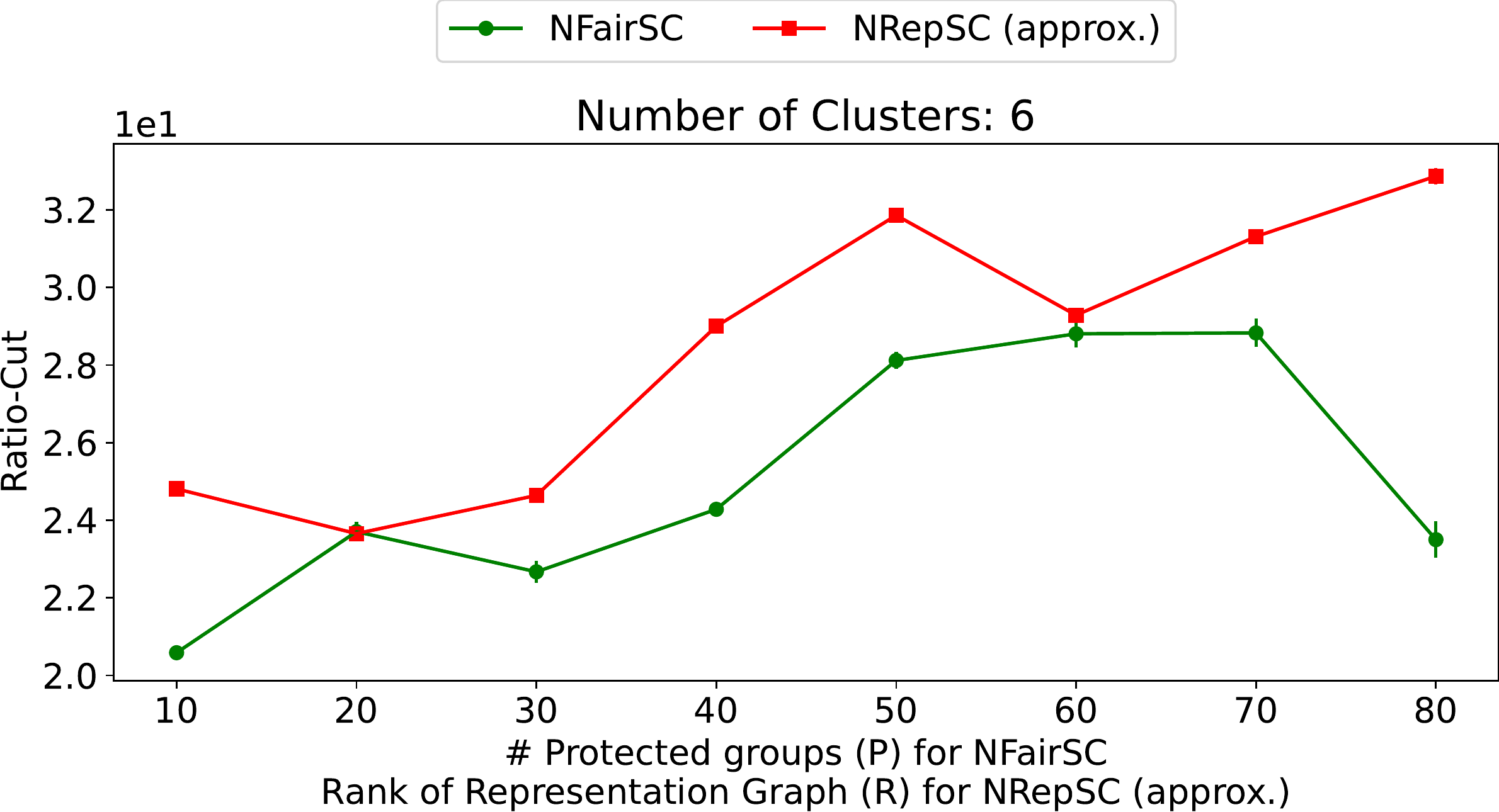}}

    \subfloat[][Average balance, $K=8$]{\includegraphics[width=0.45\textwidth]{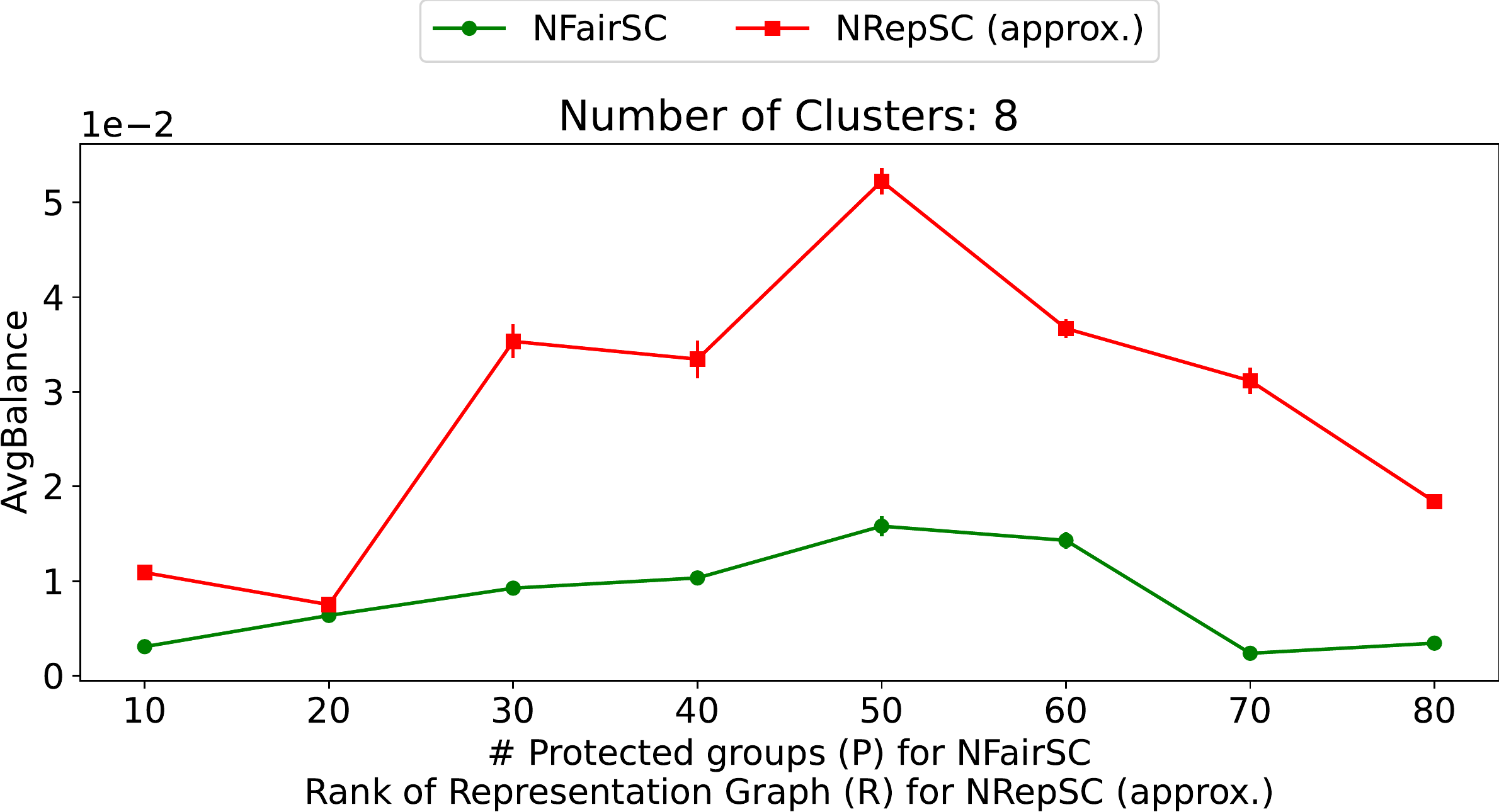}}%
    \hspace{1cm}
    \subfloat[][Ratio-cut, $K=8$]{\includegraphics[width=0.45\textwidth]{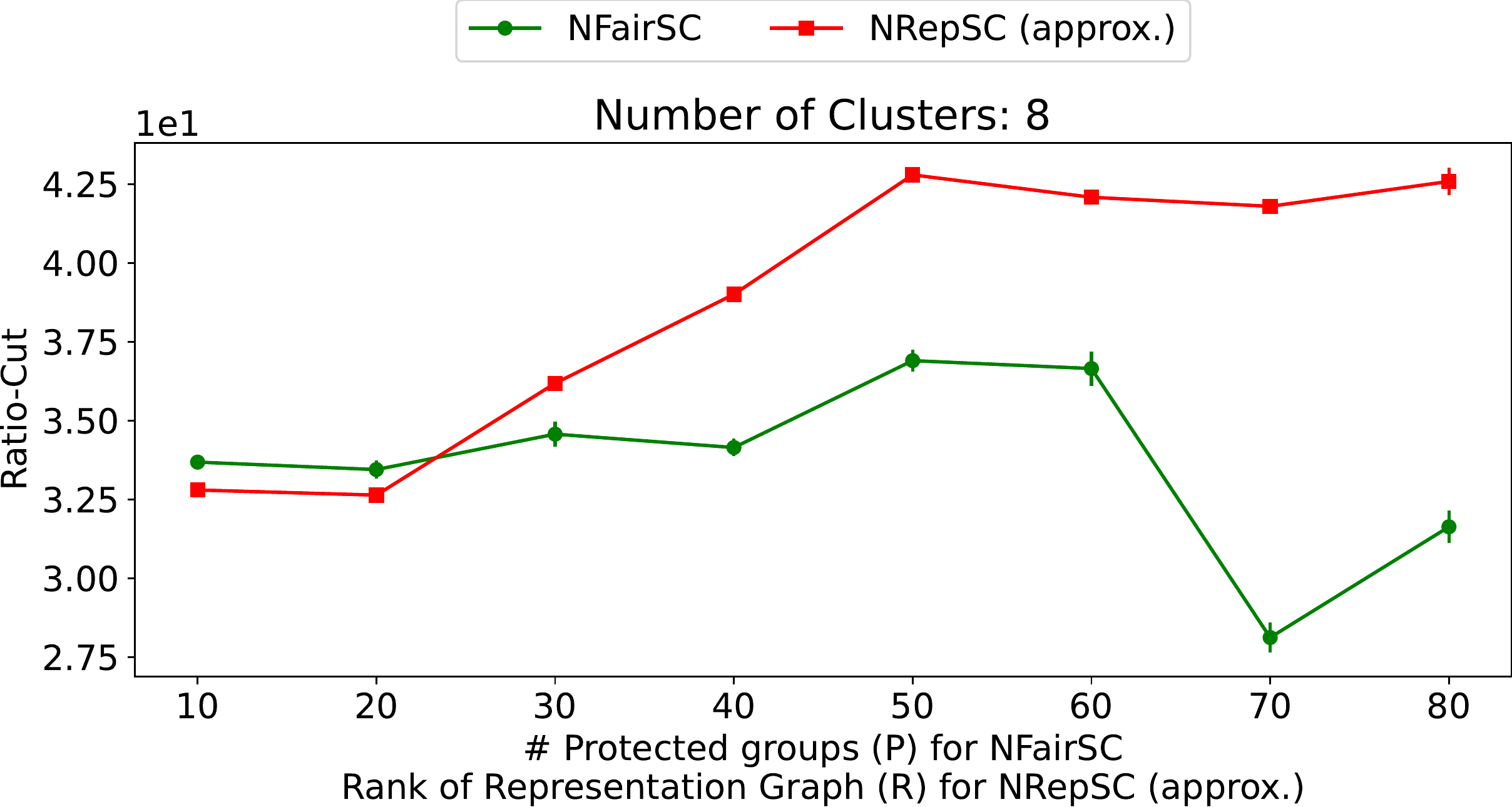}}

    \caption{\rebuttal{Comparing \textsc{NRepSC} with other ``normalized'' algorithms on the air transportation network.}}
    \label{fig:atn_comparison_norm_separated}
\end{figure}

\rebuttal{\Cref{fig:d_reg_unnorm_separated} corresponds to \Cref{fig:d_reg_unnorm}, \Cref{fig:d_reg_norm_separated} to \Cref{fig:d_reg_norm}, \Cref{fig:sbm_comparison_unnorm_separated} to \Cref{fig:sbm_comparison_unnorm_additional}, \Cref{fig:sbm_comparison_norm_separated} to \Cref{fig:sbm_comparison_norm}, \Cref{fig:real_data_comparison_unnorm_separated} to \Cref{fig:real_data_comparison_unnorm_additional}, \Cref{fig:real_data_comparison_norm_separated} to \Cref{fig:real_data_comparison_norm}, \Cref{fig:atn_comparison_unnorm_separated} to \Cref{fig:atn_comparison_unnorm}, and \Cref{fig:atn_comparison_norm_separated} to \Cref{fig:atn_comparison_norm}. As expected, individual fairness, which is a stricter requirement than group fairness, often comes at a higher cost in terms of ratio-cut. However, the difference in ratio-cut is competitive in most cases with a much higher gain in terms of average balance. As before, when the approximate variants of our algorithms are used, one can choose the rank $R$ used for approximation in a way that trades-off appropriately between a quality metric like the ratio-cut and a fairness metric like the average balance.}

